\documentclass[a4paper,11pt,twoside]{book}

\newcommand{\linespacing}{1.2}
\renewcommand{\baselinestretch}{\linespacing}

\usepackage[authoryear]{natbib}
\usepackage{pifont}
\usepackage{colortbl}
\usepackage{latexsym}
\usepackage{datetime}
\usepackage{booktabs}
\usepackage{color,soul}
\usepackage{graphicx,color}
\usepackage{epstopdf}
\usepackage[british]{babel}
\usepackage{pdflscape}
\usepackage{afterpage}
\usepackage{adjustbox}
\usepackage{multirow}
\usepackage{rotating}
\usepackage{lettrine}

\usepackage{multicol}
\usepackage{graphicx}
\usepackage{caption, subcaption, float}
\usepackage{longtable}
\usepackage[normalem]{ulem}
\useunder{\uline}{\ul}{}
\usepackage{pgfplots}
\pgfplotsset{compat=1.18}
\usepackage{rotating}
\usepackage{amsmath,amssymb}
\usepackage[ruled,linesnumbered,resetcount,algochapter]{algorithm2e}
\usepackage[noend]{algpseudocode}
\usepackage{enumitem}
\usepackage[utf8]{inputenc}
\usepackage{afterpage}

\usepackage{multirow}
\usepackage{makecell}
\usepackage{tabularx}
\usepackage{mathtools}
\usepackage{comment}
\usepackage{csquotes}
\usepackage{gensymb} 
\usepackage{listings} 
\usepackage{stmaryrd}
\usepackage[T1]{fontenc}
\usepackage{lmodern}

\usepackage[a4paper,top=2.5cm,bottom=2.5cm,left=4cm,right=2cm,headsep=10pt]{geometry}
\usepackage{fancyhdr}
\usepackage{fancybox}
\usepackage{tablefootnote}
\usepackage{todonotes}
\usepackage{multirow}

\newtheorem{definition}{Definition}[section]
\newtheorem{proposition}{PROPOSITION}[section]
\newtheorem{proof}{Proof}[section]
\newtheorem{prob}{Problem Statement}
\newtheorem{lemmasec}{Lemma}[subsection]

\newtheorem{exmp}{Example}[subsection]

\newcommand{\xra}[1]{\overset{#1}{\rightsquigarrow}}

\newcommand{\ttle}{Explainable Planning for Hybrid Systems}
\usepackage{xcolor,colortbl}
\usepackage{textcomp} 

\definecolor{Gray}{gray}{0.85}
\definecolor{LightCyan}{rgb}{0.88,1,1}
\definecolor{corn}{rgb}{0.98, 0.93, 0.36}
\definecolor{naplesyellow}{rgb}{0.98, 0.85, 0.37}
\definecolor{lightyellow}{rgb}{1.0, 1.0, 0.88}
\definecolor{lightblue}{rgb}{0.31,0.34,0.34}
\definecolor{aliceblue}{rgb}{0.94, 0.97, 1.0}
\definecolor{blizzardblue}{rgb}{0.67, 0.9, 0.93}

\newcommand{\uls}{\begin{itemize}[leftmargin=*]}
\newcommand{\ule}{\end{itemize}}
\newcommand{\ols}{\begin{enumerate}[leftmargin=*]}
\newcommand{\ole}{\end{enumerate}}

\usepackage[colorlinks,pagebackref,pdfusetitle,urlcolor=blue,citecolor=blue,linkcolor=blue,bookmarksnumbered,plainpages=false]{hyperref}

\begin{document}



\thispagestyle{empty}

\begin{center}

\vspace{3cm}


\huge\textbf{Explainable Planning for Hybrid Systems}
\vspace{3cm}

\Large \textbf{Thesis Submitted for the Degree of\\
Doctor of Philosophy} \\
\vspace{.5cm}
\Large \textbf{by}\\
\vspace{.75cm}

\textbf{Mir Md Sajid Sarwar} \\
\textbf{Registration No. 2022 03 05 01 02 073}\\
\vspace{2cm}

\begin{figure}[htb]
\begin{center}
\includegraphics[width=4.5cm]{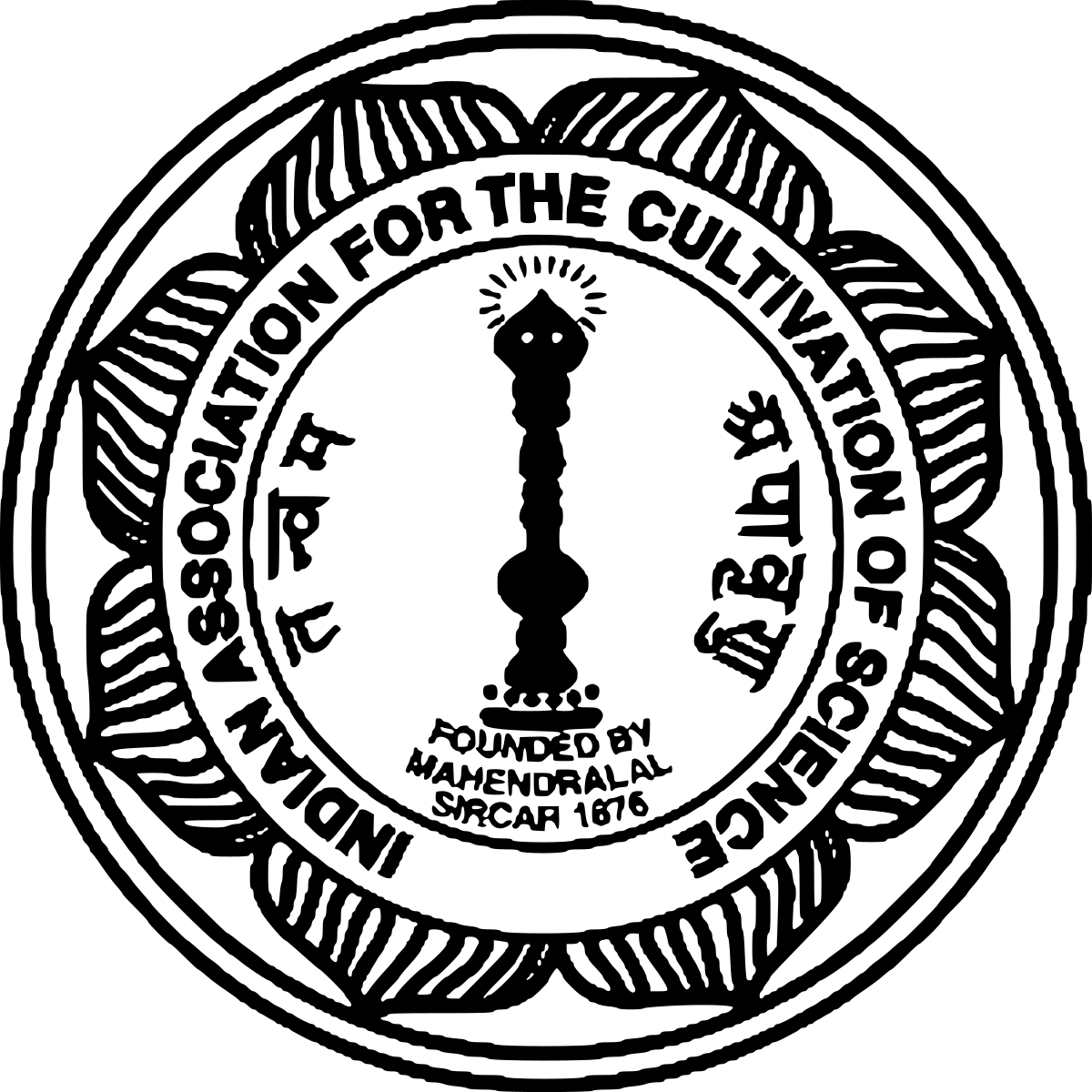}
\end{center}
\end{figure}

\vspace{1.5cm}
\Large \textbf{School of Mathematical \& Computational Sciences}\\
\textbf{Indian Association for the Cultivation of Science}\\
\textbf{Kolkata, India}\\
\Large
\textbf{June 2025}
\end{center}

\newpage\null\thispagestyle{empty}


\chapter*{}
\thispagestyle{empty}
\begin{center}\textbf{Indian Association for the Cultivation of Science \\ KOLKATA-700032, INDIA} \end{center}

\begin{flushright}
\end{flushright}

\begin{enumerate}
    \item \textbf{Title of the Thesis:} \\ \ttle
    \item \textbf{Name, Designation \& Institution of the Supervisor/s:}\\ Dr. Rajarshi Ray\\ Associate Professor,\\ School of Mathematical \& Computational Sciences, \\ Indian Association for the Cultivation of science,\\ Kolkata - 700 032, India\\
   
    \item \textbf{List of Publications:}
        \begin{enumerate}[label=(\Alph*)]
            \item \textbf{Journal Publications (2):}
            \begin{enumerate}
                \item {\bf Sarwar, Mir Md Sajid}; Ray, Rajarshi; and Banerjee, Ansuman. ``Contrastive Plan Explanation Framework for Hybrid System Models'', {\em ACM Transactions on Embedded Computing Systems}, Volume 22, Issue 2, Article 22, Pages 1-51, Year (March 2023). [doi:10.1145/3561532].

                \item {\bf Sarwar, Mir Md Sajid}; and Ray, Rajarshi. ``Exploring Inevitable Waypoints for Unsolvability Explanation in Hybrid Planning Problems'', {\em ACM Transactions on Embedded Computing Systems}, volume 24, Issue 6, Article 163, Pages 1-20, Year (October 2025). [https://doi.org/10.1145/3767745].
                       
            \end{enumerate}
        
            \item \textbf{Conference Publications (4):}
        
            \begin{enumerate}
            
                \item {\bf Sarwar, Mir Md Sajid}, Ray, Rajarshi; and Banerjee, Ansuman. ``Contrastive Plan Explanation Framework for Hybrid System Models.'' {\em In Proceedings of the 18th ACM-IEEE International Conference on Formal Methods and Models for System Design (MEMOCODE 2020)}. [doi:10.1109 /MEMOCODE51338.2020.9315040].

                \item {\bf Sarwar, Mir Md Sajid}, Ray, Rajarshi; and Banerjee, Ansuman. ``Explaining Unsolvability of Planning Problems in Hybrid Systems with Model Reconciliation.'' {\em In Proceedings of the 21st ACM-IEEE International Conference on Formal Methods and Models for System Design (MEMOCODE 2023)}. [doi:10.1145/3610579.3611082]

                \item {\bf Sarwar, Mir Md Sajid}; Yadav, Rajeshwar; Samanta, Sudip; Ray, Rajarshi; Halder, Raju; Banda, Gourinath; Bhattacharya, Ansuman; and Thakur, Atul. ``A Robotic Software Framework for Autonomous Navigation in Unknown Environment.'' {\em In Proceedings of the International Symposium of Asian Control Association on Intelligent Robotics and Industrial Automation (IRIA 2021)}. [doi:10.1109/IRIA53009.2021.9588693]

                \item {Dey, Devdan}; {\bf Sarwar, Mir Md Sajid}; Ray, Rajarshi; and Banerjee, Ansuman. ``A Contrastive Explanation Tool for Plans in Hybrid Domains.'' {\em In Proceedings of the 17th Innovations in Software Engineering Conference (ISEC 2024)}. [doi:10.1145/3641399.3641424].
             \end{enumerate}
    \end{enumerate}
    

    
    \item \textbf{List of Presentations in National / International / Conferences/ Workshops/ Symposiums:}
    \begin{itemize}
      \item  18th ACM-IEEE International Conference on Formal Methods and Models for System Design (MEMOCODE 2020), Virtual Event, December 02-04, 2020, Jaipur, India.
      \begin{itemize}
         \item  Contrastive Plan Explanation Framework for Hybrid System Models. (\textbf{Oral})
      \end{itemize}
     
      \item  21st ACM-IEEE International Conference on Formal Methods and Models for System Design (MEMOCODE 2023), September 21-22, 2023, Hamburg, Germany.
      \begin{itemize}
         \item  Explaining Unsolvability of Planning Problems in Hybrid Systems with Model Reconciliation. (\textbf{Oral})
      \end{itemize}
     
      \item  17th Innovations in Software Engineering Conference (ISEC 2024), February 22-24, 2024, Bangalore, India.
      \begin{itemize}
         \item  A Contrastive Explanation Tool for Plans in Hybrid Domains. (\textbf{Oral})
      \end{itemize}
    
      \item  International Symposium of Asian Control Association on Intelligent Robotics and Industrial Automation (IRIA 2021), Virtual Event, September 20-21, 2021, Goa, India.
      \begin{itemize}
         \item   Robotic Software Framework for Autonomous Navigation in Unknown Environment. (\textbf{Oral})
      \end{itemize}
      
      \item  Formal Methods Update Meeting, July 4-5, 2022, IITDelhi, India.
      \begin{itemize}
         \item  Contrastive Plan Explanation Framework for Hybrid System Models. (\textbf{Oral})
      \end{itemize}

      \item Research Highlights in Programming Languages, December 16-18, 2024, IIT Gandhinagar, India.
      \begin{itemize}
         \item Explaining Unsolvability of Planning Problems in Hybrid Systems with Model Reconciliation. (\textbf{Oral})
         \item Exploring Inevitable Waypoints for Unsolvability Explanation in Hybrid Planning Problems. (\textbf{Poster})
      \end{itemize}

     \end{itemize}

\item \textbf{List of Additional Publications (Not Relevant to PhD Thesis):}
        \begin{enumerate}
            
                \item Iqbal, Sk Asif; \textbf{Sarwar, Mir Md Sajid}; and Ray, Rajarshi. ``Explaining Unsolvability of Planning Problems in Cyber-Physical Systems'' {\em In Preceedings of the 17th Innovations in Software Engineering Conference (ISEC) 2025)}. ACM, Article No.: 14, Pages 1 - 11. [https://dl.acm.org/doi/10.1145/3717383.3717395]

                \item \textbf{Sarwar, Mir Md Sajid}; Samanta, Sudip; and Ray, Rajarshi. ``AUTONAV: A Tool for Autonomous Navigation of Robots''. arXiv:2504.12318. Year 2025. [https://doi.org/10.48550/arXiv.2504.12318]
            
             \end{enumerate}
\end{enumerate}

\frontmatter
\pagenumbering{roman}
\setcounter{page}{1}

\begin{figure}[ht]
\centering
\includegraphics[width=\textwidth, height=1.2\textheight, keepaspectratio]{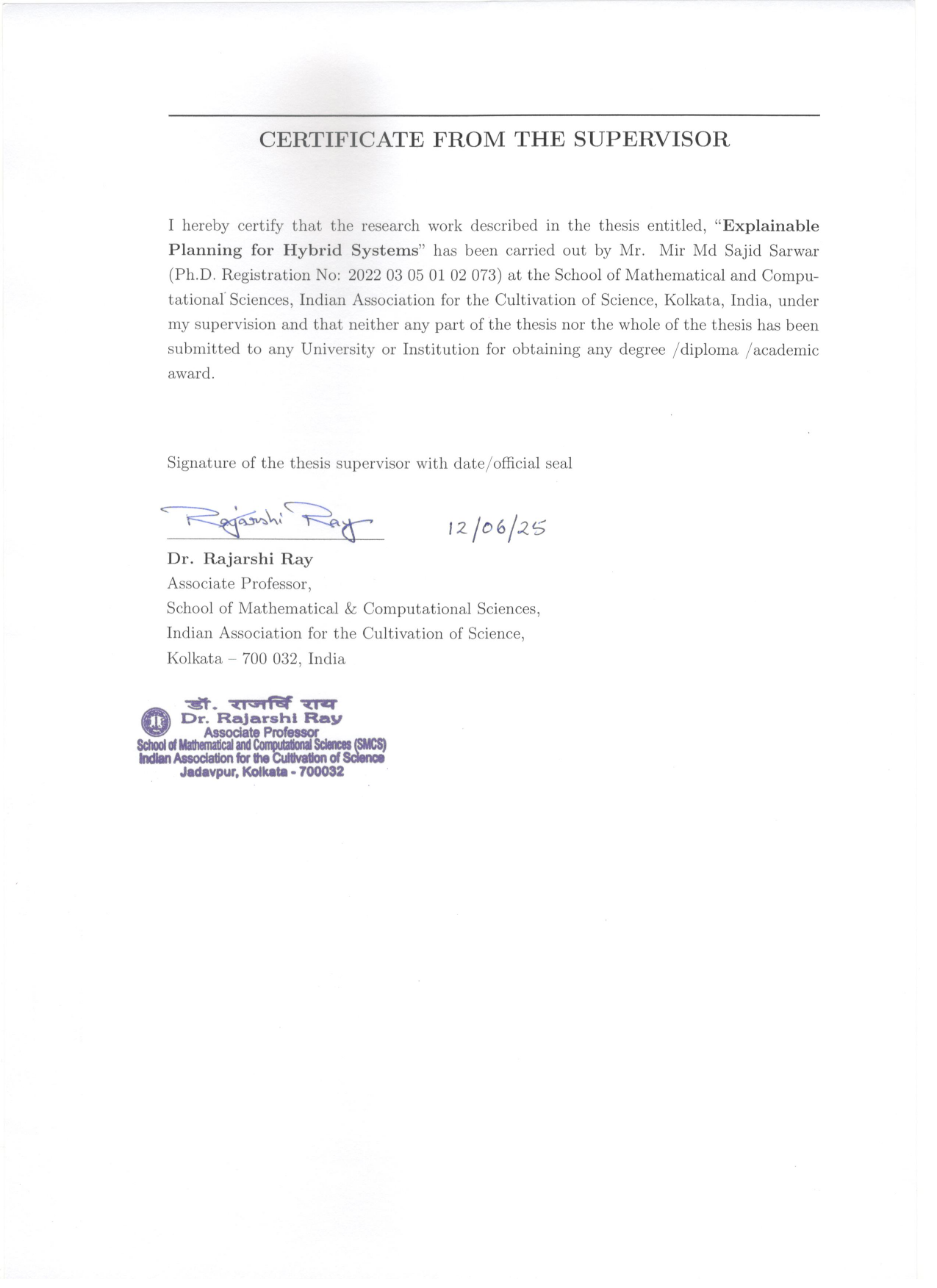}
\end{figure}

\begin{figure}[ht]
\centering
\includegraphics[width=\textwidth, height=1.2\textheight, keepaspectratio]{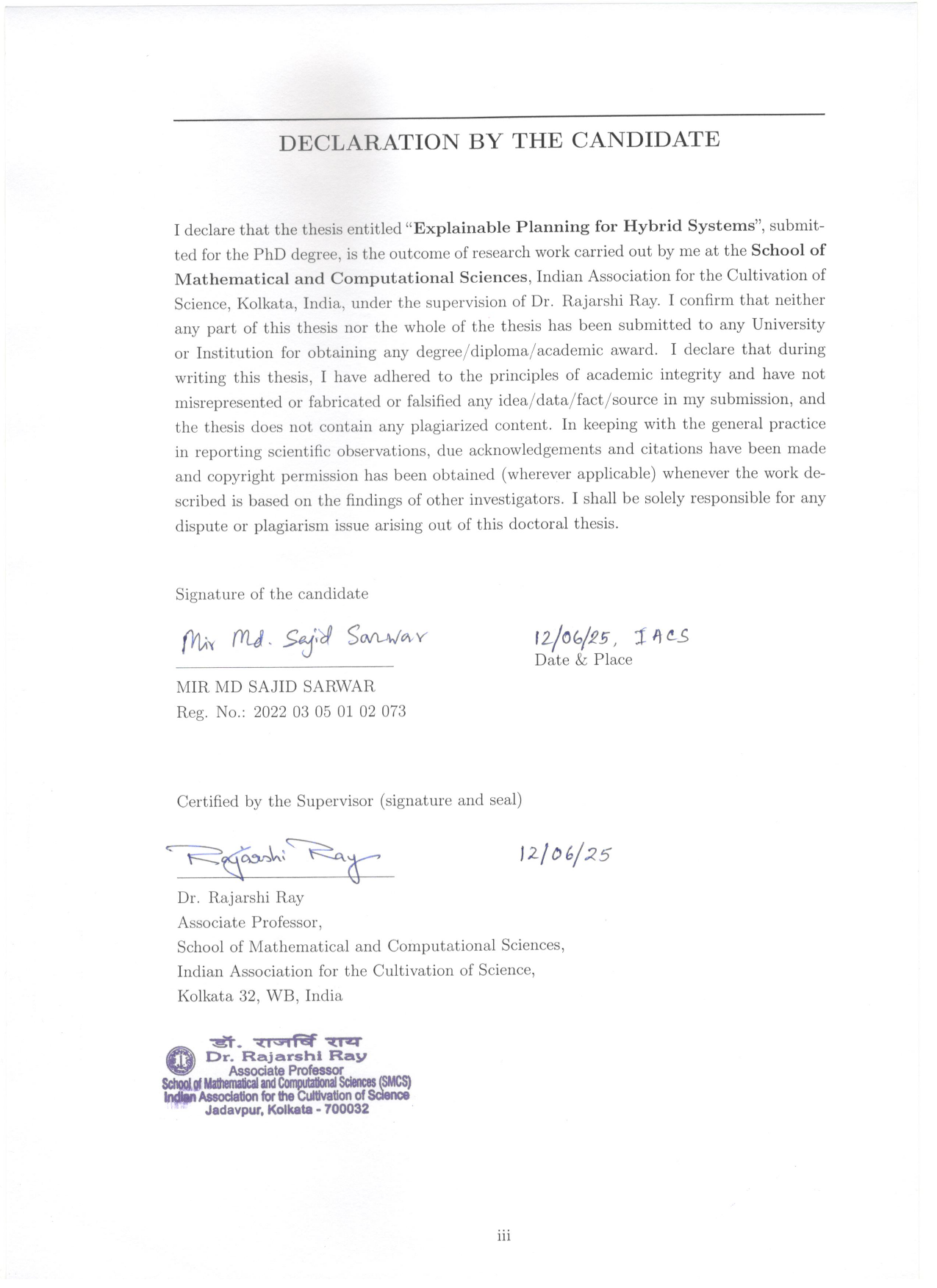}
\end{figure}

\newpage\null\thispagestyle{empty}\newpage

\chapter*{}
\thispagestyle{empty}
\begin{center}
\vskip5cm
\huge\textbf{Dedicated to\\ My Parents, My Wife, My Daughter\\ \& My Elder Sister.}
\vskip5cm
\end{center}

\newpage\null\thispagestyle{empty}\newpage

\chapter*{\centerline{Acknowledgements}}
\addcontentsline{toc}{chapter}{Acknowledgements}
This thesis marks the culmination of my enriching Ph.D. journey at the Indian Association for the Cultivation of Science (IACS), which began in October 2020. I'm delighted to express my gratitude to everyone who contributed to making this period a successful and transformative experience.

I extend my deepest gratitude to my supervisor, Dr. Rajarshi Ray, of the School of Mathematical \& Computational Sciences, Indian Association for the Cultivation of Science, Kolkata. His exceptional guidance, continuous support, and profound insights have been pivotal in shaping my research and pushing the boundaries of my academic capabilities. He truly taught me how to conduct rigorous research and consistently inspired me with innovative ideas and unwavering encouragement. I'm equally grateful to Dr. Ansuman Banerjee of the Indian Statistical Institute, Kolkata. It was a privilege to collaborate closely with him, and his invaluable mentorship, constructive feedback, and collaborative spirit significantly enriched my academic experience and the writing of my dissertation.

I'm immensely thankful for my parents' unwavering support. Their enduring belief in me has been my anchor on this arduous journey, and my gratitude for them knows no bounds.

I cannot imagine my research journey without the wonderful companionship and collaborative spirit of my lab mates, juniors, and friends here at IACS. A heartfelt thank you goes to Atanu, Devdan, Asif, Suman, Payel, Prantika, Madhusudan, Samiran, Shrimon, Avishek, Anal, Atanu (math), Goirika, Maitree, Sudip, Sarmistha, and Tanushree. Your insightful discussions, invaluable suggestions, and the sheer joy you brought made this journey truly special.

My sincere thanks go to all the faculty members, teaching and non-teaching staff, and the administration for their unwavering help and support throughout this endeavor. I am also deeply grateful for the financial and infrastructural support provided by the IMPRINT-2 project and the Institute Fellowship, generously funded by IACS and the Department of Science and Technology (DST), Government of India. This research would not have been possible without their collective contributions.

Lastly, I thank all those whom I have missed out on the above list.

\newpage\null\thispagestyle{empty}\newpage

\chapter*{\centerline{Abstract}}
\addcontentsline{toc}{chapter}{Abstract}
\renewcommand{\baselinestretch}{\linespacing}
\small\normalsize

The recent advancement in artificial intelligence (AI) technologies facilitates a paradigm shift toward automation. Autonomous systems are fully or partially replacing manually crafted ones. At the core of these systems is automated planning. With the advent of powerful planners, automated planning is now applied to many complex and safety-critical domains, including smart energy grids, self-driving cars, warehouse automation, urban and air traffic control, search and rescue operations, surveillance, robotics, and healthcare. There is a growing need to generate explanations of AI-based systems, which is one of the major challenges the planning community faces today. The thesis presents a comprehensive study on explainable artificial intelligence planning (XAIP) for hybrid systems that capture a representation of real-world problems closely.

While XAIP is the main focus of the thesis, we start investigating motion planning problems in hybrid systems and acquaint ourselves with the literature in planning. We propose a motion-planning algorithm for a robot in an unknown environment based on iterative constraint-solving. An integrated software framework consisting of a simultaneous localization and mapping module, a motion planning module, and a plan execution module specially designed for a \emph{lizard-inspired quadruped robot} has been designed and developed in collaboration with our colleagues. The techniques of motion planning and plan execution are the contributions of this thesis.

When an AI agent generates a plan for a given task, that plan needs to be understandable to its user. Contrastive approach has been a promising methodology from social sciences for explaining when there is a solution to a given planning problem. We propose a contrastive plan explanation framework for a hybrid system that builds a hypothetical model from the user's questions on the plan given by a planner and generates a hypothetical plan to provide an explanation by highlighting its contrast with the original plan. We discuss a few classes of contrastive questions. The proposed framework, based on the \emph{re-model and re-plan} idea, advocates the construction of a hypothetical model for each of these contrastive questions in such a way that any valid plan, if generated on this model, will imitate the contrastive question. In addition, we propose to prove the absence of a plan (\emph{no plan}) in a hybrid domain using bounded reachability analysis. Since planning problems are undecidable for hybrid systems in general, when a planner fails to generate a valid plan for a given problem instance, it is not clear whether this is due to the underlying undecidability or because the problem instance does not admit a valid plan.

With the objective of giving a user-friendly access to our explanation algorithms, we develop a tool with a web-based graphical user interface. The tool incorporates the iterative re-modeling and re-planning algorithm of our previous work to find a competitive contrastive plan with respect to the comparison metrics (e.g., plan makespan, plan length) that the underlying planner can provide. It provides provisions to experiment with two planners for hybrid domains and compare and contrast the explanations produced thereof. This work has been done in collaboration, where framework design and the algorithmic backbone of the tool are the contributions of this thesis.

\emph{No plan} situation, when there is no solution to a given planning problem, motivates us to look for an explanation of unsolvability. In this direction, \emph{model reconciliation problem} (MRP) is a successful approach to explanation when there is a mismatch of knowledge between the AI agent and the human. The reconciliation process brings these knowledge bases closer by providing model updates. We present a \emph{model reconciliation} framework for explaining the unsolvability of planning problems in hybrid domains, which facilitates explanation through a path-based reconciliation process. To this effect, we use a mix of \emph{graph traversal} and \emph{path analysis}, along with \emph{linear programming}, to carry out the reconciliation process. In particular, we use the concept of \emph{irreducible infeasible sets} (IIS) to generate explanations. The AI agent updates the human about the causes of the unsolvability of the planning problem through the explanation.

The divide and conquer strategy has a significant impact on many computer science problems. This approach has been widely applied to plan generation and automated problem solving to decompose tasks into sub-problems that help progressively converge towards the goal. We propose to adopt the same philosophy of sub-problem identification as a mechanism for analyzing and explaining the unsolvability of planning problems in hybrid systems. We present a framework that decomposes an unsolvable planning task into sub-problems through a novel waypoint identification method by casting it to an instance of \emph{longest common subsequence problem}, a widely popular approach in computer science, typically considered as an illustrative example for the dynamic programming paradigm.

This thesis explores the area of explainable planning for hybrid systems under the broader field of XAIP. As hybrid systems closely represent real-world problems, by attempting to explain the behaviour of such systems, we address the issues that touch everyday lives. The  \emph{contrastive explanation framework} provides explanations for the plans in Hybrid Systems. The \emph{contrastive explanation tool} provides a provision to experiment with different planning domains in hybrid systems and plug in different hybrid system planners as a plan generation engine. The \emph{model reconciliation framework} and \emph{sub-problem framework} can draw the causes of unsolvability. We believe these frameworks can be useful to the hybrid systems planning community for explaining the behaviour of autonomous systems.


\thispagestyle{empty}
\chapter*{}
\addcontentsline{toc}{chapter}{Title of the Thesis}
\begin{center}
\vskip2cm
\huge\textbf{Title of the Thesis}
\vskip2cm
\huge\textbf{\ttle}
\vskip2cm
\end{center}

\newpage\null\thispagestyle{empty}\newpage



\setcounter{tocdepth}{3}
\tableofcontents

\newpage\null\thispagestyle{empty}\newpage

\addcontentsline{toc}{chapter}{\listtablename}
\listoftables

\addcontentsline{toc}{chapter}{\listfigurename}
\listoffigures
\newpage\null\thispagestyle{empty}\newpage

\mainmatter

\pagenumbering{arabic}
\setcounter{page}{1}

\setcounter{secnumdepth}{3}


\chapter{Introduction}\label{ch1}
{\hypersetup{linkcolor=[rgb]{0.0, 0.0, 0.0}} 
}

\lettrine[lraise=0.5, nindent=0em, slope=-.5em]AI planning has been an active area of research for several decades. Autonomous systems are envisioned to automate and replace manually crafted ones, where planning remains a core component of such systems. Classical AI planning is concerned with finding a sequence of feasible actions from an initial configuration of a system to a desired goal. Planning has been an active area of research for AI practitioners, leading to the development of planners for diverse planning domains, goal descriptions, and varied optimization objectives. Techniques ranging from graph traversals to recent developments around constraint solvers for efficient and scalable planning have been widely explored.
As AI techniques mature, the number of application areas in which humans and autonomous agents collaborate increases. In such scenarios, cooperative plans derived from mutual trust and understanding are important for achieving a desired objective. Indeed, such systems' safety, robustness, and trustworthiness are analyzed before deployment. However, the major challenge that the planning community is facing today is interfacing with its users; that is, any AI-based system must be able to explain its reasoning to humans in the loop (\cite{DBLP:journals/aim/GunningA19}).
With automated planning being applied in safety-critical systems, the need for explanation and trust in the agent's behaviour has become ever more important. The ability to explain the rationale behind a decision of an autonomous agent is widely regarded as one of the precursors needed for humans to engage in trustworthy collaborations with autonomous agents.
Further, the autonomous agent often generates a plan based on an opaque model of the environment that is not transparent to the user. Thus, a plan explanation is crucial.

\section{AI Planning}



Planning is the branch of AI that seeks to automate reasoning about plans, most importantly, the reasoning that goes into formulating a plan to come up with a series of actions or procedures to accomplish a particular goal.
It is crucial for AI applications because it allows machines to \emph{think ahead}, \emph{adapt to changes}, and \emph{act autonomously}~\footnote{\url{https://www.geeksforgeeks.org/what-is-the-role-of-planning-in-artificial-intelligence/}}.
Just like humans plan their daily tasks with a goal in mind, AI systems use planning algorithms to assess the situation, identify the intended outcome, and develop a strategy that specifies the steps to take to get there. It is a model-based approach: a planning system takes a model of the environment, a description of the initial situation, the actions available to change it, and the goal condition as inputs and outputs a plan composed of those actions that will accomplish the goal when executed from the initial situation.

\begin{itemize}
    \item \textbf{Types of planning:} There are several types of planning approaches in AI, each suited to different tasks and environments:
    \begin{itemize}
        \item \emph{Classical planning} (\cite{FIKES1971189,DBLP:books/daglib/0014222}) is the traditional form of planning approaches that restrict the view of the world over which planning is performed. It assumes that the environment is fully observable and static, where all actions are deterministic. The AI agent has complete knowledge of the world and operates with a fixed goal, attempting to find a sequence of actions that leads from an initial state to a goal state.
        
        \item \emph{Probabilistic planning} (\cite{DBLP:journals/ai/KushmerickHW95}) extends classical planning by incorporating uncertainty and randomness into the planning process, allowing for more realistic and adaptable solutions in complex, unpredictable environments. The AI system must account for the fact that actions may have different possible outcomes with associated probabilities. Probabilistic planning often uses models like \emph{Markov Decision Processes} (MDPs) or \emph{Partially Observable Markov Decision Processes} (POMDPs) to manage this uncertainty.
        
        \item \emph{Hierarchical planning} (\cite{DBLP:conf/aips/ErolHN94}) breaks down complex tasks into simpler, smaller sub-tasks and creates a plan for each sub-task. This hierarchical approach is especially useful for solving large-scale problems where goals can be divided into manageable steps. It often involves decomposing high-level tasks into sequences of lower-level actions.
        
        \item \emph{Reactive planning} (\cite{DBLP:conf/ijcnn/Schmidhuber90}) denotes a group of autonomous agents' action selection techniques in highly dynamic and unpredictable environments. Rather than following a pre-defined plan, the AI agent continuously reacts to changes in the environment in real time. This approach doesn’t rely on creating a full plan ahead of time but focuses on immediate responses to the current situation.
               
        \item \emph{Temporal planning} (\cite{DBLP:conf/aips/GhallabL94,Haslum2019}) considers time restrictions and inter-dependencies between actions for reasoning about events and their temporal relationships. It ensures that the plan is workable within a certain time limit by taking into account the duration of tasks.

    \end{itemize}

    \item \textbf{Applications of planning:} AI planning is now being applied in diverse fields, demonstrating its adaptability and efficiency. A few significant applications are:

    \begin{itemize}
        \item \emph{Robotics:} (\cite{DBLP:journals/arcras/KarpasM20}) Automated planning allows robots to navigate efficiently in environments, avoid obstacles, and perform tasks autonomously. For example, an Amazon warehouse robot~\footnote{\url{https://www.aboutamazon.com/news/operations/amazon-robotics-robots-fulfillment-center/}} can plan its path to pick up items without collisions.

        \item \emph{Search and rescue operations:} (\cite{620182}) AI is revolutionizing search and rescue missions by enabling efficient planning, resource allocation, and real-time data analysis, ultimately improving the speed and precision of rescue operations.

        \item \emph{Autonomous Vehicles:} Self-driving cars (\cite{DBLP:journals/bdcc/GarikapatiS24}) use planning to navigate roads, make turns, stop at traffic signals, and avoid collisions with pedestrians or other vehicles.

        \item \emph{Healthcare:} Planning systems are used in treatment (\cite{https://doi.org/10.1002/acm2.13837}), where algorithms suggest optimal therapies for patients based on various factors like medical history, current health, and probability of success.

        \item \emph{Gaming:} In video games (\cite{DBLP:conf/cf/Yannakakis12}), planning is used to simulate intelligent behavior in non-player characters (NPCs). NPCs can plan their strategies in real-time, providing more challenging and unpredictable gameplay.

        \item \emph{Logistics:} Planning in AI optimizes logistics, inventory, and transportation, helping businesses improve efficiency and reduce costs. It can plan the most cost-effective routes for shipping goods or the best times to restock inventory.
    \end{itemize}
    
    \item \textbf{Challenges of planning:} While AI planning has many advantages, many issues need to be resolved. Typical challenges include:

    \begin{itemize}
        \item \emph{Complexity:} Planning, especially in complex environments, can be computationally expensive. Finding the optimal sequence of actions in large, dynamic systems can take a significant amount of processing power and time.

        \item \emph{Uncertainty:} In uncertain or unpredictable environments, creating a plan that can handle every possible outcome is challenging. Probabilistic and reactive planning methods aim to address this, but it remains a difficult problem.

        \item \emph{Scalability:} As the size of the problem or task increases, so does the difficulty of planning. Scaling up planning algorithms to handle large datasets or environments with numerous variables is a technical hurdle.

        \item \emph{Trust:} Trust in AI hinges on transparency, reliability, and accountability. For AI to be trustworthy, its autonomous actions must be consistent and dependable. Organizations should clearly show how their AI systems plan and operate to foster transparency and build explainable systems that earn user confidence.
    \end{itemize}
\end{itemize}

    
\subsection{Planning in Hybrid Systems}
Planning in hybrid systems involves creating plans for systems with both discrete and continuous variables, which are common in real-world applications such as autonomous navigation, control systems, industrial production processes, power systems, robotics, etc. It requires specialized planning techniques to handle these complex dynamics. Therefore, planning in hybrid systems is essential for designing and controlling these systems effectively, ensuring they achieve desired goals while considering constraints and dynamics. We discuss the relevant background in Chapter~\ref{ch6}.

\begin{itemize}
    \item \textbf{Modeling hybrid systems:} Popular techniques in literature for modeling hybrid systems are as follows:
    \begin{itemize}
        \item \emph{PDDL+} (Planning Domain Definition Language +) (\cite{PDDL+}) is an extension of PDDL designed to model hybrid systems, allowing for the representation of continuous processes and events alongside discrete actions.

        \item \emph{Hybrid Automata} (HA) (\cite{10.1007/3-540-57318-6_30,ALUR19953}) are a formal framework for modeling hybrid systems, combining discrete state transitions with continuous dynamics. 

    \end{itemize}

    \item \textbf{Planning approaches in hybrid systems:}
    \begin{itemize}
        \item \emph{Satisfiability Modulo Theories} (SMT) planning approach for hybrid systems encodes planning problems as \textit{first-order logic formulae} in a theory, allowing for formal analysis and planning of systems with both discrete and continuous dynamics. SMT solvers (\cite{10.1007/978-3-540-78800-3_24}) provide solutions to such encoding, allowing for efficient plan generation and verification (\cite{DBLP:journals/jair/CashmoreMZ20}).

        \item \emph{Mixed Integer Linear Programming} (MILP) is a mathematical optimization technique used to find the best solution to problems with both continuous and discrete variables. It can be used to formulate and solve planning problems in hybrid systems, such as path planning (\cite{DBLP:conf/amcc/RichardsH02}), scheduling, and resource allocation by modeling constraints and objectives as linear equations or inequalities.

        \item \emph{Heuristic search} has been used in hybrid systems where planning problems can be cast as sequential decision-making problems, provided some time discretizations (\cite{DBLP:conf/ecai/ScalaHTR16}).
        Such methods exploit the planning-via-discretizations approach where the continuous dynamics of a model is approximated with uniform time steps and step-functions (\cite{DBLP:phd/ethos/Piotrowski18}).

        \item \emph{Reinforcement learning} (\cite{DBLP:journals/arc/BusoniuBTKP18}) can be used to learn optimal policies for controlling hybrid systems in dynamic environments. It focuses on using AI algorithms to learn optimal strategies for managing complex systems with multiple components. This is particularly useful in areas like energy management for hybrid electric vehicles (HEVs) and microgrids, where RL can adapt to changing conditions and optimize resource allocation
    \end{itemize}
\end{itemize}

\subsection{A Case Study: Motion-planning Problem in Robotics from a Hybrid System Perspective}

The motion planning problem for robots traditionally consists of finding a state trajectory and associated inputs, connecting the initial state to a state in the goal region while satisfying the system dynamics and safety criterion, for example, avoiding obstacles. It has been widely applied to many real-world applications, such as self-driving cars (\cite{DBLP:journals/tiv/TengHDLLAYLXZC23}), unmanned aerial vehicles (UAV) (\cite{DBLP:conf/iros/LiuAMK17}), free-floating space manipulator (\cite{DBLP:journals/robotica/Rybus20}), biped robots (\cite{DBLP:journals/trob/HuangYKKAKT01}), and robotic arms (\cite{Liu2021ARO}), etc., to perform complex tasks. 
Various approaches to address the motion-planning problem exist in the literature, such as RRT (Rapidly-exploring Random Tree) (~\cite{Lavalle98rapidly-exploringrandom}), A$^*$ (\cite{4082128}), SAT, and SMT-based path-planning (~\cite{6906597,6942758}). However, a class of algorithms that have been particularly successful in solving such problems for robot models with differential constraints are the sampling-based algorithms (\cite{Choset2005PrinciplesOR,DBLP:books/daglib/0016830}).

\begin{itemize}
    \item \textbf{Motion planning in unknown or partially known environment:} The problem of finding a safe (i.e., collision-free) path from an initial state to a goal state when the navigational space is a priori unknown and is incrementally observable with the robot's movement in the environment through line-of-sight perception has ubiquitous applications. However, motion-planning problems in such a setup have received relatively little theoretical investigation as compared to the problems where the environment is known. Most of the motion-planning algorithms assume that the environment is known and thus can not be used directly (\cite{DBLP:books/daglib/0016830}). A graph-based approach based on D$^*$ algorithm is proposed in (\cite{Stentz93optimaland}) for path-planning in unknown, partially known, and changing environments. It models the environment as a graph where nodes represent the robot's states and arcs represent the cost of moving between two states. (\cite{DBLP:journals/corr/abs-1804-05804}) introduced re-planning and forward-looking biasing in a self-contained framework while avoiding \emph{inevitable collision states} (ICS) (\cite{doi:10.1163/1568553042674662}) for motion-planning in a partially known environment. Sampling-based methods like RRT$^*$ (\cite{doi:10.1177/0278364911406761}), and its many variants have had considerable success for kinematic motion-planning. However, the complexity of such methods heavily relies on robot dynamics, and the global computation over the entire state space for high-dimensional systems (like legged robots) in cluttered environments renders them slow. A similar approach, SweepingRRT (\cite{9172596}), uses a global and a local plan for motion-planning when complete environment information is not available. Global plan is computed less frequently, observing the large obstacles which are a priori known, while the local plan is computed frequently as smaller obstacles are incrementally detected on the global path. 

    \item \textbf{Motion planning from a hybrid systems perspective:}
    Motion planning requires consideration of both continuous dynamics and discrete dynamics. The continuous dynamics arise from the robot's mechanical design, whereas the discrete dynamics arise from its internal logic/timer to resolve tasks and interactions with the environment.
    Motion planning for continuous-time systems (known as kinodynamic planning aims to generate trajectories) and discrete-time systems (aims to generate discrete poses) are well-studied problems (\cite{DBLP:books/daglib/0016830}). Most existing algorithms (\cite{Lavalle98rapidly-exploringrandom}) incrementally construct a search tree (rooted in the initial state) in the state space by adding random samples while trying to find a path that connects the initial and final states in the search tree. These algorithms use operations, such as concatenation, well-defined for continuous and discrete-time systems. However, the complex domain structure inherent in hybrid systems makes it exceedingly difficult to establish well-defined operations on trajectories. Hybrid systems may exhibit the following behaviors: 1) evolves continuously, 2) makes discrete transitions (or jumps) all the time, 3) evolves continuously and exhibits one or multiple jumps at times, or 4) exhibits Zeno behavior (where a system undergoes an infinite number of discrete transitions within a finite amount of time. HyRRT (\cite{DBLP:journals/corr/abs-2406-01802}) attempts to provide mathematical definitions of operations and their analysis for motion planning algorithms for hybrid systems. (\cite{DBLP:conf/cdc/BhatiaKV10}) considers the problem of motion planning for mobile robots with nonlinear hybrid dynamics. It uses a multi-layered framework for solving planning problems. At the higher level, it employs a discrete abstraction of the hybrid system and suggests high-level plans, and at the lower level, a sampling-based planner uses the dynamics of the hybrid system and the suggested high-level plans to explore the state-space for feasible solutions. (\cite{DBLP:journals/fmsd/PlakuKV09}) approach motion-planning from a hybrid system falsification perspective, where robot motion planning is viewed as a search problem for a witness trajectory that satisfies certain invariants, such as ensuring that the robot motion respects dynamics constraints and avoids collision with obstacles.
  
\end{itemize}


\section{Explainable AI Planning (XAIP)}

As AI is increasingly being adopted into \emph{application-solutions}, the challenge of supporting interaction with humans is becoming more apparent. Partly, this is to support integrated working styles, in which humans and intelligent systems cooperate in problem-solving. It is also a necessary step in building trust with AI-based systems.
The need for \emph{explainable AI} (XAI) first became prominent in machine learning, where the lack of understandable decision rationales is particularly daunting. XAI concerns the challenge of shedding light on opaque models in contexts for which transparency is important, i.e., where these models could be used to solve analysis or synthesis tasks. While XAI at large is primarily concerned with learning-based approaches, model-based approaches are well-suited for explanation, and \emph{explainable AI planning} (XAIP) (\cite{DBLP:journals/corr/abs-1709-10256,DBLP:conf/atal/Kambhampati19}) can play an important role in addressing complex decision-making procedures.
One of the recent developments towards this end is the establishment of the XAIP Workshop~\footnote{https://kcl-planning.github.io/XAIP-Workshops/} at the International Conference on Automated Planning and Scheduling (ICAPS), the premier conference in the field.
From its inception, XAIP has garnered increasing interest due to its role in designing explainable systems that bridge the gap between theoretical and algorithmic planning literature and real-world applications (\cite{DBLP:conf/ijcai/ChakrabortiSK20}).


\subsection{XAIP Perspectives} \label{secCH1:perspective}
\begin{figure}[htbp]
    \centering
    \includegraphics[width= 0.5\textwidth]{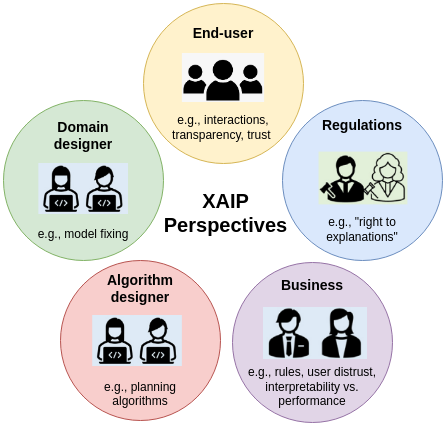}
    \caption{XAIP perspectives}
    \label{fig:xaip-perspective}
\end{figure}

XAIP can be discussed from different perspectives involved while considering the persona of the explainee (\cite{DBLP:conf/ijcai/ChakrabortiSK20}) as shown in Figure~\ref{fig:xaip-perspective}. The perspective groups below are to some extent based on the work in (\cite{DBLP:conf/fat/GadeGKMT20}, \cite{DBLP:journals/kbs/SaeedO23}), which are true for XAI in general but also acknowledged to be crucial to the XAIP scene as well (\cite{pat2019}, \cite{DBLP:conf/ijcai/ChakrabortiSK20}).

\begin{itemize}
    \item \textit{End user:} These are the individuals who will use or be impacted by the implementation of new technology and processes. They interact with the system in the form of a user. For example, this may be a passenger on an autonomous car, or a human teammate in a human-robot team (\cite{DBLP:conf/ijcai/ChakrabortiSK19}) who is affected by, or is a direct stakeholder in the agent's plans, or a user who collaborates with an automated planner in a decision support setting (\cite{DBLP:journals/hhci/GroverSCMK20}).

    \item \textit{Domain designer:} XAIP can be important in checking that the system adheres to desired properties in the pre-deployment phase. It helps a domain expert design better models by explaining the system's behaviour while interacting with the environment during test runs. For example, a designer of goal-oriented conversation systems~(\cite{DBLP:conf/aips/SreedharanCMKK20}) can take help from XAIP during the model acquisition process.

    \item \textit{Algorithm designer:} Algorithms play a crucial role in system performance, and XAIP could be useful as a feedback system in designing better algorithms. The role of an algorithm designer is distinct and may not even have any overlap in expertise with a domain designer~(\cite{DBLP:conf/aips/SreedharanCMKK20}): e.g., in the context of automated planning systems, this could be someone who works on an informed search.

    \item \textit{Regulations:} As AI-based autonomous systems are being used in many areas of our daily lives, it could result in unacceptable decisions being taken by such systems in certain situations. Such decisions need an explanation to the user. Especially those that may lead to legal effects. For example, suppose that an AI system rejects one’s application for a loan. In that case, the applicant has the right to request justifications behind that decision to ensure that the system adheres to the laws and regulations~(\cite{DBLP:series/lncs/SamekM19}). Thus, it poses a new challenge to the legislation. The General Data Protection Regulation (GDPR)~\footnote{\url{https://www.privacy-regulation.eu/en/r71.htm}} of the European Union establishes regulations for what is called the 'right to explanation', by which a user is entitled to request an explanation of the decision made by the algorithm that considerably influences them~(\cite{DBLP:journals/aim/GoodmanF17}).

    \item \textit{Business:} Winning user trust in any AI-based system is a major challenge for the industry. XAIP plays a crucial role in building a robust, transparent, and trustworthy system that can explain its behaviour to the end user~(\cite{DBLP:conf/ijcai/ChakrabortiSK20}). It helps in gaining the user's trust. However, it can increase development and deployment costs.


\end{itemize}

\subsection{Relevant XAIP concepts} \label{secCH1:xaip-concepts}
A \textbf{planning problem} $\Pi$ is a sequential decision-making problem for an AI agent. We can define $\Pi$ as a \textbf{transition function} $\delta_{\Pi}$ : $A \times S \rightarrow S \times \mathbb{R}$, where $A$ is the set of actions (or capabilities) available to the agent, and $S$ is the set of states in which it can be. The real number $\mathbb{R}$ denotes the transition cost. Thus, a transition $\delta \in \delta_{\Pi}$ defines the agent's behaviour in terms of actions, i.e., the prerequisites of actions and how it changes the state of the world. The planning algorithm $\mathbb{A}$ solves $\Pi$ subject to a desired property $\tau$ to produce a plan $\phi$, i.e. $\mathbb{A}$ : $\Pi \times \tau \mapsto \phi$. Here, $\tau$ may represent different properties such as soundness, optimality, and so on. 
A \textbf{plan} $\phi$ can be defined as a sequence of actions $\langle a_1, a_2, \dots, a_n \rangle$, $a_i \in A$ that transforms the current state $I \in S$ of the agent to its goal $G \in S$, that is, $\delta_{\Pi}(\phi,\ I)$ = $\langle G,\ \sum_{a_i \in \phi} c_i \rangle$, where $c_i$ is the cost of applying the action $a_i$.

\begin{itemize}
    \item \textbf{\textit{Explanation process}} in XAIP proceeds with a question from the explainee about a current solution or about unsolvability in the case when there is no solution for a given planning problem $\Pi$, and the explainer (the XAIP system) comes up with an explanation for it~(\cite{DBLP:conf/ijcai/ChakrabortiSK20}):
    
    \begin{itemize}
        \item Q: "\textit{Why} $\phi$?" or "\textit{Why not} $\phi'$?" - when $\Pi$ is solvable.
        
        Here, $\phi'$ is an alternate plan or a foil~(\cite{DBLP:journals/ai/Miller19}) which may be explicitly, implicitly, or even partially stated in the questions. Examples of foils would be:
        \begin{itemize}
            \item “\textit{Why} $a \in \phi$?” is a partial foil where all plans with action $a$ in them are the foils.
            \item The original question “\textit{Why} $\phi$?” where the implicit foil is “\textit{as opposed to all other plans} $\phi'$”.
         \end{itemize} 

        \item Q: "\textit{Why no plan}?" - when $\Pi$ is unsolvable.

        It is interesting when a planning problem is unsolvable for a planner. This may happen for two reasons: the planning problem is unsolvable and there is no feasible solution possible, or the planner cannot solve the problem due to its limitation and does not know whether the problem is solvable. In both cases, the XAIP system should provide reasonable explanations to the user.

        \item A: An explanation $\mathcal{E}$ ensures that the explainee can compute and verify that:

        \begin{itemize}
            \item $\mathbb{A}$ : $\Pi \times \tau \not\xmapsto{\mathcal{E}} \phi'$, or $\mathbb{A}$ : $\Pi \times \tau \xmapsto{\mathcal{E}} \phi'$, but $\phi \equiv \phi'$ or $\phi > \phi'$ (the comparison criterion may be cost, preferences, etc.).
            \item $\mathbb{A}$ : $\Pi \times \tau \xmapsto{\mathcal{E}} \emptyset$, when the planning problem is unsolvable.
        \end{itemize}

    \end{itemize}
     The Q $\&$ A continues until the explainee is satisfied, as (\cite{10.5555/2900929.2901038}) highlights that this approach to explanation is an iterative process.
    
    \item \textbf{\textit{Properties of explanation}:} The need for explanations arises when there is a mismatch between a proposed solution or the absence of a solution and the user's expectation.
    This might be because the user may not have formed an expected plan or because a plan was successfully constructed but does not match the proposed solution. The explanations attempt to bridge the gap between these mismatched positions.
    Explanations can be \textbf{local}, regarding a specific plan and its properties (\cite{DBLP:conf/kdd/Ribeiro0G16,DBLP:conf/indiaSE/DeySRB24}), or \textbf{global}, focusing on the assumptions on which the plan rests, the process by which it was constructed, or how the planning system works in general (\cite{DBLP:conf/icml/KimWGCWVS18}). (\cite{DBLP:journals/ai/Miller19}) provides an insightful view on explanations from the social sciences. It outlines three key properties for consideration: \textbf{social} in being able to model the expectations of the explainee, \textbf{selective} in being able to select explanations among several competing hypotheses, and \textbf{contrastive} in being able to differentiate properties of two competing hypotheses. The contrastive property, in particular, has received much attention (\cite{DBLP:conf/rweb/HoffmannM19,DBLP:journals/ker/Miller21}) in the XAIP community. \textbf{Abstraction} is a useful concept in which explanations given on an abstract model of a complex decision-making system are more helpful to the explainee (\cite{DBLP:conf/ijcai/SreedharanSSK19,DBLP:conf/kdd/Ribeiro0G16}).
    
\end{itemize}

\section{XAIP in Hybrid Systems}
Planning for hybrid systems is an important area in AI planning, mainly motivated by the need to deal with real-world problems. Hybrid systems closely model these problems involving continuous and discrete behaviour. Such systems, also known as Cyber-Physical Systems, have hybrid dynamics, subject to (continuous) physical effects and controlled by (discrete) digital equipment.
These systems are complex as they need to model complex domains involving continuous nonlinear methods, differential equations, fluid dynamics, etc.
As a result, the behaviour of such systems is also complex. Therefore, any automated plan in these domains needs explanations to the end user.
\begin{figure}[htbp]
    \centering
    \includegraphics[width= 0.5\textwidth]{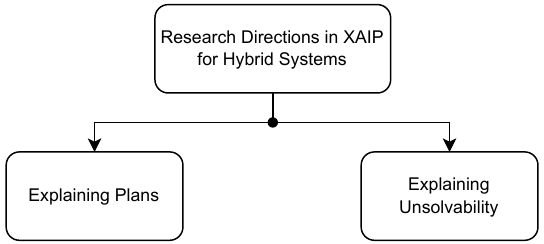}
    \caption{The proposed organization to discuss the challenges and research directions in XAIP for hybrid systems.}
    \label{fig:xaip-direction}
\end{figure}
XAIP plays a crucial role in designing and planning in the pre-deployment phase for such a system and explaining its behaviour to the users afterward. The current research directions in XAIP for hybrid system planning problems can be divided into two classes, as shown in Figure~\ref{fig:xaip-direction}.


\subsection{Explaining Plans in Hybrid Systems}
Explaining plans is the oldest branch of XAIP.
It aims to help humans understand the inner workings of a plan suggested by the AI system (\cite{DBLP:conf/flairs/McGuinnessGWS07,DBLP:conf/aips/KhanPB09,DBLP:conf/mkwi/BidotBHMNS10,DBLP:conf/aaai/SohrabiBM11,DBLP:conf/aips/SeegebarthMSB12,DBLP:conf/aips/BercherBGHNRS14,DBLP:conf/sigdial/NothdurftBBBM15}).
Different XAIP perspectives (discussed in Section~\ref{secCH1:perspective}) will require different types of explanations. These explanations can be classified into two primary classes: \textbf{algorithm-based explanations} and \textbf{model-based explanations} (\cite{DBLP:conf/ijcai/ChakrabortiSK20}).

Algorithm-based explanations attempt to explain the innards of the underlying planning algorithm and are generally useful for experts (i.e., algorithm designers). For example, (\cite{DBLP:books/sp/20/MagnaguagnoPMM20}) developed a cloud-based planning tool with state-space visualization to illustrate the operation of the planning process and how the domain dynamics evolve during the execution of the plan. It can also visualize fail planning instances, which is useful in debugging.

On the contrary, the majority of works in XAIP consider model-based explanations. This category consists of algorithm-agnostic methods for generating explanations since the properties of a solution can be evaluated independently of the method used to come up with them. Unlike debugging, requiring detailed algorithm-specific analysis, end users are primarily interested in model-based, algorithm-agnostic explanations so that services (\cite{cashmore2019towards}) can be built around it.

\subsection{Explaining Unsolvability in Hybrid Systems}
A special kind of “why” question is: “why didn’t you find a solution to this problem?” (\cite{DBLP:journals/jair/KrarupKMLC021}). While there has been a lot of research on generating explanations for planning problems, most of the earlier works in explanation generation have focused on explaining why a given plan or action was chosen (\cite{DBLP:conf/ijcai/ChakrabortiSZK17,DBLP:conf/aips/ChakrabortiKSSK19,DBLP:journals/jair/KrarupKMLC021}). However, explaining the unsolvability of a given planning problem remains a largely open and understudied problem.
The recent works that focus on explaining the unsolvability of planning problems have primarily concentrated on generating certificates or proofs of unsolvability (\cite{DBLP:conf/aips/ErikssonRH17,DBLP:conf/aips/ErikssonRH18}), these approaches, which are more oriented towards automatic verification, may fall short in adequately explaining unsolvability in complex planning domains. (\cite{DBLP:conf/aips/GobelbeckerKEBN10}) argues that excuses can be made for why a plan cannot be found by identifying counterfactual alterations to the original planning task to make it solvable. (\cite{DBLP:conf/ijcai/SreedharanSSK19}) use hierarchical model abstractions to generate the reason for unsolvability of planning problems. These hierarchical model abstractions relax a planning problem until a solution can be found. Then, they look for landmarks of this relaxed problem that cannot be satisfied in less relaxed versions of the problem. The unsatisfiability of these landmarks provides a succinct description of critical propositions that cannot be satisfied. \cite{DBLP:conf/aaai/EiflerC0MS20} derives plan properties that must be exhibited by all possible plans that could serve as explanations in case of unsolvability. However, most of these works are on classical planning problems. \textcolor{blue}{To the best of our knowledge, not much work addresses this issue for the hybrid system planning problems.}

\section{Popular Approaches to Explanation Generation in XAIP} 
In this section, we discuss a few important approaches to explanation generation in XAIP upon which we build our work in this thesis. However, the majority of the works in this direction are on discrete systems; we will highlight those that are relevant to hybrid systems whenever possible.

\subsection{Contrastive Explanation Approach}
When there is a mismatch between a proposed plan from an automated planner and the user's expectation, reconciliations are often required. The discrepancy could arise from either the user's failure to develop a predictive plan or a mismatch between their predicted plan and the one presented. Explanations serve to reconcile these differences. In (\cite{DBLP:journals/corr/abs-1709-10256}), the authors discuss how to achieve the goal of providing reasonable answers to user questions through explanations. Among the explanation properties highlighted in Section~\ref{secCH1:xaip-concepts}, contrastive explanations received significant research interest in the literature. (\cite{DBLP:journals/corr/abs-1709-10256,DBLP:conf/rweb/HoffmannM19}) highlights the role of contrastive questions in XAIP. Below, we discuss a few terminologies: 

\begin{itemize}
    \item \textbf{Contrastive questions:}  An important type of question in XAIP takes the form:
    \begin{itemize}
        \item “Why action A instead of action B?”
    \end{itemize}
    (\cite{DBLP:journals/corr/abs-1902-01876}) has shown that users tend to ask "why" questions when seeking explanations about a specific part of the plan, referred to as \emph{local questions}, while "how" or "what" questions are asked when seeking explanations about the plan as a whole, referred to as \emph{global questions}. Insights from social sciences suggest that these "why" questions are often contrastive (\cite{DBLP:journals/ai/Miller19}).

    \item \textbf{Contrastive plans:} A contrastive plan or a hypothetical plan incorporates the user-suggested foil. This is done by first deriving the constraints from the contrastive question posed by the user. These constraints are then imposed on the planning system such that any plan generated by the planner must adhere to the user suggestions. 

    \item \textbf{Contrastive explanations:} When a contrastive question is posed about a plan, a contrastive explanation can be given, highlighting how the original plan differs from an alternative plan that incorporates the user's suggested foil. Offering contrastive explanations is both an effective way to improve understanding and a simpler approach than providing a full causal analysis (\cite{DBLP:journals/ai/Miller19}). Furthermore, their inherent structure facilitates comparisons between the original plan and the one incorporating the user's suggested alternative.      
\end{itemize}

\noindent (\cite{DBLP:conf/aaai/EiflerC0MS20}) provides contrastive explanations by deriving plan properties that must hold if a contrast case was in the plan. (\cite{krarupPDDL2.1CE}) focuses on local explanations of temporal and numeric planning problems, formally describing the compilation from user questions to constraints in a PDDL2.1 planning setting, and explaining why a planner has made a certain decision. (\cite{DBLP:conf/ijcai/KimMSAS19}) introduce a Bayesian inference framework of linear temporal logic specifications to generate differences between plan traces for inferring contrastive explanations. (\cite{DBLP:conf/aips/BercherBGHNRS14}) gives contrastive explanations to user queries to help them assemble a home theatre by providing the reasons for an action's inclusion in the plan. (\cite{DBLP:conf/iccps/ZhaoS19}) discuss how such approaches can have interesting applications in cyber-physical systems (CPSs). Contrastive explanation approach has also been applied to explain machine learning based models. \cite{DBLP:conf/nips/DhurandharCLTTS18} proposes a \emph{contrastive explanations method} (CEM) to generate explanations for differentiable models such as deep neural networks, where one has complete access to the model. In \cite{2019arXiv190600117D}, a \emph{model agnostic contrastive explanations method} (MACEM) is proposed to generate contrastive explanations for any classification model where one can query only the class probabilities for a desired input. 

\subsection{Model Reconciliation Approach}
In most human-AI interaction scenarios, humans often have their own preconceived notions and expectations regarding a system (\cite{CARROLL198845}), potentially leading them to evaluate plans based on their own models, which may not align with the system's assessment of the result or quality. In this context, a recurring theme is the \emph{model reconciliation problem} (MRP) (\cite{DBLP:conf/ijcai/ChakrabortiSZK17}), a paradigm that empowers an agent (the explainer) to generate explanations by considering the ``mental model" of the human user (the explainee), drawing on the \emph{theory of mind} (ToM) (\cite{Premack_Woodruff_1978}) from human psychology. These model-based explanations aim to explain a plan by transferring a minimum number of necessary updates from the agent's model to the user, effectively bringing the model of the user closer to the agent’s model (\cite{DBLP:conf/ijcai/ChakrabortiSZK17,DBLP:conf/aips/SreedharanCK18}). The process of explanations is thus a reconciliation of the agent's model $\Pi^A$ and the human mental model $\Pi^H$ so that both can agree on the property $\tau$ of the decision being made. The model reconciliation process requires that:
\begin{align*}
     & Given:\ \mathbb{A}:\ \Pi^A \times \tau \mapsto \phi \\
     & \Pi^H + \mathcal{E} \mapsto \hat{\Pi}^H\ such\ that\ \mathbb{A}:\ \hat{\Pi}^H \times \tau \mapsto \phi.
\end{align*}

\noindent Empirical evidence suggests that model reconciliation is a natural and effective approach for explaining classical planning problems to humans (\cite{DBLP:journals/corr/abs-1802-01013,DBLP:conf/hri/ZahediOCSK19}). Using map visualizations of a planning problem, these studies specifically showed that human users understood and believed model reconciliation explanations were necessary for explaining (classical planning) plans. (\cite{DBLP:conf/ijcai/ChakrabortiSZK17}) assumes that the user's model is known and proposes a method to generate minimally complete and monotonic explanations that update the user's model to accept a plan. Conversely, (\cite{DBLP:conf/aips/SreedharanCK18}) produces conformant explanations applicable to multiple potential user models when the exact user model is unknown. Both of these approaches consider only optimal solutions in classical planning.
An AI agent here creates the best possible plan based on its model $\Pi^A$, and a human interprets this plan using their own understanding $\Pi^H$. Explanations become necessary when the AI's ``best'' plan isn't also the best plan from the human's perspective.
However, the necessity of optimal plans for explanation is generally questionable, and optimal planning for hybrid systems is undecidable (\cite{DBLP:journals/tcs/AlurCHHHNOSY95}). (\cite{DBLP:conf/atal/KulkarniZCVZK19}) compute the plan distances between the agent's plan and a user expected one. Thereafter, it uses a machine learning based regression model on human-annotated plans and the plan distances to compute explicability distance that is then used as the heuristic to search for explicable plans.

\subsection{Logic-Based Approach}
A classical planning problem can be translated into a propositional satisfiability (SAT) problem with formulas representing the initial state, goal, and action dynamics over a maximum of $n$ time steps, where $n$ is usually the upper bound on the horizon of plan length (\cite{DBLP:conf/ecai/KautzS92}). Similarly, a hybrid system planning problem can be formalized as an SMT (Satisfiability Modulo Theories) formula in first-order logic interpreted in the theory of quantifier-free linear real arithmetic (\cite{DBLP:conf/aips/CashmoreFLM16}).
These logic-based frameworks offer attractive features that are desirable in explanation generations, such as \emph{Expressivity} and \emph{Traceability} (\cite{DBLP:journals/jair/VasileiouYSKCM22}).

\begin{itemize}
    \item \textit{Expressivity} refers to the expressive power of logical languages to describe various phenomena in a principled and axiomatic way and the ability to distinguish between certain structures defined in them.
    For instance, propositional logic uses a finite set of propositions $P = \{p_1, \ldots, p_n\}$ and models $M = \{\mu_1, \ldots, \mu_k\}$ representing truth assignments to $P$. In the case of a classical planning problem $\Pi$, each proposition $p_i$ encodes states, actions, and transitions up to a time horizon $n$. Each model $\mu_j$ assigns Boolean values to these propositions, describing the truth of states and actions at a specific time step within $\Pi$'s execution. A knowledge base of these propositions can then explain events that occurred during that time.

    \item \textit{Traceability} implies that given a logical description of a problem, it is easy to trace the reasons for particular ``behavior". For example, if a knowledge base KB encodes a planning problem $\Pi$, a valid plan $\phi$ is logically implied by KB. Therefore, deductive inference can trace the reasons for $\phi$'s validity, and these reasons, expressed in the logic of KB, can explain why $\phi$ is valid.
\end{itemize}

\noindent (\cite{DBLP:journals/jair/VasileiouYSKCM22}) presents a logic-based extension to MRP problems based on knowledge representation and reasoning for mixed discrete-continuous domains. It provides a framework for finding a subset of the knowledge base of the agent with which to reconcile the human knowledge base for explanations. (\cite{DBLP:conf/aips/BercherBGHNRS14}) uses a logic-based approach to answer the question ``why the action $a \in \phi$?" by deducing a causal link chain originating at $a$ that can be traced to the goal. There has been a long history of using such information to characterize plans in the context of plan modification and reuse. (\cite{DBLP:conf/aips/SeegebarthMSB12}) presents a formal approach to plan explanation. Information about plans is represented as first-order logic formulae, and explanations are constructed as proofs in the resulting axiomatic system.

\subsection{Divide and Conquer Strategy}
A well-known insight into human thinking and problem solving is that humans tend to decompose a problem into sub-problems that help in progressively converging towards the goal.
Many AI systems mimic this notion in the way they solve problems. For example, the main feature of the pioneering automated theorem prover \emph{logic theorist} is the use of a problem-subproblem hierarchy (\cite{LT-1956}).
This has been a popular approach in many other domains, such as robotics (\cite{24200}) and AI (\cite{DBLP:journals/ai/SuttonPS99}) apart from planning (\cite{DBLP:journals/jair/HoffmannPS04,DBLP:conf/ecai/LipovetzkyG12,DBLP:conf/aaai/RichterHW08}).
Authors in (\cite{DBLP:journals/jair/HoffmannPS04,DBLP:conf/ecai/LipovetzkyG12}) find sub-problems for a solvable planning problem of the discrete domains in terms of ordered landmarks. Landmarks are facts given as propositional formulas that must be true at some point in every valid solution plan.
An innovative technique for the identification of subproblems relevant to explaining the unsolvability of a planning problem in domains with discrete dynamics has been proposed in (\cite{DBLP:conf/ijcai/SreedharanSSK19}).


\section{Decidability of Planning Problems in Hybrid Systems}

Verifying the solvability of planning problems is undecidable for hybrid systems in general (\cite{DBLP:journals/tcs/AlurCHHHNOSY95}). When a planner fails to generate a valid plan for a problem, it cannot be asserted whether it is due to the underlying undecidability or that the problem is insoluble. However, some special classes of problems within the hybrid system are known to be decidable.
The state reachability problem for \emph{timed automata} (TA) is decidable (\cite{DBLP:journals/tcs/AlurD94}), which makes this an interesting sub-class of \emph{linear hybrid automata} (LHA). Its complexity class is \emph{PSPACE-complete}. However, some problems, like the general \emph{language inclusion problem} and the \emph{determinisability problem} for certain types of TA, are undecidable (\cite{DBLP:conf/concur/Clemente0P20}). 
Decidability results for TA are generalized to \emph{multirate automata} (MA), another subclass of LHA, with variables that run at any constant positive slopes (\cite{DBLP:conf/hybrid/NicollinOSY92,DBLP:journals/fmsd/AlurCH97}). Its complexity class is \emph{PSPACE-complete}.
The decidability problem for an initialized \emph{rectangular automata} (RA) is decidable under two restrictions: 1) whenever the activity of a variable changes, the value of the variable is reinitialized; 2) the values of two variables with different activities are never compared (\cite{DBLP:journals/jcss/HenzingerKPV98}). Its complexity class is \emph{PSPACE}. A RA is a multirate automaton (MA) if \textit{act(v)} (flow) is a singleton for all vertices $v$ of RA. An MA is a timed automaton (TA) if each variable of MA is either a \emph{clock} or a \emph{memory cell}. A variable is a memory cell if it has a slope of 0 at every vertex of RA. A variable is a clock if c has a slope of 1 at every vertex. A two-slope variable with slopes 0 and 1 is a \emph{stopwatch}. The reachability problem is undecidable in TA for a single stopwatch (\cite{DBLP:journals/jcss/HenzingerKPV98}).

The decision problems for classical planning, also known as PlanSAT problems, pose the question of whether there exists any plan that solves a planning problem, and are decidable (\cite{DBLP:books/aw/RN2020}). The proof follows from the fact that the number of states is finite. However, introducing function symbols expands the state space to infinity, rendering the problem only semi-decidable. This means we can devise an algorithm that correctly solves any solvable instance but might run indefinitely for unsolvable ones. Notably, the Bounded PlanSAT problem retains its decidability even when function symbols are included. For the formal proofs of these claims, refer to (\cite{DBLP:books/daglib/0014222}). PlanSAT and its bounded variant both reside within the complexity class \emph{PSPACE}, a significantly more challenging class than \emph{NP}. Problems in \emph{PSPACE} are solvable by a deterministic Turing machine using a polynomial amount of memory. Even under substantial constraints, these problems remain hard; for instance, eliminating negative effects of actions still leaves them \emph{NP-hard}. Interestingly, if we further restrict the problems by disallowing negative preconditions, PlanSAT's complexity drops to \emph{P}.

Verifying the unsolvability of planning problems in the \emph{temporal-planning} (TP) domain is decidable under the ANSO (action non-self-overlapping) assumption (\cite{DBLP:conf/aaai/PanjkovicMC22}). Its complexity class is \emph{PSPACE-complete}. In general, the complexity of TP depends on the domain of time. If time is interpreted as a discrete quantity, TP is \emph{EXPSPACE-complete} (\cite{DBLP:conf/aips/Rintanen07}). Instead, if time is interpreted as a dense quantity, TP is undecidable (\cite{DBLP:journals/ai/GiganteMMS22}).


\section{Thesis Outline}
The research directions discussed in the preceding sections motivate the research plan presented in this thesis. The rest of the chapters are organized as follows:

\begin{itemize}
    \item In Chapter~\ref{ch6}, we first present the background concepts of hybrid systems essential to this thesis. It then presents a motion-planning problem framed from a hybrid system's perspective. The primary goal of this exercise is to familiarize ourselves with the planning literature. We propose a motion-planning algorithm for a robot in an unknown environment based on iterative constraint-solving. An integrated software framework consisting of a simultaneous localization and mapping module, a motion planning module, and a plan execution module, specifically designed for a lizard-inspired quadruped robot, has been developed in collaboration. The techniques of motion planning and plan execution are the contributions of this thesis. We present performance of the the algorithm for planning tasks in simulation settings. The contents of this work have been published in \cite{9588693}.

    \item In Chapter~\ref{ch2}, we propose a contrastive plan explanation framework for hybrid system planning problems that builds a hypothetical model from the user's questions and generates a hypothetical plan to explain by highlighting its contrast with the original plan.
    We discuss a few classes of contrastive questions. The proposed framework, based on the \emph{re-model and re-plan} idea, advocates constructing a hypothetical model for each of these contrastive questions so that any valid plan, if generated on this model, will imitate the contrastive question.
    In addition, we provide a framework that verifies unsolvable planning problems (\emph{no plan} instances) and proves the absence of a plan using bounded reachability analysis.
    The contents of this work have been published in \cite{DBLP:journals/tecs/SarwarRB23} and \cite{DBLP:conf/memocode/Sarwar0B20}.

    \item In Chapter~\ref{ch3}, we present a contrastive explanation tool for plans in hybrid domains.
    The tool consists of (1) A web-based interactive GUI for selecting questions, viewing contrastive plans and the generated explanations, and (2) A back-end implementing an iterative \emph{re-modeling and re-planning} algorithm.
    The tool offers a collection of contrastive questions over a plan for users to select. An explanation is produced by contrasting the original plan against an alternative that meets the user's expectation implicit in the question. The tool has the provision to contrast with the best alternative that the underlying planner can generate in terms of plan metrics.
    The contents of this work have been published in \cite{DBLP:conf/indiaSE/DeySRB24}.


    \item In Chapter~\ref{ch4}, we propose a \emph{model reconciliation} framework for explaining unsolvability of hybrid system planning problems. We assume that the agent has a complete model of the environment, while the human has a partial or erroneous model and expects a plan for the planning problem when there is none. The explanation problem is presented as a process of continuous reconciliation between these two entities (agent and human) to make the human domain consistent with that of the agent. We use a mix of \emph{graph traversal} and \emph{path analysis}, along with \emph{Linear programming}, to carry out the reconciliation process. In particular, we use the concept of Irreducible Infeasible Sets (IIS) to generate explanations.
    The contents of this work have been published in \cite{DBLP:conf/memocode/SarwarRB23}.

    \item In Chapter~\ref{ch5}, we present a framework to decompose an unsolvable planning task in a hybrid system into sub-problems for analyzing and explaining unsolvability. In particular, for a given unsolvable planning problem, we propose a novel method to the waypoint identification problem by casting it to an instance of the \emph{longest common subsequence problem}. As waypoints appear on every path from the source to the planning goal, this work envisions such waypoints as sub-problems of the planning problem, and the unreachability of any of these waypoints as an explanation for the unsolvability of the original planning problem.
    The contents of this work have been published in \cite{10.1145/3767745}.

    \item Finally, in Chapter~\ref{ch7}, we summarize the presented methods, emphasizing their usefulness in explanation generation and developing XAIP systems from the hybrid system perspectives, and our observations on major issues of hybrid system planning problems such as scalability, complexity, and decidability. We conclude by outlining future research directions of this work.
\end{itemize}

\section{List of Contributions}
The following are the primary contributions in terms of novel frameworks that have been developed as part of this thesis:
\begin{itemize}
    \item A \emph{robotic software framework} that integrates SLAM, motion planning, and control for autonomous navigation for a \emph{lizard-inspired quadruped} robot. We present a \emph{motion planning algorithm} for the robot through \emph{constraint-solving}.

    \item A \emph{Contrastive Plan Explanation Framework} for explaining plans for hybrid system planning problems through the \emph{re-modeling and re-planning} strength of the framework for iterative users' questions.

    \item A framework for verifying \emph{no-plan} instances through bounded reachability analysis of the planning problem.

    \item A web-based interactive \emph{Contrastive Explanation Generation Tool} that facilitates experimenting with the explanation generation process for hybrid system planning problems with integrated \emph{state-of-the-art} hybrid system planners.

    \item A path-based \emph{Continuous Model Reconciliation} framework for explaining unsolvability of the hybrid system planning problems through a mix of \emph{graph traversal} and \emph{path analysis}.

    \item A novel approach to decompose an unsolvable planning task into sub-problems through \emph{waypoint identification} and explanation generation for the unsolvability via the \emph{reachability analysis} of the sub-problems.
\end{itemize}


\chapter{Hybrid System Planning} \label{ch6}

\textbf{\textsc{Chapter Abstract:}} \textit{This chapter first presents the background concepts of hybrid systems essential to the remainder of this thesis. Building upon this background, we introduce a motion-planning problem in robotics from a hybrid system perspective. This leads to a case study on the autonomous navigation of a \emph{lizard-inspired quadruped robot in an unknown environment}, where we present an integrated software framework comprising three key modules: a simultaneous localization and mapping (SLAM) module utilizing visual odometry, a motion-planning module based on constraint-solving, and a plan-execution module designed specifically for the robot. To demonstrate the framework's efficacy, we present the results of several navigation tasks conducted in various indoor simulation settings, utilizing a specific model of the quadruped robot.}


\section*{}
\lettrine[lraise=0.3, nindent=0em, slope=-.5em]Hybrid systems are controllable physical systems (such as robots or production plants) that combine discrete and continuous behavior. This dual nature is modeled by integrating continuous dynamics (described by time derivatives over state variables, such as $\dot{x}=v$) with discrete dynamics (instantaneous state changes). The standard modeling techniques include - \emph{PDDL+} (Planning Domain Definition Language +) (\cite{PDDL+}): An extension of \emph{PDDL} used in planning to represent continuous processes/events alongside discrete actions, and \emph{Hybrid Automata} (HA) (\cite{10.1007/3-540-57318-6_30,ALUR19953}): A formal framework that models hybrid systems by combining discrete state transitions with continuous dynamics, often used for verification and analysis.
The system's behavior is modeled by three main components:
\begin{itemize}
    \item Processes: Dictate the continuous dynamics, often using differential equations (effects are sets of time-derivative functions).
    \item Events: Formalize discrete changes that happen spontaneously in the environment when their conditions are met.
    \item Actions: Formalize the agent's decisions and what it can actively do (effects are assignments like $x:=\xi$).
\end{itemize}

Planning in hybrid systems is challenging for planners to solve due to several intersecting factors~(\cite{DBLP:conf/aaai/PiotrowskiFLMM16}):
\begin{itemize}
    \item Undecidability: Continuous variables cause the reachability problem to become undecidable~(\cite{DBLP:journals/tcs/AlurCHHHNOSY95}).
    \item Search Space Explosion: The combination of discrete state variables causing state explosion and complex system dynamics (often involving non-linear behaviors) results in immense search spaces.
\end{itemize}
The goal of a hybrid system planning problem remains to find a timed plan of actions that successfully transitions the system from an initial state to a goal state.

Section~\ref{ch6background} discusses background concepts of hybrid systems and planning. Section~\ref{ch6motion-planning} introduces the motion-planning problem in robotics from a hybrid system perspective. Section~\ref{ch6CaseStudy} presents a case study of autonomous navigation in an unknown environment.

\section{Background} \label{ch6background}

This section provides an overview of the background concepts essential to this thesis. More detailed definitions will be introduced as needed in subsequent chapters. We begin with introducing the hybrid automaton, a mathematical model used to describe hybrid systems.

\begin{definition} \label{ch2def:HA}
A hybrid automaton (HA) is a seven tuple $HA$=\big($Loc$, $Var$, $Flow$, $Init$, $Lab$, $Edge$, $Inv$\big) where:
\begin{itemize}
    \item $Loc$ is a finite set of vertices called locations.
    \item $Var$ is a finite set of real-valued variables. A {$valuation\ v$} for the variables is a function that assigns a real-value {$v(x)\in \mathbb{R}$} to each variable $x \in Var$. We write $V$ for the set of valuations.
    \item $Flow$ is a mapping from each location $l \in Loc$ to a set of differential equations $\{\dot{x} = f(x_1,\ldots,x_{|Var|}) \mid x \in Var\}$, where $\dot{x}$ denotes the rate of change of variable $x$.
    \item $Init$ is a tuple $\langle l_{ini}, S \rangle$ such that $l_{ini} \in Loc$ and $S \subseteq V$.
    \item $Lab$ is a finite set of labels.
    \item $Edge$ is a finite set of transitions $e = (l, a, g, r, l')$, each consisting of a source location $l\in Loc$, a target location $l'\in Loc$, label $a\in Lab$, a guard $g \subseteq V$ and a reset map $r: \mathbb{R}^{|Var|}\rightarrow \mathbb{R}^{|Var|}$.
    \item $Inv$ is a mapping from each location $l\in Loc$ to an invariant $Inv(l) \subseteq V$.
\end{itemize}
\end{definition}

\noindent A state in an HA is a pair $(l,v)$ consisting of a location $l\in Loc$ and a valuation $v\in V$. The locations $Loc$ and the transitions between them via the transitions in $Edge$ model the discrete dynamics, whereas $Flow(l)$ models the continuous change in a hybrid system. $Inv(l)$ are constraints on the HA states requiring that $v \in Inv(l)$, for every state $(l,v)$ of the HA. A HA behaviour is defined by a \emph{run}:

\begin{definition}[Run]\label{ch2def:run}
A \emph{run} of a hybrid automaton is a sequence
\begin{align*}
(\ell_0,x_{0}) \xrightarrow{\tau_0}(\ell_0,y_{0})\xrightarrow{e_0} (\ell_1,x_{1}) \xrightarrow{\tau_1} (\ell_1,y_{1}),\ldots, \xrightarrow{e_{N-1}}(\ell_N,x_{N})\xrightarrow{\tau_{N}} (\ell_N,y_{N})
\end{align*}
such that for all $i=0,\ldots,N-1$, (i) $(\ell_0, x_{0}) \in Init$; 
(ii) in each step $i$, the labeling function $e_i(l)$ maps the start location of the edge $e_i$ to $\ell_{i}$ and $e_i(l')$ maps the end location of the edge $e_i$ to $\ell_{i+1}$, where $\ell_{i},\ell_{i+1} \in Loc$.
(iii) $\forall t\in [0, \tau_i]$, $flow_{\ell_i}(x_i, t)\in Inv(\ell_i)$;
(iv) $flow_{\ell_i}(x_i,\tau_i) = y_i$;
(v) $y_i \in e_{i}(g)$, where $e_{i}(g)$ maps to the guard $g_i$ of the edge $e_i$;
(vi) $x_{i+1} = e_i(r)(y_i)$ and $y_N=flow_{\ell_N}(x_N, \tau_N)$ such that $\forall t\in [0, \tau_N]$, $flow_{\ell_N}(x_N, t)\in Inv(\ell_N)$.
The times $\tau_i$ are called the dwell times of the system in respective locations $\ell_i$. $\Box$
\end{definition}

\noindent A planning problem in a hybrid system requires two components: a domain description and a problem description. The domain is modeled using standard techniques (such as \emph{PDDL+} or \emph{Hybrid Automata}) and defines the system's dynamics. The problem description configures the initial and goal states for the specific planning task. We formally define a hybrid system planning problem as follows:

\begin{definition} \label{ch6def:PI}
A \emph{planning problem} $\Pi$ for a hybrid system is a pair ($Dom$, $Prob$), where $Dom$ defines a planning domain represented as hybrid automata \emph{HA}/PDDL+ domain, and $Prob$ represents a problem description defining the initial and goal configurations. $\Box$
\end{definition}

\noindent A system's state evolves under two dynamic modes:
\begin{enumerate}
    \item Continuous Evolution (Time Passage): State variables evolve according to the flow defined in the current location.
    \item Discrete Transition: An instantaneous change occurs when a labeled action/transition is applied. This requires the state's valuation to satisfy the transition's guard condition. The valuation of the resulting state is then dictated by the transition's reset map.
\end{enumerate}
Given these dynamics, we now define a plan for a planning problem $\Pi$ as follows:

\begin{definition}\label{ch6def:plan}
A plan for a planning problem $\Pi$ is a tuple $\langle$ $\lambda_n$, $makespan$ $\rangle$ where $\lambda_n$ is a finite sequence of $n$ pairs $\langle t_i, a_i \rangle$. In the pair, $t_i \in \mathbb{R}^+$ is the time instance of executing the action $a_i \in$ \emph{Lab}. In the sequence, $t_i$ is non-decreasing. The $makespan$ is the duration of the plan. $\Box$
\end{definition}
A planning problem $\Pi$ is solvable if a valid plan exists; otherwise, it is unsolvable.


\section{Motion Planning in Robotics from a Hybrid System Perspective} \label{ch6motion-planning}
Motion planning~\footnote{\url{https://en.wikipedia.org/wiki/Motion_planning}}, also known as path planning or the navigation problem, is a computational task that involves finding a valid sequence of configurations to move an object from a starting point to a destination. This concept is used in fields like computational geometry, computer animation, robotics, and computer games. For example, navigating a mobile robot to a distant waypoint inside a building, while avoiding walls and stairs, is a task that motion planning addresses. A motion planning algorithm accepts these task descriptions as inputs and outputs the speed and turning commands for the robot's wheels.
These algorithms tackle intricate scenarios, such as multi-joint robots (like industrial arms), object manipulation tasks, diverse constraints (e.g., a car's forward-only movement), and uncertainties in both the environment and robot models.
Motion planning has several robotics applications, such as autonomy, automation, navigation, and robotic surgery etc. It requires consideration of both continuous dynamics and discrete dynamics (\cite{DBLP:journals/corr/abs-2406-01802}). For instance, the position and velocity of a collision-resilient multicopter system in (\cite{DBLP:conf/icra/ZhaM21}) evolves continuously in open space, yet exhibits discrete state changes upon collision with a wall. In these situations, neither a purely continuous nor a purely discrete-time model is adequate for capturing the system's behavior. A hybrid system model is therefore essential, capable of capturing purely continuous, purely discrete, and combined behaviors. The continuous dynamics arise from the robot's mechanical design, whereas the discrete dynamics arise from its internal logic/timer to resolve tasks and interactions with the environment.
In this chapter, we present an integrated software framework for the autonomous navigation of a \emph{lizard-inspired quadruped robot} (\cite{lizardrobot}) designed for a stealth surveillance operation~\footnote{This work is a part of the Science and Engineering Research Board (SERB) project with File No. IMP/2018/000523 for developing a surveillance robot for security operations.}, while emphasis remains on the motion-planning problem of the robot. We design a hybrid controller that controls the navigation of the robot along a projected path.

\section{A Case Study for Autonomous Navigation in an Unknown Environment} \label{ch6CaseStudy}
Autonomous navigation is of central importance in robotics, with increasing use of robots in various applications, such as search and rescue operations~(\cite{620182}), warehouse automation~(\cite{BERTAZZI2013255}), surveillance~(\cite{10.1007/s10514-015-9503-7}), etc. Many of these applications require planning and control in unknown environments. Autonomously navigating a robot in an unknown scene comprises tasks such as: \textit{mapping} and \textit{localization} from the percepts received via sensors, safe motion-planning under partial knowledge of the scene, and finally, plan execution utilizing the available motion primitives of the robot. Though many software systems support these sub-tasks, there is a lack of integrated software that supports all the sub-tasks for autonomous navigation.

In this work, we present a robotic software architecture comprised of SLAM (simultaneous localization and mapping), motion-planning, and control modules~\footnote{This work is part of a collaboration. Motion planning and control modules are part of this thesis.}. The control module is mainly designed for driving a \emph{lizard-inspired quadruped robot} designed for a stealth surveillance operation. We show the utility of this robotic software in addressing various navigation tasks, emphasizing the motion-planning and control parts. To exemplify one such application, consider a hostage scenario, depicted in Figure \ref{fig:ch6-scenario}, where navigation and map generation in an unknown hostage environment by lizard-like robots enable the security forces to take timely measures stealthily.  Though navigation in robotics is a heavily explored area, navigation in an unknown environment remains relatively less explored and calls for a different approach than the classical algorithms. To summarize, the main contributions of this work are:

\begin{figure}[t]
    \centering
    \includegraphics[width=0.66\textwidth]{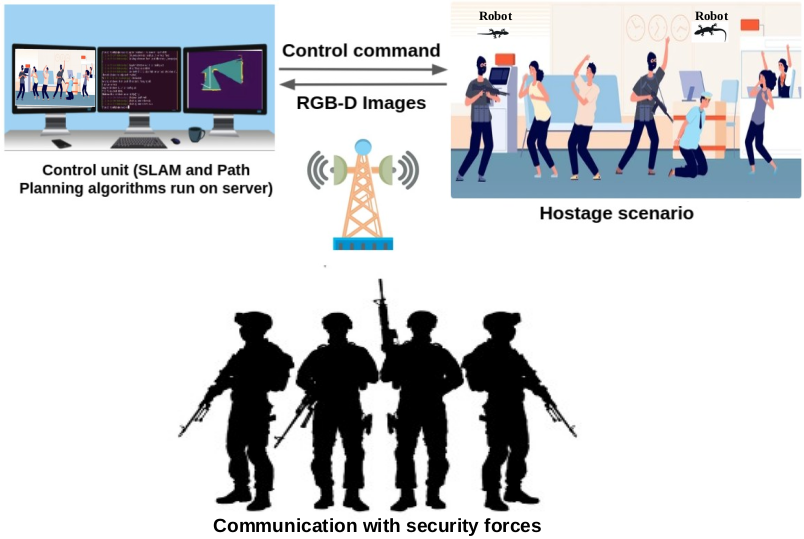}
    \caption{Stealth surveillance system for monitoring hostage scenario}
    \label{fig:ch6-scenario}
\end{figure}

\begin{itemize}
    \item We design an integrated robotic software framework for autonomous navigation of \emph{lizard-inspired quadruped robot} in an unknown environment; 
    
    \item For navigation in unknown environments, the SLAM, motion-planning, and the control components of the software run in synergy to generate a safe path-planning and simultaneously drive the robot to the goal. The SLAM module is not part of this thesis, which has been moved to the preliminaries of this work.
    
    \item We show the reduction of the motion-planning problem as a constraint satisfaction problem and solve it by a state-of-the-art constraint solver and
    
    \item Finally, we develop a working prototype of the proposed framework in  CoppeliaSim with support via ROS Interface. We present experimental results on several environments, ranging from simple to complex, to demonstrate the efficacy of the framework.
\end{itemize}
 
\noindent The rest of the chapter is organized as follows. Section \ref{ch6-sec-relwork} gives an overview of related works on motion-planning problems. Section \ref{ch6-sec-prilims} provides preliminaries. In Section \ref{ch6framework}, we describe in detail our proposed robotic software framework. Section \ref{ch6result} presents experimental results in a number of simulation scenarios. Section \ref{ch6conclusion} concludes the work.
\subsection{Related Works}\label{ch6-sec-relwork}
Various approaches to address the motion-planning problem  exist in the literature, such as RRT (Rapidly-exploring Random Tree)~(\cite{Lavalle98rapidly-exploringrandom}), A$^*$~(\cite{4082128}), SAT and SMT-based path-planning~(\cite{6906597,6942758}). 
Most of the motion-planning algorithms assume that the environment is known and thus can not be used directly (\cite{DBLP:books/daglib/0016830}). In (\cite{Stentz93optimaland}), a graph-based approach based on D$^*$ algorithm is proposed for path-planning in unknown, partially known and changing environments. 
Although this model works well in a partially known environment, it is computationally expensive in an unknown environment. 
Current approaches on sampling-based planning methods like RRT$^*$ (\cite{doi:10.1177/0278364911406761}), and its many variants have had considerable success for kinematic motion-planning. However, the complexity of such methods heavily relies on robot dynamics, and the global computation over the entire state space for high-dimensional systems (like legged robots) in cluttered environments becomes slow. A similar approach, SweepingRRT (\cite{9172596}) use a global and a local plan for motion-planning when complete environment information is not available. Global plan is computed less frequently, observing the large obstacles which are a priori known, while the local plan is computed frequently as smaller obstacles are incrementally detected on the global path. 

\subsection{Preliminaries} \label{ch6-sec-prilims}

\subsubsection{SLAM Module}
The SLAM module consists of two sub-modules: (1) \emph{Localization} and (2) \emph{Mapping}.
Localization is responsible for estimating the robot's pose in a given environment, whereas mapping generates a 2D occupancy grid map of the environment.

\paragraph{\textbf{Localization}} 
Localizing a robot in an environment at a time-instance $t$ requires the following: Measurement data $z_t$ and Control data $u_t$. Measurement data $z_t$ provides information about the robot's pose $p_t$ at time $t$. Since we are only interested in 2D pose of the robot, we define $p_t$ as follows:
$p_t = [x(t), y(t), \theta(t)]$, where ($x(t)$, $y(t)$) and $\theta(t)$ are the estimated position and orientation of the robot at time $t$. We use an RGB-D camera mounted on the robot to capture RGB image and depth information of the environment. This depth information is then used to extract point cloud for each pixel in the image. Once a point cloud of an image is extracted as its 3D representation, we use Iterative Closest Point (ICP) algorithm (\cite{segal2009generalized}) to compute $z_t$ by transforming the current point cloud so that it will be aligned with the previous point cloud. Control data $u_t$, on the other hand, provides information about how much the robot moves in $x$ and $y$ directions and the robot's turning angle around its joint. 
In practice, sensors and actuators in a robot yield uncertainties due to the presence of noise in $z_t$ and $u_t$. For example, depth information from an RGB-D camera may have noise due to improper reflection of the depth signal from objects, incorrect image taken when the camera is unstable, etc. Similarly, actions taken by the robot may have noise due to slippage, mechanical errors, etc., which yield uncertainties in control data. To overcome this, we use a sample-based probabilistic approach to estimate the robot's pose and to generate the map of the environment. In particular, we use Particle Filter algorithm (\cite{Thrun02d}) which takes control data $u_t$, measurement data $z_t$, and a set $S_{t-1}$=\{$p_{t-1}^{[1]}, p_{t-1}^{[2]}, \dots, p_{t-1}^{[K]}$\} of $K$ pose samples at time $t-1$ as input and returns robot's pose $p_t$ at time $t$ as output.  

\paragraph{\textbf{Mapping}}  
Since the robot is capable of walking only on the floor, we are interested in identifying the obstacles that touch the ground and have a height greater than the ground clearance from the robot's base link. This phase considers the point clouds that satisfy the aforesaid conditions, and the position of these point clouds in discretized 2D space is calculated using the robot's position. The map is generated in the form of 2D occupancy grid by marking each cell with one of the values from \{-1, 0, 1\}, where -1, 0, and 1 denote the cell as \emph{unexplored}, \emph{free}, or \emph{occupied} respectively.
Since measurement data may have noise, the grid cells that should be marked with 0 can be erroneously marked as 1 or vice versa. To overcome this, we use the notion of occupancy probability of grid cells (\cite{thrun2002probabilistic}), which defines the probability of a cell being occupied. Initially, the occupancy probability of each cell is set to 0.5, indicating all of them as unexplored. During the robot's navigation, the robot observes the nearby cells repeatedly over a period of time, and accordingly, the occupancy probabilities of the cells either increase or decrease. When the probability of a grid cell increases and reaches the threshold $P_{occupied}$, the corresponding cell value is set to 1. Similarly, when the probability decreases and reaches the threshold $P_{free}$, the corresponding cell-value is set to 0. Values for the cells having occupancy probability between  $P_{occupied}$ and $P_{free}$ are set to -1. This way, eventually, an estimate of the environmental map closer to the actual map is achieved by the robot.
\begin{algorithm}[htbp]
  \scriptsize
  \caption{\textsf{Mapping}}
  \LinesNumbered
  \SetAlgoLined
  \KwIn{Robot pose $(p_t)$ and depth map  $(d_t)$ at time $t$, probabilistic partial map \textit{PM}$_{t-1}$  at $t - 1$ , camera\_intrinsic\_parameters.\\}
  \KwOut{Probabilistic partial map \textit{PM}$_{t}$, partial global map $M_{t}$, snapshot map $L$. \\}
  
  \Begin{
   
    Initialize, $L_i$ = $-1$, $\forall L_i\in L$ ; $g_i$ = $-1$, $\forall g_i\in M_t$; \textit{PM}$_{t}$ = \textit{PM}$_{t-1}$ ;\\
    Initialize \texttt{obstacle\_set} = \texttt{obstacle\_grids} = \texttt{free\_grids} = $\emptyset$; \\
    \textit{PC} = compute\_point\_cloud($d_t$, camera\_intrinsic\_parameters);\\
    \textit{PC}$_{transformed}$ = transform\_point\_cloud(\textit{PC},~$p_t$);\\
   
    \For {\textit{pc}$_i$ $\in$  \textit{PC}$_{transformed}$}  { 
       \If {z-axis of $pc_i > $ ground\_clearance of robot base}{
           insert $pc_i$ in \texttt{obstacle\_set};
        }
    }
  
    \texttt{obstacle\_grids} = grid\_location(\texttt{obstacle\_set}); \\
    
    \texttt{free\_grids} = find\_free\_grids(\texttt{obstacle\_grids} , $p_t$ ); \\
    
    \For{all cell $m_i\in$ \textnormal{\texttt{obstacle\_grids}} or \textnormal{\texttt{ free\_grids}}} {
       $l_{0}$ = $\log (probability(m_i = 1)/probability(m_i = 0))$
       
       \textit{PM}$_{t, i}$ = \textit{PM}$_{t-1, i}$ + inverse\_sensor\_model($m_i, p_t, d_t$) - $l_{0}$; \\
      }
    \For{all cell \textit{pm}$_{i}$ $\in$ \textit{PM}$_{t}$, $g_i \in M_t$} {
    
        \If {\textit{pm}$_{i}$ $\geq$  $P_{occupied}$ }{ 
           $g_i = 1 $ \tcc*[r]{obstacle} 
        }
        
        \ElseIf {$pm_i$ $\leq$ $P_{free}$}{ 
           $g_i = 0 $ \tcc*[r]{free space}
        }
    }
    
    \For{all cell $L_i \in L$ , $g_i \in M_t$} { 
        \If{$L_i \in$ \textnormal{\texttt{ free\_grids}} or \textnormal{\texttt{obstacle\_grids}}}{ 
        $L_i = g_i$;}
        }
    return \textit{PM}$_{t}$, $M_t$ , $L$;
 }
\end{algorithm}
\noindent The overall mapping algorithm is depicted in Algorithm 1. The algorithm takes robot pose $p_t$, depth map $d_t$ at time $t$ and the probabilistic partial map \textit{PM}$_{t-1}$ at time $t-1$ as inputs, and it generates probabilistic partial map \textit{PM}$_{t}$, partial map $M_t$ at time $t$ and snapshot map $L$. The algorithm begins with computing the probabilistic partial global map \textit{PM}$_{t}$ in steps 1-18, and then it generates $M_t$ and $L$ in steps 19-30. Steps 2-5 deal with initialization, where each cell in $L$ and $M_t$ is set to -1 indicating unexplored, \textit{PM}$_{t}$ is set to \textit{PM}$_{t-1}$  and the variables \texttt{obstacle\_set}, \texttt{obstacle\_grids} and \texttt{free\_grids} for storing valid obstacles, obstacle grid locations and free grid locations respectively are set to $\emptyset$. Step 6 generates a point cloud in the robot’s frame of reference, which is then transformed to the initial frame of reference in step 7. The point clouds, which are just above the ground clearance of the robot pose, are considered as the obstacles in steps 8-12. Once obstacle grid locations and free grid locations are determined in steps 13-14, their occupancy probabilities are computed using \emph{inverse\_sensor\_model} (\cite{thrun2002probabilistic}) and the log odd ratio of prior occupancy in steps 15-18. Steps 19-30 construct partial global map $M_t$ and snapshot map $L$ depending upon whether the occupancy probabilities in the corresponding cells in \textit{PM}$_{t}$ and $L$ meet the threshold $P_{occupied}$ and $P_{free}$.
\subsection{Robotic Navigation Framework}\label{ch6framework}

\begin{figure}[b]
  \centering
  \scalebox{0.8}{\includegraphics[width=0.40\textwidth]{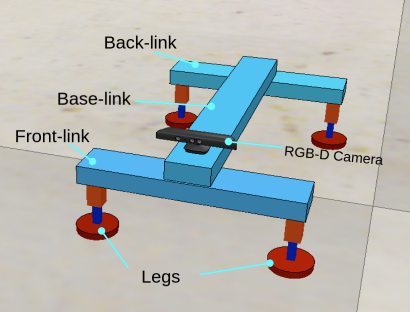}}
  \caption{CoppeliaSim model of the quadruped robot}
  \label{fig:lizard}
\end{figure}

Figure \ref{fig:lizard} depicts the design of the Lizard-inspired quadruped robot. It has a \textsf{front-link} and a \textsf{back-link} attached to a \textsf{base-link}. The \textsf{front-link} and \textsf{back-link} can rotate upto $45\degree$ w.r.t \textsf{base-link}. Two legs that can be lifted or grounded are attached to \textsf{front-link} and \textsf{back-link} each. The length, width, and height of the robot are $15cm$ by $10cm$ by $5cm$, respectively. A \textsf{RGB-D camera} is also mounted on the robot.

The overall architecture of the proposed framework is shown in Figure \ref{fig:controller}, which comprises two essential modules: (1) SLAM and (2) Planning and Control.
The SLAM  module is responsible for mapping the unknown environment and detecting the robot's position using the odometry data and depth information from the RGB-D camera mounted on the robot. However, the SLAM module is not part of this thesis. A description of it is given in section~\ref{ch6-sec-prilims}. In this section, we describe the Planning and Control module in detail:

\begin{figure}[t]
        \centering
        \scalebox{1.0}{\includegraphics[width=0.75\textwidth]{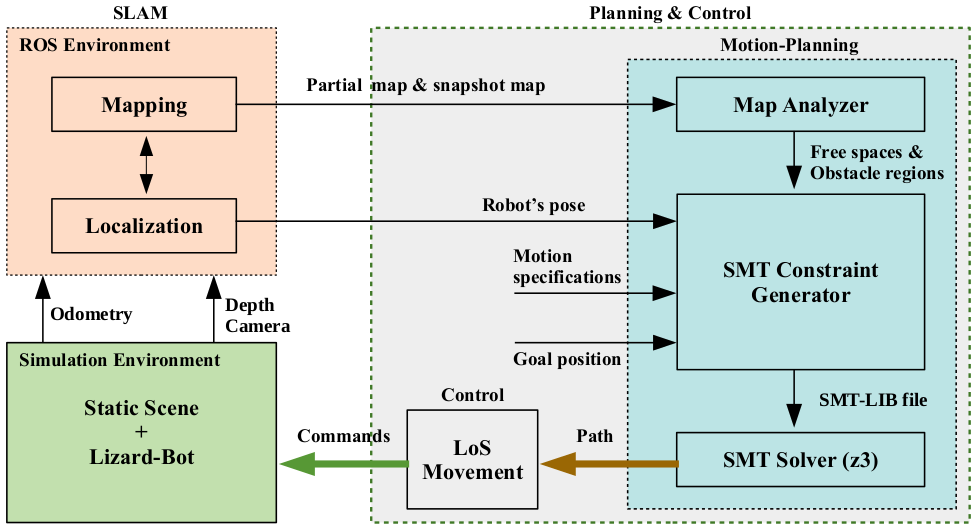}}
        \caption{The robotic software architecture for SLAM, path-planning, and control.}
        \label{fig:controller}
\end{figure}

\subsubsection{Planning and Control Module}
Planning and control module has two sub-modules, namely \emph{motion-planning} and \emph{control}. We now discuss each in detail:

\paragraph{Motion-planning:}
A motion-plan for a robot is a finite sequence of \emph{way-points} in the $2$D environment that, when traced, leads the robot to the assigned destination. We say that a motion-plan is \emph{safe} when the path that it gives (the path formed by joining the consecutive way-points through straight lines) is obstacle-free. The motion-planning problem can thus be defined as the problem of finding a finite sequence of way-points such that it is safe. In the following, we briefly illustrate its reduction to a satisfiability problem of a first-order logic formula. We reduce the motion-planning problem as a \textbf{constraint-satisfaction-problem} by encoding the initial, goal positions, the movement of the robot, and the obstacles in the map as first-order-logic formulae in the \textit{theory of quantifier-free nonlinear real arithmetic} (\cite{6942758}). The $t^{th}$ way-point in a motion-plan is represented by a pair of real variables $x_t$ and $y_t$. The number of way-points in a motion plan is upper bounded by a constant $M$, i.e., $0\leq t\leq M$. Our planner is restricted to generating piece-wise-linear (PWL) paths. The number of line-segments in the path and hence the precision can be tuned with the value of $M$.

\paragraph{Initial and Goal State:} The initial and the goal positions of the robot are tuples $\langle x_{init}, y_{init}\rangle$ and $\langle x_g, y_g\rangle$ respectively which are represented by the following constraints: 

\begin{align*} 
\begin{split}
& Init: (x_0=x_{init})\land(y_0=y_{init}) \\
& Goal: \big(\bigvee_{1\leq t\leq M} (x_t=x_g)\land(y_t=y_g)\big) \bigwedge \\
& \big( \bigwedge_{1\leq t < M} (x_t=x_g) \land (y_t=y_g) \implies (x_{t+1} = x_g) \land (y_{t+1} = y_g) \big).
\end{split}
\end{align*}
It encodes that the first way-point must be the initial position, and at least one of the way-points is the goal position. The second clause encodes that the robot remains in the goal position after reaching there.

\paragraph{Obstacles:} Each obstacle \emph{Obs} represents a rectangular region in the map bounded by the four corner points, where $(x_{tl},y_{tl})$, $(x_{tr},y_{tr})$, $(x_{bl},y_{bl})$ and $(x_{br},y_{br})$ denote the top-left, top-right, bottom-left and bottom-right corner points, respectively. We further inflated each obstacle region by $r$ grid units on each side, where $r>$ radius of the circumscribed circle for our robot. The $i^{th}$ obstacle $obs_i$ is defined as follows:
\begin{align*} 
\begin{split}
& obs_ix_{tl}=obs_ix_{tl}-r,\ \land \ obs_iy_{tl}=obs_iy_{tl}+r\ \land\ obs_ix_{tr}=obs_ix_{tr}+r\ \land\\
& obs_iy_{tr}=obs_iy_{tr}+r\ \land\ obs_ix_{bl}=obs_ix_{bl}-r\ \land \ obs_iy_{bl}=obs_iy_{bl}-r\ \land \\
& obs_ix_{br}=obs_ix_{br}+r\ \land \ obs_iy_{br}=obs_iy_{br}-r.
\end{split}
\end{align*}

\paragraph{Obstacle Free Path:} The constraints given by \emph{Obs\_freepath} ensures an obstacle free path given by the way-points. If $(x_t, y_t)$ and $(x_{t+1}, y_{t+1})$ are any two consecutive way-points in the projected path, it must be ensured that the line joining them does not pass through any obstacle region. Let $(obs_jx_{tl},obs_jy_{tl})$, $(obs_jx_{tr},obs_jy_{tr})$, $(obs_jx_{bl},obs_jy_{bl})$ and $(obs_jx_{br},obs_jy_{br})$ are the corner points of the $j^{th}$ obstacle respectively. For every rectangular obstacle, say $obs_j$, the idea is to search for a separating line $a_j.x + b_j.y + c_j = 0$ such that any pair of way-points $(x_t,y_t)$ and $(x_{t+1},y_{t+1})$ lie on one side of the line whereas the four corner points of $obs_j$ are on the other side of the line. This ensures that the line joining the way-points does not pass through the obstacle. 
\begin{align*} 
\begin{split}
& Obs\_freepath: \bigwedge_{1\leq t<M} \bigg [  \bigwedge_{0\leq j<N} \bigg [ \bigg (  \\ 
& (a_{tj}x_{t-1} + b_{tj}y_{t-1} + c_{tj} <0) \land (a_{tj}x_t + b_{tj}y_t + c_{tj} <0) \land \\
& (a_{tj}obs_jx_{bl} + b_{tj}obs_jy_{bl} + c_{tj} > 0) \land (a_{tj}obs_jx_{br} + b_{tj}obs_jy_{br} + c_{tj} > 0) \land  \\
& (a_{tj}obs_jx_{tl} + b_{tj}obs_jy_{tl} + c_{tj} > 0) \land (a_{tj}obs_jx_{tr} + b_{tj}obs_jy_{tr} + c_{tj} > 0) \bigg )  \\
& \bigvee \bigg ( (a_{tj}x_{t-1} + b_{tj}y_{t-1} + c_{tj} >0) \land (a_{tj}x_t + b_{tj}y_t + c_{tj} >0) \land  \\
& (a_{tj}obs_jx_{bl} + b_{tj}obs_jy_{bl} + c_{tj} < 0) \land (a_{tj}obs_jx_{br} + b_{tj}obs_jy_{br} + c_{tj} < 0)\land  \\
& (a_{tj}obs_jx_{tl} + b_{tj}obs_jy_{tl} + c_{tj} < 0) \land (a_{tj}obs_jx_{tr} + b_{tj}obs_jy_{tr} + c_{tj} < 0) \bigg )\bigg ] \bigg ] 
\end{split}
\end{align*}
where $t \in \{1,\ldots, M-1\}, j \in \{0,\ldots, N-1\}$,\ $a_{tj},b_{tj},c_{tj} \in \mathbb{R}$ and $N$ denotes the number of obstacle regions. Additionally, the robot's movement between two consecutive way-points is bounded by $v$ units in both $X$ and $Y$ axes, which is encoded as
\begin{align*} 
\begin{split}
 Mov:\bigwedge_{0\leq t<M} \bigg [[abs(x_{t+1} - x_t) < v] \land [abs(y_{t+1}- y_t) < v] \bigg ] \\
\end{split}
\end{align*}
where $v \in \mathbb Z^{+}$ is less than the minimum of height and width of the environment. This constraint ensures that consecutive way-points are within a reasonable distance.

The \emph{Map-Analyzer} generates a bounding-box approximation of the obstacle regions and returns a list of obstacles from the partial map, assuming the unknown regions as free spaces. This allows for an easy representation of obstacles as constraints, in the \emph{SMT-Constraint-Generator} sub-module. An SMT-LIB file is generated from \emph{SMT-Constraint-Generator} sub-module representing the free-spaces, obstacle regions in the partial map, the robot's initial position, goal position, and safe-movement as constraints. Then, the constraints in the SMT-LIB file are solved for satisfiability by the Z3~(\cite{10.1007/978-3-540-78800-3_24}) SMT-Solver. When satisfiable, a motion plan is extracted from the satisfying assignments as a sequence of way-points.

\paragraph{Control:}
This sub-module is responsible for moving the \emph{lizard-inspired quadruped robot} along the computed way-points using a \emph{LoS algorithm} (see Algo.~\ref{algo:los}) and a hybrid controller module. However, before discussing the algorithm and the controller module, we first give a brief description of the motion-primitives of the robot.

\paragraph{Motion-primitives:} Motion primitives are a set of control actions that can be provided as commands to our robot for execution. Our robot has four basic primitives: \emph{Move-forward} and \emph{Move-backward} for linear movement and \emph{Rotate-clockwise (RotClk)}, \emph{Rotate-anti-clockwise (RotAclk)} for rotational movements. Rotational movement has four discretization angles- $6\degree$, $4.5\degree$, $3\degree$, and $1.5\degree$.
Each motion-primitive has two phases, and when applied on the \emph{lizard-inspired quadruped robot} will perform one linear or rotational movement. The details of the phases of motion-primitives \emph{Move-forward} and for one discretization angle of \emph{RotClk} (see in Figure \ref{ch6fig:robot_movement}) is given below:

\noindent 1) {Move-forward:} The motion primitive \emph{Move-forward} is used to perform a forward direction movement for the robot. In the first phase, the robot lifts its front-right and back-left legs together while front-joint and back-joint are rotated by $30\degree$ and $-30\degree$ angles, respectively. As front-left and back-right legs are grounded at that time, the rotations of front-joint and back-joint make the front-right and back-left legs move forward (see Figure \ref{fig:fwd_movement} first phase, where $\theta_f = -\theta_b = 30\degree$). In the second phase, the robot lifts its front-left and back-right legs together while front-joint and back-joint are rotated with $-30\degree$ and $30\degree$ angles, respectively (see Figure \ref{fig:fwd_movement} second phase, where $-\theta_f = \theta_b = 30\degree$). Now, these two legs move forward and thus complete one forward movement. As a result of one such forward movement, the effective displacement of the robot is $6.46$ cm (calculated via simulation).

\begin{figure}[t]
    \centering
      \begin{subfigure}[b]{0.45\textwidth}
         \centering
         \includegraphics[height=4cm, width=\textwidth]{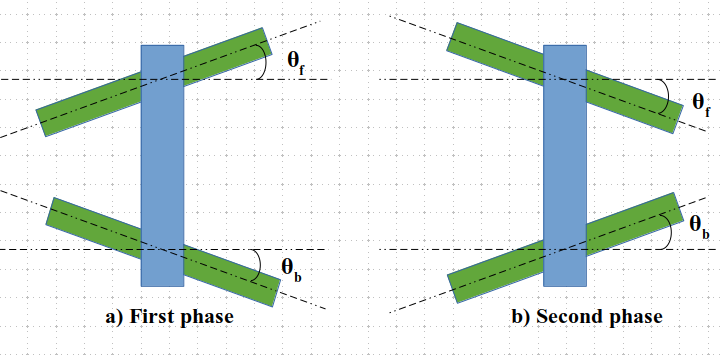}
         \caption{Move-forward}
         \label{fig:fwd_movement}
      \end{subfigure}
      \hfill
      \begin{subfigure}[b]{0.45\textwidth}
         \centering
         \includegraphics[height=4cm, width=\textwidth]{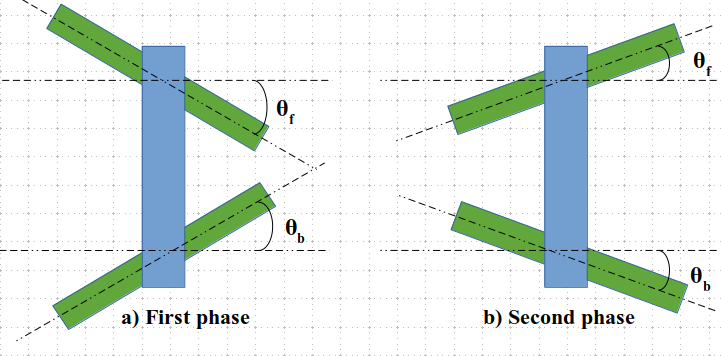}
         \caption{\emph{RotClk}}
         \label{fig:right_turn}
      \end{subfigure}
    \caption{Phases of \emph{Move-forward} and \emph{Rotate-Clockwise} maneuvering of the robot}
    \label{ch6fig:robot_movement}
\end{figure}

\noindent 2) {Rotate-clockwise (RotClk):} The motion primitive \emph{RotClk} will cause the robot to make a clockwise angular rotation. We now present the mechanism for $6\degree$ clockwise rotation (\emph{RotClk-6}): In the first phase, the robot lifts its front-left and back-right legs together while front-joint and back-joint are rotated with $\theta_f$ and $\theta_b$, respectively (see Figure \ref{fig:right_turn} first phase, where $-\theta_f = \theta_b = 45\degree$). In the second phase, the robot lifts its front-right and back-left legs together while front-joint and back-joint are rotated with $20\degree$ and $-20\degree$ angles, respectively (see Figure \ref{fig:right_turn} second phase, where $\theta_f = -\theta_b = 20\degree$). \emph{RotClk} movement will rotate the robot clockwise $\alpha$ degrees which is the effective angular rotation. This $\alpha$ depends on $\gamma$, given as: $\gamma = K * \alpha$, where $\gamma = \|\theta_1\|-\|\theta_2\|$, $\theta_1$ and $\theta_2$ are angle of rotation of front-joint and back-joint in the first and second phase respectively while $K$ is the proportional gain and obtained by simulations. In this case, $\|\alpha\|=6\degree$ as $\gamma=25\degree$.

\begin{table}[htbp]
    \centering 
    \begin{tabular}{|c|c|c|c|c|}
    \hline

         Rot ($\|\theta_1\|$, $\|\theta_2\|$) & Rot-diff ($\gamma$) & Eff-rot ($\alpha$) & Ratio ($K$) \\
    \hline
         $45\degree$ , $20\degree$ & $25\degree$ & $5.947\degree\approx 6\degree$ & $4.204$\\\cline{1-4}
         $40\degree$ , $20\degree$ & $20\degree$ & $4.535\degree\approx 4.5\degree$ & $4.414$\\\cline{1-4}
         $35\degree$ , $20\degree$ & $15\degree$ & $3.072\degree\approx 3\degree$ & $4.883$\\\cline{1-4}
         $30\degree$ , $20\degree$ & $10\degree$ & $1.485\degree\approx 1.5\degree$ & $6.734$\\\cline{1-4}
         $30\degree$ , $30\degree$ & $0\degree$ & $0\degree$ & $-$\\
    \hline 
    \end{tabular}
    \caption{Clockwise and anti-clockwise rotations. $\theta_1$ and $\theta_2$ are angles of rotation of the joints in the first and second phases, respectively.}
    \label{ch6tab:table1}
\end{table}
In Table \ref{ch6tab:table1}, we have shown different discretization of clockwise and anti-clockwise rotational movements of our robot, which are achieved by applying different $\theta_1$ and $\theta_2$ rotations to the joints in both phases of the movements. \emph{Rot-diff} denotes rotational difference in both phases, \emph{Eff-rot} denotes effective rotation of the robot, and $K$ is the ratio of $\gamma/\alpha$.

\paragraph{LoS algorithm:} The \emph{LoS algorithm} is responsible for the line-of-sight navigation of the robot between the consecutive way-points on a projected path. The algorithm receives the robot's positional information ($x_t$, $y_t$) from the current pose and fetches the next way-point $w_p$ ($w_x$, $w_y$) from the sequence of way-points $wps$ produced by the \emph{motion-planning} sub-module. The algorithm begins by calculating the angle $\phi$ between the robot's current position and the waypoint with respect to the positive \emph{X-axis}. It then calculates the effective angle $\theta$, which is the angle the robot needs to rotate considering its current orientation $\theta_t$.
Then, repeatedly apply rotational motion-primitives to rotate the robot towards the waypoint accordingly. Each rotational movement is followed by a \emph{Move-backward} movement to minimize positional changes during rotation due to the unavailability of in-place rotation for our robot.
The algorithm then calculates the Euclidean distance $dist$ between ($x_t$, $y_t$) and ($w_x$, $w_y$) and the number of \emph{Move-forward} motion primitives it needs to apply to reach the waypoint $w_p$ by dividing $dist$ with a constant value of $6.46cm$ which is the robot's effective displacement for one application of \emph{Move-forward} movement.
It calls the \emph{hybrid controller}, which we discuss in the next section, in synergy with the movement to ensure the robot does not deviate from its path.

\begin{algorithm}[htbp]
  \caption{LoS algorithm}
  \label{algo:los}
  \LinesNumbered
  \SetAlgoLined
  \KwIn{Robot's pose $p_t$ and way-point $w_p$. \\}
  \Begin{
     Calculate the angle $\phi$;\\
     Calculate the effective angle, $\theta \gets \phi-\theta_t$; \\
     \If{$\theta > 180$}{$\theta \gets \theta - 360$;}
     Apply \emph{rotational motion-primitives} to cover $\theta$ \tcc*{\scriptsize Each rotational movement is followed by a move backward movement}
     Calculate the distance $dist \gets \sqrt{(w_x-x_t)^2 + (w_y-y_t)^2}$;\\
     Calculate the $n\_steps \gets dist/6.46$ \tcc*{\scriptsize The number of linear movements needed to cover the distance}
     \For {$i=0$; $i$ $<$ $n\_steps$; $i$++}{
        Apply the \emph{move-forward motion-primitive};\\
        Calls \emph{hybrid controller} \tcc*{\scriptsize Calculates the deviation from the path, and moves accordingly.}
     }
  }
\end{algorithm}

\begin{figure}[htb]
    \centering
    \scalebox{0.66}{\includegraphics[width=0.5\textwidth]{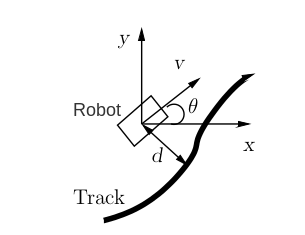}}
    \caption{Robot movement}
    \label{ch6fig:robot_movement1}
\end{figure}

\paragraph{Designing a hybrid controller for the Lizard-inspired quadruped robot:} We propose designing a hybrid controller that controls the navigation of the robot along a projected path. Let ($x,y$) and $\theta$ define the robot's current position and orientation in the environment, respectively, and the robot has a fixed speed $v$ as shown in Figure \ref{ch6fig:robot_movement1}. Thus, the velocity of the robot in the X and Y axes direction can be derived as $v*\cos{\theta}$ and $v*\sin{\theta}$, respectively. The robot can rotate with an angular speed $w$, for simplicity, which can be either of three discrete values $-6$, $0$, and $6$ degrees/second. When $w=0$, the direction $\theta$ remains unchanged, and the robot is going straight. When $w=-6$, the direction $\theta$ is decreasing, and thus the robot is attempting to turn right. Conversely, when $w=6$, the direction $\theta$ is increasing, and thus the robot is attempting to turn left.

\begin{figure}[htb]
    \centering
    \scalebox{0.5}{\includegraphics[width=\textwidth]{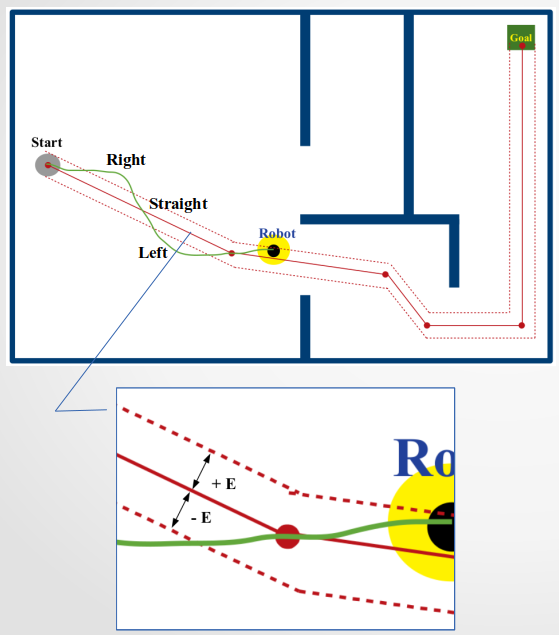}}
    \caption{Robot's trajectory}
    \label{fig:trajectory}
\end{figure}

The movement of the robot following a projected path is shown in Figure \ref{fig:trajectory}, which is made for illustration and may not be an actual path in the environment. The green trace represents the robot's trajectory. The red points are the waypoints, and the lines joining them represent the projected path from start to goal location. 

\begin{figure}[htbp]
    \centering
    \scalebox{0.5}{\includegraphics[width=\textwidth]{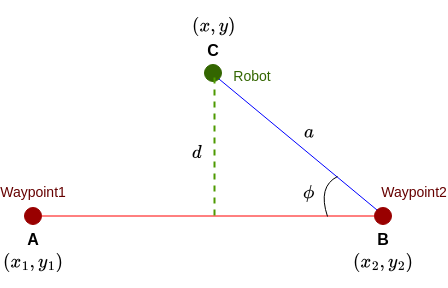}}
    \caption{Robot's deviation from the path}
    \label{fig:d_cal}
\end{figure}

\noindent We assume a buffer zone of distance $\pm E$ around the projected path as shown in Figure \ref{fig:trajectory}. We continuously monitor the robot's deviation $d$ from the projected path as shown in Figure \ref{fig:d_cal}, where the line \textit{AB} joins the current waypoint and the next waypoint on the projected path and the line \textit{CB} joins the robot's current position and the next waypoint. We calculate the slopes $m_1$ and $m_2$ of the lines \textit{AB} and \textit{CB} with X-axis as below:
\begin{equation*}
\begin{split}
 & m_1 = \frac{y_2 - y_1}{x_2 - x_1} \\
 & m_2 = \frac{y_2 - y}{x_2 - x}
\end{split}
\end{equation*}

\noindent The angle $\phi$ between the line \textit{AB} and \textit{CB} can be derived as:
\begin{equation*}
\begin{split}
 & \phi = \arctan {\|\frac{m_2 - m_1}{1 + {m_2}*{m_1}}\|} \\
\end{split}
\end{equation*}
The length $a$ of \textit{CB} is calculated as:
\begin{equation*}
\begin{split}
 & a = \sqrt{(x_2 - x)^2 + (y_2 - y)^2} \\
\end{split}
\end{equation*}
The robot's deviation from the projected path is calculated as:
\begin{equation*}
\begin{split}
 & d = a*\sin{\phi} \\
\end{split}
\end{equation*}

\begin{figure}[htb]
    \centering
    \scalebox{0.8}{\includegraphics[width=\textwidth]{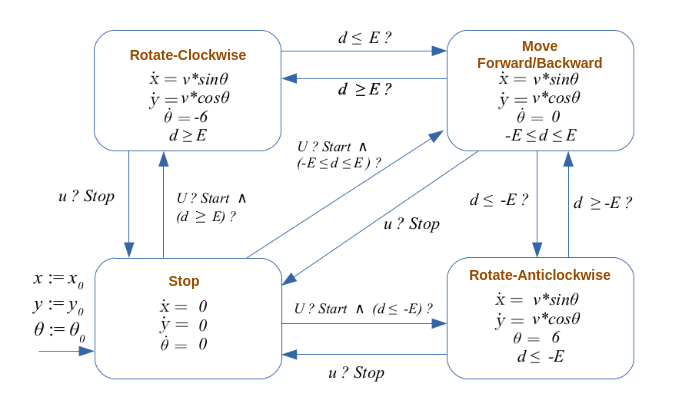}}
    \caption{Hybrid controller}
    \label{fig:hybrid_controller}
\end{figure}

\paragraph{Hybrid controller:} The hybrid controller for our robot is shown in Figure \ref{fig:hybrid_controller}. It has four modes- \textit{Move Forward/Backward}, \textit{Rotate-Clockwise}, \textit{Rotate-Anticlockwise}, and \textit{Stop}. The \emph{LoS algorithm} calls the hybrid controller to control the robot's motion dynamics to minimize deviation from the projected path. 

\subsubsection{Integrating SLAM, Planning and Control}
The Navigation algorithm controls the navigation of our \emph{lizard-inspired quadruped robot} in an unknown environment by integrating the \emph{SLAM}, \emph{Motion-planning}, and \emph{Control} modules. Through \emph{SLAM} module, it incrementally builds a partial global map of the observable terrain and localizes the robot in that space. It then uses the \emph{motion-planning} module to find a safe path-plan to a given goal position through iterative re-planning applied on the progressively gathered knowledge about the environment. Finally, the \emph{control} module drives the robot along the path to reach the goal. These modules work in synergy to generate a safe path-plan and simultaneously drive the robot to the goal. The details of this integration are shown in Algorithm \ref{algo:integration}.
The algorithm takes the goal position ($x_g$, $y_g$), a set of pose samples $S_{t-1}$, measurement data $z_t$, control data $u_t$, depth map $d_t$ and probabilistic partial map \textit{PM}$_{t-1}$ as inputs. The algorithm begins with initializing the time $t$, $S_{t-1}$, \textit{PM}$_{t-1}$, and the problem instance \textit{prob} in line 2.
In line 3, it receives the robot's current pose $p_t$, pose samples set $S_t$, partial global map $M_t$, snapshot map $L$, and probabilistic partial map \textit{PM}$_t$ from the SLAM module. Lines 4-15 perform motion-planning task which starts with \emph{Map-Analyzer} to build a list of obstacle regions $obs_{list}$ using the information in $M_t$ in line 4. Line 5 initializes two variables $max\_wps$ and $max\_dist$ that correspondingly set the limits for the maximum number of way-points in a plan and the maximum distance between the two consecutive way-points. In lines 6-10, the algorithm checks iteratively whether the problem instance is satisfiable with a maximum of $10$ way-points in a plan starting with an initial value of $2$. For each iteration in line 7, \emph{SMT-Constraint-Generator} formulates the constraint-satisfaction problem by encoding the constraints $obs_{list}$, robot's current pose $p_t$, goal position ($x_g$, $y_g$), $max\_wps$ and $max\_dist$ as a first-order-logic formulae $\phi$ such that the satisfiability of $\phi$ implies the existence of a safe motion plan. The formula $\phi$ is then solved by the \emph{SMT-Solver} \textsc{Z3} in line 8, which returns either of the two values \textit{'SAT'} or \textit{'UNSAT'} to indicate whether the problem instance is satisfiable or not in the current iteration.
When satisfiable, the motion plan is constructed from the satisfying assignment as a sequence of way-points $wps$ in line 12. Lines 16-24 are responsible for driving the robot along the path to reach the goal. In line 17, every way-point $w_p$ in $wps$ is processed sequentially and checked in line 19 against the snapshot map $L$, to verify whether $w_p \in$ \textit{free\_spaces} (\textit{free\_spaces} is the set of free spaces in $L$). If the $w_p$ is found to be in free space, the \emph{LoS algorithm} is called in line 20 to generate commands in the form of motion-primitives to move the robot towards the way-point and control its movement along a path by using the hybrid controller. Otherwise, the algorithm replans with the updated $M_t$, which is continuously being updated with the robot's movement in the environment. 

\begin{algorithm}[t]
  \caption{\textsf{Navigation}}
  \label{algo:integration}
  \LinesNumbered
  \SetAlgoLined
  \KwIn{Goal position ($x_g$,$y_g$), a set of pose samples $S_{t-1}$, control data $u_t$, measurement data $z_t$, depth map $d_t$ and probabilistic partial map \textit{PM}$_{t-1}$. \\}
  \Begin{
    $t = 1$; $S_0$ = $\emptyset$; \textit{PM}$_0$ = $\emptyset$; $prob$ = \textit{UNSAT};\\
    \{$p_t$, $S_t$\} = \textsf{Localization}($S_{t-1}$, $u_t$, $z_t$); \{$M_t$, $L$, \textit{PM}$_t$\} = \textsf{Mapping}($p_t$, $d_t$, \textit{PM}$_{t-1}$); \\
    $obs_{list}$ = \textsf{Map-Analyzer}($M_t$); \\
    $max\_wps$ = $2$; $max\_dist$ = $100$\tcc*{\scriptsize Max way-points in a plan and max distance between two consecutive way-points.}
    \While{($max\_wps \leq 10$) $\land$ (prob == UNSAT)}{
    $\phi$ = \textsf{SMT-Constraint-Generator}($obs_{list}$, $p_t$, ($x_g$, $y_g$), $max\_wps$, $max\_dist$);\\
    $prob$ = \textsf{SMT-Solver}($\phi$); \\
    $max\_wps$ = $max\_wps + 1$;
    }
    \eIf{prob==SAT}
        {Construct $wps$ from \textit{SAT} assignment of $\phi$;}
        {return no-plan;}
    \While{$wps \neq \emptyset$}{
        $w_p$ = $wps$.delete;\\
        $t$ = $t+1$;\\
        \eIf{$w_p \in free\_spaces$}
          {Call \textsf{LoS}($p_t$, $w_p$);}
          {Go to step 2;
          }
      }
  }
\end{algorithm}
\subsection{Simulation and Experimental Evaluation}\label{ch6result}

\begin{table*}[t]
\centering
\caption{Simulation Results}
\resizebox{\textwidth}{!}{
\begin{tabular}{|c|c|c|c|c|c|c|c|c|c|c|c|c|c|}
\hline
\multicolumn{4}{|c|}{Environment} & \multirow{2}{*}{\textit{obs}}  & \multirow{2}{*}{\textit{var}} & \multirow{2}{*}{\textit{cnstr}} & \multirow{2}{*}{\textit{turns}} & \multirow{2}{*}{\textit{replans}}  & $T_{plan}$ & $dist$ & $T_{traverse}$ & \multirow{2}{*}{\textit{cov}(\%)} & \multirow{2}{*}{Remarks} \\\cline{1-4}
Scene & \emph{Dim} & \emph{Ins} & Complexity & & & & & & (in Sec) & (in Meter) & (in Sec) & & \\

\hline
       \multirow{4}{*}{Scene-1} & \multirow{4}{*}{$10\times10m^2$} & $ins1$ & Complex & $225$  & $2779$ & $1807$ & $7$ & $3$ & $123.81$ & $13.29$ & $919$ & $36.59$ & Success\\\cline{3-14}
       & & $ins2$ & Complex & $117$ & $3056$ & $1661$ & $7$ & $6$ & \textsf{Time Out} & $2.61$ & $526$ & $21.16$ & Failure\\\cline{3-14}
       & & $ins3$ & Simple & $60$ & $4$ & $8$ & $1$ & $1$ & $0.12$ & $5.32$ & $211$ & $20.21$ & Success\\\cline{3-14}
       & & $ins4$ & Complex & $136$ & $2405$ & $1565$ & $5$ & $2$ & $61.98$ & $7.36$ & $539$ & $37.63$ & Success\\
\hline

       \multirow{3}{*}{Scene-2} & \multirow{3}{*}{$5\times5m^2$} & $ins1$ & Complex & $116$ & $2246$ & $1611$ & $5$ & $3$ & $17.1$ & $6.92$ & $510$ & $61$ & Success\\\cline{3-14}
       & & $ins2$ & Simple & $55$ & $454$ & $331$ & $2$ & $1$ & $1.27$ &$4.12$ & $290$& $26.12$ & Success \\\cline{3-14}
       & & $ins3$ & Moderate & $43$ & $594$ & $431$ & $4$ & $3$ & $10.1$ & $3.7$ & $6.1$ & $33.1$ & Success\\
\hline

       \multirow{4}{*}{Scene-3} & \multirow{4}{*}{$10\times10m^2$} & $ins1$ & Complex & $192$ & $2439$ & $1587$ & $4$ & $2$ & $4$ & $10.2$ & $432$ & $26$ & Success\\\cline{3-14}
       &  & $ins2$ & Complex & $201$ & $10730$ & $7671$ & $9$ & $5$ & $11.12$ & $9.4$ & $1200$ & $39.11$ & Success\\\cline{3-14}
       &  & $ins3$ & Simple & $101$ & $103$ & $89$ & $1$ & $1$ & $1.2$ & $2.6$ & $130$ & $20.63$ & Success\\\cline{3-14}
       &  & $ins4$ & Moderate & $125$ & $1518$ & $1091$ & $3$ & $4$ & $8.4$ & $4.27$ & $367$ & $23.2$ & Success\\
\hline

\end{tabular}
}
\label{ch6table:table2}
\end{table*}

\begin{figure}[tbh]
    \centering
      \begin{subfigure}[b]{0.3\textwidth}
        \centering
        \includegraphics[height=4.5cm, width=\textwidth]{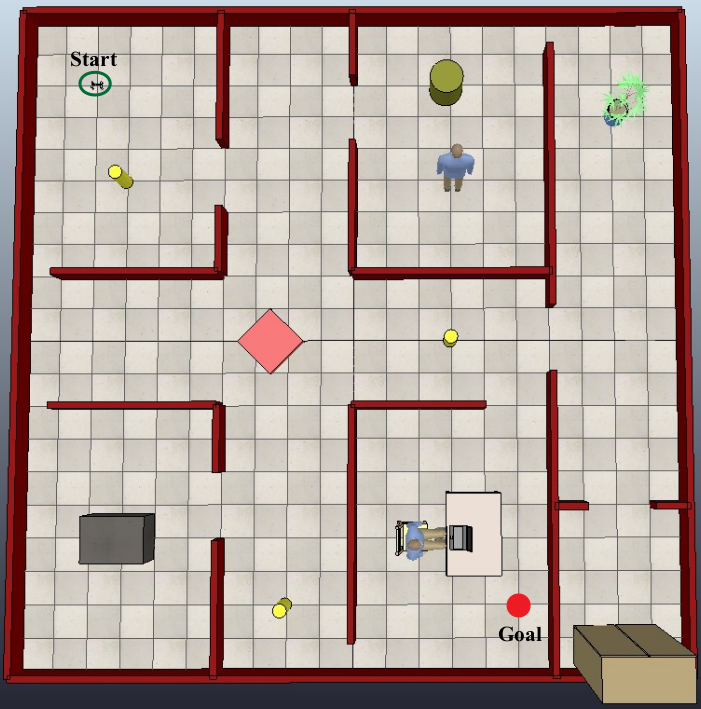}
        \caption{}
        \label{fig:scene3}
      \end{subfigure}
    \hfill
      \begin{subfigure}[b]{0.3\textwidth}
        \centering
        \includegraphics[height=4.5cm, width=\textwidth]{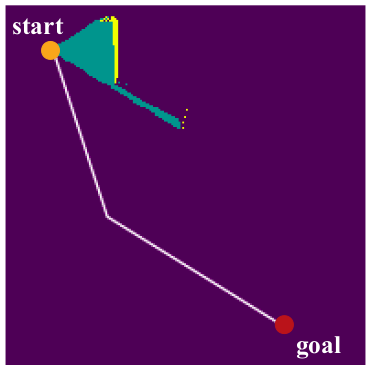}
        \caption{}
        \label{fig:plan}
      \end{subfigure}
    \hfill
      \begin{subfigure}[b]{0.3\textwidth}
        \centering
        \includegraphics[height=4.5cm, width=\textwidth]{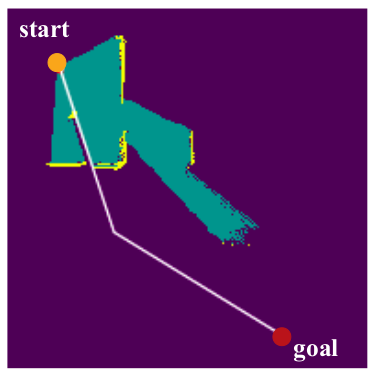}
        \caption{}
        \label{fig:obstacle}
      \end{subfigure}      
    \newline
      \begin{subfigure}[b]{0.3\textwidth}
        \centering
        \includegraphics[height=4.5cm, width=\textwidth]{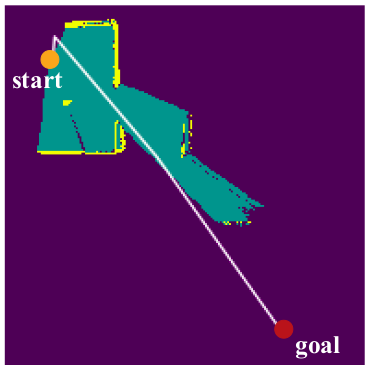}
        \caption{}
        \label{fig:plan2}
      \end{subfigure}
    \hfill
      \begin{subfigure}[b]{0.3\textwidth}
        \centering
        \includegraphics[height=4.5cm, width=\textwidth]{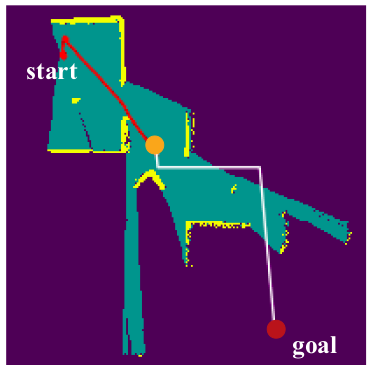}
        \caption{}
        \label{fig:plan3}
      \end{subfigure}
    \hfill
      \begin{subfigure}[b]{0.3\textwidth}
        \centering
        \includegraphics[height=4.5cm, width=\textwidth]{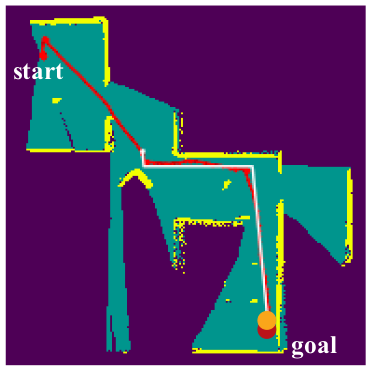}
        \caption{}
        \label{fig:traversed_path}
      \end{subfigure}
     \caption{Planning and Navigation: simulation results of plan generation and navigation in an office-like environment using CoppeliaSim and ROS: (a) is shown here. The initial plan is given in (b) when most of the environment was unknown, assuming the unknown as free spaces. (c) shows detection of obstacles on the path, (d) and (e) present two re-planning steps at different time points during navigation, and (f) depicts the robot's actual traversed path while following the generated plans.}
     \label{fig:navigation}
      \vspace{-10pt} 
\end{figure}

We now present our simulation results~\footnote{The source codes of our implementation are available at: \url{www.iitp.ac.in/~halder/IRIA2021/Codes.zip}} on various indoor environments. The simulations are performed on CoppeliaSim\footnote{https://www.coppeliarobotics.com/} and ROS\footnote{https://www.ros.org/} integrated platform, on a system with Intel® Core™ i5-8250U CPU @ 1.60GHz, 8 cores, 8GB RAM, Ubuntu 18.04. We use the standard SMT-LIB format to represent our constraints for motion-planning. These constraints are solved by using Z3 solver\footnote{https://github.com/Z3Prover/z3}. Figure \ref{fig:navigation} shows various stages of our navigation software at work in an indoor simulation environment, where plan generation and navigation of our robot are shown at different time points. The robot's starting position and goal position are marked by the green and red circles, respectively, in the environment (in Figure \ref{fig:scene3}). Figures \ref{fig:plan}-\ref{fig:traversed_path} show the robot's perception of the environment at different time-instances during planning and navigation steps. The orange dot indicates the robot's current position in the perceived environment. The purple and cyan areas are, respectively, the unknown and free spaces, whereas the yellow regions represent obstacle regions. The white trajectories are the projected path from the robot's current position to the goal after plan-generation, considering the partial information gathered till that point, and the red trajectories represent the traversed path by the robot along the projected path.   
%
In Table \ref{ch6table:table2}, we present the results of our simulation experiments on different scenarios of varying complexity. \emph{Dim} and \emph{Ins} represent the dimension and problem instances in the scene. \emph{obs} denotes the number of obstacle regions detected in the environment, whereas \emph{var} and \emph{cnstr} denote the numbers of SMT variables and constraints used, respectively, to solve the problem instances. \emph{turns} reports the number of turns taken by the robot during its traversal to the goal, \textit{re-plans} records the number of re-plans during this traversal, and $T_{plan}$ is the accumulated duration of time for the generation of these plans. \emph{dist} and $T_{traverse}$ are the distance and time to traverse by the robot from the initial to the goal position. The percentage of the total area covered by the robot during this navigation is presented as \emph{cov}. We consider the `Complexity' of an environment based on the dimensions of the environment, the number of obstacles detected, and the number of turns needed to reach the goal position.
We classify the problem instances as `Simple', `Moderate' and `Complex' by observing the no. of constraints and variables used in their first-order-logic formula encoding (\ref{ch6framework}(B)), given as: Simple: $(var < 1000) \land (cnstr < 1000)$, Moderate: $(1000 < var < 2000) \land (1000 < cnstr < 1500)$, Complex: $(var > 2000)$ $\land$ $(cnstr > 1500)$.
We report whether the robot successfully reaches the goal position within the plan-generation timeout of $60$ seconds under `Remarks'. Our navigation software successfully solves 10 out of the 11 navigation tasks.

\subsection{Conclusion}\label{ch6conclusion}
The work presents an integrated software framework for autonomous navigation of a \emph{lizard-inspired quadruped robot} in an unknown environment. The \emph{SLAM}, \emph{motion-planning}, and \emph{control} components of the framework run in synergy to generate a safe path plan and simultaneously drive the robot to the goal. The motion-planning problem is solved by reducing the constraint-satisfaction problem, which in turn is solved with a state-of-the-art constraint solver, Z3. The robotic software architecture has been tested with several planning problem instances in various indoor simulation scenarios, and the results met our objectives. 

\chapter{A Contrastive Plan Explanation Framework for Hybrid System Models} \label{ch2}
\textbf{\textsc{Chapter Abstract:}} \textit{In AI planning, having an explanation of a plan given by an AI planner is often desirable. The ability to explain various aspects of a synthesized plan to an end-user not only brings trust in the AI-based system but also reveals insights into the planning domain and the planning process. Contrastive explanation has been a popular approach in the literature (\cite{DBLP:journals/corr/abs-1709-10256,DBLP:conf/rweb/HoffmannM19}). It is not only an effective way for enhancing understanding but also simpler to generate than full causal analyses (\cite{DBLP:journals/ai/Miller19}). Additionally, their inherent design makes it easy to compare the original plan with the user's alternative. Contrastive questions such as “Why action A instead of action B?” can be answered with a contrastive explanation that compares the properties of the original plan containing A against the contrastive plan containing B.}

\textit{In this chapter, we explore a set of contrastive questions that a user of a planning tool may raise and propose a \emph{re-model and re-plan} framework to provide explanations to such questions. Earlier work has reported this framework on planning instances for discrete problem domains described in the Planning Domain Definition Language (PDDL) and its variants.
This work proposes an extension for planning instances described by PDDL+ for hybrid systems that portray a mix of discrete-continuous dynamics. Specifically, given a mixed discrete-continuous system model in PDDL+ and a plan describing the set of desirable actions on the same to achieve a destined goal, we present a framework that can integrate contrastive questions in PDDL+ and synthesize alternate plans. We present a detailed case study on our approach and propose a comparison metric to compare the original plan with the alternate ones.}


\section*{}
\lettrine[lraise=0.3, nindent=0em, slope=-.5em]Artificial Intelligence (AI) planning has been an active area of research for several decades, and is a core component of all autonomous systems today. Classical AI planning concerns with finding a sequence of feasible actions from an initial configuration of a system to a desired goal. Planning has been an active area of research for AI practitioners, leading to the development of planners for diverse planning domains, goal descriptions, and varied optimization objectives. Techniques ranging from graph traversals to recent developments around constraint solvers for efficient and scalable planning have been explored widely. Planning with hybrid domains modelled in PDDL+ (\cite{PDDL+}) has been gaining research interest in the automated planning community in recent years. Hybrid domains capture a more accurate representation of real-world problems that involve an interplay of continuous and discrete processes. However, solving problems represented as PDDL+ domains is very challenging due to the complex system dynamics, including non-linear processes and events. In planning literature over the past years, a number of languages for expressing planning problem specifications have evolved (e.g. STRIPS \cite{FIKES1971189}, ADL \cite{10.5555/112922.112954}, PDDL \cite{aeronautiques1998pddl} and PDDL+) and sophisticated planners (e.g. \textsc{GraphPlan} \cite{10.5555/295240.295918}, \textsc{SatPlan} \cite{kautz2006satplan}, \textsc{FF} \cite{DBLP:journals/aim/Hoffmann01, DBLP:journals/jair/HoffmannN01}, \textsc{FastDownward} \cite{DBLP:journals/jair/Helmert06}, \textsc{HSP} \cite{DBLP:conf/ecp/BonetG99}, \textsc{ENHSP} \cite{DBLP:conf/ecai/ScalaHTR16}, \textsc{UPMurphi} \cite{DBLP:journals/apin/PennaMM12}, \textsc{DiNo} \cite{DBLP:conf/aaai/PiotrowskiFLMM16}, \textsc{SMTPlan+} \cite{DBLP:journals/jair/CashmoreMZ20} and \textsc{LPG} \cite{DBLP:conf/aips/GereviniS02}) have advanced the field of planning. 

Explainable Planning (XP), also termed as Explainable AI Planning (XAIP)~(\cite{DBLP:conf/rweb/HoffmannM19}) has picked up as a problem of immense recent importance, given the multitude of application domains where autonomous plans are being envisioned to automate and replace manually crafted ones. Indeed, questions like safety, robustness, and trustworthiness of automatically learned and synthesized plans are being analyzed before deployment. More importantly, explainable planning has been an interesting area of research in recent times, given the increasing number of application areas in which humans and autonomous agents collaborate with mutual trust and where mutual understanding and cooperative plans are important for achieving a desired objective. With the advent of automated planning being applied in safety-critical systems, the need for plan explanation to a human expert responsible for implementing the plan has become ever more important. Before accepting and executing the plan, the human expert ought to be convinced about the safety and rationality of the plan. In the first place, there may be a mismatch between the knowledge of the human agent about the planning domain and the domain knowledge of the autonomous planning agent. Therefore, many obvious choices of actions anticipated by the human agent might not be apparent to the autonomous agent.
Further, the autonomous agent often generates a plan based on an abstract model of the domain. Due to such an abstraction, the plan may not be executable in the physical world. Explanation of plans is thus crucial. To address the explanation concern, a popular approach in recent literature is that of contrastive explanations. In particular, contrasting a given plan with alternate ones, in order to produce an argument about why the generated plan is to be chosen for execution over the possible alternatives, has been proposed in~(\cite{krarupPDDL2.1CE}), in the context of discrete systems and problem domains. This forms a foundation for the motivation of our work.

In recent times, with the advent of Cyber-Physical Systems (CPS) and Internet-of-Things, there has been a renewed research interest on planning for hybrid systems that exhibit an interplay of discrete and continuous dynamics. Planning in hybrid systems poses particular challenges to classical AI planners, due to the interplay of continuous and discrete dynamics. This has inspired recent research on planning for hybrid systems, and hybrid system planners (e.g. \textsc{SMTPlan+, \textsc{UPMurphi}} and \textsc{{ENHSP}}) have been at the forefront of planning research in recent literature. These tools accept the planning problem description in the PDDL+ (\cite{PDDL+}) modelling language, which allows to describe processes with mixed (continuous and discrete) dynamics. 

\textbf{Contributions:} In this chapter, we propose a contrastive explanation paradigm for plans in hybrid systems, as done for their discrete counterparts, by highlighting the contrasts of the original plan with an alternate one. We aim to explain a plan by a contrastive explanation framework that provides answers to contrastive questions posed by a plan user. Through such answers, a user gains insights as to why a particular plan can be trusted and deployed, or on the other-hand, why there may be a need for re-planning. Our framework works 
by generating alternate plans to the planning problem by constructing a hypothetical planning problem, and this construction is such that any valid plan for the hypothetical problem will meet the user's expectation phrased in the contrastive question. We present the construction of such hypothetical planning problems for 8 classes of contrastive questions and present proofs of their soundness. In summary, the contributions of this work are as follows:
\begin{enumerate}
    \item We propose a set of eight contrastive questions on plans for hybrid systems. This set of questions is certainly not exhaustive and only portrays some of the critical questions that we believe a user of a planning tool may be interested in posing on plans for hybrid systems.
    
    \item An explanation framework based on the \emph{re-model and re-plan} idea is proposed that advocates the construction of a hypothetical model for each of these contrastive questions in such a way that any valid plan, if generated on this model, will imitate the contrastive question.
    
    \item We introduce a set of metrics for the comparison of a contrastive plan with that of the original plan, thus generating contrastive explanations for the user's questions that may help in understanding the planning domain and the behaviour of the planner in a better way.
    
    \item In addition, we propose to prove the absence of a plan in a hybrid domain using verification tools such as \textsc{dReach} (\cite{DBLP:journals/corr/GaoKCC14}). Since planning problems are undecidable for hybrid systems in general (\cite{ALUR19953}), when a planner fails to generate a valid plan for a given problem instance, it is not clear whether this is due to the underlying undecidability or because the problem instance does not admit a valid plan. To explain this to a user, we leverage bounded reachability analysis of the domain.
    
    \item To illustrate our explanation framework, we consider a car-domain which is modeled in PDDL+ and we synthesize plans using the \textsc{SMTPlan+} planner. The compilation of the hypothetical models are shown for each of the questions, and the resulting contrasting plans are compared with the original plan following a set of metrics similar to the ones introduced in (\cite{cashmore2019towards}). Experimental evaluation of the framework has been presented for two other domains, namely generator-events and planetary-lander, along with the car domain using the three planners \textsc{SMTPlan+}, \textsc{UPMurphi}, and \textsc{ENHSP}.

\end{enumerate}

\noindent 
The rest of this chapter is organized as follows. Section~\ref{sec2} presents an overview of relevant preliminaries and an example problem context. Section~\ref{sec3} illustrates our framework of contrastive explanations, and Section~\ref{secCaseStudy} elucidates a case study on an example domain. Section~\ref{sec4} elaborates on the explanation of no-plans, while Section~\ref{sec5} shows implementation and results. Section~\ref{sec6} presents related literature and Section~\ref{sec7} discusses the limitations of our approach. Section~\ref{sec8} summarizes the contributions and our findings from this work.

\section{Preliminaries} \label{sec2}
\noindent 
In this section, we present an overview of some background concepts that are needed for this work. We begin with a brief description of the Planning Domain Description Language PDDL+. PDDL+ extends PDDL2.1 for representing mixed discrete-continuous domains and planning problems on them. The key features additionally supported in PDDL+ are the ability to model exogenous events and continuous changes as processes. We begin with the definition of a planning instance in PDDL+ and then touch upon each of the above-mentioned features that it brings forth using an example of a planning problem.

\begin{definition} \label{def:PI}
A \emph{planning instance} $\Pi$ in PDDL+ (\cite{PDDL+}) is a pair ($Dom$, $Prob$), where $Dom$   is a 6-tuple $\langle$\textit{Fs}, \textit{Rs}, \textit{As}, \textit{Es}, \textit{Ps}, \textit{arity}$\rangle$ called the domain consisting of a finite set of function symbols \textit{Fs}, predicate symbols \textit{Rs}, action symbols \textit{As}, event symbols \textit{Es}, process symbols \textit{Ps} and an \textit{arity} function that maps each of these symbols to their respective arities. The arity of a symbol specifies the number of arguments it takes. $Prob$ is a triplet $\langle Os, I, G \rangle$ called the planning problem consisting of the set of objects $Os$ in the planning instance, the initial state $I$ and the goal condition $G$. $\Box$
\end{definition}

\noindent We consider a domain consisting of a car that has to travel a specified distance, respecting certain constraints. The planning domain represented in PDDL+ is shown in \textbf{Listing 1}. It has six functions symbols: $d$, $v$, $a$, $upLimit$, $downLimit$ and $runningTime$, three predicate symbols: $running$, $engineBlown$ and $goalReached$, three action symbols \emph{accelerate}, \emph{decelerate} and \emph{stop}, one event symbol \emph{engineExplode} and one process symbol \emph{moving}. The $arity$ maps each of these symbols to $0$. The problem instance that we have used in this domain is shown in \textbf{Listing 2} where $I$ is the initial state and $G$ is the goal condition as discussed later.

\begin{lstlisting}[basicstyle=\footnotesize,caption={The car domain in PDDL+.}, label={lst:Listing1}]
(define (domain car)
(:predicates (running) (engineBlown) (goalReached))
(:functions (d) (v) (a) (upLimit) (downLimit) (runningTime))
(:process moving
  :parameters ()
  :precondition (and (running))
  :effect (and (increase (v) (* #t (a))) (increase (d) (* #t (v)))
	  (increase (runningTime) (* #t 1))))
(:action accelerate
  :parameters()
  :precondition (and (running) (< (a) (upLimit)))
  :effect (and (increase (a) 1)))
(:action decelerate
  :parameters()
  :precondition (and (running) (> (a) (downLimit)))
  :effect (and (decrease (a) 1)))
(:event engineExplode
  :parameters ()
  :precondition (and (running) (>= (a) 1) (>= (v) 100))
  :effect (and (not (running)) (engineBlown) (assign (a) 0)))
(:action stop
  :parameters()
  :precondition (and (= (v) 0) (>= (d) 30) (not (engineBlown)))
  :effect(goalReached)))
\end{lstlisting}

\noindent Grounding of any symbol in PDDL+ is defined as follows:
\begin{definition}
Given a planning instance $\Pi$, \textbf{grounding} of a symbol $\chi \in$ \textit{Fs} $\cup$ \textit{Rs} $\cup$ \textit{As} $\cup$ \textit{Es} $\cup$ \textit{Ps} is formed by instantiating the arguments of the symbol with its actual parameters while respecting the arity of the symbol. $\Box$
\end{definition}
\noindent A set of atomic $propositions$ $P$ is obtained from grounding the predicate symbols in $Rs$. For example, the atomic proposition $(running)$ can be directly obtained from the predicate symbol $running$ since its arity is 0. An atomic proposition ($available$ $unit1$) is an example of a grounded predicate of arity 1 in the \emph{Planetary Lander} domain described in Appendix 11.2. Here, $available$ is a predicate and $unit1$ is an object that is an argument to the predicate $available$.
\noindent The set of \emph{Primitive\ Numeric\ Expressions} (\textit{PNEs}) (\cite{PDDL+}) is defined in the following.
\begin{definition}
For a given planning instance $\Pi$, the \textbf{Primitive Numeric Expressions} (\textit{PNE}s) are the terms constructed from the grounded function symbols of the domain with the number of parameters given by their arities. $\Box$
\end{definition}
\noindent Like grounded predicate symbols result in atomic propositions, grounded function symbols result in numeric values.
An example of a \textit{PNE} in our domain is $(a)$, which denotes a numeric value modeling the acceleration of the car at an instance of time. A domain may contain numeric expressions formed with arithmetic operations on the \textit{PNEs}, such as $a=0$ or $a=a+1$, which denote initialization of acceleration or increase in acceleration of the car or formation of a constraint in the domain, etc. We denote the set of grounded actions by $A$ where a ground action is defined in the following:

\begin{definition} \label{def:grd-action} 
A \textbf{ground action} (\cite{PDDL+}) $act$ is obtained from an action symbol in $As$ by substituting objects for each of its parameters. The components of $act$ consist of a precondition \textit{Pre($act$)} and an effect \textit{Eff($act$)}. \textit{Pre($act$)} is a formula that needs to be satisfied to make the action applicable. An \textit{Eff($act$)} is the action's postcondition consisting of assignments to atomic propositions and \textit{PNEs}. {The effect \textit{Eff} can be \textit{Eff}$^{+}$ or \textit{Eff}$^{-}$. \textit{Eff}$^{+}$ signifies addition of an effect with the existing ones. Similarly, \textit{Eff}$^{-}$} signifies deletion of an effect from the existing ones. $\Box$
\end{definition}

\noindent An example of a ground action in our domain is \emph{accelerate}, which has the precondition $running~\land~(a < upLimit)$ as its precondition where $\land$ denotes Boolean AND. The postcondition is an assignment of $a+1$ to the \textit{PNE} $a$, shown with the $increase$ construct in \textbf{Listing 1}.
PDDL+ supports discrete instantaneous actions as well as durative actions, i.e., actions which occur for a finite duration. Durative actions can also be mapped into an equivalent start-process-stop representation. The reason for performing the mapping is in order to give durative actions a semantics in terms of the underlying constructs of PDDL+, which are themselves given an interpretation in terms of hybrid automata as discussed in (\cite{PDDL+}). However, the semantics of PDDL+ can also be defined without going through hybrid automata, as explained by (\cite{DBLP:journals/ai/ShinD05}) and (\cite{DBLP:conf/aips/PercassiSV21}). The definition of a \textbf{ground event} is the same as Definition \ref{def:grd-action}, with the restriction that events are required to have at least one numeric precondition, which makes them a special case of actions (\cite{PDDL+}). An example of an event in our car domain is \emph{engineExplode}. Actions in PDDL+ are executable by the planner, whereas the events are not, since they are triggered by the environment. In our car domain, the execution of actions \emph{accelerate} and \emph{decelerate} respectively increases and decreases the acceleration of the car by one unit. The event \emph{engineExplode} is triggered when the velocity and acceleration of the car reach a certain threshold. The set of $ground\ processes$ $pr$ is obtained from the process symbols in $Ps$. A ground process is defined as follows:

\begin{definition}
A \textbf{ground process} (\cite{PDDL+}) $proc$ is an instantiation of a process symbol in $Ps$ having a name together with its actual parameters. It consists of a precondition \textit{Pre($proc$)} and an effect \textit{Eff($proc$)}. \textit{Pre($proc$)} is a propositional precondition for the process's activation and consists of atomic propositions formed over either the ground atoms in the planning domain or else relational terms constructed from arithmetic operations applied to \textit{PNEs} or real values. An \textit{Eff($proc$)} is a numeric postcondition that is a conjunction of additive assignment propositions, the values of which are expressions that are of the form $(*\ \#t\ exp)$ where exp is $\#t$ free ($exp$ does not contain any sub-expression that uses $\#t$). Continuous effects are represented by update expressions that refer to the special variable $\#t$. $\Box$
\end{definition}

\noindent The variable \textit{$\#t$} in the postcondition is syntactic and signifies that the effect of a process is time-dependent, which represents the continuous system dynamics. An example of a ground process in our domain is $moving$ with the atomic proposition $running$ as its activation precondition. The process has a postcondition \textit{$increase\ (v)\ (*\ \#t\ (a))$}, that is semantically equivalent to $\frac{dv}{dt}$ = $a$. Similarly, the continuous change in distance ($d$) covered by the car and the time elapsed while moving ($runningTime$) are modeled with differential equations.

\noindent A state $s$ of a PDDL+ domain consists of a time $t \in\mathbb{R}$, a logical constituent $s_l\subseteq P$, and a numeric constituent $s_v$ that describes the values for the \textit{PNE}s at that state. The goal condition $G$ is a proposition that can include both atoms formed from the relation symbols and objects of the planning instance and numeric propositions between primitive numeric expressions and numbers (\cite{PDDL+}). The initial state $I$ is the state of the model at time $t$ = $0$. The set of objects $Os$ is the entity of interest in the domain. An example of a planning problem in PDDL+ is shown in \textbf{Listing 2}. The problem has an empty set of objects $Os$, the initial condition $I$ specifies that the predicate $running$ is initially $true$, the initial value of the functions $a$, $v$, $d$ and $runningTime$ are assigned 0 and that of $upLimit$, $downLimit$ are assigned 1 and -1 respectively. The goal for the car is to travel a minimum distance of 30 units in less than 50 units of time while avoiding an engine explosion, which is caused when the velocity of the car is greater than or equal to 100 units and the acceleration is greater than or equal to 1 unit. This is specified in $G$ with the predicates $goalReached$ and the negation of $engineBlown$, together with the numeric condition $runningTime \leq 50$. 

\lstset{
    escapeinside={(@}{@)}
 }
\begin{lstlisting}[basicstyle=\footnotesize,caption={The planning problem in the car domain in PDDL+.}, label={Listing2}]
(define (problem car_prob)
   (:domain car)
   (:init (running) (= (runningTime) 0) (= (upLimit) 1) (= (downLimit) -1)
         (= d 0) (= a 0) (= v 0))
  (@\textbf{\ \ (\ :\ goal\ (\ and\ (\ goalReached\ )\ (\ not\ (\ engineBlown\ )\ )
    \ \ (\ $\leq$\ (\ runningTime\ ) 50\ )\ )\ )} @)
\end{lstlisting}

\noindent Hybrid automata will be presented pictorially alongside the PDDL+ representation of planning domains in the work, for ease of illustration. We now briefly describe a mapping between the PDDL+ constructs and the components of HA. The variables $x\in Var$ in HA map to the \textit{PNEs} of a PDDL+ model. Recall that a \textit{PNE} is a grounded function and assumes numeric values. Each location $l \in Loc$ maps to a logical state of the PDDL+ model, i.e., a subset of predicates $P$ which are true. The tuple $Init= \langle l_{ini}, S \rangle$ maps to the initial condition $I$ of the PDDL+ model. $l_{ini}$ represents the predicates in $P$ which are true initially and $S$ represents initial assignment of values to the \textit{PNEs}. The flow in a location $Flow(l)$ maps to the continuous change due to a process in PDDL+. The invariant in a location $Inv(l)$ maps to a subset of \textit{PNEs} in PDDL+ that is used to specify the condition on the numeric state of the PDDL+ model that should hold throughout the execution of a process. The set of labels $Lab$ maps to the set of ground actions and events in PDDL+. Note that there is no distinction between an action and an event in an HA. A transition $e$ = $(l,a,g,r,l')$ maps to a discrete change of logical state from $l$ to $l'$ in PDDL+ due to any action or event $a$. The guard $g$ maps to the precondition, and the reset $r$ maps to the postcondition of the action/event on the \textit{PNEs} respectively.
We present a representation of the car domain as a hybrid automaton in Figure \ref{fig:domain} below:
\begin{figure}[htbp]
    \centering
    \includegraphics[width=0.66\textwidth]{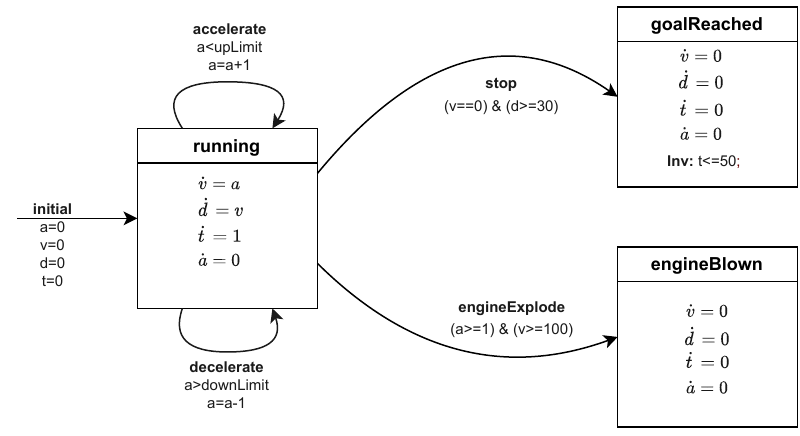}
    \caption{{The hybrid automaton model of the Car domain}}
    \label{fig:domain}
\end{figure}


\begin{definition}\label{ch2def:plan}
A plan $\phi$ is a tuple $\langle \lambda, makespan \rangle$. For a planning instance with a set of ground actions $A$, $\lambda$ is a finite set of triplets $\langle t, act, dur \rangle$ {together with the plan \emph{makespan} $\in \mathbb{R}$}. In the triplet $\lambda$, $t \in \mathbb{R}^+$ is the time instant of executing the action $act \in A$ and $dur \in \mathbb{R}^+$ is the duration for which the action $act$ remains active in the plan. {The \emph{makespan} is the overall duration of the plan}. $\Box$
\end{definition}

\begin{figure}[htbh]
    \centering
      \begin{subfigure}[b]{0.6\textwidth}
        \centering
        \scalebox{0.7}{\includegraphics[width=\textwidth]{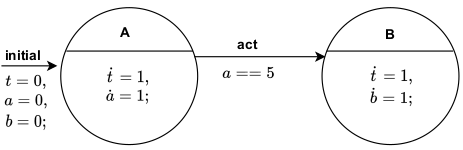}}
        \caption{Case1}
        \label{fig:case1}
      \end{subfigure}
    \hfill
      \begin{subfigure}[b]{0.27\textwidth}
        \centering
        \scalebox{0.7}{\includegraphics[width=\textwidth]{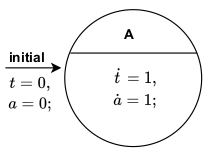}}
        \caption{Case2}
        \label{fig:case2}
      \end{subfigure}
     \caption{Example scenarios where \emph{makespan} is useful. In the figure, $t$ encodes time, and $a$ and $b$ are variables.}
     \label{fig:fig1}
\end{figure}

\noindent Note that a plan does not report events and processes. This is because neither the occurrence of an event nor its duration can be controlled by a planner. For the same reason, since processes model continuous changes in the system, they are not under the control of the planner. The duration of instantaneous actions in a plan is always zero. \emph{makespan} captures the total time spent by a system during which a plan is active. A system may spend time dwelling due to the application of processes. As processes are not visible in the plan, such dwelling times are captured in the passage of time between consecutive applications of actions. The dwell time before any action appears in the plan is captured by the time point at which the first action appears. However, the dwell time that may appear after the application of the last action in the plan before reaching the goal state needs to be derived from the \emph{makespan}. Consider Case 1 in Figure \ref{fig:fig1} where for a given goal state such as \{$(B)\land (b==5)$\}, a plan may consist only of an application of the action \emph{act}. However, the system needs to dwell 5 time units even after the application of the action \emph{act} to reach the goal state, which is not visible in the plan but captured in the \emph{makespan}. Another typical situation may arise when a goal state could be reached only by dwelling for a duration. In such a case, the plan may not consist of any action. For example, consider Case 2 in Figure \ref{fig:fig1} where a goal state \{$(A)\land (a==5)$\} can be reached only by dwelling 5 time units in the location \emph{A}. The plan in this case consists of an empty list of actions. Here, the dwelling time of the system is captured by the \emph{makespan} for such an empty plan.

\noindent In our contrastive plan explanation framework, which is discussed in Section \ref{sec3}, some of the contrastive questions are concerning the sequence of actions appearing in a plan. We now therefore define an action sequence in a plan $\phi$.

\begin{definition} \label{def:ASeq}
Given a planning instance $\Pi$ and a plan $\phi$, an \textbf{action sequence} is an ordered set of ground actions in $\phi$ ordered by their time of appearance in $\phi$. $\Box$ 
\end{definition}

\noindent Listing 3 shows the generated plan on the planning problem shown in Listing 2 on the car domain of Listing 1, generated by \textsc{SMTPlan+}, a planning tool for hybrid systems.

\lstset{
    escapeinside={(@}{@)}
}
\begin{lstlisting}[basicstyle=\footnotesize,caption={A plan generated by \textsc{SMTPlan+} on the car domain on a planning problem in Listing 2. The plan duration is 32.0 units.}, label={Listing3}]
    (@\textbf{\underline{Time}}@)         (@\textbf{\underline{Action}}@)              (@\textbf{\underline{Duration}}@)
    0.0:        (accelerate)        [0.0]
    1.0:        (decelerate)        [0.0]
    31.0:       (decelerate)        [0.0]
    32.0:       (stop)              [0.0]
    (@\textbf{\emph{makespan}:} = 32.0 units.@)
\end{lstlisting}

\noindent In the following, we describe the intuition behind contrastive questions for hybrid system plans, and discuss how the questions can be translated to PDDL+ for use in \textsc{SMTPlan+} for alternative plan generation. The hybrid automata description of the corresponding PDDL+ models are also shown for clarity on model semantics and as a visual aid for the readers.
\section{Explanation Framework} \label{sec3}



\noindent In this section, we address contrastive questions on a plan in a hybrid system, such as (a) why execute action A and not action B at a point in the plan? (b) why apply an action for $\tau$ duration at a point in the plan and not more or less? (c) why does an action sequence $a_i,a_{i+1},\dots,a_{i+n}$ appear in the plan and not any other sequence? (d) why does a plan have a given duration/length and not less? 
To address the class (a) questions, there are finitely many contrastive plan alternatives that are to be considered, given that the number of discrete actions is finitely many. We propose to address explanations of contrastive questions over the actions by a contrastive explanation framework, where the human agent may ask contrastive questions over the action space, and our framework answers with alternative plans and their costs until the human agent builds trust over the generated plan.
Addressing contrastive questions of class (b) requires other novel techniques since we cannot explicitly consider all possible alternate dwell times and re-plan options as there are infinitely many. Here, we intend to explore solutions by building hypothetical models, with added time variables and constraints on them, and then, re-plan over these hypothetical ones to obtain an explanation. For example, we may introduce a new time variable $\pi$ and add a constraint such as $(\pi < \tau)$ in the original model to obtain a hypothetical model H. Similarly, we can also obtain another hypothetical model H’ by adding the constraint $(\pi > \tau)$ and executing a re-plan step. Re-planning on H and H’ and comparing the results with the original plan may provide an explanation of (b).
The class (c) questions require preserving the action sequence that appears before and after the action sequence $a_i,a_{i+1},\dots,a_{i+n}$ while allowing any sequence of actions other than the one specified to happen in between. The contrastive questions of class (d) relate to queries to investigate if there are plans of shorter duration or shorter length. Such questions can use the iterative strength of our explanation framework to find an optimal plan in terms of duration/length.

In this work, we propose the following collection of contrastive questions and techniques to provide their explanation:
\begin{enumerate}
    \item Why did the planner choose to do action A and not B instead? 
    \item Why did the planner not choose to do an action later in the plan?
    \item Why did the planner not choose to do an action earlier in the plan?
    \item Why did the planner choose to do an action in the plan, instead of not doing it?
    \item Why not have fewer occurrences of an action in the plan?
    \item Why is the accumulative duration of the plan not less?
    \item Why did an action sequence appear in the plan?
    \item Why is the length of the plan not less?
\end{enumerate}


\subsection{Contrastive Explanation Framework} \label{sec3.2}
{In this section, we present an explanation framework for contrastive questions about a planning instance for a hybrid system. The framework is an iterative and collaborative model based on (\cite{DBLP:journals/corr/abs-2103-15575}), where the collaboration comes as the four-stage mixed-initiative process as:} 
\noindent{(i) a user asks a contrastive question by observing a plan; (ii) a formal question is formed by deriving the constraints from the user question; (iii) these constraints are compiled into a hypothetical planning model; (iv) a solution for the hypothetical model is formed. The alternative solution thus derived is called a hypothetical plan. It contains the contrast cases expected by the user, and hence can be compared with the original plan. A comparison between the plans serves as an explanation for the user's query.}

\begin{figure}[htb]
    \centering
    \includegraphics[width=0.5\textwidth]{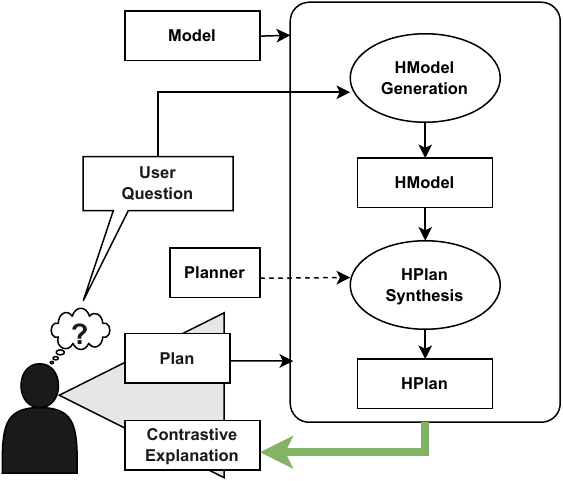}
    \caption{{The Contrastive Plan Explanation Framework.}} 
    \label{fig:remodel-replan-scheme}
\end{figure}

However, before initiating the discussion on our contrastive plan explanation framework for a hybrid system, we formally define an explanation problem as below:
\begin{definition} \label{def:explanation-problem}
An \textbf{explanation problem} is a tuple $E=\langle \Pi,\phi,Q\rangle$ (\cite{cashmore2019towards}), where $\Pi$ represents the planning instance (see Def. \ref{def:PI}), $\phi$ is the plan generated by a planner (see Def. \ref{ch2def:plan}), and $Q$ represents the contrastive question posed by the user. $\Box$ 
\end{definition}
\noindent To bring forth explanations to the above-mentioned set of contrastive questions, we propose a \emph{Contrastive Plan Explanation Framework} based on~(\cite{krarupPDDL2.1CE}) which is given in the context of discrete systems and problem domains. We extend their framework to hybrid system models. Additionally, we propose a set of algorithms for each contrastive question that constructs the hypothetical models from the corresponding user questions. The framework takes an \emph{explanation problem $E$} and produces a \emph{contrastive explanation $CE$} (see Def. \ref{def:CE}), by contrasting a given plan with alternate ones produced with hypothetical models, as an argument about why the generated plan is to be chosen for execution over the possible alternatives. A schematic of this framework is shown in Figure \ref{fig:remodel-replan-scheme}.

\noindent Based on the contrastive question, the original model is modified to a hypothetical model that enforces the planner to generate a contrastive plan as expected by the user. We denote the original planning instance as \textit{Model} $\Pi$ and the hypothetical model formed after incorporating the necessary changes in the original model as \textit{HModel} $\Pi'$. A plan obtained for an \textit{HModel} is a hypothetical plan denoted as \textit{HPlan} $\phi'$.
To formalize the discussion of contrastive plan explanation, we now introduce the relevant definitions. A constraint property is formed from the question $Q$, which represents some user-imposed constraints over a plan $\phi$ and can be defined as follows.
\begin{definition} \label{def:const-prop}
    For a planning instance $\Pi$ and a plan $\phi$, a \textbf{constraint property} (\cite{DBLP:journals/corr/abs-2103-15575}) is a quantifier-free first-order logic predicate $\psi$ over $\phi$ representing constraints from a user question $Q$. $\Box$
\end{definition}

\noindent The constraint operator, which encapsulates a constraint property within a planning model, is given as follows.
\begin{definition} \label{def:const-op}
(\cite{DBLP:journals/corr/abs-2103-15575}) A \textbf{constraint operator} $\times$ is defined so that, for a planning instance $\Pi$ and any constraint property $\psi$, $\Pi \times \psi$ constructs a \textit{HModel} $\Pi'$, satisfying the condition that any plan for $\Pi'$ is a plan for $\Pi$ that also satisfies $\psi$. $\Box$
\end{definition}

\begin{definition}
\textbf{\textit{HModel}} is a new planning instance $\Pi'$ constructed from the planning instance $\Pi$ which encapsulates a constraint property $\psi$ of a user question $Q$. $\Pi'$ can be defined as:
\begin{equation*}
    \Pi' = (\langle Fs', Rs', As', Es', Ps', arity'\rangle,\langle Os', I', G'\rangle)
\end{equation*}
where \textit{Fs'}, \textit{Rs'}, \textit{As'}, \textit{Es'}, \textit{Ps'}, \textit{arity'}, \textit{Os'}, \textit{I'} and \textit{G'} indicate modifications on the corresponding components. $\Box$
\end{definition}
\begin{definition}

\textbf{\textit{HPlan}} is a hypothetical plan $\phi'$ produced by the planner over an \textit{HModel} $\Pi'$ which satisfies the constraint(s) posed in question $Q$ by a user. $\Box$
\end{definition}

\noindent \emph{Contrastive Explanation Metrics:} we compare the \textit{HPlan} $\phi'$ with the original plan $\phi$. The comparison is done by adopting a contrastive explanation metric defined as follows.

\begin{definition} \label{def:CE}
To show the differences between $\phi$ and an  \textit{HPlan} $\phi'$, a contrastive explanation \textit{CE} (\cite{cashmore2019towards}) of a plan $\phi$ on a hybrid-system domain $D$ for a given question $Q$ is a tuple $\langle E, Q \rangle$, where $E$ consists of the following six components:

\begin{itemize}
    \item \emph{\textit{Remove ($\phi$,$\phi'$)}}: The set of actions removed from the original plan.
    \item \emph{\textit{Add ($\phi$,$\phi'$)}}: The set of actions added to the alternate plan.
    \item \emph{\textit{Common ($\phi$,$\phi'$)}}: The set of actions present in both the original and the alternate plan.
    \item \textit{dwell-diff}: is the set $\{ (p,t) \mid p \in Ps \textit{ and } t = t_{orig}\ - t_{alt} \}$. The set contains a pair $(p,t)$ for every $p$ in the set of process symbols $Ps$ where the corresponding $t$ shows the difference of the dwell time in process $p$ due to this alternate plan. Here, $t_{orig}$ and $t_{alt}$ denote the dwell time in $p$ in the original plan $\phi$ and the alternate plan $\phi'$, respectively.
    \item $\textit{diff-cost}_{\tau}$: The \emph{makespan} of the alternate plan minus the \emph{makespan} of the original plan.
    \item $\textit{diff-cost}_{len}$: The number of occurrences of actions in the alternate plan minus the number of occurrences of actions in the original plan.
\end{itemize}
\end{definition}

\noindent Now in the light of PDDL+, from a given planning instance $\Pi$, and a contrastive question $Q$, we denote a compilation of a hypothetical planning instance by $Compilation(\Pi, Q)$. The result of this compilation is a modified planning instance $\Pi'$ = ($\langle Fs'$, $Rs'$, $As'$, $Es'$, $Ps'$, $arity'\rangle$,$\langle O's$, $I'$, $G'\rangle$). Here, the HModel $\Pi'$ is derived as $\Pi \times \psi$, where $\psi$ is the constraint property derived from $Q$. $Fs'$, $Rs'$, $As'$, $Es'$, $Ps'$, $arity'$, $O's$, $I'$, and $G'$ represent modifications on the corresponding elements. However, if the user wants to ask questions iteratively about the model, then $\Pi'$ can also be used as an input iteratively.
This allows the user to stack questions, further increasing the understanding of the plan through iterative questioning.

\section{Contrastive Questions} \label{secCaseStudy}

In this section, we consider the case study of the car domain (Listing 1) and an instance of a planning problem (Listing 2) to illustrate our explanation framework for each of the earlier discussed contrastive questions. We use \textsc{SMTPlan+} as the planner for our experiments. 
We start by taking a user question, a formal question is formed by deriving the constraints, and these constraints are compiled into a hypothetical planning model \emph{HModel}. Then, a solution for the \emph{HModel} is formed. The alternative solution derived is called a hypothetical plan \emph{HPlan}.
The compilation of the hypothetical model (\textit{HModel}) is shown for the corresponding question, and the resulting \textit{HPlan} generated by \textsc{SMTPlan+} is compared with the original plan. 
A contrastive explanation, as per our comparison metric, is reported for each case to show the differences between the original plan and the alternate one.
These comparisons between the plans serve as an explanation for the user's query.

\subsection{Replacing an Action by Another in the Plan} \label{secA}
\noindent A user might question the appearance of an action at a certain stage of the plan, and may look for an alternate option. For example, for a given plan $\phi$, a contrastive question Q can be asked of the form:
\begin{quote}
Why did the planner choose to perform action A in state S rather than action B?
\end{quote}
\noindent {We are given a plan $\phi$ consisting of an action sequence $\langle a_1$, $a_2$, \ldots, $a_i$, \ldots, $a_n\rangle$ that leads the system from an initial state $I$ to the goal $G$.
Let the system be in state $S$ after the application of the grounded action $a_{i-1}$. Then, the question can be more precisely asked as \enquote{why was the action $ground(a_i)$ at state $S$ chosen to be applied, rather than the action $ground(b)$?}.}
\noindent {For example, by observing the example plan in Listing 3, the user might ask the question:}
\begin{quote}
    {\enquote{Is it possible to replace the first instance of \emph{decelerate} with \emph{accelerate}}?}
\end{quote}
\noindent {An intuitive argument supporting the preference to replace the \emph{decelerate} action with the \emph{accelerate} action early in the plan might be due to the fact that it can enable the car to travel faster and the appearance of \emph{decelerate} action earlier in the plan can slow down the car.}

{To generate the \emph{HPlan} $\phi'$, a compilation is formed such that the action $ground(b)$ appears in the plan in place of the action $ground(a_i)$ resulting in a new state $S'$, followed by a new sub-plan $\phi_s$ containing an arbitrary sequence of grounded actions $\langle c_1$, $c_2$, \ldots, $c_m\rangle$ leading to the goal $G$. The \emph{HPlan} is then the initial set of actions of the original plan $\phi$ concatenated with $b$ and the new sub-plan $\phi_s$ as below: 
\begin{equation*}
    \phi': \langle a_1, a_2, \ldots, a_{i-1}, b, c_1, c_2, \ldots, c_m\rangle
\end{equation*}
To obtain the state $S'$, we need to preserve the sequence of grounded actions $\langle a_1$, $a_2$, \ldots, $a_{i-1}\rangle$, which is followed by $ground(b)$ in a valid plan. For that, the corresponding \emph{HModel} is constructed as follows:}
{For each action $a_j$ in $\langle a_1$, $a_2$, \ldots, $a_{i-1}\rangle$, we introduce a new action $act_j$ which can only appear in the preserved sequence and a new predicate $has\_done\_act_j$ which represents that the action $act_j$ has been applied.
For the action $b$, we introduce a new action $b1$ and a new predicate $has\_done\_b$. $b1$ will only appear to replace $a_i$ in a plan, whereas $b$ may appear in any other place in a plan. The goal is extended to include the grounded predicate $has\_done\_b$ to enforce that the user suggested action $ground(b)$ replaces the action $ground(a_i)$ in the original plan $\phi$. This is achieved by constructing \emph{HModel} $\Pi'$ as follows.}

\begin{equation*}
    {\Pi'=(\langle Fs, Rs', As', Es, Ps, arity'\rangle,\langle Os, I, G'\rangle)}
\end{equation*}
where
\begin{itemize}
    \item {{$Rs'$: $Rs \cup \{(has\_done\_act_j$, $\forall j \in [1,i-1]), has\_done\_b\}$;}}
    \item {{$As'$: $As \cup \{(act_j$, $\forall j \in [1,i-1]), b1\}$;}}
    \item {{$arity'(act_j)$ = $arity(a_j)$, $\forall j \in [1,i-1]$;}}
    \item {{$arity'(b1)$ = $arity(b)$;}}
    \item {{$arity'(has\_done\_act_j)$ = $0$, $\forall j \in [1,i-1]$;}}
    \item {{$arity'(has\_done\_b)$ = $0$;}}
    \item {{$G'$: $G\land ground(has\_done\_b)$;}}
\end{itemize}
{where the first action in the sequence $act_1$ has been extended with an added precondition $\neg(has\_done\_act_1)$}
\begin{equation*}
    {\textit{Pre}(act_1)=\textit{Pre}(a_1)\land \neg(has\_done\_act_1)}
\end{equation*}
{and an added effect ($has\_done\_act_1$), which indicates that the action has been applied in the plan,}
\begin{equation*}
    {\textit{Pre}(act_1)=\textit{Pre}(a_1)\land (has\_done\_act_1)}
\end{equation*}
for all $j\in [2,i-1]$, each $act_j$ has been extended with added precondition $has\_done\_act_{j-1}$ and $\neg(has\_done\_act_j)$ which ensures that the previous action in the sequence has been applied already,
\begin{equation*}
    {\textit{Pre}(act_j)=\textit{Pre}(a_j)\land (has\_done\_act_{j-1}) \land \neg(has\_done\_act_j)}
\end{equation*}
{and added effect $has\_done\_act_{j}$, which indicates that the action has been applied in the plan,}
\begin{equation*}
    {\textit{Eff}^{+}(act_j)=\textit{Eff}(a_j)\cup \{has\_done\_act_{j}\}}
\end{equation*}
{Action $b1$ extends the action $b$ with an added precondition $has\_done\_a_{i-1}$ and an added effect $has\_done\_b$. The precondition $has\_done\_a_{i-1}$ ensures that $b1$ will appear only after the sequence $\langle act_1$, $act_2$, \ldots, $act_{i-1}\rangle$ and the effect $has\_done\_b$ would enforce its application in a valid plan.}
\begin{equation*}
    {{\textit{Pre}(b1)=\textit{Pre}(b)\land (has\_done\_act_{i-1})}}
\end{equation*}
\begin{equation*}
    {{\textit{Eff}^{+}(b1)=\textit{Eff}(b)\cup \{has\_done\_b\}}}
\end{equation*}
All the existing actions of the original model $\Pi$ are extended with the added precondition $has\_done\_b$, which indicates that they may appear in the rest of the plan.
\begin{equation*}
    {{\textit{Pre}(a)=\textit{Pre}(a)\land (has\_done\_b), \forall a \in As'}}
\end{equation*}
All the added actions have semantics similar to their original counterparts in the domain; we only restrict where they can appear in a valid plan. Therefore, $act_j$ in the \emph{HPlan} $\phi'$ represents $a_j$ in the original plan $\phi$ and should not be treated as a different action. Similarly, $b1$ and $b$ represent the same action in the \emph{HPlan} except for a different precondition.
Now, we show that the above construction of the \emph{HModel} is sound, in other words, it produces an \emph{HPlan} which is expected by the user.
\begin{lemmasec}
The construction of the \emph{HModel} is sound.
\end{lemmasec}

\noindent \textsc{Proof:} A valid \emph{HPlan} $\phi'$ for the \emph{HModel} $\Pi'$ consists of an action sequence $\langle act_1$, $act_2$, \ldots, $act_{i-1}$ $b1\rangle$ which is followed by some sequence of actions in $As'$. As ($has\_done\_b$) is included as a goal constraint to ensure that $b1$ appears in a plan whereas it's precondition {$has\_done\_act_{i-1}$} ensures that it can appear only after the action sequence $\langle act_1$, $act_2$, \ldots, $act_{i-1}\rangle$ has appeared in the plan. Moreover, the precondition $\textit{Pre}(act_j)=\textit{Pre}(a_j)\land (has\_done\_act_{j-1}) \land \neg(has\_done\_act_j)$ ensures that actions appearing in the preserved sequence do not repeat. Other domain actions have the precondition $has\_done\_b$, which means that they can appear only after $b1$ has been applied in a plan. Hence, any valid plan must start with an action sequence $\langle act_1$, $act_2$, \ldots, $act_{i-1}\rangle$ that is followed by $b1$ which replaces the action $a_i$ in the original plan $\phi$ and finally, succeeded by some action sequence $\langle c_1$, $c_2$, \ldots, $c_m\rangle$ leading to the goal. This shows that our construction is sound. $\Box$

\begin{exmp}
{Consider the user question above where $b$ is \emph{accelerate} and $a_i$ is the grounded \emph{decelerate} action {(the second ground action)} in the original plan $\phi$. The plan $\phi$ consists of a sequence of grounded actions:
\begin{equation*}
    \phi: \langle accelerate, decelerate, decelerate, stop\rangle  
\end{equation*}
To reflect the user suggestions in the \emph{HPlan} $\phi'$, a valid plan must preserve the first grounded action \emph{accelerate} in the plan whereas the second grounded action \emph{decelerate} must be replaced with another grounded \emph{accelerate} action.}

To generate the \emph{HPlan}, the compilation is formed as discussed above.
{We introduce a new action \emph{accelerate1} which appears only at the start of a valid plan preserving the place of the first ground action \emph{accelerate} in the original plan $\phi$ and a new predicate $has\_done\_acc1$ which represents the fact that action \emph{accelerate1} has been applied. Similarly, we introduce another new action \emph{accelerate2} to replace the second grounded action \emph{decelerate} in the original plan $\phi$ to form a valid \emph{HPlan} $\phi'$ and a predicate $has\_done\_acc2$. The latter indicates that the grounded action \emph{accelerate2} has been applied to replace the grounded \emph{decelerate} action from a valid \emph{HPlan}. All existing actions of the domain have been extended with the precondition $has\_done\_acc2$. The modifications are shown below:}

\begin{itemize}
    \item {\textit{Pre}$(accelerate1)$ = \textit{Pre}$(accelerate)\land\ \neg(has\_done\_acc1)$}
    \item {\textit{Eff}$^{+}(accelerate1)$ = \textit{Eff}$(accelerate)\cup\ \{(has\_done\_acc1)\}$}
    \item {\textit{Pre}$(accelerate2)$ = \textit{Pre}$(accelerate)\land\ (has\_done\_acc1)\land\ \neg(has\_done\_acc2)$}  
    \item \textit{Eff}$^{+}(accelerate2)$ = \textit{Eff}$(accelerate)\cup \{(has\_done\_acc2)\}$
    \item \textit{Pre}$(accelerate)$ = \textit{Pre}$(accelerate)\land (has\_done\_acc2)$
    \item \textit{Pre}$(decelerate)$ = \textit{Pre}$(decelerate)\land (has\_done\_acc2)$
    \item \textit{Pre}$(stop)$ = \textit{Pre}$(stop)\land (has\_done\_acc2)$
\end{itemize}

The modified domain is shown in Listing 4:
\begin{lstlisting}[basicstyle=\footnotesize,caption={Modified domain with \emph{accelerate}, \emph{accelerate1} and \emph{accelerate2} action.}]
(:predicates (running) (engineBlown) (goalReached)
             (has_done_acc1) (has_done_acc2))
(:functions (d) (v) (a) (upLimit) (downLimit) (runningTime) )
    ...
(:action accelerate1
  :parameters()
  :precondition (and (running) (< (a) (upLimit)) (not (has_done_acc1)))
  :effect (and (increase (a) 1) (has_done_acc1)))
(:action accelerate2
  :parameters()
  :precondition (and (running) (has_done_acc1) (not (has_done_acc2)))
  :effect (and (increase (a) 1) (has_done_acc2)))
(:action accelerate
  :parameters()
  :precondition (and (running) (< (a) (upLimit)) (has_done_acc2))
  :effect (and (increase (a) 1)))
(:action decelerate
  :parameters()
  :precondition (and (running) (> (a) (downLimit)) (has_done_acc2))
  :effect (and (decrease (a) 1)))
(:action stop
  :parameters()
  :precondition (and (= (v) 0) (>= (d) 30) (not (engineBlown))
                (has_done_acc2))
  :effect(goalReached)))
\end{lstlisting}
The initial and goal states are modified as shown below:
\begin{lstlisting}[basicstyle=\footnotesize]
(:init (running) (= (runningTime) 0) (= (upLimit) 1) (= (downLimit) -1)
         (= d 0) (= a 0) (= v 0) (= c 0))
(:goal (and (goalReached) (not(engineBlown))
            (<= (runningTime) 50) (has_done_acc2)))
\end{lstlisting}
The hybrid automaton model of the modified domain is shown in Figure \ref{fig:domain_A}, where two new actions \emph{accelerate1} and \emph{accelerate2} are added to the model. It may be noted that the precondition $(a<upLimit)$ of the \emph{accelerate} action in the original model is omitted for the \emph{accelerate2} action in the \emph{HModel} as it restricts two consecutive occurrences of \emph{accelerate} actions in the plan.
\begin{figure}[htbp]
    \centering
    \includegraphics[width=0.66\textwidth]{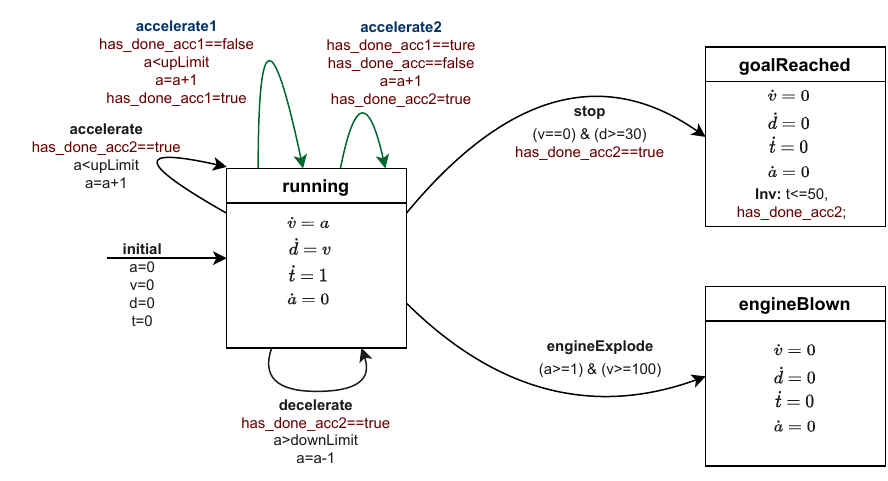}
    \caption{The hybrid automaton for the corresponding \textit{HModel} of the question is addressed in sect 4.1.}
    \label{fig:domain_A}
\end{figure}


The generated \emph{HPlan by \textsc{SMTPlan+} is:}
\begin{lstlisting}[basicstyle=\footnotesize]
    (@\underline{Time}@)         (@\underline{Action}@)               (@\underline{Duration}@)
    0.0:        (accelerate1)        [0.0]
    1.0:        (accelerate2)        [0.0]
    2.0:        (decelerate)         [0.0] 
    3.0:        (decelerate)         [0.0]
    8.0:        (decelerate)         [0.0]
    12.0:       (stop)               [0.0]
    (@\emph{makespan}: = 12.0 units.@)
\end{lstlisting}
\end{exmp}

\noindent We observe that the synthesized plan is shorter in \emph{makespan} but longer in length, with a greater number of actions.

Contrastive Explanation: The \emph{contrastive explanation} for the above user question is as follows:
\begin{itemize}
\item{\textit{Remove ($\phi$,$\phi'$)}: \emph{accelerate} action is removed from the original plan;}
\item{\textit{Add ($\phi$,$\phi'$)}: \emph{accelerate1} and \emph{accelerate2} are added to the alternate plan;}
\item{\textit{Common ($\phi$,$\phi'$)}: \emph{decelerate} and \emph{stop};}
\item{\textit{dwell-diff}: \{$moving,\ 20$\};}
\item{\textit{diff-cost}$_{\tau}$: $-20$;}
\item{\textit{diff-cost}$_{len}$: $2$.}
\end{itemize}
A user of the framework can draw the following conclusion from the contrastive explanation:
\begin{quote}
Conclusion: Replacing \emph{decelerate} with an \emph{accelerate} action at time instant 1 provides a shorter plan in terms of the \emph{makespan} but a longer plan in terms of the number of applied actions.
\end{quote}

\subsection{Restricting an Action to Appear After a Certain Time} \label{secB}
A user might be interested in seeing the consequences of delaying the appearance of an action in a plan. In such a case, a formal question $Q$ can be:
\begin{quote}
Why did the planner choose to do an action at a particular time instant, why not later?
\end{quote}
{For example, given a plan $\phi$ which consists of an ordered sequence of ground actions $\langle a_1$, $a_2$, \ldots, $a_i$, \ldots, $a_n\rangle$, the user might ask \enquote{why the \emph{ground($a_i$)} appeared in the plan at time $t$, why not later?}.}
\noindent For the example plan in Listing 3, a question might be:
\begin{quote}
    \enquote{Why is the ground action \emph{(decelerate)} used at time 1.0, why not later?}
\end{quote}
\noindent Since the goal is to travel a distance of at least 30 units within 50 units of time, this question attempts to understand the rationale of a \emph{decelerate} action that will slow down the car, which may result in either the car taking more than 50-time units to cover the distance or the car coming to a halt before traveling a distance of 30 units.

To generate the \emph{HPlan} $\phi'$ to answer such a question, a compilation is formed such that the action \emph{ground($a_i$)} is restricted to appear after the time $t$ in a plan to be valid. Let $T_v$ be the time variable that captures the time elapsed in the system during which a plan is active. For that, we create a separate process that remains active throughout the execution of the plan and updates the value of $T_v$ using the $\#t$ literal.
A new constraint $(T_v > t)$ is included, which is used to extend the precondition of the action $a_i$, forcing it to appear only after the time $t$. The compiled \emph{HModel} $\Pi'$ is:
\begin{equation*}
    {{\Pi' = (\langle Fs', Rs, As', Es, Ps, arity'\rangle,\langle Os, I, G\rangle)}}
\end{equation*}
{where
\begin{itemize}
    \item $Fs'$ = $Fs \cup \{T_v\}$
    \item $As'$ = $\{a_{i}'\} \cup As \setminus \{a_i\}$, 
    (here, $\setminus$ represents the set difference operation).
    \item $arity'(T_v)$ = $arity'(a_{i}')$ = $arity(a_i)$
    \item $arity'(a_{i}')$ = $arity(a_i)$
\end{itemize}
where the new action $a_{i}'$ extends $a_i$ with the added precondition $(T_v > t)$, i.e.}
\begin{equation*}
    {{\textit{Pre}(a_{i}')=\textit{Pre}(a_i)\land (T_v > t)}}
\end{equation*}

\noindent{The action $a_{i}'$ represents the original action $a_i$ in the domain and should be considered as the same action. It is only restricted to appear in a plan after a certain time. The construction proposed for the \emph{HModel} is sound, and it produces an \emph{HPlan} which is expected by the user.}
\begin{lemmasec}
{The construction of the \emph{HModel} is sound.}
\end{lemmasec}

\noindent{\textsc{Proof:} A valid \emph{HPlan} must apply a ground ($a_i'$) action after a time $t$. As the action $a_i'$ has an added precondition ($T_v>t$), it can appear only after time $t$ in any plan. $\Box$}

\begin{exmp}
Consider the user question above. Let $a_i$ be the grounded \emph{decelerate} action that appeared at time $1.0$ in the plan $\phi$. To answer this question, a compilation to form the \emph{HModel} follows the above discussion, where  \emph{decelerate1} replaces the \emph{decelerate} action in the domain. The grounded \emph{decelerate1} action is constrained to appear later than the time $1.0$ in the \emph{HPlan}. The function symbol $runningTime$, which is updated by the process \emph{moving}, encodes the elapsed time in the domain. The precondition of the \emph{decelerate1} action is extended with an additional constraint $(runningTime > 1)$.
\begin{equation*}
    \textit{Pre (decelerate1)} = \textit{Pre (decelerate)} \land (runningTime > 1)  
\end{equation*}
where \textit{decelerate} $\in As$. The changes encoded in PDDL+ are shown below:
\lstset{
    escapeinside={(@}{@)}
}
\begin{lstlisting}[basicstyle=\footnotesize,caption={The modified \emph{decelerate} action with the added constraint.}]
(:action decelerate1
  :parameters()
  :precondition (and (running) (> (a) (downLimit)) (> (runningTime) 1))
  :effect (and (decrease (a) 1)))
\end{lstlisting}

\noindent The modified hybrid automaton model is shown in Figure \ref{fig:domain_B} where $runningTime$ is mapped to the variable $t$. The constraint (t $>$1) is added in the self-transition with the label \emph{decelerate1}.
\begin{figure}[htbp]
    \centering
    \includegraphics[width=0.66\textwidth]{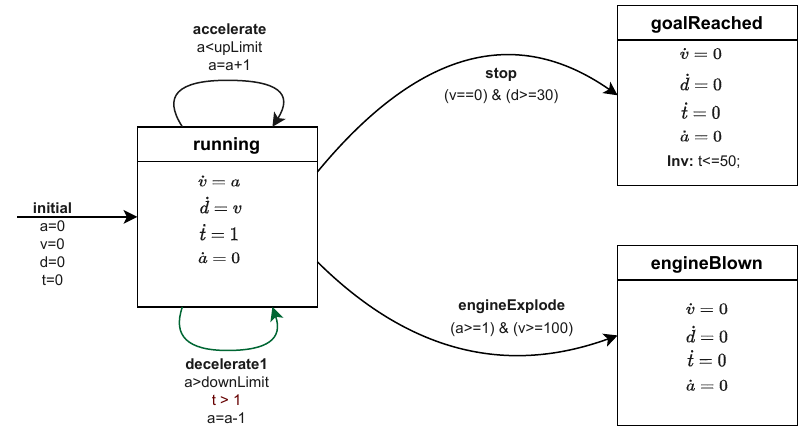}
    \caption{The hybrid automaton for the corresponding \textit{HModel} of the question is addressed in sect. 4.2.}
    \label{fig:domain_B}
\end{figure}

The resulting \emph{HPlan} given by \textsc{SMTPlan+} is:
\begin{lstlisting}[basicstyle=\footnotesize]
    (@\underline{Time}@)         (@\underline{Action}@)               (@\underline{Duration}@)
    0.0:        (accelerate)        [0.0]
    2.0:        (decelerate1)       [0.0] 
    16.0:       (decelerate1)       [0.0]
    18.0:       (stop)              [0.0]
    (@\emph{makespan}: = 18.0 units.@)
\end{lstlisting}
\end{exmp}

\noindent We observe that the plan originally synthesized by the planner is not optimal on the plan duration as the contrastive plan is of shorter \emph{makespan}.

\noindent Contrastive Explanation: The \emph{contrastive explanation} for the above user question is as follows:
\begin{itemize}
\item{\textit{Remove ($\phi$,$\phi'$)}: The \emph{decelerate} action is removed from the original plan;}
\item{\textit{Add ($\phi$,$\phi'$)}: The \emph{decelerate1} action is added to the alternate plan;}
\item{\textit{Common ($\phi$,$\phi'$)}: \emph{accelerate}, and \emph{stop};}
\item{\textit{dwell-diff}: \{$moving,\ 14$\};}
\item{\textit{diff-cost}$_{\tau}$: $-14$;}
\item{\textit{diff-cost}$_{len}$: $0$, as the original plan has same length.}
\end{itemize}

\noindent A user of the framework can draw the following conclusion from the contrastive explanation:
\begin{quote}
Conclusion: The original plan generated by the planner is not optimal with respect to the duration of the plan for the planning problem instance. The alternate plan is better than the original, having a lesser \emph{makespan}.
\end{quote}

\subsection{Restricting an Action to Appear Before a Certain Time} \label{secC}
\noindent In contrast to the question in Section~\ref{secB}, a user might be interested in hastening the appearance of an action in a plan. In such a case, for a given plan $\phi$, a formal question $Q$ can be asked of the form:
\begin{quote}
    Why did the planner choose to do an action at a particular time instant, why not earlier?
\end{quote}
{The user may be interested to know whether the goal could be achieved by choosing to do an action earlier in the plan. For example, given a plan $\phi$ consisting of ground actions $\langle a_1$, $a_2$, \ldots, $a_i$, \ldots, $a_n\rangle$, the user might ask \enquote{why the $ground(a_i)$ appears in the plan at time $t$, why not earlier?}. For the plan in Listing 3, a pertinent question can be:}
\begin{quote}
    \enquote{Why is the action \emph{(decelerate)} used at time instant 1.0, why not earlier?}
\end{quote}

\noindent {For a contrastive explanation of a question of this type, a compilation is formed such that an action $ground(a_i)$ strictly does appear before time $t$ in \emph{HPlan}. To construct the \emph{HModel}, we extend the original action $a_i$ to $a_{i1}$ while also retaining $a_{i}$ in the domain. The action $a_{i1}$ is constrained to appear in an \emph{HPlan} strictly before time $t$. We introduce a new function symbol $T_v$ to encode the system time, which is then used to form a constraint $(T_v < t)$.
Similar to the previous discussion, a process that remains active throughout the plan duration is created to update the value of $T_v$.
The precondition of the action $a_{i1}$ is extended with the constraint $(T_v < t)$, and a new predicate symbol \textit{do\_before\_t} is introduced, which is used to extend the effect of the action $a_{i1}$. The predicate is also used as a goal constraint to enforce that the action $a_{i1}$ always appears before time $t$ in a valid plan. The action $a_{i}$ acts similarly as in the original model, such that it may also be available to use later in the plan. The compiled \emph{HModel} $\Pi'$ is:}
\begin{equation*}
    {{\Pi' = (\langle Fs', Rs', As', Es, Ps, arity'\rangle,\langle Os, I, G'\rangle)}}
\end{equation*}
{where}
\begin{itemize}
    \item $Fs'$ = $Fs \cup \{T_v\}$
    \item $Rs'$ = $Rs \cup \{\textit{do\_before\_t}\}$
    \item $As'$ = $As \cup \{a_{i1}\}$
    \item $arity'(T_v)$ = $0$
    \item $arity'(\textit{do\_before\_t})$ = $0$
    \item $arity'(a_{i1})$ = $arity(a_i)$
    \item $G'$ : $G \land ground(\textit{do\_before\_t})$
\end{itemize}
where the new action $a_{i1}$ extends $a_i$ with the added precondition and effect, i.e.
\begin{equation*}
    {{\textit{Pre}(a_{i1})=\textit{Pre}(a_i)\land (T_v < t)}}
\end{equation*}
\begin{equation*}
    {{\textit{Eff}^{+}(a_{i1})=\textit{Eff}(a_i)\cup \{(\textit{do\_before\_t})\}}}
\end{equation*}

\noindent{The added action $a_{i1}$ has similar semantics as the original action $a_i$ in the domain; it is only restricted to appear in a plan before a certain time. The actions $a_{i1}$ and $a_i$ should be viewed as actions with the same interpretation except that their preconditions and effects are different. The constructed \emph{HModel} is sound and it produces an \emph{HPlan} which is expected by the user.}
\begin{lemmasec}
{The construction of \emph{HModel} is sound.}
\end{lemmasec}

\noindent{\textsc{Proof:} A valid \emph{HPlan} must apply a ground ($a_{i1}$) action before a time $t$. As the action $a_{i1}$ has an added precondition ($T_v<t$), it can be applied only before the time instance $t$ in a plan. $\Box$}

\begin{exmp}
Consider the user question above, let $a_i$ be the grounded \emph{decelerate} action that appears at time 1.0 in the plan $\phi$ in Listing 3. To answer this question, the compilation to form the \emph{HModel} follows the above discussion. A new action \emph{decelerate1} is introduced. The grounded \emph{decelerate1} action is constrained to appear earlier than the time 1.0 in an \emph{HPlan}. The function symbol $runningTime$, which is updated by the process \emph{moving}, encodes the elapsed time in the domain. The precondition of the \emph{decelerate1} action is extended with an additional constraint $(runningTime<1)$.
\begin{equation*}
    \textit{Pre}(decelerate1) = \textit{Pre}(decelerate) \land (runningTime < 1)
\end{equation*}
and the effect of \emph{decelerate1} is extended with the predicate $(\textit{do\_before\_1})$.
\begin{equation*}
    \textit{Eff}^{+}(decelerate1) = \textit{Eff}^{+}(decelerate) \cup \{(\textit{do\_before\_1})\}
\end{equation*}
where \emph{decelerate} $\in As$ and $\textit{do\_before\_1}$ is a new predicate symbol introduced. The goal is also extended with an additional constraint $(\textit{do\_before\_1})$.
The modifications in the domain are shown in Listing 6:
\begin{lstlisting}[basicstyle=\footnotesize,caption={The modified domain with the \emph{decelerate1} action.}]
(:action decelerate1
  :parameters()
  :precondition (and (running) (> (a) (downLimit)) (< (runningTime) 1))
  :effect (and (decrease (a) 1) (do_before_1)))
\end{lstlisting}

\noindent The goal state is modified as shown below:
\begin{lstlisting}[basicstyle=\footnotesize]
(:goal (and (goalReached) (not(engineBlown)) (do_before_1) 
       (<= (runningTime) 50)))
\end{lstlisting}

\begin{figure}[htbp]
    \centering
    \includegraphics[width=0.66\textwidth]{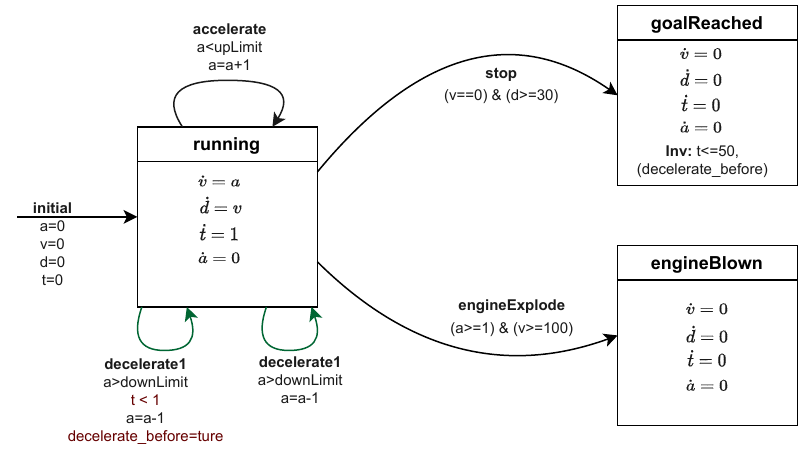}
    \caption{The hybrid automaton for the corresponding \textit{HModel} of the question is addressed in sect. 4.3.}
    \label{fig:domain_C}
\end{figure}


\noindent The model as a modified hybrid automaton is shown in Figure \ref{fig:domain_C} where the action \emph{decelerate1} is introduced. {We map $runningTime$ to the variable $t$.} The proposition $(\textit{do\_before\_1})$ is used in the goal state to force the ground action \emph{decelerate1} to appear in the plan before time $t$. The resulting \emph{HPlan} given by \textsc{SMTPlan+} is shown below:
\begin{lstlisting}[basicstyle=\footnotesize]
    (@\underline{Time}@)         (@\underline{Action}@)               (@\underline{Duration}@)
    0.0:        (accelerate)        [0.0]
    0.75:       (decelerate1)       [0.0] 
    40.75:      (decelerate)       [0.0]
    41.5:       (stop)              [0.0]
    (@\emph{makespan}: = 41.5 units.@)
\end{lstlisting}
\end{exmp}

\noindent We see that the \emph{decelerate} action now appears earlier. However, the plan is of a longer \emph{makespan}. This observation stands as an explanation of why the decelerate action was taken at time instance 1 in the original plan.

Contrastive Explanation:
\begin{itemize}
\item{\textit{Remove ($\phi$,$\phi'$)}: No action is removed;}
\item{\textit{Add ($\phi$,$\phi'$)}: \emph{decelerate1} is added;}
\item{\textit{Common ($\phi$,$\phi'$)}: \emph{accelerate}, \emph{decelerate} and \emph{stop};}
\item{\textit{dwell-diff}: \{$moving,\ -9.5$\};}
\item{\textit{diff-cost}$_{\tau}$: $9.5$;}
\item{\textit{diff-cost}$_{len}$: $0$, as the original plan has same length.}
\end{itemize}

\noindent A user of the framework can draw the following conclusion from the contrastive explanation:
\begin{quote}
    Conclusion: The application of \emph{decelerate} action later than 1-time unit results in a longer plan in terms of the \emph{makespan}.
\end{quote}

\subsection{Barring an Action from Appearing in the Plan} \label{secD}
\noindent The appearance of a certain action may seem confusing to a user at times, and barring the action from appearing in a plan may seem reasonable. To simulate such a scenario, the following contrastive question $Q$ can be asked:
\begin{quote}
    Why is an action used in the plan, rather than not being used?
\end{quote}
{For example, given a plan $\phi$ of ground actions $\langle a_1$, $a_2$, \ldots, $a_i$, \ldots, $a_n\rangle$, the user might ask \enquote{why is the action $a_i$ used in the plan, rather than not being used?}.}
To construct the \emph{HModel} to answer this question, a compilation is formed such that the action $ground(a_i)$ is barred from appearing in a generated \emph{HPlan}. To do this, dropping the action $a_i$ from the original model will suffice to serve the purpose. The compiled \emph{HModel} $\Pi'$ is:
\begin{equation*}
    {{\Pi' = (\langle Fs, Rs, As', Es, Ps, arity\rangle,\langle Os, I, G\rangle)}}
\end{equation*}
where $As'$ = $As$ $\setminus$ \{$a_i$\}, ($\setminus$ represents the set difference operation).

\noindent The constructed \emph{HModel} is sound and will produce an \emph{HPlan} which is expected by the user, if there exists any.
\begin{lemmasec}
The construction of the \emph{HModel} is sound.
\end{lemmasec}
\noindent{\textsc{Proof:} A valid \emph{HPlan} must not contain the specified action $a_i$. As the action $a_i$ is deleted from the set of available actions in the domain, the planner is bound to choose actions only from the available set of actions in a valid plan. $\Box$}

\begin{exmp}
In the plan of Listing 3, the action $a_i$ can be the \emph{decelerate} action in the domain. A user might think the action \emph{decelerate} is slowing down the car and may be detrimental to an efficient plan towards achieving the goal of traveling 30 units in a short time. The contrastive question that could be asked:
\begin{quote}
    \enquote{Why is the decelerate action used in the plan, rather than not being used?}
\end{quote}
To answer this contrastive question, the \emph{HModel} is formulated following the above discussion. We drop the action \emph{decelerate} from our domain shown in Listing 1 so that it cannot be used in the \emph{HPlan}.
The modified hybrid automaton model of the domain is shown in Figure \ref{fig:domain_D}, where the transition with the label \emph{decelerate} is now removed.
\begin{figure}[htbp]
    \centering
    \includegraphics[width=0.66\textwidth]{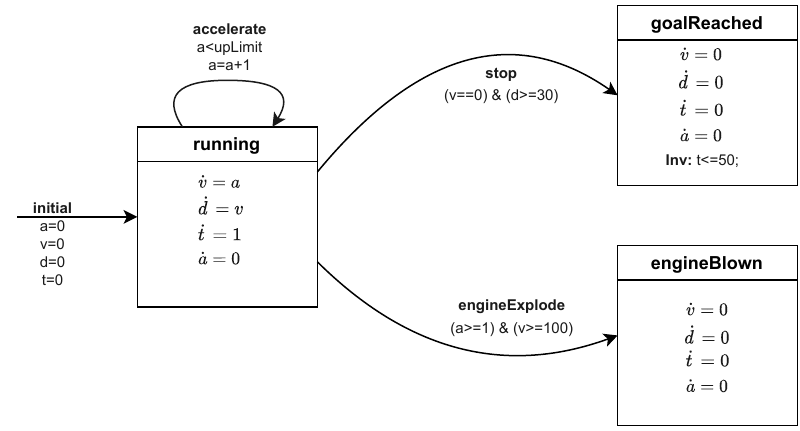}
    \caption{{The hybrid automaton for the corresponding \textit{HModel} of the question addressed in sect 4.4}}
    \label{fig:domain_D}
\end{figure}
\end{exmp}

\noindent As a result of this change in the model, the planner is \emph{unable to generate a valid plan}. The contrastive explanation is shown below:

\begin{itemize}
\item{\textit{Remove ($\phi$,$\phi'$)}: Not available, as there is no alternate plan;}
\item{\textit{Add ($\phi$,$\phi'$)}: Not available, as there is no alternate plan;}
\item{\textit{Common ($\phi$,$\phi'$)}: Not available, as there is no alternate plan;}
\item{\textit{dwell-diff}: Not available;}
\item{\textit{diff-cost}$_{\tau}$: $\infty$;}
\item{\textit{diff-cost}$_{len}$: $\infty$.}
\end{itemize}

\noindent We observe that not having the \emph{decelerate} action in the model results in no valid plan being synthesized by the planner. Hence, in the car domain, a user of the framework can draw the following conclusion from the contrastive explanation:

\begin{quote}
    Conclusion: Barring the \emph{decelerate} action from appearing in the plan leads to no valid plan in this domain.
\end{quote}

\noindent \textbf{\textit{Remark:}} The planning problem is undecidable for hybrid systems in general (\cite{ALUR19953}). When the planner cannot generate a valid plan for a planning problem, it cannot be asserted whether it is due to the underlying undecidability or that the planning problem admits no valid plan. This leads to a limitation in our explanation framework since incorrect explanations may result in such cases. To address this limitation to a certain extent, we propose a technique in Section \ref{sec4}. 

\subsection{Restricting an Action to Occur Less Than a Certain Number of Times} \label{secE}
At times, from a human perspective, it may not be trivial to figure out why an action is performed a certain number of times. Thus, it is legitimate to ask the following contrastive question:
\begin{quote}
    Why did the planner choose to do an action $n$ number of times, why not fewer?
\end{quote}
\noindent For example, given a plan $\phi$ of ground actions, a grounded action $b$ may appear multiple times in the plan. Let the action $b$ appear $n$ times in $\phi$. Then, a user might ask \enquote{why did the planner choose to do the action $ground(b)$ $n$ number of times, why not fewer?}. From our plan in Listing 3, the question can be:
\begin{quote}
    \enquote{Why is the action (\emph{decelerate}) taken twice in the plan, why not once?}
\end{quote}

\noindent {To generate the \emph{HPlan}, a compilation is formed such that the number of occurrences of the grounded action $b$ is less than $n$ in a valid plan. A new function symbol $s$ is introduced, which keeps track of the number of grounded actions $b$ in a plan, and is initially set to 0. The action $b'$ extends $b$ with an additional add effect $(s=s+1)$. The goal condition is also extended with an additional constraint $(s < n)$ such that a valid plan will always contain fewer than $n$ grounded $b$ actions. The compiled \emph{HModel} $\Pi'$ is:
\begin{equation*}
    {\Pi'=(\langle Fs', Rs, As', Es, Ps, arity'\rangle,\langle Os, I', G'\rangle)}
\end{equation*}
where
\begin{itemize}
\item {$Fs'$ = $Fs$ $\cup$ \{$s$\}}
\item {$As'$ = $\{b'\} \cup As \setminus \{b\}$, (here, $\setminus$ represents the set difference operation)}
\item {$arity'(s)$ = $0$}
\item {$arity'(b)$ = $arity(b)$}
\item {$I'$ = $I \cup \{\textit{PNE}(s=0)$, $s\in Fs'\}$}
\item {$G'$ = $G \land (s<n)$}
\end{itemize}
where $b'$ extends $b$ with an additional add effect, i.e.
\begin{equation*}
    {{\textit{Eff}^{+}(b')=\textit{Eff}(b) \cup \{(s=s+1)\}}}
\end{equation*}
}


\noindent{The action $b'$ represents the original action $b$ in the domain; it is only restricted to appear less than a specified number of times in a plan. The constructed \emph{HModel} is sound and will always produce an \emph{HPlan} where $b$ appears less than $n$ times, if any such plan exists.}
\begin{lemmasec}
{The construction of the \emph{HModel} is sound.}
\end{lemmasec}
\noindent{\textsc{Proof:} In a valid \emph{HPlan} the $ground (b')$ action must appear less than $n$ times. On each occurrence of the action $b'$ in a plan, the added effect of $b'$ increases the value of the function symbol $s$ by 1, which is initially set to 0. Hence, the added goal constraint ($s<n$) ensures that only those plans are generated where $b'$ has appeared less than $n$ times. $\Box$}

\begin{exmp}
\noindent Consider the user question above, where $b$ is the \emph{decelerate} action. The compilation to form the \emph{HModel} follows the above discussion. A new function symbol $s$ is introduced, whose initial value is set to 0 in the initial state. Let $n$ be two here, which implies that the appearance of \emph{decelerate} action should be less than twice in a plan. The action \emph{decelerate} is modified to \emph{decelerate1}, which is allowed to occur only once in the plan.
The action \emph{decelerate1} now has an added effect to increase the value of $s$, together with the other effects of the \emph{decelerate} action in the original domain:
\begin{itemize}
    \item {\textit{Pre} (\emph{decelerate1}) = \textit{Pre} (\emph{decelerate})}
    \item {\textit{Eff}$^{+}$ (\emph{decelerate1}) = \textit{Eff}(\emph{decelerate}) $\cup$ \{$(s=s+1)$, $s\in$ \textit{Fs}\}}.
\end{itemize}
Additionally, a constraint $(s<2)$ is added in the goal state such that the action \emph{decelerate1} appears only once in a plan. The changes made in the domain are shown below:
\begin{lstlisting}[basicstyle=\footnotesize]
(:functions (d) (v) (a) (upLimit) (downLimit) 
            (runningTime) (s))
    ...
(:action decelerate1
  :parameters()
  :precondition (and (running) (> (a) (downLimit)))
  :effect (and (decrease (a) 1) (increase (s) 1)))
\end{lstlisting}
The constraint $(s<2)$ is added in the goal state as shown below:
\begin{lstlisting}[basicstyle=\footnotesize]
(:goal (and (goalReached) (not(engineBlown)) (<= (runningTime) (< (s) 2)))
\end{lstlisting}
\noindent Figure \ref{fig:domain_E} represents the relevant modifications in the hybrid automaton representation of the domain.
\begin{figure}[htbp]
    \centering
    \includegraphics[width=0.66\textwidth]{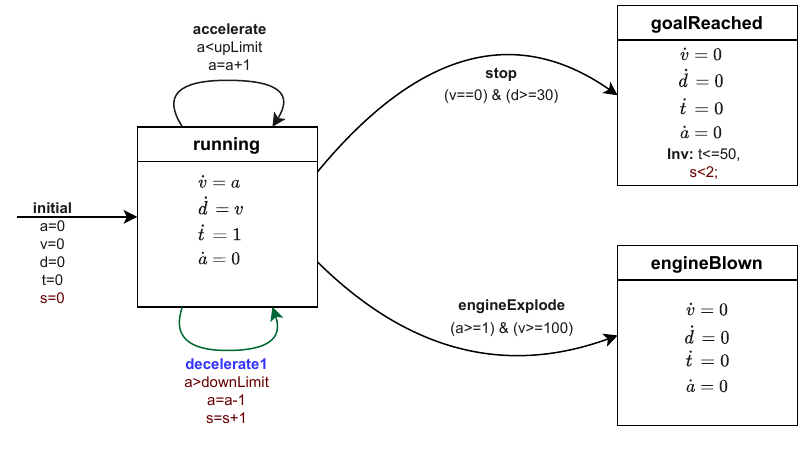}
    \caption{{The hybrid automaton for the corresponding \textit{HModel} of the question addressed in sect. 4.5.}}
    \label{fig:domain_E}
\end{figure}
\end{exmp}

\noindent As a result of these modifications, no valid plan can be generated by the planner. The contrastive explanation is:
\begin{itemize}
\item{\textit{Remove ($\phi$,$\phi'$)}: Not available, as there is no alternate plan;}
\item{\textit{Add ($\phi$,$\phi'$)}: Not available, as there is no alternate plan;}
\item{\textit{Common ($\phi$,$\phi'$)}: Not available, as there is no alternate plan;}
\item{\textit{dwell-diff}: Not available;}
\item{\textit{diff-cost}$_{\tau}$: $\infty$;}
\item{\textit{diff-cost}$_{len}$: $\infty$.}
\end{itemize}

\noindent A user of the framework can draw the following conclusion from the contrastive explanation:
\begin{quote}
    Conclusion: In this domain, no valid plan can be generated that has fewer than two executions of the action \emph{decelerate}.
\end{quote}

\subsection{Questioning the Optimality of Plan Duration} \label{secF}
A user might want to know whether the observed plan is optimal in terms of duration or whether there exists any other plan with a shorter plan-duration. The following contrastive question can be considered:
\begin{quote}
  Why is the \emph{makespan} of the plan not less?
\end{quote}
Recall that \emph{makespan} is the duration of a plan.
\noindent As we have seen earlier, the planner may not provide an optimal plan. Therefore, an iterative question \enquote{Why the \emph{makespan} is $t$ to complete a task, and not less?} where $t$ is the \emph{makespan} of the generated plan in the last successful iteration, can help in questioning the optimality of the plan. In our car domain, a user might ask:
\begin{quote}
    \enquote{Why is the \emph{makespan} of the plan not less than 32?}
\end{quote}

\noindent For such a question, we propose an iterative \textit{HModel} formation that addresses a sequence of user questions.
A user may interact iteratively in this framework and can successively view generated \textit{HPlans} and seek counter explanations by imposing additional constraints on the planning problem. The collection of \textit{HModels} that can be successively built forms a hierarchical structure rooted at the original model and extended by the incremental addition of new constraint properties (Definition \ref{def:const-prop}).
{An \emph{iterative Hmodel formation} is defined as follows.
}
\begin{definition}
{\textbf{Iterative HModel Formation}: For a planning instance $\Pi$ and a plan $\phi$, let $\psi_i$ be the set of user imposed constraint properties derived from $\phi_{i-1}$ which is initially empty, i.e. $\psi_0 = \emptyset$. Each stage $i$ (initially 0) of this process starts with the planner producing a \textit{HPlan} $\phi_i^{'}$ for the \textit{HModel} $\Pi_i^{'} = \Pi \times \psi_i$, where $\times$ is the constraint operator (Definition \ref{def:const-op}) which encapsulates $\psi_i$ in $\Pi_i^{'}$. $\Box$}
\end{definition}

{Algorithm \ref{algo:algo6} constructs the iterative \textit{HModel} for the iterative user queries.}
\begin{algorithm}[htpb]
  \scriptsize
  \caption{\textsf{Iterative HModel Construction}}
  \label{algo:algo6}
  \LinesNumbered
  \SetAlgoLined
  \KwIn{The \emph{HModel} $\Pi_{i-1}'$, and the \emph{HPlan} $\phi_{i-1}'$ where $\Pi_0'$ and $\phi_0'$ are the original model $\Pi$ and the original plan $\phi$ correspondingly;\\}
  \KwOut{The \emph{HModel} $\Pi_i'$}
  \Begin{
   $\times$: The constraint operator;\\
   $i$ = 1;\\
   \While{true}{
    \eIf{($\phi_{i-1}'$ $\neq$ \textit{no-plan})}{
        $\Pi_i'$ = $\Pi_{i-1}' \times \psi_i(\phi_{i-1}')$ \tcc*{$\psi_i(\phi_{i-1}')$ is the user imposed constraint from $\phi_{i-1}'$}
        $i=i+1$;\\
    }
    {return \textit{HModel} cannot be constructed.}
   }
  }
\end{algorithm}

\begin{exmp}
Consider the user's question on the optimality of the makespan of a plan. The constraint $\psi_i$ can be updated from a valid plan HPlan $\phi_{i-1}'$ generated from the previous iteration, as shown below.
\begin{align*}
    if (\phi_{i-1}'\neq\ & \textit{no-plan})\ \{
     \psi_i = (runningTime < makespan(\phi_{i-1}'))
    \}
\end{align*}
where \emph{makespan}($\phi_{i-1}'$) denotes the duration of the plan $\phi_{i-1}'$.
For example, in the original problem (see Listing 2), the constraint $\psi$ = ($runningTime\leq50$) leads to a plan of a duration of 32.0 (see Listing 3). So in the next iteration, the constraint $\psi_1$ in $\Pi_1'$ is modified to $runningTime<32$ to find a plan with a duration of less than 32.0-time units to check if any such plan exists. This iterative process is continued till we can conclude that no plan exists for the given $runningTime$ constraint. The iterative \emph{HModel} $\Pi_i'$ is given below:
\begin{equation} \label{eq:rec}
    \Pi_i'=\Pi'_{i-1} \times \{\psi_i(\phi'_{i-1})\}
\end{equation}
where ${\Pi_{i}'}$ is the \emph{HModel} formed in the $i$-th iteration, $\phi_{i-1}'$ is the \emph{HPlan} generated on ${\Pi_{i-1}'}$ and  $\psi_i(\phi_{i-1}')$ defines a constraint generated by observing $\phi_{i-1}'$. The new $\Pi_i'$ is formed from ${\Pi_{i-1}'}$ by imposing the constraint $\psi_i$. The constraint-operator $\times$ is defined in Definition \ref{def:const-op}.

\noindent{The constructed iterative \emph{HModel} is sound and will produce in each iteration a valid \emph{HPlan} with shorter \emph{makespan} than the \emph{HPlan} of the previous round.}
\begin{lemmasec}
{The construction of the iterative \emph{HModel} is sound.}
\end{lemmasec}
\noindent{\textsc{Proof:} In each iteration a valid \emph{HPlan} must have a shorter \emph{makespan} than the \emph{HPlan} of the previous round. The goal constraint in the $i$-th iteration is $\psi_i$ which imposes the constraint that the $runningTime$ of $\phi_i'$ should be less than the \emph{makespan} of $\phi_{i-1}'$ ($runningTime$ $<$ \emph{makespan}($\phi_{i-1}'$)). Thus, it ensures that $\phi_i'$ has a shorter \emph{makespan} than $\phi_{i-1}'$. $\Box$}

In Table \ref{ch2tab:table1}, we present the results in which \emph{makespan} is the duration of the plan, \emph{constr} is the upper-bound on the constraint on $runningTime$ for each iteration, whereas \emph{dwell-diff} and \emph{{diff-cost}$_{\tau}$} are the components of contrastive\ explanation.
\begin{table}[htbp]
    \centering
    \begin{tabular}{|c|c|c|c|c|}
    \hline
         Round & \emph{constr} & \emph{makespan} & \emph{dwell-diff} & \emph{diff-cost}$_{\tau}$\\
    \hline
         1 & $\leq 50$ & 32 & $-$ & $-$\\
         2 & $< 32$ & 31.5 & \{$moving,\ 0.5$\} & $-0.5$\\
         3 & $< 31$ & 18 & \{$moving,\ 13.5$\} & $-13.5$\\
         4 & $< 18$ & 17.5 & \{$moving,\ 0.5$\} & $-0.5$\\
         5 & $< 17$ & 14 & \{$moving,\ 3.5$\} & $-3.5$\\
         6 & $< 14$ & 13.5 & \{$moving,\ 0.5$\} & $-0.5$\\
         7 & $< 13$ & 12 & \{$moving,\ 1.5$\} & $-1.5$\\
         8 & $< 12$ & 11.75 & \{$moving,\ 0.25$\} & $-0.25$\\
         9 & $< 11$ & 10.96 & \{$moving,\ 0.79$\} & $-0.79$\\
        10 & $< 10.96$ & 10.95 & \{$moving,\ 0.01$\} & $-0.01$\\
         11 & $< 10.95$ & $\infty$ & NA & $\infty$\\
    \hline 
    \end{tabular}
    \caption{The \emph{makespan} of the plan for each re-plan iteration}
    \label{ch2tab:table1}
\end{table}
\end{exmp}

\noindent We can see in the table that for $runningTime~\leq~10.95$, the planner is unable to find a valid plan for the planning instance. Therefore, via the contrastive questions, a user may improve the quality of the plan in terms of the \emph{makespan} metric. The contrastive explanation is:

\begin{itemize}
\item{\textit{Remove ($\phi$,$\phi'$)}: No action is removed from the original plan;}
\item{\textit{Add ($\phi$,$\phi'$)}: No new action is added in each iteration;}
\item{\textit{Common ($\phi$,$\phi'$)}: \emph{accelerate}, \emph{decelerate} and \emph{stop};}
\item {The \textit{dwell-diff} and \textit{diff-cost}$_{\tau}$ of \emph{HModel} $\Pi'$ for each iteration are given in Table \ref{ch2tab:table1}}
\item{The \textit{diff-cost}$_{len}$ for each iteration is 0 as all alternate plans are of the same length except for the no-plan where the length is denoted by $\infty$.}
\end{itemize}

\noindent A user of the framework can draw the following conclusion from the contrastive explanation:
\begin{quote}
    Conclusion: The \emph{makespan} of the original plan is not optimal, and there can be alternate plans with lesser \emph{makespan}.
\end{quote}

\subsection{Question on the Sequence of Actions in the Plan} \label{secG}
A certain sequence of actions in a plan may seem costly to a user, or the user might have a preference in mind for a sequence of actions, and the absence of that sequence in a plan may give rise to the question of:
\begin{quote}
 {Why this sequence of actions, why not any other instead?}
\end{quote}


\noindent An action sequence is defined in Definition \ref{def:ASeq}. To address such user questions on a sequence of actions appearing in a plan, the following contrastive question can be asked: 
\begin{quote}
  Why does the action sequence $a_i-a_{i+1}-\ldots-a_{i+k}$ appear in the plan $\langle a_1-a_2-\ldots-a_{i-1}-a_{i}-a_{i+1}-\ldots-a_{i+k}-a_{i+k+1}-\ldots-a_m\rangle$, why not some other sequence?
\end{quote}

\noindent In the plan in Listing 3, a question can be asked:
\begin{quote}
  \enquote{Why does the action sequence \emph{decelerate-decelerate} appear in the plan $\langle$\emph{accelerate-decelerate-decelerate-stop}$\rangle$, why not some other sequence?}
\end{quote}

\noindent To explain this question, we intend to design a hypothetical model such that any hypothetical plan in the model preserves the action sequence that appears before and after the sequence in the question in the original plan. We refer to these as \textbf{\emph{pre-sequence}} and \textbf{\emph{post-sequence}} respectively.  For example, in the question above, the \textit{pre-sequence} $a_1-a_2-\ldots-a_{i-1}$ and the \textit{post-sequence} $a_{i+k+1}-a_{i+k+2}-\ldots-a_m$ are needed to be preserved in the hypothetical plan. In addition, we need to ensure that the action sequence in the question $a_{i}-a_{i+1}-\ldots-a_{i+k}$ does not appear in the hypothetical plan. In the following section, we first discuss a general approach to sequence modification and then come back to our original question in the car domain. Note that our proposed solution introduces primed actions where necessary.
For example, $a$ is an action in the domain, $a'$ is a primed action newly introduced into the \emph{HModel}.
These primed actions have the same interpretation as their unprimed counterparts, except that they may have different preconditions and effects.

\noindent \textbf{Preserving \textit{pre-sequence}:} {To preserve the \textit{pre-sequence} in the plan, we treat each action that appears in the \textit{pre-sequence} as an independent action that can only appear in the \textit{pre-sequence} of a plan. We modify each action that appears in the sequence $a_1-a_2-\ldots-a_{i-1}$ in the following way:}

\noindent For each action $a_j$, $\forall j \in \{1, 2, \ldots, i-1\}$, a new action $a_j'$ is constructed while also retaining the $a_j$ in the \emph{HModel}. The action $a_j'$ will appear only in the \textit{pre-sequence} while the action $a_j$ may appear later in a plan.
\textcolor{black}{For each action $a_j'$ ($j' \in \{1,2,\ldots,i-1\}$), new predicate symbols $has\_done\_a_j'$ are used to preserve their order of appearance in the plan.}
A predicate symbol $has\_done\_preseq$ is introduced, which represents that the \textit{pre-sequence} has been applied to the plan.
\textcolor{black}{The action $a_1$ is extended to $a_1'$ with an added precondition $\neg(has\_done\_a_1')$ and an added effect $has\_done\_a_1'$, i.e.}
\begin{equation*}
    \textcolor{black}{{\textit{Pre}(a_1')=\textit{Pre}(a_1) \land \neg(has\_done\_a_1')}}
\end{equation*}
\begin{equation*}
    \textcolor{black}{{\textit{Eff}^{+}(a_1')=\textit{Eff}(a_1) \cup \{(has\_done\_a_1')\}}}
\end{equation*}
\textcolor{black}{For each action $a_j$ ($j \in \{2,3,\dots,i-1\}$), $a_j$ is extended to $a_j'$ with added precondition $has\_done\_a_{j-1}'$ and $\neg(has\_done\_a_j')$ and an added effect $has\_done\_a_j'$. This ensures that the actions of the \textit{pre-sequence} appear in order.}
\begin{equation*}
    \textcolor{black}{{\textit{Pre}(a_j')=\textit{Pre}(a_j) \land (has\_done\_a_{j-1}') \land \neg(has\_done\_a_j')}}
\end{equation*}
\begin{equation*}
    \textcolor{black}{{\textit{Eff}^{+}(a_j')=\textit{Eff}(a_j) \cup \{(has\_done\_a_j')\}}}
\end{equation*}
\textcolor{black}{Additionally, the action $a_{i-1}'$ has an added effect $has\_done\_preseq$.}
\begin{equation*}
    \textcolor{black}{{\textit{Eff}^{+}(a_{i-1}')=\textit{Eff}(a_{i-1}) \cup \{(has\_done\_a_{i-1}'), (has\_done\_preseq)\}}}
\end{equation*}

\noindent {All other domain actions are modified with an added precondition \textit{has\_done\_preseq} such that they may appear in a plan only after the \textit{pre-sequence}.
$has\_done\_preseq$ is also used as a goal constraint to ensure the \textit{pre-sequence} is preserved in a plan.}
\\
\\
\noindent \textbf{Restricting the \textit{sequence} in the question:}
{To restrict the action sequence $a_{i}-a_{i+1}-\ldots-a_{i+k}$ in a plan, we modify each action that appears in the sequence in the following way:}

\noindent We introduce a new function symbol $c$ that represents a counter indicating how much of the forbidden sequence has been seen so far. It is set to -1 initially.
\textcolor{black}{For the first action $a_i$ in the sequence, we create two actions $a_i'$ and $a_i''$. $a_i'$ is extended with added preconditions $has\_done\_preseq$ and $c=-1$ indicating that it can appear only at the beginning of a sequence after the \textit{pre-sequence} has been applied in the plan and an added effect to assign $c$ to the value 1 to indicate the first action of the forbidden sequence has appeared in the plan.}
\begin{equation*}
    \textcolor{black}{{\textit{Pre}(a_{i}')=\textit{Pre}(a_{i}) \land (has\_done\_preseq) \land (c==-1)}}
\end{equation*}
\begin{equation*}
    {{\textit{Eff}^{+}(a_{i}')=\textit{Eff}(a_{i}) \cup \{(c=1)\}}}
\end{equation*}
\textcolor{black}{$a_i''$ can appear at any other place in the sequence when appearing out of order. $a_i''$ is extended with added preconditions $has\_done\_preseq$ and $c\neq-1$ to indicate that it can appear after the \textit{pre-sequence} has been applied in the plan and an added effect to assign $c$ to the value $0$ to indicate that it is appearing out of order and hence the forbidden sequence is broken.}
\begin{equation*}
    \textcolor{black}{{\textit{Pre}(a_{i}'')=\textit{Pre}(a_{i}) \land (has\_done\_preseq) \land (c\neq-1)}}
\end{equation*}
\begin{equation*}
    \textcolor{black}{{\textit{Eff}^{+}(a_{i}'')=\textit{Eff}(a_{i}) \cup \{(c=0)\}}}
\end{equation*}


\noindent For $\forall j$ in ${1,2,\ldots,k-1}$, each $a_{i+j}$ is flattened to $a_{i+j}'$ and $a_{i+j}''$. $a_{i+j}'$ has been extended with added preconditions $has\_done\_preseq$ and $(c=j)$ and an added effect $(c=j+1)$. This is how the consecutive appearance of the actions in the forbidden sequence are remembered.
$a_{i+j}''$ has been extended with added preconditions $has\_done\_preseq$ and $(c\neq j)$, and an added effect $(c=0)$ which resets the counter.
The last action in the sequence $a_{i+k}$ is barred from appearing in a plan when all previous actions of the forbidden sequence have appeared in order. $a_{i+k}$ is modified to $a_{i+k}'$ with added preconditions $has\_done\_preseq$ and $(c\neq k)$ \textcolor{black}{ to ensure that it can appear only out of order, and an added effect $(c=0)$ to indicate that the forbidden sequence has not appeared in the plan}. The modifications are shown below:

\begin{itemize} 
    \item \textit{Pre}$(a_{i+j}')$ = \textit{Pre}$(a_{i+j})\land $has\_done\_preseq$ \land (c==j)$,
    \item \textit{Eff}$^{+}(a_{i+j}')$ = \textit{Eff}$(a_{i+j})\cup \{(c=j+1)\}$,
    \item \textit{Pre}$(a_{i+j}'')$ = \textit{Pre}$(a_{i+j})\land (c\neq j)\land (has\_done\_preseq)$,
    \item \textit{Eff}$^{+}(a_{i+j}'')$ = \textit{Eff}$(a_{i+j})\cup \{(c=0)\}$,
    \item \textit{Pre}$(a_{i+k}')$ = \textit{Pre}$(a_{i+k})\land (c\neq k)\land (has\_done\_preseq)$,
    \item \textit{Eff}$^{+}(a_{i+k}')$ = \textit{Eff}$(a_{i+j})\cup \{(c=0)\}$. 
\end{itemize}
The rest of the actions in the domain are modified to include an added effect $(c=0)$ to indicate that the forbidden sequence has not appeared in the plan.
\\
\\
\noindent \textbf{Preserving \textit{post-sequence}:} {Similar to the \textit{pre-sequence}, the \textit{post-sequence} also needs to be preserved in a plan, and we treat each action that appears in the \textit{post-sequence} as an independent action that can appear only at the \textit{post-sequence} in a plan.}
We modify each action that appears in the sequence $a_{i+k+1}-a_{i+k+2}-\ldots-a_m$ as given below:

For each action $a_j$, $\forall j \in \{i+k+1,i+k+2,\ldots,m\}$, a new action $a_j'$ is constructed while also retaining the $a_j$ in the \emph{HModel}. {The action $a_j'$ will appear only in the \textit{post-sequence} while the action $a_j$ may appear elsewhere in a plan.}
\textcolor{black}{A new predicate symbol $start\_postsequence$ is introduced, which is used to mark the beginning of the \textit{post-sequence} in a plan.}
\textcolor{black}{For each action $a_j'$ ($j' \in \{i+k+1,i+k+2,\ldots,m\}$), a new predicate symbol $has\_done\_a_j'$ is introduced to ensure their order of appearance in the plan. Finally, another predicate symbol $has\_done\_postsequence$ is introduced, which represents that the \textit{post-sequence} has been applied to the plan.}
\textcolor{black}{The first action $a_{i+k+1}$ in the \textit{post-sequence} is extended to $a_{i+k+1}'$ with added preconditions $has\_done\_preseq$, $\neg(has\_done\_a_{i+k+1}')$ and $\neg(start\_postsequence)$, and with added effects $has\_done\_a_{i+k+1}'$ and $start\_postsequence$. $has\_done\_preseq$ ensures that it can appear only after the \textit{pre-sequence}, while $start\_postsequence$ marks the beginning of the \textit{post-sequence}. The duality of the precondition $\neg(has\_done\_a_{i+k+1}')$ and the effect $has\_done\_a_{i+k+1}'$ prevents it from further repetitions in a plan.}
\begin{equation*}
    \begin{split}
        \textit{Pre}\ (a_{i+k+1}')= & \textit{Pre}\ (a_{i+k+1}) \land (has\_done\_preseq) \land \neg(start\_postsequence) \land \\
        & \neg(has\_done\_i_{i+k+1}')
    \end{split}
\end{equation*}

\begin{equation*}
    \textit{Eff}\ ^{+}(a_{i+k+1}')=\textit{Eff}\ (a_{i+k+1}) \cup \{(has\_done\_a_{i+k+1}'), (start\_postsequence)\}
\end{equation*}
\noindent \textcolor{black}{For all $j$ in $\{i+k+2,i+k+3,\ldots,m\}$, $a_j$ is extended to $a_j'$ with added preconditions $(has\_done\_a_{j-1}')$ and $\neg(has\_done\_a_j')$ and an added effect $(has\_done\_a_j')$, i.e.}
\begin{equation*}
    \textcolor{black}{\textit{Pre}\ (a_{j}')=\textit{Pre}\ (a_{j}) \land (has\_done\_a_{j-1}') \land \neg(has\_done\_a_j')}
\end{equation*}
\begin{equation*}
    \textcolor{black}{\textit{Eff}\ ^{+}(a_{j}')=\textit{Eff}\ (a_{j}) \cup \{(has\_done\_a_{j}')\}}
\end{equation*}
The action $a_m$ is extended to $a_m'$ with an added effect $has\_done\_postsequence$, i.e.
\begin{equation*}
    {{\textit{Eff}\ ^{+}(a_{m}')=\textit{Eff}\ (a_{m}) \cup \{(has\_done\_a_m'),(has\_done\_postsequence)\}}}
\end{equation*}
To enforce that the \textit{post-sequence} is preserved in a plan, the constraint $(has\_done\_postsequence)$ is also used as a goal condition. 
\textcolor{black}{All other actions in the domain are extended with added precondition $\neg(start\_postsequence)$ so that they cannot be applied after the \textit{post-sequence} begins in a plan.}
The \emph{HModel} $\Pi'$ is:
\begin{equation*}
    {{\Pi' = (\langle Fs', Rs', As', Es, Ps, arity'\rangle,\langle Os, I', G'\rangle)}}
\end{equation*}
{where
\begin{itemize}
    \item $Fs'$ = $Fs \cup \{c\}$
    \item $Rs'$ = $Rs$ $\cup$ \{$has\_done\_preseq$, $has\_done\_postsequence$, \textcolor{black}{$start\_postsequence$},\\
    \{$has\_done\_a_j'$, $\forall j'$ $\in$ \{$1,2,\ldots,i$\} $\cup$ \{$i+k+1,i+k+2,\ldots,m$\}\}\}
    \item $As'$ = \{$a_{1}', a_{2}',\ldots, a_{i}', a_{i+1}', a_{i+1}'',\ldots, a_{i+k-1}', a_{i+k-1}'', a_{i+k}', a_{i+k+1}',\ldots, a_m'$\}\\
    $\cup$ $As$ $\setminus$ \{$a_i, a_{i+1}, \ldots, a_{i+k}$\}, (here, $\setminus$ represents the set difference operation)
    \item $arity'(As')$ = $arity(As)$
    \item $I'$ = $I$ $\cup$ \{($c=-1$)\}
    \item $G'$ : $G$ $\land$ $ground(has\_done\_preseq)$ $\land$ $ground(has\_done\_postsequence)$
\end{itemize}
}


\noindent{The constructed \emph{HModel} of restricting an action sequence in a plan is sound and will produce an \emph{HPlan} as expected by the user.}
\begin{lemmasec}
{The construction of the \emph{HModel} is sound.}
\end{lemmasec}

\noindent \textsc{Proof:} A valid \emph{HPlan} must preserve the \textit{pre-sequence} and the \textit{post-sequence} while there can be an arbitrary action sequence in between them in a plan. The \textit{pre-sequence} and the \textit{post-sequence} are preserved in a valid \textit{HPlan} through the use of the goal constraints $has\_done\_preseq$ and $has\_done\_postsequence$. A sequence of actions other than the forbidden sequence may appear in between the \textit{pre-sequence} and the \textit{post-sequence}.
In the \textit{pre-sequence}, each action is treated uniquely, and necessary preconditions and effects are added to bar their repeated occurrences in the pre- and post-sequences. While the primed actions will appear only in the \textit{pre-sequence}, their unprimed counterparts may appear later in the plan.
The forbidden sequence without a prefix or suffix cannot appear in a valid plan, as we remember the actions in their order of occurrence. If a sequence matches all the actions of the forbidden sequence except the last one, we forbid the last action of the forbidden sequence to appear in the plan. The primed actions of that section appear in the plan when they appear in the same order as the forbidden sequence and hence increase the counter, whereas the double-primed counterparts appear when they break the sequence. Whenever an action appears out of order, it erases the counter to indicate that the forbidden sequence is broken.
Similar to the \textit{pre-sequence}, each action in \textit{post-sequence} is treated uniquely. The primed actions will appear only in the \textit{post-sequence}, whereas their unprimed counterparts may appear elsewhere in the plan. $\Box$
\begin{exmp}
\noindent Considering the user question on action sequence in our car domain shown above, we have the following constructs:
\begin{itemize}
    \item \textit{pre-sequence}: \emph{accelerate}
    \item \textit{forbidden-sequence}: \emph{decelerate-decelerate}
    \item \textit{post-sequence}: \emph{stop}
\end{itemize}
To preserve the \textit{pre-sequence} in a plan following the above discussion, we introduce a new action \emph{accelerate'} which extends the \emph{accelerate} action while also retaining it in the \emph{HModel}. A new predicate symbol $has\_done\_preseq$, which represents the \textit{pre-sequence}, is added in the planning instance.
As our \textit{pre-sequence} consists of only one grounded \emph{accelerate} action, we do not need any additional predicate symbol to maintain the order.
The accelerate' action is extended with an \textcolor{black}{added precondition $\neg(has\_done\_preseq)$} and an added effect $(has\_done\_preseq)$.
\begin{equation*}
    \textcolor{black}{\textit{Pre}\ (accelerate')=\textit{Pre}\ (accelerate) \land \neg(has\_done\_preseq)}
\end{equation*}
\begin{equation*}
    {\textit{Eff}\ ^{+}(accelerate')=\textit{Eff}\ (accelerate) \cup \{(has\_done\_preseq)\}}
\end{equation*}
The constraint $(has\_done\_preseq)$ is also added as a goal condition which enforces the \textit{pre-sequence} into a plan.
All other domain actions include a precondition ($has\_done\_preseq$) such that they can appear in a plan only after the \textit{pre-sequence}. To restrict the appearance of the \textit{forbidden-sequence} \emph{decelerate}-\emph{decelerate} in a plan, we modify the actions of the domain. The first action in the sequence \emph{decelerate} is modified to \emph{decelerate'}, which can only appear at the start of a sequence that succeeds the \textit{pre-sequence}. We introduce a new function symbol $c$ used as a counter variable, which is initially set to -$1$. The \emph{decelerate'} action has added preconditions ($has\_done\_preseq$) and ($c=-1$) where ($has\_done\_preseq$) represents it can appear in a plan only after the \textit{pre-sequence} and a negative value of $c$ suggests the forbidden sequence has not started yet, respectively, and an added effect which assign $c$ to the value 1.
We further modify the \emph{decelerate} action to \emph{decelerate''} which includes preconditions ($has\_done\_preseq$), ($c\neq1$), and \textcolor{black}{($c\neq-1$)}, to indicate that it appears after the \textit{pre-sequence} and not preceding a \emph{decelerate'} action respectively, and an added effect $c=0$ indicates that the forbidden sequence is broken.
The rest of the actions in the domain, \emph{accelerate} and \emph{stop}, are modified to include an added precondition ($has\_done\_preseq$) such that they may appear only after the \textit{pre-sequence} and an added effect ($c=0$) to indicate the forbidden \textit{sequence} has not appeared in the plan. The modifications are shown below:

\begin{itemize}
    \item \textit{Pre} $(decelerate')$ = \textit{Pre} $(decelerate)\land (has\_done\_preseq)\land (c==-1)$,
    \item \textit{Eff} $^{+}(decelerate')$ = \textit{Eff} $(decelerate)\cup \{(c=1)\}$,
    \item \textit{Pre} $(decelerate'')$ = \textit{Pre} $(decelerate)\land (has\_done\_preseq)\land (c\neq1) \land \textcolor{black}{(c\neq-1)}$,
    \item \textit{Eff} $^{+}(decelerate'')$ = \textit{Eff} $(decelerate)\cup \{(c=0)\}$,   
    \item \textit{Pre} $(accelerate)$ = \textit{Pre} $(accelerate)\land (has\_done\_preseq)$,
    \item \textit{Pre} $(stop)$ = \textit{Pre} $(stop)\land (has\_done\_preseq)$,
    \item \textit{Eff} $^{+}(accelerate)$ = \textit{Eff} $(accelerate)\cup \{(c=0)\}$,
    \item \textit{Eff} $^{+}(stop)$ = \textit{Eff} $(stop)\cup \{(c=0)\}$.
\end{itemize}
To preserve the \textit{post-sequence} in a plan, a new predicate symbol $has\_done\_postsequence$ is introduced, which represents the \textit{post-sequence} that has been applied to the plan. As the \textit{post-sequence} consists of only one grounded \emph{stop} action, we do not need any additional predicate symbol to maintain the order. The original \emph{stop} action in the domain is extended to \emph{stop'}. 
{The added preconditions ($has\_done\_preseq$) and $\neg(has\_done\_postsequence)$ ensure that the action \emph{stop'} can be applied in a plan only after the \textit{pre-sequence} and it does not repeat itself, respectively.}
The added effect $(has\_done\_postsequence)$ ensures that the \textit{post-sequence} always appears in a plan, i.e.
\begin{equation*}
    \textcolor{black}{{\textit{Pre}\ (stop')=\textit{Pre}\ (stop) \land (has\_done\_preseq) \land \neg(has\_done\_postsequence)}}
\end{equation*}
\begin{equation*}
    {{\textit{Eff}\ ^{+}(stop')=\textit{Eff}\ (stop) \cup \{(has\_done\_postsequence)\}}}
\end{equation*}
The constraint $(has\_done\_postsequence)$ is used as a goal condition to enforce the appearance of the \textit{post-sequence} in a plan.
\textcolor{black}{All other actions in the domain are extended with an additional precondition $\neg(has\_done\_postsequence)$ to prevent them from appearing after the \textit{postsequence} in a plan.}
The modifications in the domain are shown below:
\begin{itemize}
    \item \textit{Fs'} = \textit{Fs} $\cup\ \{c\}$,
    \item \textit{Rs'} = \textit{Rs} $\cup \{has\_done\_preseq, has\_done\_postsequence\}$
    \item \textit{As'} = $\{accelerate', decelerate',  decelerate'', stop'\} \cup$ \textit{As} $\setminus \{decelerate\}$,
    \item \textit{arity'(As')} = \textit{arity(As)},
    \item \textit{I'} = \textit{I} $\cup$ \{\textit{PNE} $(c=-1)$\},
    \item \textit{G'} = \textit{G} $\land (has\_done\_preseq) \land (has\_done\_postsequence)$.
\end{itemize}
The modifications in the domain are shown below:
\begin{lstlisting}[basicstyle=\footnotesize]
(:predicates (running) (engineBlown) (goalReached) (has_done_presequence)
             (has_done_postsequence)
(:functions (d) (v) (a) (upLimit) (downLimit) (runningTime) (c))
    ...
(:action accelerate
  :parameters()
  :precondition (and (running) (< (a) (upLimit)) (has_done_presequence)
            (not (has_done_postsequence)))
  :effect (and (increase (a) 1) (assign (c) 0)))
(:action accelerate'
  :parameters()
  :precondition (and (running) (< (a) (upLimit)) 
            (not (has_done_presequence)))
  :effect (and (increase (a) 1) (has_done_presequence)))
(:action decelerate'
  :parameters()
  :precondition (and (running) (> (a) (downLimit)) 
        (has_done_presequence) (= (c) -1) (not (has_done_postsequence)))
  :effect (and (decrease (a) 1) (assign (c) 1) ))                
(:action decelerate''
  :parameters()
  :precondition (and (running) (> (a) (downLimit)) (not (= (c) 1))
    (not (= (c) -1)) (has_done_presequence) (not (has_done_postsequence)))
  :effect (and (decrease (a) 1) (assign (c) 0)))
(:action stop
  :parameters()
  :precondition (and (= (v) 0) (>= (d) 30) (not (engineBlown))
                 (has_done_presequence) (not (has_done_postsequence)))
  :effect(and (goalReached) (assign (c) 0) ))
(:action stop'
  :parameters()
  :precondition (and (= (v) 0) (>= (d) 30) (not (engineBlown)
                 (has_done_presequence) (not (has_done_postsequence)))
  :effect (and (goalReached) (has_done_postsequence)))
\end{lstlisting}
The initial and goal states are modified as shown below:
\begin{lstlisting}[basicstyle=\footnotesize]
(:init (running) (= (runningTime) 0) (= (upLimit) 1) (= (downLimit) -1)
    (= d 0) (= a 0) (= v 0) (= c -1))
(:goal (and (goalReached) (not(engineBlown)) (<= (runningTime) 50)
        (has_done_presequence) (has_done_postsequence)))
\end{lstlisting}
The modifications made are shown in the corresponding hybrid automata representation of the \emph{HModel} in Figure \ref{fig:domain_G}:
\begin{figure}[htbp]
    \centering
    \includegraphics[width=0.66\textwidth]{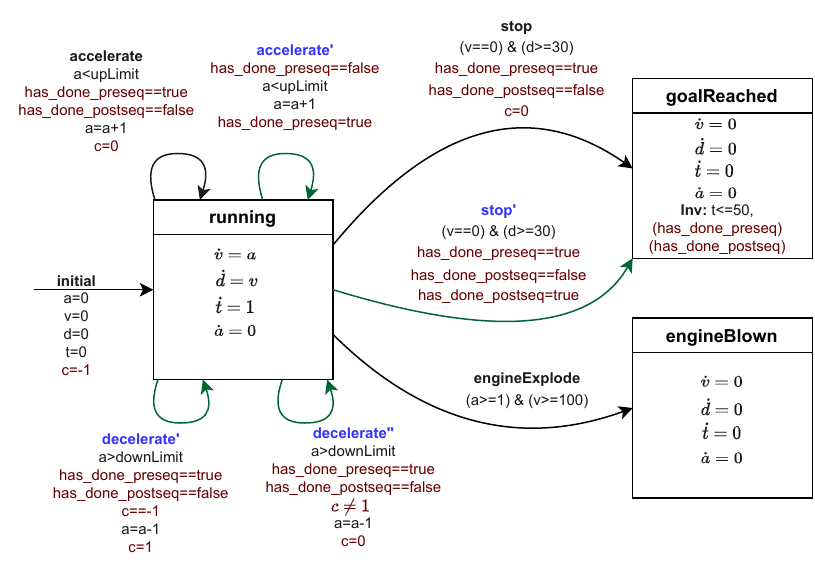}
    \caption{The hybrid automaton for the corresponding \emph{HModel} of the question addressed in sect. 4.7.}
    \label{fig:domain_G}
\end{figure}

The resulting \emph{HPlan} given by \textsc{SMTPlan+} is:

\begin{lstlisting}[basicstyle=\footnotesize]
    (@\underline{Time}@)         (@\underline{Action}@)               (@\underline{Duration}@)
    0.0:        (accelerate')           [0.0]
    1.0:        (decelerate')           [0.0] 
    3.0         (accelerate)            [0.0]
    5.0         (decelerate'')          [0.0]
    13.0:       (decelerate'')          [0.0]
    16.0:       (stop')                 [0.0]
    (@\emph{makespan}: = 16.0 units.@)
\end{lstlisting}
\end{exmp}

\noindent Recall that primed actions have the same interpretation as their unprimed counterparts, except that they may have different preconditions and effects.
\noindent The contrastive explanation is:
\begin{itemize}
\item{\textit{Remove ($\phi$,$\phi'$)}: \emph{stop} and \emph{decelerate} are removed in the alternate plan;}
\item{\textit{Add ($\phi$,$\phi'$)}: \emph{accelerate'}, \emph{decelerate'}, \emph{decelerate''} and \emph{stop'} are added in the alternate plan;}
\item{\textit{Common ($\phi$,$\phi'$)}: \emph{accelerate};}
\item{\textit{dwell-diff}: \{$moving$, $16$\};}
\item{\textit{diff-cost}$_{\tau}$: $-16$;}
\item{\textit{diff-cost}$_{len}$: $2$.}
\end{itemize}

\noindent A user of the framework can draw the following conclusion from the contrastive explanation:
\begin{quote}
    Conclusion: For the given problem instance, there exist plans that consist of action sequences other than the sequence \textit{decelerate-decelerate}, however, the plan length is greater than that of the original one.
\end{quote}

\subsection{Questioning the Optimality of Plan Length} \label{secH}
Similar to the question in Section~\ref{secF}, a user might be interested in a plan with a shorter plan length. For an explanation of the plan length, the following contrastive question can be asked,
\begin{quote}
  Why is the length of the plan not less?
\end{quote}
\noindent Similar to addressing the question in Section \ref{secF}, an iterative question \enquote{Why is the length of the plan $k$ and not less?} where $k$ is the length of the valid plan successfully generated by the planner in the last iteration, can help to address the optimality of the planner in terms of the plan length. In our car domain, a user might ask:
\begin{quote}
    \enquote{Why is the length of the plan not less than 4?}
\end{quote}
{To answer such a contrastive question, the planning instance is compiled to introduce a new function symbol $s$ set to $0$ initially. Each action $n\in As$ is modified to have a new added effect that increases the value of $s$ by 1 on each occurrence of the action in a plan.
\begin{equation*}
    {\textit{Eff}^{+}(n)' = \textit{Eff}(n)\cup\{\textit{PNE}(s=s+1), s\in \textit{Fs}\},\ n\in As}
\end{equation*}
where \textit{Eff}$^{+}(n)'$ indicates the modified post-condition of the action $n$.
A constraint $\textit{PNE}(s<k)$ is added as a goal condition to check whether a plan of length less than $k$ exists. Iteratively lowering the value of $k$ using a model similar to Equation \eqref{eq:rec} can provide an optimal plan in terms of plan length. The new planning instance is:
\begin{equation*}
   \Pi' = (\langle Fs', Rs, As, Es, Ps, arity'\rangle,\langle Os, I', G'\rangle) 
\end{equation*}
where
\begin{itemize}
\item {\textit{Fs}$'$ = \textit{Fs} $\cup\{s\}$}
\item {\textit{arity}$'(s)$ = $0$}
\item {$I'$ = $I$ $\cup$ \{\textit{PNE} ($s=0$)\}}
\item {$G'$ = $G$ $\cup$ \{\textit{constraint}(\textit{PNE} ($s<k$))\}}
\end{itemize}
}
{The constructed iterative \emph{HModel} is sound and will produce an \emph{HPlan} having a plan-length shorter than the \emph{HPlan} of the previous round, if there exists any.}
\begin{lemmasec}
{The construction of the iterative \emph{HModel} is sound.}
\end{lemmasec}
\noindent{\textsc{Proof:} In each iteration a valid \emph{HPlan} must be shorter in plan-length than the \emph{HPlan} of the previous round. The function symbol $s$ keeps track of the applied action in a plan. The goal constraint ($s<k$) imposes that the plan-length of \emph{HPlan} $\phi_i'$ in the $i$-th iteration should be less than $k$, where $k$ is the plan-length of the \emph{HPlan} of the previous round. Thus, it ensures that $\phi_i'$ has a shorter plan-length than $\phi_{i-1}'$. $\Box$}
{
\begin{exmp}
Consider the user question above, where we look for a plan with a plan-length less than 4. For that, a new function symbol $s$ is introduced and initialized to 0. Each action in the domain is extended with an added effect to increase the value of $s$ by one on its appearance in a plan. The goal condition is additionally extended with the constraint ($s<k$), where $k$ is the plan-length of the \emph{HPlan} obtained in the previous iteration. For example, in the first iteration, the \emph{HModel} $\Pi_1'$ contains the goal constraint ($s<4$).
\end{exmp}
}

\begin{figure}[htbp]
    \centering
    \includegraphics[width=0.66\textwidth]{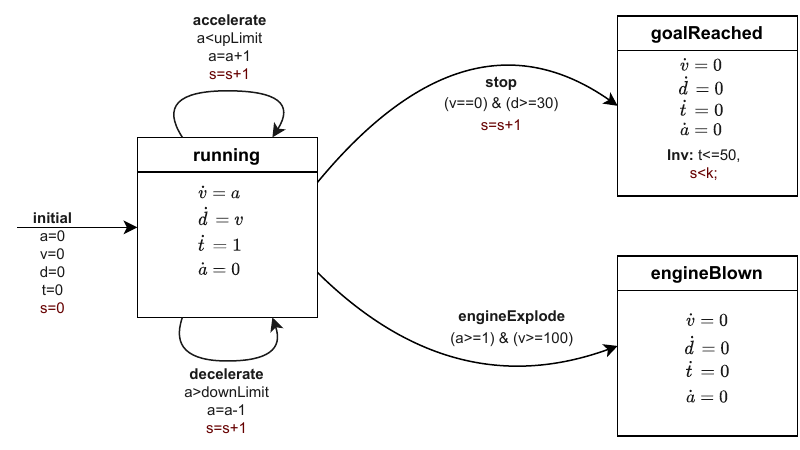}
    \caption{{The hybrid automaton for the corresponding \emph{HModel} for question 8.}}
    \label{fig:domain_H}
\end{figure}

The corresponding modification in the hybrid automaton representation of the domain is shown in Figure~\ref{fig:domain_H}.

\noindent Restricting the plan length to be less than 4 results in no plan being generated by the planner. The contrastive explanation is:
\begin{itemize}
\item{\textit{Remove ($\phi$,$\phi'$)}: Not available, as there is no alternate plan;}
\item{\textit{Add ($\phi$,$\phi'$)}: Not available, as there is no alternate plan;}
\item{\textit{Common ($\phi$,$\phi'$)}: Not available, as there is no alternate plan;}
\item{\textit{dwell-diff}: Not available;}
\item{\textit{diff-cost}$_{\tau}$: $\infty$;}
\item{\textit{diff-cost}$_{len}$: $\infty$.}
\end{itemize}

\noindent A user of the framework can draw the following conclusion from the contrastive explanation:
\begin{quote}
    Conclusion: For the problem instance, no valid plan can be generated that has a plan length less than 4.
\end{quote}

\noindent \textbf{Summary Table:} A summary of the contrastive explanations of the above-discussed user questions is provided in Table \ref{ch2table:table5}. In the table, \emph{Q. No.} marks the section number and represents the particular contrastive question that we discussed in that section. \emph{Rem}, \emph{Add}, and \emph{Com} are the number of actions that are removed from the original plan, the number of new actions that appear in the alternate plan \emph{HPlan}, and the number of actions that appear in both the original and the alternate plans, respectively. \emph{Dwell-diff} exhibits the dwell time differences of the processes in the original and the \emph{HPlan}, where a positive value indicates that the original plan has a longer dwell time in the respective location. \emph{DC{$_\tau$}} and \emph{DC{$_{l}$}} show the differences of \emph{makespan} and plan-length of the original and the \emph{HPlan}. A positive value of \emph{DC{$_\tau$}} and \emph{DC{$_{l}$}} indicates the \emph{HPlan} has a longer \emph{makespan} and a greater plan-length than the original. \emph{Remark} draws the conclusion on the quality of the generated \emph{HPlan} and the experimental observations.

\begin{table*}[htbp]
  \centering\tiny
  \resizebox{\textwidth}{!}{
    \begin{tabular}{|c|c|c|c|c|c|c|c|}
    \hline
         Q. No. & Rem & Add & Com  & Dwell-diff & DC{$_\tau$} & DC{$_{l}$} & Remark\\
    \hline
           \multirow{3}{*}{\ref{secA}} & \multirow{3}{*}{0} & \multirow{3}{*}{2} & \multirow{3}{*}{3} & \multirow{3}{*}{\{\emph{moving}, 20\}} & \multirow{3}{*}{-20} & \multirow{3}{*}{2} & \emph{HPlan} has a shorter \emph{makespan} but a\\
           & & & & & & & longer plan in terms of applied actions.\\
           & & & & & & & The original plan is better.\\
    \hline
           \multirow{3}{*}{\ref{secB}} & \multirow{3}{*}{1} & \multirow{3}{*}{1} & \multirow{3}{*}{2} & \multirow{3}{*}{\{\emph{moving}, 14\}} & \multirow{3}{*}{-14} & \multirow{3}{*}{0} & The original plan is not optimal\\
           & & & & & & & with respect to the duration of the plan.\\
           & & & & & & & The \emph{HPlan} is better.\\
    \hline
           \multirow{3}{*}{\ref{secC}} & \multirow{3}{*}{0} & \multirow{3}{*}{1} & \multirow{3}{*}{2} & \multirow{3}{*}{\{\emph{moving}, -9.5\}} & \multirow{3}{*}{9.5} & \multirow{3}{*}{0} & The \emph{decelerate} action being applied\\
           & & & & & & & later than 1-time unit, resulted in a longer\\
           & & & & & & &  plan. The original plan is better.\\
    \hline
           \multirow{3}{*}{\ref{secD}} & \multirow{3}{*}{NA} & \multirow{3}{*}{NA} & \multirow{3}{*}{NA} & \multirow{3}{*}{NA} & \multirow{3}{*}{$\infty$} & \multirow{3}{*}{$\infty$} & Barring the \emph{decelerate} action from\\
           & & & & & & & appearing in the plan leads to no\\
           & & & & & & & valid plan in this domain.\\
    \hline
           \multirow{3}{*}{\ref{secE}} & \multirow{3}{*}{NA} & \multirow{3}{*}{NA} & \multirow{3}{*}{NA} & \multirow{3}{*}{NA} & \multirow{3}{*}{$\infty$} & \multirow{3}{*}{$\infty$} & No valid plan can be generated\\
           & & & & & & & that has less than two executions\\
           & & & & & & & of the action \emph{decelerate}.\\
    \hline
           \multirow{3}{*}{\ref{secF}} & \multirow{3}{*}{0} & \multirow{3}{*}{0} & \multirow{3}{*}{3} & \multirow{3}{*}{see Table \ref{ch2tab:table1}} & \multirow{3}{*}{see Table \ref{ch2tab:table1}} & \multirow{3}{*}{0} & The \emph{makespan} of the original plan\\
           & & & & & & & is not optimal and there can be alternate\\
           & & & & & & & plans with shorter \emph{makespan}.\\
    \hline
           \multirow{3}{*}{\ref{secG}} & \multirow{3}{*}{2} & \multirow{3}{*}{4} & \multirow{3}{*}{1} & \multirow{3}{*}{\{\emph{moving}, 16\}} & \multirow{3}{*}{-16} & \multirow{3}{*}{2} & The \emph{HPlan} consisting of action sequence\\
           & & & & & & & other than \textit{decelerate-decelerate} has greater\\
           & & & & & & & plan-length than the original one.\\
    \hline
           \multirow{2}{*}{\ref{secH}} & \multirow{2}{*}{NA} & \multirow{2}{*}{NA} & \multirow{2}{*}{NA} & \multirow{2}{*}{NA} & \multirow{2}{*}{$\infty$} & \multirow{2}{*}{$\infty$} & No valid plan can be generated that\\
           & & & & & & & has a plan length less than 4.\\
    \hline
    \end{tabular}
  }
  \caption{A brief summary of the contrastive explanations generated for the user questions discussed above.}
  \label{ch2table:table5}
\end{table*}

\section{Proving the Absence of a Plan} \label{sec4}
The planning problem is undecidable for hybrid systems in general (\cite{ALUR19953}). When a planner fails to generate a valid plan for a planning problem, it cannot be asserted whether it is due to the underlying undecidability or that the planning problem admits no valid plan. This leads to a limitation in our explanation framework since incorrect explanations may result in such cases. To address this issue, we attempt to prove the non-existence of a plan by \emph{reachability analysis} in the hybrid domain of the planning problem for which we get no-plan (we refer to them as no-plan-models in the text that follows). Reachability analysis checks whether the system can reach a target state from an initial state under the dynamics of the system. Observe that reachability analysis can be useful in proving the non-existence of a plan because unreachability of the goal-state from the initial state under the dynamics of the planning domain proves and also explains, in some sense, why a sound planner does not return a valid plan for a given planning problem instance. Note that reachability analysis is undecidable for hybrid systems in general. However, allowing $\delta$ approximation in the analysis may often lend to decidability as shown in (\cite{DBLP:conf/lics/GaoAC12}). We leverage this observation in the explanation of no-plan being returned by a planner by proving the absence of a valid plan. A brief discussion on the notion of $\delta$ approximation in reachability analysis follows in the text. The key idea of $\delta$-approximate reachability analysis is to create a bounded abstraction of the system and reduce it to the satisfiability problem of a first-order logic formula over reals with Type-2 computable functions (\cite{10.1007/978-3-642-31365-3_23, DBLP:conf/lics/GaoAC12}). The presence of bounded quantifiers and the infusion of a $\delta$-relaxed notion of satisfiability in the resulting formula ensures decidability for a positive rational number $\delta$. We leverage this $\delta$-approximate bounded reachability analysis feature of \textsc{dReach} (\cite{DBLP:journals/corr/GaoKCC14}) on our no-plan-models to explain the non-existence of plans in the no-plan instances that we encounter with \textsc{SMTPlan+}. In our reachability analysis of the no-plan-models, we have set a bound of k-depth (we refer to as plan-depth) as it corresponds to checking the existence of any plan upto k-length. In the following sections, we first describe how the $\delta$-approximation is applied and its significance in reachability analysis. We then discuss the explanations for the no-plan-models and also provide an algorithm for the same.

\noindent \paragraph{Remark:} Note that with $\delta$-approximate bounded reachability analysis, we can only provide a proof that the problem instance does not admit a valid plan. Further insights on why there is no valid plan is not addressed by this method. There are notable works which provide reasons for unsolvability of planning problem instances for discrete domains such as (\cite{DBLP:conf/aips/ErikssonH20}, \cite{DBLP:conf/aips/ErikssonRH17}). For hybrid domains, we are not aware of any such work, and we plan to address this as future work.
For certain PDDL+ problem instances, planners like \textsc{ENHSP} (\cite{DBLP:conf/ecai/ScalaHTR16}) can deduce unsolvability and report it. In problem instances beyond the scope of \textsc{ENHSP}, the proposed reachability analysis based method can be used with \textsc{dReach}, which can analyze a large class of non-linear systems.

\subsection{$\delta$-Approximation}
\noindent The $\delta$-approximation of a system can be best explained in terms of showing the $\delta$-weakening (\cite{DBLP:journals/corr/GaoKCC14}) of the corresponding hybrid automata. Let $\delta \in \mathbb{Q}^+\cup \{0\}$ be an arbitrary rational number and $H$=\big($Loc$, $Var$, $Flow$, $Init$, $Lab$, $Edge$, $Inv$\big) represent a hybrid automaton. The $\delta$-weakening of $H$ can be obtained by weakening the flow and invariants in each location, initial valuation of variables, guards, and assignments of each transition of the automaton. We represent a $\delta$-weakened hybrid system as $H^{\delta}$:
\begin{align*}
    H^{\delta}=\big(Loc, Var, Flow^{\delta}, Init^{\delta}, Lab, Edge^{\delta}, Inv^{\delta} \big)
\end{align*}
\noindent It may be noted from the syntactic construction described above that $H^{\delta}$ represents an over-approximation of $H$. Thus, the set of runs in $H^{\delta}$ subsumes the set of runs in $H$. Let $H(runs)$ denote the set of all runs in $H$. $H(runs)\subseteq H^{\delta}(runs)$. Consider the no-plan-model of Section \ref{secD}, for which the corresponding hybrid automaton is shown in Figure \ref{fig:domain_D}. $\delta$-weakening of the model in Section \ref{secD} corresponds to the weakening of flows and invariants in each location, guards and assignments of each transition by the value $\delta$.
In bounded reachability analysis, for a given $\delta\in \mathbb{Q}^+$ and a bounded depth $k$, if \textsc{dReach} concludes that a given problem instance is unsatisfiable, then it implies the following:
\begin{enumerate}
    \item The problem instance is also unsatisfiable for all values $\leq\delta$ for a given bound $k$.
    \item The problem instance is also unsatisfiable for all the depths $<$ $k$ for a given $\delta$.
\end{enumerate}
Next, it can be shown that if there is no plan in $H_{\delta}$ for any given problem instance, there is no plan in $H$ as well.
\begin{lemmasec}
    For a given problem instance $\Pi$, the absence of any plan in $H^{\delta}$ implies the absence of any plan in $H$. $\Box$
\end{lemmasec}
\noindent\textsc{Proof}: Since $H(runs)\subseteq H^{\delta}(runs)$ and a plan $\phi$ is an instantiation of a run $r\in H(runs)$, it is self-evident that for the given problem instance $\Pi$, a precision $\delta$, and plan-depth $k$ if there exists no plan in $H^{\delta}$ there exists no plan in $H$. $\Box$

\section{Results} \label{sec5}

In this section, we present the performance statistics of our contrastive explanation framework. We have chosen three benchmark PDDL+ domains: the Car domain (\cite{PDDL+benchmarks}), the Generator-events domain (\cite{PDDL+benchmarks}) and the Planetary-lander domain (\cite{PDDL+}) to present a comparative look on the scalability of the contrastive explanation framework on these planning problems over these domains. 
Additionally, we evaluate the planning problem instances over some of the state-of-the-art planners, which further shows the applicability of our framework across planners.
For a problem instance, we use different planners to generate plans. Contrastive questions are formed based on those plans, and corresponding \emph{HModels} are constructed for each contrastive question. These \emph{HModels} are used with the planners to generate the \emph{HPlans}. Below, we provide a brief description of the planning domains.

\subsection{Benchmarks}\label{sec:benchmarks}

\textcolor{black}{\textit{Car Domain (CD):} The planning problem consists of a car (initially at rest) that has to cover a given distance within a specified time bound. Further, it is needed to ensure that the car has a zero velocity at the end as well. The domain comprises of six function symbols: $d$, $v$, $a$, $upLimit$, $downLimit$ and $runningTime$, three predicate symbols: \emph{running}, \emph{engineBlown} and \emph{goalReached}, three instantaneous actions \emph{accelerate}, \emph{decelerate} and \emph{stop}, one event \emph{engineExplode} and one process \emph{moving}. The function symbols 
$d$, $v$, and $a$ respectively represent the distance, velocity, and acceleration of the car, whereas $upLimit$ and $downLimit$ specify the upper and lower bounds on the acceleration. $runningTime$ encodes the elapsed time of the car. The predicates \emph{running}, \emph{engineBlown}, and \emph{goalReached} define the states of the car. The actions \emph{accelerate} and \emph{decelerate} respectively increase and decrease the acceleration of the car by one unit. The \emph{stop} action is applicable when the car covers the specified distance and has a velocity of zero. The event \emph{engineExplode} is triggered when the velocity and acceleration of the car reach a certain threshold and the car is in the \emph{engineBlown} state. The process \emph{moving} is activated when the car is in the \emph{running} state specified by a predicate \emph{running}. This process updates the distance and the velocity of the car as a function of time. Initially, the car is in the \emph{running} state with a zero velocity. The goal is to traverse a distance of 30 units within 50 units of time. The planning domain represented in PDDL+ is shown in \textbf{Listing 1}. The initial condition \emph{init} specifies that the predicate \emph{running} is initially \emph{true}, the initial values of the functions $a$, $v$, $d$, and $runningTime$ are assigned 0.
Across the problem instances, we vary the value of $upLimit$ and $downLimit$, which specify the bounds on the acceleration of the car.
For problem instance 1 (see \textbf{Listing 2}), $upLimit$ and $downLimit$ are assigned 1 and -1, respectively, whereas for problem instance 2 (see \textbf{Appendix \ref{appendix1}}) these values are set to 2 and -2, respectively. The goal for the car is to travel a minimum distance of 30 units in less than or equal to 50 units of time while avoiding an engine explosion, which is caused when the velocity is greater than or equal to 100 units and the acceleration is greater than or equal to $upLimit$. This is specified as the goal condition \emph{goal} with the predicates \emph{goalReached} and the negation of \emph{engineBlown}, together with the numeric condition $runningTime \leq 50$.}
\\

\noindent \textit{Generator-Events Domain (GD):} \textcolor{black} {The generator-events domain consists of a generator that has to run for a specified amount of time. The domain comprises of four function symbols: \emph{fuelLevel}, \emph{capacity}, \emph{fuelInTank} and \emph{ptime}, four predicate symbols: \emph{generator-ran}, \emph{available}, \emph{using} and \emph{safe}, one durative-action \emph{generate} which operates for a fixed duration of 1000 time units, one instantaneous action \emph{refuel}, one process \emph{refuelling} and two events: \emph{tankEmpty} and \emph{generatorOverflow}. We need to instantiate objects of types generator and tank, which will be set as parameters to different domain symbols while respecting their arities. The \emph{fuelLevel} and the \emph{capacity} are associated with the generator and indicate the current fuel level and fuel storage capacity of the generator, whereas the \emph{fuelInTank} and the \emph{ptime} are associated with the tanks, indicating fuel reserve in a tank and fuel pouring time from a tank while in use, respectively. The \emph{generator-ran} indicates whether the generator ran successfully for a specified amount of time (1000 time units), \emph{available} indicates tanks that are in store for use, \emph{using} indicates tanks that are being used by the generator, and \emph{safe} indicates if the generator is operating safely. The durative-action \emph{generate} \textcolor{black}{has precondition \emph{fuelLevel} $\geq$ 0 and \emph{safe} to indicate that the current fuel level in the generator needs to be greater than 0 and the generator needs to be in the safe mode. It runs for a fixed duration of time (1000 units) and has an effect to decrease the \emph{fuelLevel} at a constant rate. At the end, this makes \emph{generator-ran} to be true to reflect that the generator ran successfully for the specified duration.} The instantaneous-action \emph{refuel} enables the generator to use an available tank and activates the process \emph{refuelling}, which in turn increases the \emph{fuelLevel} at a rate proportional to the pouring time \emph{ptime} while decreasing the \emph{fuelInTank} at the same rate. The event \emph{emptyTank} triggers to indicate that a tank is unavailable for use when \emph{fuelInTank} becomes 0 for it, whereas the event \emph{generatorOverflow} drives the generator to an unsafe-mode when its \emph{fuelLevel} exceeds the \emph{capacity} threshold. For creating different problem instances, we vary the number of fuel reserve tanks and the initial \emph{fuelLevel} of the generator. For problem instance 1 (see \textbf{Appendix \ref{appendix2}}), the initial state specifies that the current \emph{fuelLevel} of the generator \emph{gen} is 940 units, whereas the \emph{capacity} to hold fuel in the generator is 1600 units. The tanks \emph{available} to use are \emph{t1} and \emph{t2}, and the fuel reserve in each tank is 40 units. In problem instance 2 (see \textbf{Appendix \ref{appendix2}}), the initial \emph{fuelLevel} of the generator \emph{gen} is 900 units, and the tanks \emph{available} to use are \emph{t1}, \emph{t2}, and \emph{t3}, and the fuel reserve in each tank is 40 units. Initially, the generator is in the \emph{safe} mode. The predicate and function symbols that are not initialized explicitly in the initial state are initialized with a default value of false and 0, respectively. In both cases, the goal condition comprises of a single constraint \emph{(generator-ran)} which needs to be true in the goal state, indicating that the generator ran successfully for a specified amount of time (1000 time units). The PDDL+ representation of the domain is shown in \textbf{Appendix \ref{appendix2}}. Additionally, a hybrid automaton model for the generator-events domain for problem instance 1 is shown for visual representation in Appendix \ref{appendix2}.}
\\

\noindent \textit{Planetary-lander domain (PD):} \textcolor{black}{The planetary-lander domain is based on a simplified model of a solar-powered lander.
The domain has 21 function symbols: \emph{demand}, \emph{supply}, \emph{soc}, \emph{charge\_rate}, \emph{daytime}, \emph{heater\_rate}, \emph{dusk}, \emph{dawn}, \emph{fullprepare\_durtime}, \emph{prepareobs1\_durtime}, \emph{prepareobs2\_durtime}, \emph{observe1\_durtime}, \emph{observe2\_durtime}, \emph{obs1\_rate}, \emph{obs2\_rate}, \emph{A\_rate}, \emph{B\_rate}, \emph{C\_rate}, \emph{D\_rate}, \emph{safeLevel} and \emph{solar\_const}, which control different parameters of the domain. It has 7 predicate symbols: \emph{day}, \emph{commsOpen}, \emph{readyForObs1}, \emph{readyForObs2}, \emph{gotObs1}, \emph{gotObs2} and \emph{available}.
The domain has five durative-actions: \emph{fullPrepare}, \emph{prepareObs1}, \emph{prepareObs2}, \emph{observe1} and \emph{observe2}. Each action performs a specific task and draws a fixed power throughout the operation. \emph{observe1} and \emph{observe2} are two observe actions, which observe two different phenomena. However, prior to performing any observation task, the system must prepare the observers either by using a single long action, called \emph{fullPrepare}, or by using two shorter actions, called \emph{prepareObs1} and \emph{prepareObs2}, which are specific to one of the observation actions. The shorter actions cumulatively have higher power requirements for their execution than the single preparation action. The lander is required to execute both observation actions before a communication link is established, which sets a deadline on the activities. These activities are all carried out against a backdrop of fluctuating power supply. The lander is equipped with a regenerative solar power resource.
The domain contains four processes: \emph{generating}, \emph{charging}, \emph{discharging} and \emph{night\_operations}, and two events: \emph{nightfall} and \emph{daybreak}.
The \emph{generating} process generates solar power for the lander, which is governed by the position of the sun. At night, there is no power generated, which rises smoothly to a peak at midday, falling back to zero at dusk. The \emph{nightfall} event stops the \emph{generating} process, and the lander enters the \emph{night\_operations} mode. This process draws a constant power requirement for a heater used to protect its instruments, in addition to any requirements for instruments. The \emph{daybreak} event stops the \emph{night\_operations} process, and the \emph{generating} process restarts. Both these events are triggered by a simple clock that is driven by the twin processes of \emph{generating} and \emph{night\_operations} and reset by the events. The lander is equipped with a battery, allowing it to store excess energy as charge through the process \emph{charging} while excess demand must be supplied from the battery through the process \emph{discharging}. The planning problem comprises an initial state, goal conditions, and object instantiations. An object of type \emph{equipment} is instantiated. The initial state specifies the initial values of the different parameters of the domain. The goal of the planning problem is to observe the two phenomena of the environment, which are specified by two predicate symbols \emph{gotObs1} and \emph{gotObs2}. These predicates are initially set to false, which indicates that those phenomena have not been observed yet. The actions \emph{observe1} and \emph{observe2} set them to true after observing those phenomena.
In this case, to create the different instances, we vary the initial state of charge \emph{soc} and the safe operating charge level \emph{safeLevel} for the lander. For problem instance 1, \emph{soc} is set to 93 units and \emph{safeLevel} is set to 15 units, whereas in problem instance 2, \emph{soc} is set to 50 units and \emph{safeLevel} is set to 25 units. The PDDL+ representation of the domain and the planning problem instances 1 and 2 are shown in \textbf{Appendix \ref{appendix3}}.}

\subsection{Evaluation}

We present here the performance of our framework with respect to the contrastive questions presented in Section \ref{sec3}, working with three benchmark PDDL+ planning domains and three state-of-the-art planners \textsc{SMTPlan+} (SM), \textsc{ENHSP} (EN), and \textsc{UPMurphi} (UP). The results are presented in Table \ref{ch2table: table4}. In the table, \emph{BM} represents the benchmark problem domains in PDDL+. \emph{Ins} represents the problem instance of a domain. For each domain, we present two problem instances in the table. \emph{PL} reports the hybrid system planner chosen for solving the planning problem instance. For each problem instance in each domain, we execute all the three planners for each of the eight questions, with a time bound of one hour and peak memory usage bound of 4 GB. In\emph{Dur} and \emph{Len} show the original plan's \emph{makespan} and plan-length for the problem instance. \emph{QN} represents the contrastive question addressed in Section \ref{secCaseStudy} (\ref{secA}-\ref{secH}). For each of these contrastive questions, an \emph{HPlan} is generated. \emph{H-dur} and \emph{H-len} show the \emph{makespan} and the plan-length of the corresponding \emph{HPlan}. \emph{Diff-cost{$_\tau$}} and \emph{Diff-cost{$_{len}$}} are the contrastive explanation (CE) metrics discussed in Definition \ref{def:CE}.
For abbreviation, we use \emph{DC$_\tau$} and \emph{DC$_l$} to denote these metrics in the table.
\emph{DC$_\tau$} presents the plan-duration difference and \emph{DC$_l$} presents the plan-length difference of the original plan and the \emph{HPlan}. A positive \emph{DC$_\tau$} indicates a longer \emph{makespan}, whereas a positive \emph{DC$_l$} indicates a longer plan-length of the \emph{HPlan} in comparison to the original one. \emph{CM} reports the number of constraint modifications that are made in the domain (inclusive of constraint additions and deletions). \emph{Time} and \emph{H-time} represent the plan generation time whereas \emph{Mem} and \emph{H-mem} report the memory usages for the original plan and the \emph{HPlan}, respectively. Finally, \emph{HQ} represents an evaluation of the quality of the \emph{HPlan} against the original plan while considering the contrastive explanation metrics \emph{DC$_\tau$} and \emph{DC$_l$}, where a $\checkmark$ indicates that the original plan is better than the \emph{HPlan}, a $\times$ indicates that the \emph{HPlan} is better than the original plan, and a $=$ indicates that both the plans are of equal quality. A $\checkmark$ in the \emph{HQ} field for a contrastive question provides an explanation on why the planner has provided such a plan over the hypothetical one. However, $\times$ in the \emph{HQ} field exhibits that there is a better plan possible for the problem instance than the original plan, considering plan makespan and length as quality evaluation metrics. A $=$ in the \emph{HQ} field indicates that the produced \emph{HPlan} is an alternate plan of the same quality with respect to the original plan. 
In some of the problem instances, the \emph{HModels} for the corresponding contrastive questions produce no plans \textcolor{black}{which is either reported by the planner (indicated as NP in the table) or the planner exceeds the resource bounds set by us (indicated as RO in the table)}. \emph{Q6} questions the optimality of plan-length of a plan (see Section \ref{secF}), and is not included in the table since it represents an iterative questioning and compilation (details can be seen in Table \ref{ch2tab:table1}). \emph{Q5} is also not included in the table for the planetary-lander domain since this question is not applicable to the considered planning problem instances in which the original plan is such that the actions that appear in the plan do not repeat.

\noindent \paragraph{Analysis of Results:} 
\textcolor{black}{In the car domain, for problem instance 1, \textsc{SMTPlan+} produces HPlan for the contrastive questions \emph{QN1}, \emph{QN2}, \emph{QN3}, \emph{QN6}, and \emph{QN7}, but is not able to produce HPlans for \emph{QN4}, \emph{QN5}, and \emph{QN8} within the time bound. We understand that \emph{QN4}, \emph{QN5}, and \emph{QN8} are actually unsolvable as \textsc{ENHSP}, which is a sound and complete planner, reports the non-existence of a plan for these instances. Additionally, it may be noted that \textsc{UPMurphi} reports no plan for \emph{QN4}, however, it reports a memory overflow for \emph{QN5} and \emph{QN8}. Similarly, for problem instance 2 of the car domain, \textsc{SMTPlan+} reports plans for the contrastive questions \emph{QN1}, \emph{QN2}, \emph{QN3}, \emph{QN6}, and \emph{QN7}, but does not produce any plan for \emph{QN4}, \emph{QN5}, and \emph{QN8} due to timeout. In this case, both \textsc{ENHSP} and \textsc{UPMurphi} produce plans for all questions except \emph{QN4}, for which both of them report a no plan. This leads us to conclude that the three planners agree on the positive questions. Further, we note the fact that \emph{QN4} is actually unsolvable for this instance. Table \ref{ch2table: table4} shows that for the generator-events domain, for problem instance 1, \textsc{SMTPlan+} produces HPlans for the contrastive questions \emph{QN1}, \emph{QN2}, \emph{QN3}, \emph{QN6}, and \emph{QN7}, but does not produce any plan for \emph{QN4}, \emph{QN5}, and \emph{QN8} due to timeout. For problem instance 2, \textsc{SMTPlan+} produces HPlans for the questions \emph{QN2}, \emph{QN3}, and \emph{QN6}, and does not produce any plan for \emph{QN1}, \emph{QN4}, \emph{QN5}, \emph{QN7} and \emph{QN8} due to timeout. In both the problem instances, both \textsc{ENHSP} and \textsc{UPMurphi} fail to produce any plan on the original domain and for all the questions as well, due to resource overflow. A similar interpretation follows for the entries on the Planetary-lander domain. 
}


\paragraph{Contrastive Explanation:} We observe that for most of the problem instances, the original plan is better than the contrasting alternate one (\emph{HPlan}) marked by $\checkmark$ in Table \ref{ch2table: table4} and thus justifies the original plan to the user. In instances marked with a $\times$, a user may consider to re-plan given that an alternate hypothetical plan is better in comparison to the original one.


\begin{table*}[htbp]
  \centering\tiny
  \resizebox{\textwidth}{!}{
    \begin{tabular}{|c|c|c|c|c|c|c|c|c|c|c|c|c|c|c|c|}
    \hline
         \multirow{2}{*}{BM} & \multirow{2}{*}{Ins} & \multirow{2}{*}{PL} & \multirow{2}{*}{Dur} & \multirow{2}{*}{Len} & \multirow{2}{*}{QN} & \multirow{2}{*}{H-dur} & \multirow{2}{*}{H-len} & \multicolumn{2}{|c|}{CE metrics} & \multirow{2}{*}{CM} & Time & H-time & Mem & H-mem & \multirow{2}{*}{HQ}\\\cline{9-10}
         & & & & & & & & DC$_\tau$ & DC$_l$ & & (in sec) & (in sec) & (in MB) & (in MB) & \\
    \hline 
           & & & & & \textcolor{black}{Q1} & \textcolor{black}{12} & \textcolor{black}{6} & \textcolor{black}{-20} & \textcolor{black}{2} & \textcolor{black}{12} & & \textcolor{black}{0.046} & & \textcolor{black}{$<$1} & $\checkmark$ \\\cline{6-11}\cline{13-13}\cline{15-16}
           & & & & & Q2 & 18 & 4 & -14 & 0 & 1 & & 0.024 & & $<$1 & $\times$ \\\cline{6-11}\cline{13-13}\cline{15-16}
           & & & & & Q3 & 41.5 & 4 & 9.5 & 0 & 4 & & 0.036 & & $<$1 & $\checkmark$ \\\cline{6-11}\cline{13-13}\cline{15-16}
           & & SM & 32 & 4 & Q4 & RO & RO & RO & RO & 1 & 0.027 & RO & $<$1 & RO & $\checkmark$ \\\cline{6-11}\cline{13-13}\cline{15-16}
           & & & & & Q5 & RO & RO & RO & RO & 3 & & RO & & RO & $\checkmark$ \\\cline{6-11}\cline{13-13}\cline{15-16}
           & & & & & \textcolor{black}{Q7} & \textcolor{black}{18} & \textcolor{black}{6} & \textcolor{black}{-14} & \textcolor{black}{2} & \textcolor{black}{21} & & \textcolor{black}{0.066} & & \textcolor{black}{$<$1} & $\checkmark$ \\\cline{6-11}\cline{13-13}\cline{15-16}
           & & & & & Q8 & RO & RO & RO & RO & 5 & & RO & & RO & $\checkmark$ \\\cline{3-16}
           & & & & & \textcolor{black}{Q1} & \textcolor{black}{50} & \textcolor{black}{28} & \textcolor{black}{11} & \textcolor{black}{12} & \textcolor{black}{7} & & \textcolor{black}{0.404} & & \textcolor{black}{70.7} & $\checkmark$ \\\cline{6-11}\cline{13-13}\cline{15-16}
           & & & & & Q2 & 40 & 14 & 1 & -2 & 1 & & 0.337 & & 66.1 & $\checkmark$ \\\cline{6-11}\cline{13-13}\cline{15-16}
           & & & & & Q3 & 39 & 22 & 0 & 6 & 4 & & 0.371 & & 78.6 & $\checkmark$ \\\cline{6-11}\cline{13-13}\cline{15-16}
           & p1 & EN & 39 & 16 & Q4 & NP & NP & NP & NP & 1 & 0.425 & 0.285 & 82.3 & 66.3 & $\checkmark$ \\\cline{6-11}\cline{13-13}\cline{15-16}
           & & & & & Q5 & NP & NP & NP & NP & 3 & & 2.351 & & 481.3 & $\checkmark$ \\\cline{6-11}\cline{13-13}\cline{15-16}
           & & & & & \textcolor{black}{Q7} & \textcolor{black}{49} & \textcolor{black}{29} & \textcolor{black}{10} & \textcolor{black}{13} & \textcolor{black}{85} & & \textcolor{black}{0.537} & & \textcolor{black}{114.1} & $\checkmark$ \\\cline{6-11}\cline{13-13}\cline{15-16}
           & & & & & Q8 & NP & NP & NP & NP & 5 & & 4.135 & & 1198.9 & $\checkmark$ \\\cline{3-16}
           & & & & & \textcolor{black}{Q1} & \textcolor{black}{9.505} & \textcolor{black}{6} & \textcolor{black}{-1.498} & \textcolor{black}{2} & \textcolor{black}{10} & & \textcolor{black}{4.67} & & \textcolor{black}{2166.4} & $\checkmark$ \\\cline{6-11}\cline{13-13}\cline{15-16}
           & & & & & \textcolor{black}{Q2} & \textcolor{black}{11.503} & \textcolor{black}{4} & \textcolor{black}{0.5} & \textcolor{black}{0} & \textcolor{black}{1} & & \textcolor{black}{4.97} & & \textcolor{black}{2161.3} & $\checkmark$ \\\cline{6-11}\cline{13-13}\cline{15-16}
           & & & & & \textcolor{black}{Q3} & \textcolor{black}{11.503} & \textcolor{black}{4} & \textcolor{black}{0.5} & \textcolor{black}{0} & \textcolor{black}{4} & & \textcolor{black}{6.47} & & \textcolor{black}{2161.5} & $\checkmark$ \\\cline{6-11}\cline{13-13}\cline{15-16}
           & & \textcolor{black}{UP} & \textcolor{black}{11.003} & \textcolor{black}{4} & \textcolor{black}{Q4} & \textcolor{black}{NP} & \textcolor{black}{NP} & \textcolor{black}{NP} & \textcolor{black}{NP} & \textcolor{black}{1} & \textcolor{black}{4.77} & \textcolor{black}{3.6} & \textcolor{black}{2159.8} & \textcolor{black}{1968.8} & $\checkmark$ \\\cline{6-11}\cline{13-13}\cline{15-16}
           & & & & & \textcolor{black}{Q5} & \textcolor{black}{RO} & \textcolor{black}{RO} & \textcolor{black}{RO} & \textcolor{black}{RO} & \textcolor{black}{4} & & \textcolor{black}{RO} & & \textcolor{black}{RO} & $\checkmark$ \\\cline{6-11}\cline{13-13}\cline{15-16}
           & & & & & \textcolor{black}{Q7} & \textcolor{black}{11.005} & \textcolor{black}{6} & \textcolor{black}{0.002} & \textcolor{black}{2} & \textcolor{black}{21} & & \textcolor{black}{5.82} & & \textcolor{black}{2168.6} & $\checkmark$ \\\cline{6-11}\cline{13-13}\cline{15-16}
        CD & & & & & \textcolor{black}{Q8} & \textcolor{black}{RO} & \textcolor{black}{RO} & \textcolor{black}{RO} & \textcolor{black}{RO} & \textcolor{black}{6} & & \textcolor{black}{RO} & & \textcolor{black}{RO} & $\checkmark$ \\\cline{2-16}
        
           & & & &  & \textcolor{black}{Q1} & \textcolor{black}{12} & \textcolor{black}{6} & \textcolor{black}{-20} & \textcolor{black}{2} & \textcolor{black}{12} & & \textcolor{black}{0.048} & & \textcolor{black}{$<$1} & $\checkmark$ \\\cline{6-11}\cline{13-13}\cline{15-16}
           & & & & & Q2 & 18 & 4 & -14 & 0 & 1 & & 0.025 & & $<$1 & $\times$ \\\cline{6-11}\cline{13-13}\cline{15-16}
           & & & & & Q3 & 41.5 & 4 & 9.5 & 0 & 4 & & 0.031 & & $<$1 & $\checkmark$ \\\cline{6-11}\cline{13-13}\cline{15-16}
           & & SM & 32 & 4 & Q4 & RO & RO & RO & RO & 4 & 0.028 & RO & $<$1 & RO & $\checkmark$ \\\cline{6-11}\cline{13-13}\cline{15-16}
           & & & & & Q5 & RO & RO & RO & RO & 3 & & RO & & RO & $\checkmark$ \\\cline{6-11}\cline{13-13}\cline{15-16}
           & & & & & \textcolor{black}{Q7} & \textcolor{black}{18} & \textcolor{black}{6} & \textcolor{black}{-14} & \textcolor{black}{2} & \textcolor{black}{21} & & \textcolor{black}{0.075} & & \textcolor{black}{$<$1} & $\checkmark$ \\\cline{6-11}\cline{13-13}\cline{15-16}
           & & & & & Q8 & RO & RO & RO & RO & 5 & & RO & & RO & $\checkmark$ \\\cline{3-16}
           & & & & & \textcolor{black}{Q1} & \textcolor{black}{45} & \textcolor{black}{61} & \textcolor{black}{19} & \textcolor{black}{46} & \textcolor{black}{15} & & \textcolor{black}{0.714} & & \textcolor{black}{155.6} & $\checkmark$ \\\cline{6-11}\cline{13-13}\cline{15-16}
           & & & & & Q2 & 50 & 37 & 24 & 22 & 1 & & 3.242 & & 872.5 & $\checkmark$ \\\cline{6-11}\cline{13-13}\cline{15-16}
           & & & & & Q3 & 15 & 19 & -9 & 4 & 4 & & 0.461 & & 79.8 & $\checkmark$ \\\cline{6-11}\cline{13-13}\cline{15-16}
           & p2 & EN & 26 & 15 & Q4 & NP & NP & NP & NP & 1 & 0.535 & 0.475 & 133.5 & 110.4 & $\checkmark$ \\\cline{6-11}\cline{13-13}\cline{15-16}
           & & & & & Q5 & 50 & 14 & 24 & -1 & 3 & & 1.023 & & 259.1 & $\checkmark$ \\\cline{6-11}\cline{13-13}\cline{15-16}
           & & & & & \textcolor{black}{Q7} & \textcolor{black}{47} & \textcolor{black}{64} & \textcolor{black}{21} & \textcolor{black}{49} & \textcolor{black}{79} & & \textcolor{black}{2.378} & & \textcolor{black}{350.9} & $\checkmark$ \\\cline{6-11}\cline{13-13}\cline{15-16}
           & & & & & Q8 & 24 & 4 & -2 & -11 & 5 & & 1.001 & & 230.5 & $\checkmark$ \\\cline{3-16}
           & & & & & \textcolor{black}{Q1} & \textcolor{black}{7.508} & \textcolor{black}{9} & \textcolor{black}{-0.498} & \textcolor{black}{2} & \textcolor{black}{15} & & \textcolor{black}{6.6} & & \textcolor{black}{2169.6} & $\checkmark$ \\\cline{6-11}\cline{13-13}\cline{15-16}
           & & & & & \textcolor{black}{Q2} & \textcolor{black}{8.506} & \textcolor{black}{7} & \textcolor{black}{0.5} & \textcolor{black}{0} & \textcolor{black}{1} & & \textcolor{black}{5.56} & & \textcolor{black}{2162.5} & $\checkmark$ \\\cline{6-11}\cline{13-13}\cline{15-16}
           & & & & & \textcolor{black}{Q3} & \textcolor{black}{8.006} & \textcolor{black}{7} & \textcolor{black}{0} & \textcolor{black}{0} & \textcolor{black}{4} & & \textcolor{black}{10.46} & & \textcolor{black}{2162.5} & $=$ \\\cline{6-11}\cline{13-13}\cline{15-16}
           & & \textcolor{black}{UP} & \textcolor{black}{8.006} & \textcolor{black}{7} & \textcolor{black}{Q4} & \textcolor{black}{NP} & \textcolor{black}{NP} & \textcolor{black}{NP} & \textcolor{black}{NP} & \textcolor{black}{1} & \textcolor{black}{8.11} & \textcolor{black}{3.8} & \textcolor{black}{2160.9} & \textcolor{black}{2161.7} & $\checkmark$ \\\cline{6-11}\cline{13-13}\cline{15-16}
           & & & & & \textcolor{black}{Q5} & \textcolor{black}{9.504} & \textcolor{black}{5} & \textcolor{black}{1.498} & \textcolor{black}{-2} & \textcolor{black}{4} & & \textcolor{black}{15.3} & & \textcolor{black}{2159.6} & $\checkmark$ \\\cline{6-11}\cline{13-13}\cline{15-16}
           & & & & & \textcolor{black}{Q7} & \textcolor{black}{8.008} & \textcolor{black}{9} & \textcolor{black}{0.002} & \textcolor{black}{2} & \textcolor{black}{41} & & \textcolor{black}{9.21} & & \textcolor{black}{2175.3} & $\checkmark$ \\\cline{6-11}\cline{13-13}\cline{15-16}
           & & & & & \textcolor{black}{Q8} & \textcolor{black}{11.003} & \textcolor{black}{4} & \textcolor{black}{2.997} & \textcolor{black}{-3} & \textcolor{black}{6} & & \textcolor{black}{7.08} & & \textcolor{black}{2161.5} & $\checkmark$ \\
    \hline
           & & & & & \textcolor{black}{Q1} & \textcolor{black}{1000} & \textcolor{black}{3} & \textcolor{black}{-64} & \textcolor{black}{0} & \textcolor{black}{4} & & \textcolor{black}{0.192} & & \textcolor{black}{32.1} & $\times$ \\\cline{6-11}\cline{13-13}\cline{15-16}
           & & & & & Q2 & 1072 & 3 & 8 & 0 & 4 & & 0.223 & & 32.2 & $\checkmark$ \\\cline{6-11}\cline{13-13}\cline{15-16}
           & & & & & Q3 & 1001 & 3 & -63 & 0 & 4 & & 0.071 & & $<$1 & $\times$ \\\cline{6-11}\cline{13-13}\cline{15-16}
           & p1 & SM & 1064 & 3 & Q4 & RO & RO & RO & RO & 1 & 0.068 & RO & $<$1 & RO & $\checkmark$ \\\cline{6-11}\cline{13-13}\cline{15-16}
           & & & & & Q5 & RO & RO & RO & RO & 3 & & RO & & RO & $\checkmark$ \\\cline{6-11}\cline{13-13}\cline{15-16}
           & & & & & Q7 & 1064 & 3 & 0 & 0 & 17 & & 5.263 & & 33.2 & $=$ \\\cline{6-11}\cline{13-13}\cline{15-16}
         & & & & & Q8 & RO & RO & RO & RO & 4 & & RO & & RO & $\checkmark$ \\ \cline{2-16}
               
         GD & & &  &  & Q1 & RO & RO & RO & RO & 3 & & RO & & RO & $\checkmark$ \\\cline{6-11}\cline{13-13}\cline{15-16}
         & & & & & Q2 & 1072 & 4 & 8 & 0 & 4 & & 14.508 & & 37.3 & $\checkmark$ \\\cline{6-11}\cline{13-13}\cline{15-16}
           & & & & & Q3 & 1001 & 4 & -63 & 0 & 4 & & 0.237 & & 32.0 & $\times$ \\\cline{6-11}\cline{13-13}\cline{15-16}
           & p2 & SM & 1064 & 4 & Q4 & RO & RO & RO & RO & 1 & 1.089 & RO & 32.7 & RO & $\checkmark$ \\\cline{6-11}\cline{13-13}\cline{15-16}
           & & & & & Q5 & RO & RO & RO & RO & 3 & & RO & & RO & $\checkmark$ \\\cline{6-11}\cline{13-13}\cline{15-16}
           & & & & & Q7 & RO & RO & RO & RO & 24 & & RO & & RO & $\checkmark$ \\\cline{6-11}\cline{13-13}\cline{15-16}
           & & & & & Q8 & RO & RO & RO & RO & 4 & & RO & & RO & $\checkmark$ \\
    \hline
           & & & & & Q1 & 18.003 & 4 & 0.001 & 1 & 8 & & 64.79 & & 2222.1 & $\checkmark$ \\\cline{6-11}\cline{13-13}\cline{15-16}
           & & & & & Q2 & 18.002 & 3 & 0 & 0 & 5 & & 42.06 & & 2224.2 & $=$ \\\cline{6-11}\cline{13-13}\cline{15-16}
           & & & & & Q3 & 18.003 & 4 & 0.001 & 1 & 5 & & 36.68 & & 2224.5 & $\checkmark$ \\\cline{6-11}\cline{13-13}\cline{15-16}
           & p1 & UP & 18.002 & 3 & Q4 & 18.003 & 4 & 0.001 & 1 & 1 & 42.304 & 28.38 & 2225.5 & 2213.1 & $\checkmark$ \\\cline{6-11}\cline{13-13}\cline{15-16}
           & & & & & Q5 & NP & NP & NP & NP & 4 & & 54.94 & & 2221.9 & $\checkmark$ \\\cline{6-11}\cline{13-13}\cline{15-16}
           & & & & & Q7 & 18.002 & 3 & 0 & 0 & 21 & & 37.67 & & 2241.3 & $=$ \\\cline{6-11}\cline{13-13}\cline{15-16}
         & & & & & Q8 & NP & NP & NP & NP & 7 & & 245.46 & & 2224.8 & $\checkmark$ \\\cline{2-16}
           
         PD & & & & & Q1 & 18.003 & 4 & 0.001 & 1 & 8 & & 114.90 & & 2225.5 & $\checkmark$ \\\cline{6-11}\cline{13-13}\cline{15-16}
        & & & & & Q2 & 18.002 & 3 & 0 & 0 & 5 & & 86.40 & & 2221.7 & $=$ \\\cline{6-11}\cline{13-13}\cline{15-16}
           & & & & & Q3 & 18.003 & 4 & 0.001 & 1 & 5 & & 65.83 & & 2224.6 & $\checkmark$ \\\cline{6-11}\cline{13-13}\cline{15-16}
           & p2 & UP & 18.002 & 3 & Q4 & 18.003 & 4 & 0.001 & 1 & 1 & 76.42 & 47.06 & 2223.3 & 2213.1 & $\checkmark$ \\\cline{6-11}\cline{13-13}\cline{15-16}
           & & & & & Q5 & NP & NP & NP & NP & 4 & & 82.80 & & 2224.7 & $\checkmark$ \\\cline{6-11}\cline{13-13}\cline{15-16}
           & & & & & Q7 & 18.002 & 3 & 0 & 0 & 21 & & 60.90 & & 2241.0 & $=$ \\\cline{6-11}\cline{13-13}\cline{15-16}
           & & & & & Q8 & NP & NP & NP & NP & 7 & & 237.11 & & 2224.6 & $\checkmark$ \\\cline{2-16}
           
    \hline
    \end{tabular}
  }
  \caption{\emph{BM} represents the problem domains in PDDL+,and \emph{Ins} represents the problem instances of a domain. \emph{PL} reports the hybrid system planner chosen for solving the instances, where SM, EN, and UP represent \textsc{SMTPlan+}, \textsc{ENHSP}, and \textsc{UPMurphi}, respectively. \emph{Dur} and \emph{Len} represent the original plan-duration and plan length for a problem instance, \emph{QN} represents the set of contrastive questions presented in Section \ref{secCaseStudy}, \emph{H-dur} and \emph{H-len} represent the plan-duration and the plan-length of the hypothetical plan \emph{HPlan}. \emph{DC$_\tau$} and \emph{DC$_l$} represent the plan-duration difference and the plan-length difference of the HPlan with the original plan, respectively; these are the contrastive explanation (CE) metrics illustrated in Def \ref{def:CE}. \emph{CM} reports the number of constraint modifications that are made in the domain (inclusive of constraint addition and deletion). \emph{Time} and \emph{H-time} represent the plan generation time whereas \emph{Mem} and \emph{H-mem} report the memory usages for the original plan and the \emph{HPlan}, respectively.  \emph{HQ} represents an evaluation on the quality of the \emph{HPlan} against the original plan.}
  \label{ch2table: table4}
\end{table*}

\begin{figure}[htbp]
    \centering
      \begin{subfigure}[b]{0.32\textwidth}
         \centering
         \includegraphics[height=4cm, width=\textwidth]{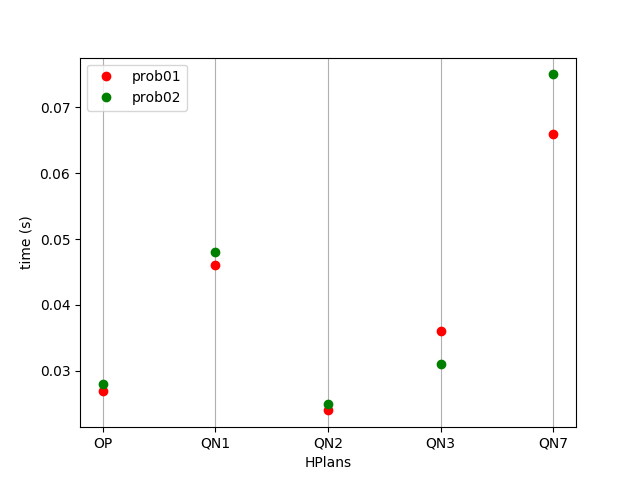}
         \caption{The \emph{HPlan} generation times using \textsc{SMTPlan+} for two problem instances in Car domain.}
         \label{fig:smt_time}
      \end{subfigure}
      \hfill
      \begin{subfigure}[b]{0.32\textwidth}
         \centering
         \includegraphics[height=4cm, width=\textwidth]{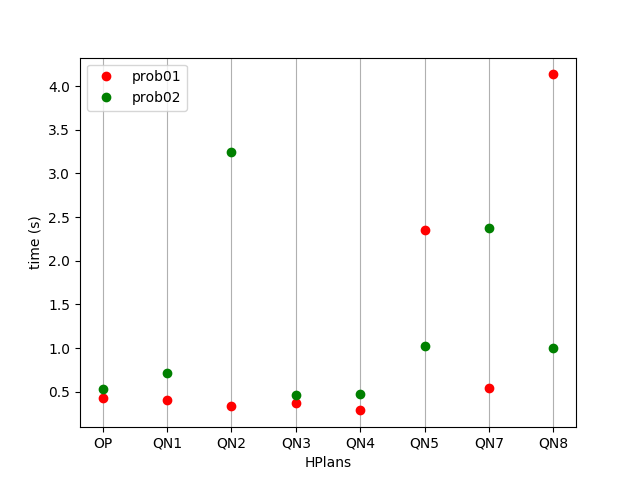}
         \caption{The \emph{HPlan} generation times using \textsc{ENHSP} for two problem instances in Car domain.}
         \label{fig:enhsp_time}
      \end{subfigure}
      \hfill
      \begin{subfigure}[b]{0.32\textwidth}
         \centering
         \includegraphics[height=4cm, width=\textwidth]{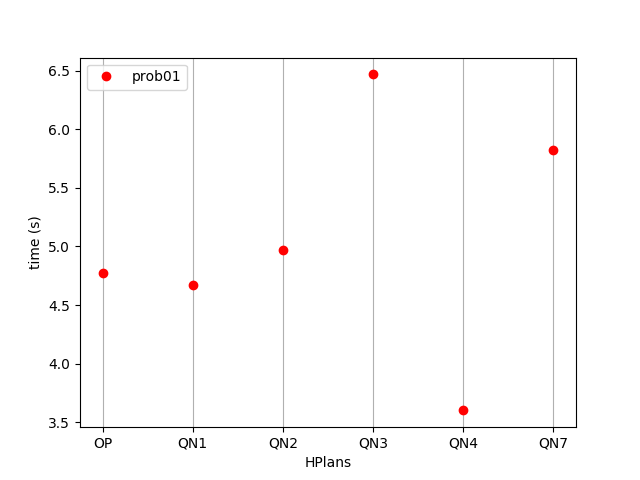}
         \caption{The \emph{HPlan} generation times using \textsc{UPMurphi} for problem instance 1 in Car domain.}
         \label{fig:upmurphi_time1}
      \end{subfigure}      
      \newline
      \begin{subfigure}[b]{0.32\textwidth}
         \centering
         \includegraphics[height=4cm, width=\textwidth]{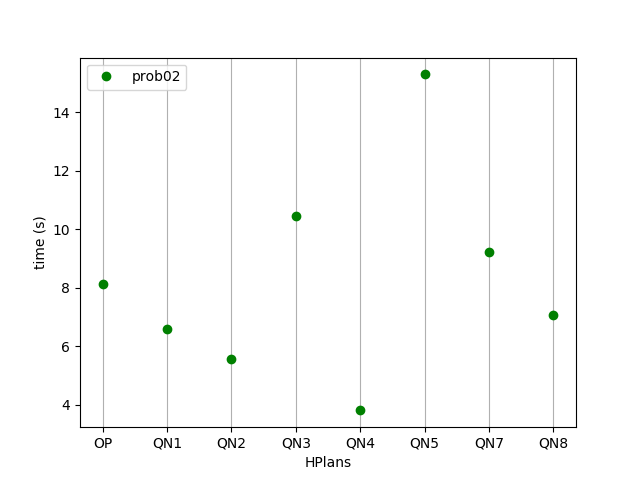}
         \caption{The \emph{HPlan} generation times using \textsc{UPMurphi} for problem instance 2 in Car domain.}
         \label{fig:upmurphi_time2}
      \end{subfigure}
      \hfill
      \begin{subfigure}[b]{0.32\textwidth}
         \centering
         \includegraphics[height=4cm, width=\textwidth]{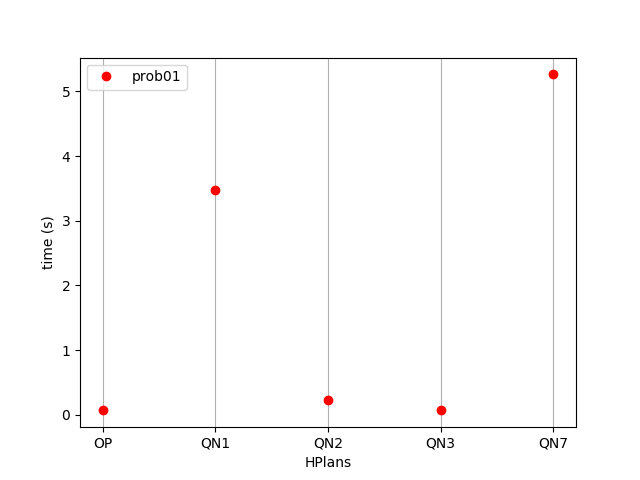}
         \caption{The \emph{HPlan} generation times using \textsc{SMTPlan+} for problem instance 1 in Generator-events domain.}
         \label{fig:smt_time1}
      \end{subfigure}
      \hfill
      \begin{subfigure}[b]{0.32\textwidth}
         \centering
         \includegraphics[height=4cm, width=\textwidth]{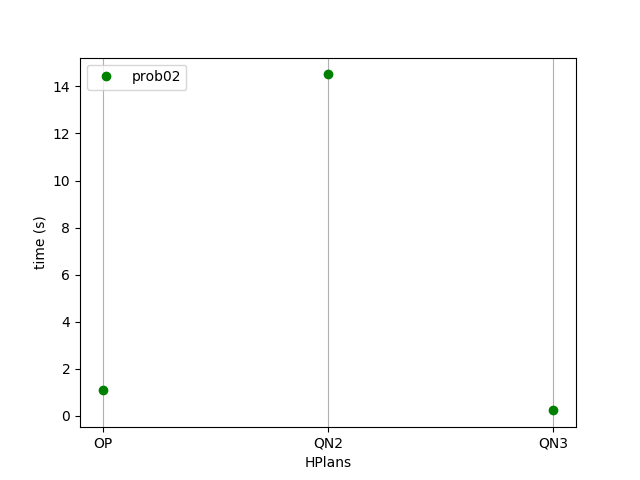}
         \caption{The \emph{HPlan} generation times using \textsc{SMTPlan+} for problem instance 2 in Generator-events domain.}
         \label{fig:smt_time2}
      \end{subfigure}
      \newline
      \begin{subfigure}[b]{0.32\textwidth}
         \centering
         \includegraphics[height=4cm, width=\textwidth]{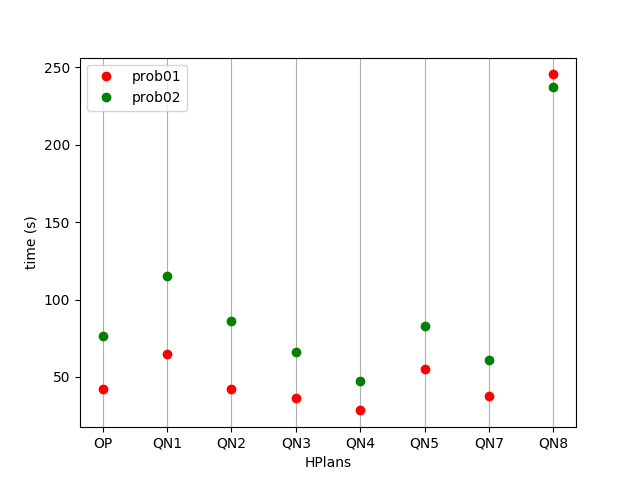}
         \caption{The \emph{HPlan} generation times using \textsc{UPMurphi} for instances in Planetary-lander domain.}
         \label{fig:upmurphi_time}
      \end{subfigure}
    \caption{Scatter graph comparing the \emph{HPlan} generation time over the original plan. \emph{OP} represents the original plan whereas \emph{QN1}, \emph{QN2}, \emph{QN3}, \emph{QN4}, \emph{QN5}, \emph{QN7} and \emph{QN8} represent the \emph{HPlans} generated against the compilations for the contrastive questions discussed in Sec. \ref{secA}, \ref{secB}, \ref{secC}, \ref{secD}, \ref{secE}, \ref{secG} and \ref{secH}, respectively.}
    \label{fig:time_plot}
\end{figure}

\begin{figure}[htbp]
    \centering
      \begin{subfigure}[b]{0.32\textwidth}
         \centering
         \includegraphics[height=4cm, width=\textwidth]{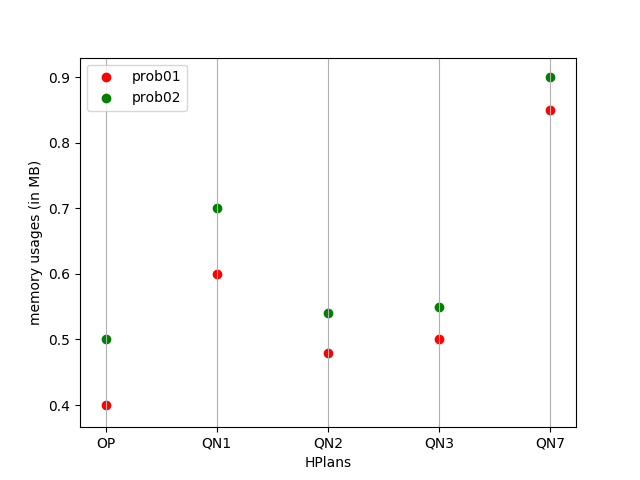}
         \caption{The memory usage in \emph{HPlan} generation using \textsc{SMTPlan+} for prob-instances in Car domain.}
         \label{fig:smt_mem}
      \end{subfigure}
      \hfill
      \begin{subfigure}[b]{0.32\textwidth}
         \centering
         \includegraphics[height=4cm, width=\textwidth]{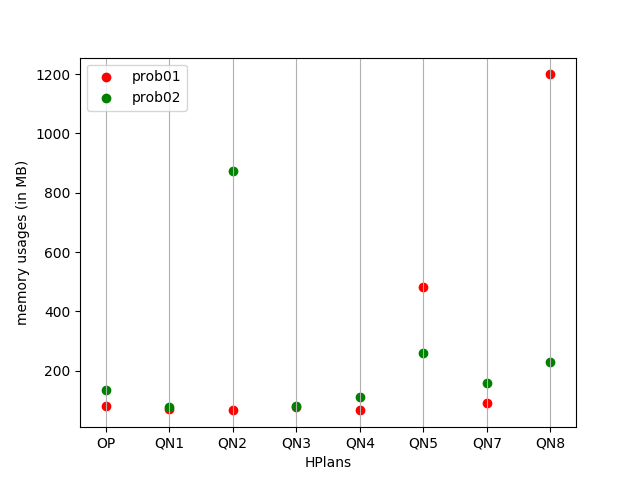}
         \caption{The memory usage in \emph{HPlan} generation using \textsc{ENHSP} for prob-instances in Car domain.}
         \label{fig:enhsp_mem}
      \end{subfigure}
      \hfill
      \begin{subfigure}[b]{0.32\textwidth}
         \centering
         \includegraphics[height=4cm, width=\textwidth]{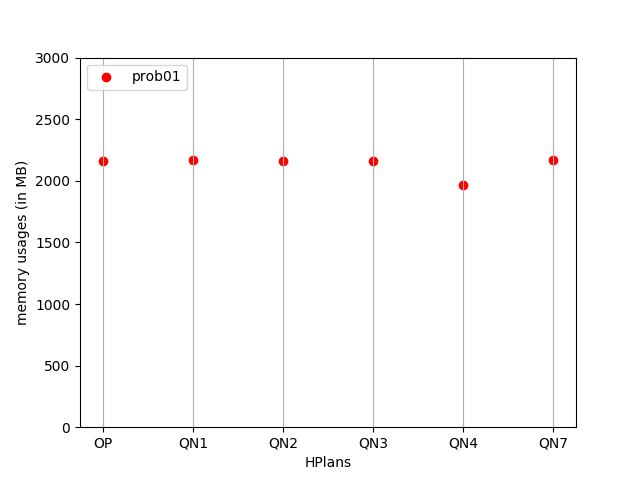}
         \caption{The memory usage in \emph{HPlan} generation using \textsc{UPMurphi} for prob-ins 1 in Car domain.}
         \label{fig:upmurphi_mem1}
      \end{subfigure}
      \newline
      \begin{subfigure}[b]{0.32\textwidth}
         \centering
         \includegraphics[height=4cm, width=\textwidth]{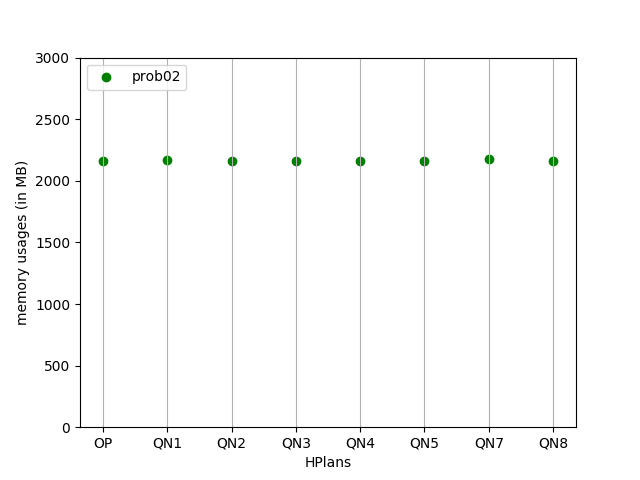}
         \caption{The memory usage in \emph{HPlan} generation using \textsc{UPMurphi} for prob-ins 2 in Car domain.}
         \label{fig:upmurphi_mem2}
      \end{subfigure}
      \hfill      
      \begin{subfigure}[b]{0.32\textwidth}
         \centering
         \includegraphics[height=4cm, width=\textwidth]{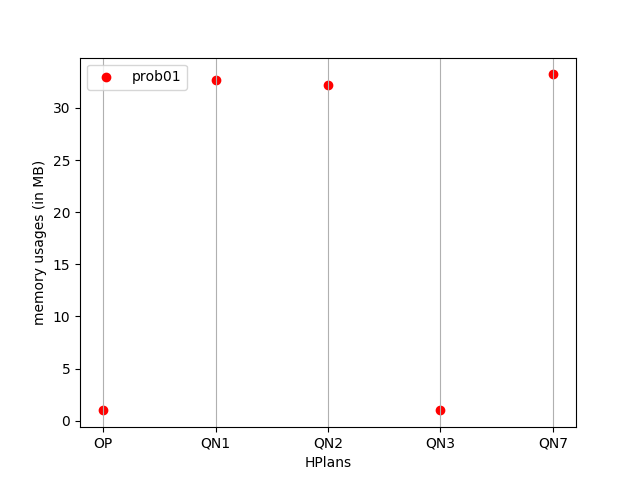}
         \caption{The memory usage in \emph{HPlan} generation using \textsc{SMTPlan+}for prob-ins 1 in Generator-events.}
         \label{fig:smt_mem1}
      \end{subfigure}
      \hfill
      \begin{subfigure}[b]{0.32\textwidth}
         \centering
         \includegraphics[height=4cm, width=\textwidth]{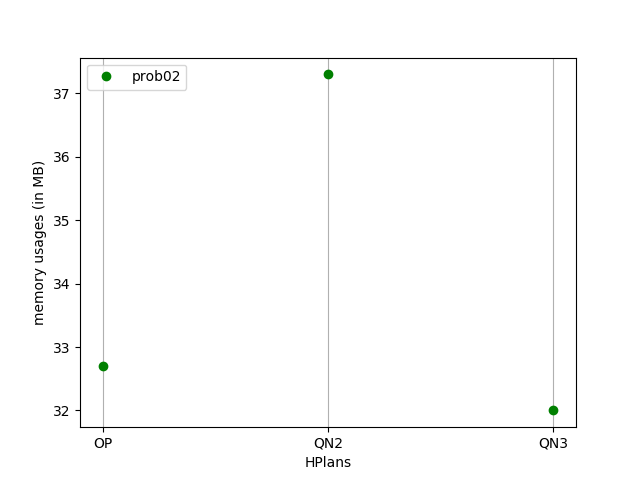}
         \caption{The memory usage in \emph{HPlan} generation using \textsc{SMTPlan+} for prob-ins 2 in Generator-events.}
         \label{fig:smt_mem2}
      \end{subfigure}
      \newline
      \begin{subfigure}[b]{0.32\textwidth}
         \centering
         \includegraphics[height=4cm, width=\textwidth]{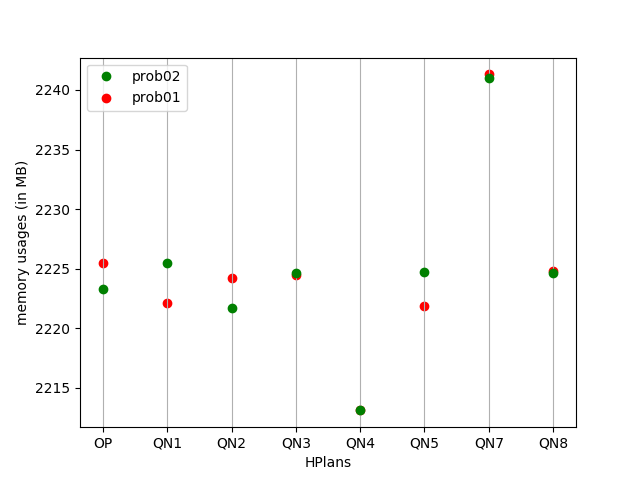}
         \caption{The memory usage of the \emph{HPlan} generation by \textsc{UPMurphi} for instances in Planetary-lander.}
         \label{fig:upmurphi_mem}
      \end{subfigure}
    \caption{Scatter graph comparing the memory usages by the planner for the generation of \emph{HPlans} and the original plan. \emph{OP} represents the original plan whereas \emph{QN1}, \emph{QN2}, \emph{QN3}, \emph{QN4}, \emph{QN5}, \emph{QN7} and \emph{QN8} represent the \emph{HPlan} corresponding to the contrastive questions addressed in Section \ref{secA}, \ref{secB}, \ref{secC}, \ref{secD}, \ref{secE}, \ref{secG} and \ref{secH}, respectively.}
    \label{fig:mem_plot}
\end{figure}

\noindent \paragraph{Performance Evaluation:} We compare the time and memory usage to generate the \emph{HPlan} with the time and memory usage to generate the original plan by the planners \textsc{SMTPlan+}, \textsc{ENHSP}, and \textsc{UPMurphi}, presented in Figure \ref{fig:time_plot} and Figure \ref{fig:mem_plot}, respectively. The y-axis represents time (in seconds) in Figure \ref{fig:time_plot} and memory usage (in MB) in Figure \ref{fig:mem_plot}, while the x-axis represents the alternate plans generated for the contrastive questions. \emph{OP} represents the original plan and \emph{QN1} to \emph{QN8} denote the questions addressed in Section \ref{secA} through \ref{secH} for which the alternate plan is generated. We have included only those planning instances where a valid plan is generated or the instance is reported as unsolvable by the planner. The original plan and the \emph{HPlans} for problem instance 1 are marked in Red, whereas the instances of problem instance 2 are marked in Green.
We can conclude that the time and memory usage to generate the contrastive alternate plans (\emph{HPlans}) by the planners are similar to the time and memory usage to generate the original plan, except in some of the cases. The contrastive explanation framework does not, therefore, incur much performance overhead. The scalability of the framework is dependent on the scalability of the planner used for plan generation. Note that the \emph{HPlan} generation for \emph{QN8} for problem instance 1 in Figure \ref{fig:enhsp_time} and \emph{QN8} for problem instance 1 and 2 in Figure \ref{fig:upmurphi_time} take considerable time than the generation time of the original plan. The \emph{HModels} corresponding to these problem instances generate no valid plans. Deducing whether there exist no plans is more performance-intensive than generating a plan for a problem instance, which justifies the added time taken to solve these instances.
\emph{QN7} for the problem instance 1 in Figure \ref{fig:smt_time1} represents the compilation that questions an action sequence in a plan, which imposes many constraints and additional actions in the domain, which may be the cause for taking the additional time to find a solution in this instance.
For similar reasons, \emph{QN8} for problem instance 1 in Figure \ref{fig:enhsp_mem}, and \emph{Q7} for problem instances 1 and 2 in Figure \ref{fig:upmurphi_mem} require additional memory in solving the planning instances.
The experiments were performed on a machine with 8 GB RAM, Intel Core i5-8250U@1.60GHz, 8 8-core processor with Ubuntu 18.04 64-bit OS. All the example domains, the problem files, and the constructed hypothetical models (\emph{HModels}) for the contrastive questions can be found at: \url{https://gitlab.com/Sazwar/contrastive-explanations}.

\subsection{Performance Evaluation of Proving Absence of Plan}

\noindent To prove the absence of a plan with bounded reachability analysis in our running example, we present the experimental observations with \textsc{dReach}. We set a $\delta$ to a reasonably small value of $0.01$, and $k$ is assigned a value of $10$ and $20$ in 2 separate experiments. For all the experiments, \textsc{dReach} reports the no-plan instances as unsatisfiable, which implies that there is no plan for these problem instances within the plan-depth $k$. The reachability analysis results of the no-plan-models are shown in Table \ref{ch2tab:table3}, where \emph{prob\_ins} represents the \textsc{dReach} encoding of the corresponding problem instances presented in Section \ref{secD}, \ref{secE} and \ref{secH}, \emph{plan-depth} is the bound on the plan length $k$, \emph{time} represents the time taken by \textsc{dReach} to complete the reachability analysis and \emph{satisfiability} reports whether the problem instance is reported as satisfiable by \textsc{dReach} (where SAT and UNSAT indicates satisfiable and unsatisfiable respectively). The planning problem instance of Section \ref{secE} did not terminate within $48^+$ hrs for a plan-depth of more than 13, and thus we could not draw any conclusion from this instance beyond a depth of 13.
\begin{table}[htbp]
    \centering
    \begin{tabular}{|c|c|c|c|}
    \hline
         prob\_ins & bound on plan-depth (k)  & time (sec) & satisfiability \\
    \hline
         \multirow{2}{*}{Sec. \ref{secD}} & 10 & 0.288 & UNSAT \\
               & 20 & 1.277 & UNSAT \\
    \hline
         \multirow{2}{*}{Sec. \ref{secE}} & 10 & 541.423 & UNSAT \\
               & 13 & 4010.524 & UNSAT \\
    \hline
         \multirow{2}{*}{Sec. \ref{secH}} & 10 & 1.931 & UNSAT \\
               & 20 & 8.152 & UNSAT \\
    \hline
    \end{tabular}
    \caption{Bounded reachability analysis of the no-plan-models for a given $\delta$ perturbation of 0.01.}
    \label{ch2tab:table3}
\end{table}

All our no-plan-models were unsatisfiable (UNSAT)
in $H^{\delta}$ which proved that there are no plans within the plan-depth $k$ in the corresponding $H$ system. However, the same can not be asserted if a problem instance is observed to be satisfiable (SAT) in the corresponding $H^{\delta}$ model for a given $\delta$ and $k$. To cover such issues, we propose an algorithm that iteratively lowers the value of $\delta$ and checks the reachability of the problem upto a certain precision (Algorithm \ref{algo:algorithm}). The algorithm takes an \emph{HModel} and plan-depth $k$ as inputs. Initially, we set the precision $\delta$ to be 0.01. It then checks the satisfiability of \emph{HModel} in the corresponding $\delta$-perturbed hybrid system and iteratively lowers the $\delta$ precision (upto $10^{-6}$) when the problem is SAT. If the problem instance becomes UNSAT, it concludes that there is no plan. However, the explanation of the non-existence of a plan remains inconclusive when the algorithm returns SAT for the $\delta$ perturbation less than $10^{-6}$.

\begin{algorithm}[htbp]
  \caption{\textsf{algorithm to prove absence of plan by bounded reachability analysis}}
  \label{algo:algorithm}
  \LinesNumbered
  \SetAlgoLined
  \KwIn{\emph{HModel}, \textit{plan-depth} $k$.\\}
  \Begin{
  $\delta$ = $0.01$;  \tcc*[f]{Initialize the value of $\delta$}\\
    \While(\tcc*[f]{Checking the value of $\delta$}){($\delta \neq 10^{-6}$)}{
        $res$ = \textsc{dReach}(\emph{HModel}, $\delta$, $k$); \tcc*[f]{\textsc{DReach} call}\\
        \eIf(\tcc*[f]{Satisfiability check}){$res$ is SAT}{
        $\delta$ = $\delta*0.1$; \tcc*[f]{Lowering the value of $\delta$ by 0.1}\\ }
        {Print "Goal state unreachable from the initial state in the domain";\\
        return;}
    }
    Print "Cannot explain the non-existence of a plan";\\
    return;
  }
\end{algorithm}



\section{Related Works} \label{sec6}
\noindent 
A roadmap for explainable artificial intelligence planning with contrastive questions such as “Why did the planner do A rather than B?” or simply “why” and “why not” questions have been addressed in (\cite{2017arXiv170910256F}) where the authors discuss things that need to be explained, features of planning that facilitate explanations and how to achieve the goal of providing reasonable answers to user questions through the explanations. The work in (\cite{2018arXiv181103163M}) extended the structural causal model approach to the case of contrastive explanations and defined contrastive explanations for two types of questions: alternative questions and congruent questions, where the author argues that an alternative question of the form “Why A rather than B?” is a contrastive question. (\cite{10.1145/1518701.1519023}) shows that these why and why not questions benefit in terms of objective understanding and feelings of trust. A detailed overview of the explainable artificial intelligence planning landscape and different terms used in this domain are introduced in (\cite{2018arXiv181109722C}), while (\cite{DBLP:conf/ijcai/ChakrabortiSZK17}) considered explanation as a model reconciliation problem assuming that the agent and the human may have possibly different models of the environment. In such a scenario, the agent explains those actions to the human that are not expected to be executed, taking the human model as a reference. Explanation here can be seen as a reconciliation between the agent and the human. In (\cite{DBLP:conf/nips/DhurandharCLTTS18}), a method, called Contrastive Explanations Method (CEM) was proposed to generate contrastive explanations for differentiable models such as deep neural networks, where one has complete access to the model. In (\cite{2019arXiv190600117D}), Model Agnostic Contrastive Explanations Method (MACEM) was proposed to generate contrastive explanations for any classification model where one is able to only query the class probabilities for a desired input which allows them to generate contrastive explanations for not only neural networks but models such as random forests, boosted trees and even arbitrary ensembles that are still among the state-of-the-art when learning on structured data. In (\cite{10.5555/2900929.2901038}), Smith put forward the challenge of planning as an iterative process for better modeling preferences and providing explanations. In our paper, we have also taken this concept of iterative explanations for our model. While improving the user’s level of understanding and building trust in the system are the main purposes of these explanations, they can be local (regarding a specific plan) or global (concerning how the planning system works in general). (\cite{krarupPDDL2.1CE}) focus on local explanations of temporal and numeric planning problems, introducing a formal description of the compilation from user questions to constraints in a PDDL2.1 planning setting, and explaining why a planner has made a certain decision. In contrast, our paper applies a similar approach to both local and global planning scenarios for modeling mixed discrete-continuous planning problems using the PDDL+ planning setting. (\cite{cashmore2019towards}) presents a prototype framework to facilitate Explainable Planning as a service, limited to discrete systems. Our framework is built as a plug-in on top of \textsc{SMTPlan+}, the state-of-the-art planner for hybrid system domains.

A different direction to explainable AI planning is explaining the unsolvability of a planning problem when there is no-plan. There are notable works in this direction trying to explain the unsolvability of a planning problem.
(\cite{DBLP:conf/aips/GobelbeckerKEBN10}) argues that excuses can be produced for why a plan cannot be found, and the work proposes a formalization of counterfactual alterations to the original planning task such that the new planning task will be solvable. Based on that, they provide an algorithm to find these excuses in a reasonable time.
(\cite{DBLP:conf/ijcai/SreedharanSSK19}) uses hierarchical model abstractions to generate the reason for the unsolvability of planning problems. This hierarchical model abstraction relaxes a planning problem until a solution can be found. Then, they look for landmarks of this relaxed problem that cannot be satisfied in less relaxed versions of the problem. The unsatisfiability of these landmarks provides a succinct description of critical propositions that cannot be satisfied.
(\cite{DBLP:conf/aaai/EiflerC0MS20}) has taken a somewhat different approach by deriving properties that must be exhibited by all possible plans that could serve as explanations in case of unsolvability.

We have observed that some of the contrastive questions result in no-plans in our example domain, which serves as the motivation to verify the unsolvability of those no-plan problem instances. Though similar work on verifying the unsolvability of the planning problem has been done previously (\cite{DBLP:conf/aips/ErikssonH20}, \cite{DBLP:conf/aips/ErikssonRH17}), these are on discrete domains. Verifying the unsolvability of the planning problems in hybrid systems comes with an additional challenge since the planning problem is undecidable in general hybrid domains. In this work, we attempt to prove the non-existence of a plan in the hybrid domain by reachability analysis. We do not provide explanations for the non-existence of a plan in this paper. Addressing this issue is left for future work.

\section{Limitation} \label{sec7}
In the car domain, we observe that \textsc{SMTPlan+} does not guarantee an optimal plan generation, and the plan produced is rather just one among the many possible plans for a given problem instance in the domain. Due to this, some of the conclusions may be weak, which are based on comparing the hypothetical plan with the original one. In the no-plan explanation framework, we check the existence of a plan using $\delta$-reachability analysis of \textsc{DReach} of the no-plan model upto a given plan-depth $k$. The non-existence of a plan for a problem instance upto $k$ depth does not imply that there does not exist a plan with length $>k$. However, the choice of $k$ has significance on the usability of a plan in real-world applications, and concluding the non-existence of plans of bounded length can be insightful to a user.

\section{Conclusion} \label{sec8}
\noindent In this chapter, we explore a set of contrastive questions that a user of a planning tool may raise, and we propose a \emph{re-model and re-plan} framework to provide explanations to such questions. Specifically, given a hybrid system model in PDDL+ and a plan describing the set of desirable actions on the same to achieve a destined goal, our framework can integrate contrastive questions in PDDL+ and synthesize alternate plans using a hypothetical model (HModel) constructed by imposing constraints drawn from the questions. We present a detailed case study on our approach, and with a comparison metric, we compare the original plan with the alternate ones. We show that our contrastive explanations can draw conclusions about the planning domain and the planning tool as well, such as identifying that the plan is not always necessarily cost-optimal, that one or some actions need to appear later in the plan, etc. Even the no-plans are helpful to figure out the critical actions for a certain problem.
Further, we provide a no-plan explanation algorithm for our no-plan models through bounded reachability analysis to verify the reachability of the problem instances.


We demonstrate experimental results on three planning domains using three state-of-the-art planners.
We believe our frameworks can be of immense importance to the hybrid systems planning community for synthesizing better, explainable plans.





\chapter{A Contrastive Explanation Tool for Plans in Hybrid Domains} \label{ch3}

\textbf{\textsc{Chapter Abstract:}} \textit{This work presents a web-based tool for generating contrastive explanations of plans in hybrid domains. The tool offers a collection of contrastive questions over a plan, such as "Why use action A?", for users to select. An explanation of the user question is produced by contrasting the original plan against an alternative that meets the user's expectation implicit from the question. The tool has the provision to contrast with the best alternative that the underlying planner can generate in terms of plan makespan (duration of a plan) and length. The current version supports two planners in the hybrid domain, namely SMTPlan+, and ENHSP, which a tool user can select. The tool consists of (1) A web-based interactive GUI for selecting questions, viewing contrastive plans and the generated explanations, and (2) A back-end implementing an iterative re-modeling and re-planning algorithm. We demonstrate the working of the tool over two case studies.}

\section*{}
\lettrine[lraise=0.3, nindent=0em, slope=-.5em]
There has been a surge of applications where automated planners are deployed and where humans and autonomous agents collaborate to achieve a desired objective. Ensuring the safety, robustness, and trustworthiness of such systems is of utmost importance.
Explainable Artificial Intelligence Planning (XAIP) (\cite{DBLP:conf/ijcai/ChakrabortiSK20,DBLP:journals/corr/abs-1709-10256,DBLP:conf/rweb/HoffmannM19,7989155}) is a promising field of research that aims to provide AI planning systems the ability to explain the rationale behind a decision of an autonomous agent to engage in trustworthy collaborations with humans. A roadmap for XAIP is proposed in (\cite{DBLP:journals/corr/abs-1709-10256}) where the authors discuss a taxonomy of user questions that should be addressed, things that need to be explained, features of planning that facilitate explanations and how to achieve the goal of providing reasonable answers to user questions through the explanations.
When there is a mismatch between a plan obtained from an automated planner and the user's expectations, reconciliations are often required. (\cite{DBLP:journals/corr/abs-1902-01876}) has shown that users tend to ask "why" questions when seeking explanations about a specific part of the plan, referred to as a \emph{local question}, while "how" or "what" questions are asked when seeking explanations about the plan as a whole, referred to as \emph{global questions}.
Insights from social sciences (\cite{DBLP:journals/ai/Miller19}) suggest that "why" questions are often contrastive, taking the form of "Why A rather than some B?" 
Based on these observations, (\cite{DBLP:journals/corr/abs-1709-10256}) demonstrates that when the planning domain is well-known to the user, it is more common to ask more local, contrastive "why" questions than global "how" or "what" questions. A contrastive question can be addressed with a contrastive explanation by virtue of its ability to highlight the differences between the original plan and an alternative plan that accommodates the user's suggested alterations. This mode of explanation has proven to be highly effective in enhancing comprehension and is also simpler to execute than a full-fledged causal analysis (\cite{DBLP:journals/jair/KrarupKMLC021}). Furthermore, contrastive explanations have several comparative advantages, such as enabling a clear and direct comparison between the planner's given plan and the user's conceived plan. The paradigm of contrastive explanations has been explored in the context of planning domains having discrete state-transition representations (PDDL2.1~\cite{DBLP:journals/jair/FoxL03}) (\cite{cashmore2019towards}). In our earlier work (\cite{Sarwar2022Contrastive}), we presented a general framework for contrastive explanation of plans in domains represented as discrete state-transition systems, together with continuous variables with their evolution given as differential equations. Such domains are known as \emph{hybrid domains} in literature and modeled with PDDL+~(\cite{Cashmore2016Compilation}).

\noindent In this work, we present an interactive web interface for a contrastive explanation of plans in hybrid domains. The explanation algorithms implemented in the tool are based on the work in (\cite{Sarwar2022Contrastive}). It offers a user-friendly interface that can be used by AI practitioners for explainable planning. 
The key contributions of this work are:
\begin{itemize}
    \item A web-based interactive tool interface supporting XAIP in hybrid domains.
    \item An iterative re-modeling and re-planning algorithm to find a competitive contrastive plan with respect to the comparison metrics (e.g., plan makespan, plan length) that the underlying planner can provide.
    \item A provision to experiment with two planners for hybrid domains and compare and contrast the explanations produced thereof.
\end{itemize}

\noindent \textbf{Note on Collaboration and Authorship:} This work was developed in collaboration and has been previously published in \cite{DBLP:conf/indiaSE/DeySRB24}. For this thesis, the framework design and the algorithmic backbone of the tool are the primary contributions. The first author primarily implemented and carried out experimental validation.


The rest of this chapter is organized as follows. In Section~\ref{sec:architecture}, we provide an overview of our framework. Section~\ref{sec:case_study} presents implementation and results. Section~\ref{sec:ch3-related-work} discusses related works, and Section~\ref{sec:ch3-conclusion} concludes this work.
\section{Tool Overview} \label{sec:architecture}
\begin{figure}[htbp]
    \centering
    \includegraphics[width= 0.5\textwidth]{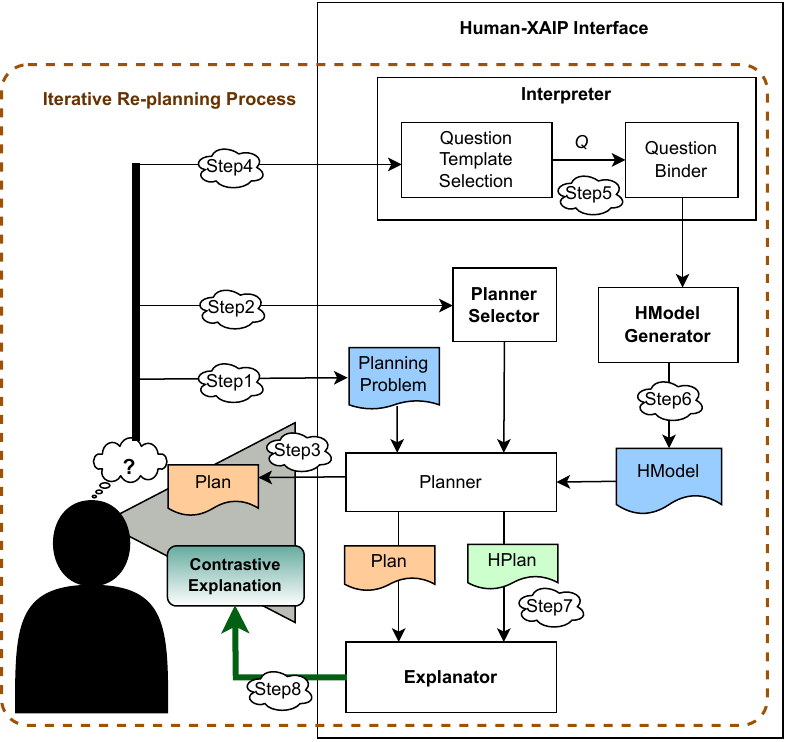}
    \caption{Tool Workflow}
    \label{fig:framework_architecture}
\end{figure}

We first present the workflow of the tool as shown in Fig \ref{fig:framework_architecture}. 

\noindent The tool takes as input a planning problem $\Pi$ (\cite{Sarwar2022Contrastive}) which comprises a pair $(Dom, Prob)$, where $Dom$ is a representation of the planning domain in PDDL+, and $Prob$ is a planning problem instance that constitutes an initial state and a goal condition. The planning problem $\Pi$ is fed to a hybrid system planner \textsc{SMTPlan+} (\cite{DBLP:journals/jair/CashmoreMZ20}) or \textsc{ENHSP} (\cite{DBLP:conf/ecai/ScalaHTR16})  based on the user's choice. The planner produces a plan $\phi$ which is presented to the user. This plan is referred to as the \emph{original plan}. A plan is a time-annotated sequence of actions, where the annotation denotes the time at which the action is to be applied (\cite{Sarwar2022Contrastive}). The user chooses a template question $Q$ on the original plan $\phi$. The tool provides a drop-down menu for choosing the question. Our tool offers the following set of seven templatized contrastive questions to the users: (a) Why did the planner choose to do action A and not B instead? (b) Why did the planner not choose to do action A later in the plan? (c) Why did the planner not choose to do action A earlier in the plan? (d) Why did the planner take action A in the plan, instead of not taking it? (e) Why not have fewer occurrences of action A in the plan? (f) Why is the accumulative duration of the plan not less? (g) Why is the length of the plan not less? 
This set of questions is not exhaustive, but we believe these questions are important in these domains.
Once the question is chosen, the user is prompted to bind the free variables in the templatized question to the corresponding values in the plan context. The original planning problem $\Pi$ is then altered to a hypothetical planning problem (\emph{HModel}) $\Pi^{'}$ based on the constructions proposed in (\cite{Sarwar2022Contrastive}) implemented in the HModel Generator module. The hypothetical planning problem $\Pi^{'}$ is fed to the same planner chosen in Step-2, and a hypothetical plan (\emph{HPlan}) $\phi^{'}$ as expected by the user is generated. 
The tool interface displays the HPlan $\phi^{'}$ along with a contrastive explanation by comparing it with the original plan $\phi$. The tool supports the following comparison metrics: (a) Plan makespan and (b) Plan length.
A shorter makespan and length of a plan signify that the goal can be achieved relatively quickly with less number of applied actions, which may incur costs.

We now discuss the major building blocks of the tool in the following subsection.

\subsection{Module Description} \label{module_description}

\begin{figure}[tbh]
 \begin{subfigure}{\textwidth}
   \centering
   \subfloat[Loading planning problem and choosing a planner.]{%
   \fbox{\includegraphics[width=0.9\textwidth]{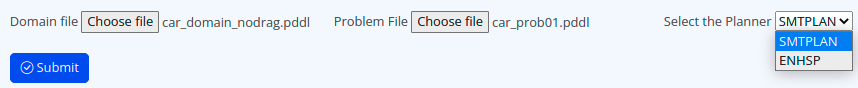}}
   \label{fig:select_planner}}
 \end{subfigure}      
 \newline
 \newline
 \newline
 \hfill
 \begin{subfigure}{\textwidth}
   \centering
   \subfloat[Observing original plan and selecting a templatized question.]{%
   \fbox{\includegraphics[width=0.9\textwidth]{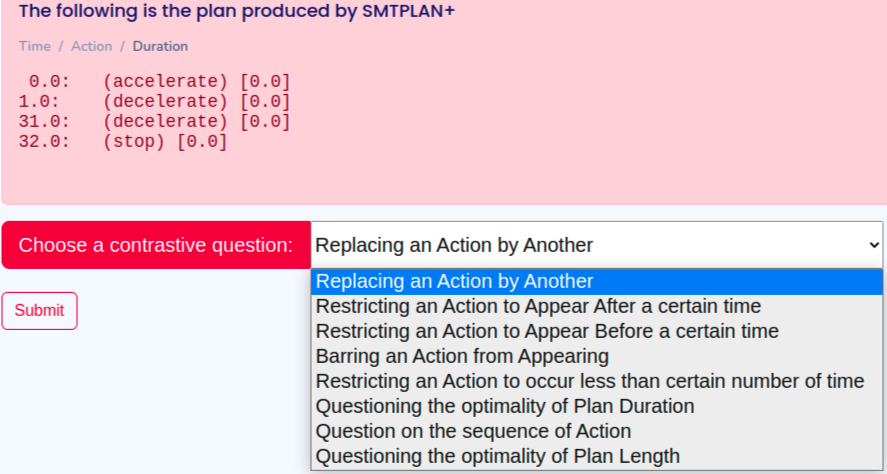}}
   \label{fig:select_question}}
 \end{subfigure}
 \caption{A snapshot of the tool's GUI}
 \label{fig:xaip_gui}
\end{figure}

\paragraph{Planner Selector:} A user can choose one among these 2 state-of-the-art hybrid system planners \textsc{SMTPlan+} and \textsc{ENHSP} via the tool interface as shown in Figure \ref{fig:select_planner}.
We try to make this interface generic for any planner that is capable of dealing with PDDL+ syntax and can be plugged in as a plan generation engine.
Consider a planning domain that models the motion of a car. The function symbols represent the distance, velocity, and acceleration of the car, while the predicates define the states of the car. The goal is to safely traverse a distance of 30 units within 50 units of time. The planning domain represented in PDDL+ is shown in Listing \ref{ch3lst:Listing1} and a problem instance that we have used in this domain is shown in Listing \ref{ch3lst:Listing2}. A detailed description of the domain can be found at \url{https://github.com/manabjamin2nadved1947/XAIP.git}.

\begin{lstlisting}[basicstyle=\scriptsize,caption={The car domain in PDDL+.}, label={ch3lst:Listing1}]
(define (domain car)
(:predicates (running) (engineBlown) (goalReached))
(:functions (d) (v) (a) (upLimit) (downLimit)
  (runningTime))
(:process moving
  :parameters ()
  :precondition (and (running))
  :effect (and (increase (v) (* #t (a))) (increase (d)
    (* #t (v))) (increase (runningTime) (* #t 1))))
(:action accelerate
  :parameters()
  :precondition (and (running) (< (a) (upLimit)))
  :effect (and (increase (a) 1)))
(:action decelerate
  :parameters()
  :precondition (and (running) (> (a) (downLimit)))
  :effect (and (decrease (a) 1)))
(:event engineExplode
  :parameters ()
  :precondition (and (running) (>= (a) 1) (>= (v) 100))
  :effect (and(not(running))(engineBlown)(assign (a) 0)))
(:action stop
  :parameters()
  :precondition (and(=(v)0)(>=(d) 30)(not(engineBlown)))
  :effect(goalReached)))
\end{lstlisting}

\begin{lstlisting}[basicstyle=\scriptsize,caption={The planning problem in the car domain in PDDL+.}, label={ch3lst:Listing2}]
(define (problem car_prob)
(:domain car)
(:init (running) (= (runningTime) 0) (= (upLimit) 1)
  (= (downLimit) -1) (= d 0) (= a 0) (= v 0))
(:goal (and (goalReached) (not (engineBlown))
  (<= (runningTime) 50))))
\end{lstlisting}

\noindent A domain and problem file is entered (Fig.~\ref{fig:select_planner}) alongside a selected planner ({SMTPlan+}), which returns a plan as follows:

\lstset{
    escapeinside={(@}{@)}
}
\begin{lstlisting}[basicstyle=\scriptsize,caption={A plan generated by \textsc{SMTPlan+} on car domain on the problem in Listing \ref{ch3lst:Listing2}. }, label={lst:Listing3}]
    (@\textbf{\underline{Time}}@)         (@\textbf{\underline{Action}}@)              (@\textbf{\underline{Duration}}@)
     0.0:      (accelerate)          [0.0]
     1.0:      (decelerate)          [0.0]
     31.0:     (decelerate)          [0.0]
     32.0:       (stop)              [0.0]
    (@\textbf{\emph{makespan}:} = 32.0 units.@)
\end{lstlisting}

\begin{figure}[htbp]
        \centering
        \fbox{\includegraphics[width= 0.9\textwidth]{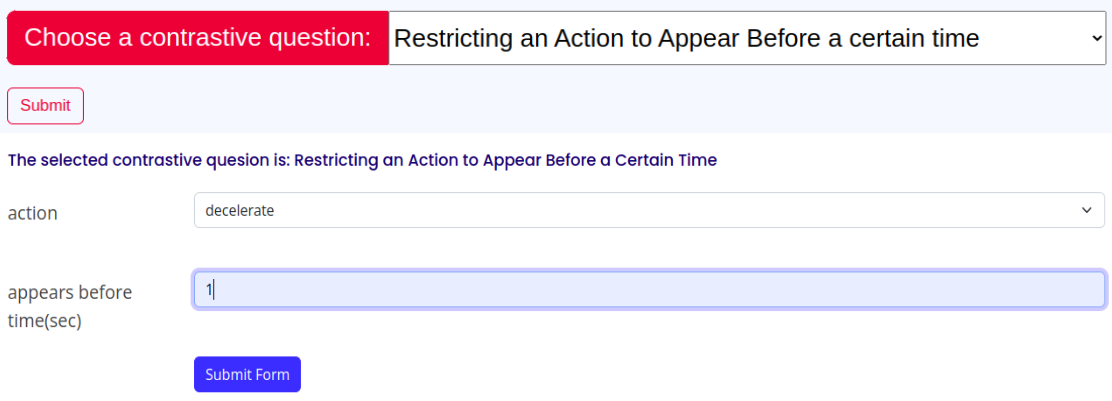}}
        \caption{Binding templatized question with instances.}
        \label{fig:binding_question}
\end{figure}

\paragraph{Interpreter:}
This module involves transforming the user's contrastive inquiry into a formalized question. This is accomplished by feeding the user-chosen templatized contrastive question together with template bindings of actions, time-instances, and frequency of actions of the domain, leading to the generation of a concrete question.
In the plan presented in Listing \ref{lst:Listing3}, a user may wish to ask {\em why the planner chose to decelerate at the $1^{st}$ time unit rather than not earlier}, anticipating that it would have yielded a better plan. We take this question as the running example to illustrate the rest of the blocks. By selecting the third templatized question from the menu as shown in Figure \ref{fig:select_question}, that is "Restricting an Action to Appear Before a Certain Time", the user is redirected to an input form to specify the relevant action and the time to bind with the placeholders in the question as shown in Figure \ref{fig:binding_question}. After gathering the user input, the question binder generates a concrete question. 
For example, when the user questions the use of the \emph{decelerate} action at time instant $1.0$ rather than earlier, the question binder concretizes the template question with the action instance \emph{decelerate} and a time instance $t$ which is less than 1.

\paragraph{HModel generator:}
This module automates the generation of HModel by receiving the concrete question generated by the Interpreter and the original planning problem as inputs. It applies pattern-matching techniques with arguments specified as regular expressions (regex) and utilizes string manipulation to compile a set of constraints in PDDL+ to generate the HModel.
In our example, the domain is updated by introducing a new action called \emph{decelerate\_new}, such that any valid plan must contain this action, which always appears before the time instance 1.0 unit (See Listing 4).

\begin{lstlisting}[basicstyle=\scriptsize,caption={The update in the domain with the \emph{decelerate\_new} action.}, label={lst:Listing4}]
...
(:action decelerate_new
:parameters()
:precondition (and (< (running_time) 1) (running)
  (> (a) (down_limit)))
:effect (and (do_before_new) (decrease (a) 1)))
...
\end{lstlisting}

\begin{figure}[htbp]
        \centering
        \fbox{\includegraphics[width= 0.9\textwidth]{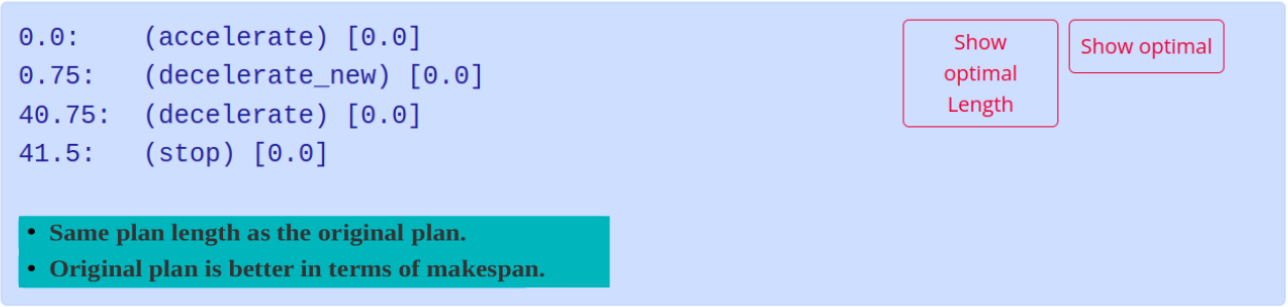}}
        \caption{An HPlan generated by \textsc{SMTPlan+} according to the contrastive question.}
        \label{fig:HPlan}
\end{figure}

\noindent The goal state in the problem file is updated with additional requirements (Listing 5):

\begin{lstlisting}[basicstyle=\scriptsize,caption={The modified goal state with the \emph{do\_before\_new} predicate.}, label={lst:Listing5}]
(:goal (and(@\textbf{(do\_before\_new)}@) (goal_reached) (not(engineBlown))
  (<= (running_time) 50) (transmission_fine)))
\end{lstlisting}

\paragraph{Explanator:}
After an HPlan is generated by the planner from the HModel, it is passed to this module along with the original plan for generating a contrastive explanation. The quality of the hypothetical plan is compared with the original plan with respect to the metrics of makespan and plan length (the number of actions appearing in the plan). When the makespan as well as the length of HPlan is longer than that of the original plan, the tool reports this to the user as a justification for using the original plan instead of the alternate user-expected hypothetical plan. On the other hand, if either the makespan or the plan length of the hypothetical plan is better than that of the original plan, the tools report this to the user, in which case a replan may be considered.
The following HPlan (see Figure \ref{fig:HPlan}) is received for the contrastive question of our running example.

Upon comparison with the original plan as shown in Listing \ref{ch3lst:Listing2}, it is evident that the original plan has the same length as the contrastive plan, but has a better makespan. This is highlighted to the user as an explanation of the original plan.
The provision of the \emph{show optimal} and \emph{show optimal length} buttons shown in Figure \ref{fig:HPlan} entails the generation of an optimal plan in terms of makespan and length, respectively, based on the user's specification for each question.
The approach employed to achieve this is described below.

\paragraph{Iterative Re-planning:}
This module harnesses the power of iterative modeling to facilitate repeated user questioning. Sometimes users may need to ask repeated questions to refine their queries and obtain better answers, particularly when it comes to optimizing plan length or duration (\emph{makespan}). For instance, a user can repeatedly ask questions (f) and (g) (Section 2, point 4) to converge towards a makespan and length optimal plan. Our tool allows for iterative execution of steps 5-8 (as outlined in Section \ref{sec:architecture}) to support these repeated queries. This process can be repeated until no further hypothetical plans can be generated for the given query.

\paragraph{Iterative Re-planning with Bisection:}\phantomsection\label{para:bisection} A limitation of the iterative re-planner module is that of repeated manual tool invocation to progressively converge towards an optimal hypothetical plan with respect to either makespan or plan length. Additionally, a tool user may wish to contrast the original plan against that hypothetical plan, which is makespan or length optimal, instead of contrasting with just any. This tool has a provision that progressively searches for better alternate plans and terminates with the best-found hypothetical plan to be contrasted with. This is achieved using a bisection algorithm (see Algorithm \ref{algo:algorithm1}). The algorithm begins by setting an upper and lower bound of the optimization objective, which is either makespan or plan length. The lower bound is set to 0, and the upper bound is set to the makespan (length) returned for a hypothetical plan by the planner. We use variables \emph{high} and \emph{low} respectively to store these values. In each iteration, the algorithm probes whether a plan exists with a makespan (length) $\le mid = \frac{low + high}{2}$. If such a plan is found, we improve the upper bound \emph{high} to the makespan (length) of that plan. Conversely, in the absence of a feasible plan, we improve the lower bound (\emph{low}) to mid. The algorithm terminates when the difference between the upper and lower bound is less than a given tolerance $tol$, and the plan having a makespan (length) equal to high is returned as the best found contrastive plan.

\begin{algorithm}[!htb]
  \caption{\small \textsf{Bisection~($\Pi$.dom, $\Pi$.prob, low, high, tol)}}
  \label{algo:algorithm1}
  \scriptsize
  \label{algo:bisection}
  \LinesNumbered
  \SetAlgoLined
  \KwIn{low, high, tol \\}
  \KwOut{makespan of the contrastive plan (high), the contrastive plan $\phi$}
  \Begin{
  \If{abs(high - low) $<$ tol}{
         modifyProblemFile($\Pi$.prob, high) \tcc*{modify makespan bound}
         $\phi$ = runPlanner($\Pi$.dom, $\Pi$.prob) \tcc*{get HPlan}
         return (high, $\phi$);
  }
   mid = (low+high)/2;\\
   modifyProblemFile($\Pi$.prob, mid);\\    
   $\phi$ = runPlanner($\Pi$.dom, $\Pi$.prob); \\
   \If{$\phi$ == null}{
   return Bisection($\Pi$.dom, $\Pi$.prob, mid, high, tol) \tcc*{No plan found}}
   \If(\tcc*[f]{A plan exists}){$\phi$ != null}{
      high = getMakespan($\phi$) \tcc*{high = getLength($\phi$) for length objective}
      return Bisection($\Pi$.dom, $\Pi$.prob, low, high, tol)
   }
  }
\end{algorithm}
\section{Implementation and Results} \label{sec:case_study}
\paragraph{Implementation:} We use \emph{Python} and its regular expression library package \emph{re}, \emph{Javascript} for designing the web interface, \emph{Nodejs} for the server environment, \emph{Express} framework for web application features, \emph{Bootstrap 5.2.3} and \emph{EJS} for designing front-end GUI.
The experiments are performed in a machine with 8 GB RAM, an Intel Core i5-8250U@1.60GHz 8-core processor, with Ubuntu 20.04 64-bit OS.

\paragraph{Case Study:} We demonstrate the tool's performance on two benchmark PDDL+ planning domains, the car domain and the generator-events domain. A brief description of the car domain is discussed in Section \ref{module_description}. The \textbf{\textit{Generator-events Domain}} consists of a generator and two fuel tanks. The generator consumes fuel while running and has a fuel capacity. The fuel in the tanks can be poured into the generator. The dynamics of fuel consumption and fuel pouring from tanks into the generator are given. The planning task is to run the generator for 1000 time units safely, that is, without an overflow or underflow of the fuel. The source code of the tool and a detailed description of the domains can be found at \url{https://github.com/manabjamin2nadved1947/XAIP.git}.

\paragraph{Evaluation:} A performance summary of generating explanations is provided in Table \ref{ch3tab:table1}. For each question, we construct an HModel, and the \emph{HModel Compilation Time} shows the time needed to generate these HModels. The mean HModel compilation time in our framework is approximately 0.055 seconds, which is significantly small. In the \textsc{SMTPlan+} and \textsc{ENHSP} columns, we report the corresponding HPlan generation time. The mean HPlan generation time for \textsc{SMTPlan+} in the car domain and the generator-events domain is, respectively, 0.05 seconds and 0.079 seconds, whereas for \textsc{ENHSP} the mean time is 0.53 seconds in the car domain.
The generator-events domain is evaluated using \textsc{SMTPlan+} only, as the domain is not supported by \textsc{ENHSP}.
However, for question 4, which specifies to exclude one action from the original plan, neither \textsc{SMTPlan+} nor \textsc{ENHSP} is able to generate a solution within 10 seconds in both domains.
We also note the time required to generate a contrastive plan by iterative re-planning with Bisection, with minimum makespan objective, shown under the \emph{Best Contrastive Plan Time} in the table. The average time taken to generate the best contrastive plan is 59.26 seconds and 98.09 seconds for the car and generator-events domains. The results indicate that the best contrastive plan generation process is expensive in comparison to HModel generation and plan execution. This can be attributed to the iterations in the Bisection algorithm.

\begin{table}[ht]
    \centering
    \resizebox{\columnwidth}{!}{%
    \begin{tabular}{|c|c|c|c|c|c|}
    \hline
        & & & & {\centering {HModel}} & {\centering {Best Contrastive} }\\
        {\centering {Benchmark}} & {\centering {Q No.}} & SMTPlan+ & ENHSP & {\centering {Compilation Time}} & {\centering{Plan Time}}\\
    \hline
          & q1 & 0.090s & 0.842s & {0.038s} & {1m2.396s}\\ \cline{2-6}
          & q2 & 0.059s & 0.520s & {0.026s} & {1m28.344s}\\ \cline{2-6}
          \multirow{2}{*}{Car} & q3 & 0.041s & 0.384s & {0.029s} & {0m51.703s}\\ \cline{2-6}
          \multirow{2}{*}{domain} & q4 & NP & NP & {0.041s} & { NP}\\ \cline{2-6}
          & q5 &  0.036s & 0.492s & {0.092s} & {0m52.123s}\\ \cline{2-6}
          &  q6 & 0.038s & 0.436s & {0.025s} & {1m30.856s}\\ \cline{2-6}
          &  q7 & 0.068s & 0.511s & {0.244s}& {10.106s}\\
    \hline
        & q1 & 0.034s & & 0.036s & 2m32.106s\\ \cline{2-3}\cline{5-6}
        & q2 & 0.034s & & 0.027s & 1m31.017s\\ \cline{2-3}\cline{5-6}
        Generator & q3 & 0.049s & \multirow{2}{*}{NA} & 0.037s & 2m23.407s\\ \cline{2-3}\cline{5-6}
        -events & q4 & NP & & 0.067s & NP\\ \cline{2-3}\cline{5-6}
        domain  &  q5 & 0.158s & & 0.034s & 1m30.850s\\ \cline{2-3}\cline{5-6}
        &  q6 & 0.039s  & & 0.036s & 1m30.924s\\ \cline{2-3}\cline{5-6}
        &  q7 & 0.164s & & 0.038s & 10.235s\\
    \hline
    \end{tabular}
    }
    \caption{Table depicting the mean response time of our tool to each of the seven contrastive questions in the car and the generator-events domain. \emph{Q. No.} is the contrastive question from the set of questions in Section \ref{sec:architecture}. The columns \emph{SMTPlan+} and \emph{ENHSP} show the time taken by these planners to generate a plan. The column \emph{HModel Compilation Time} shows the time taken to construct the Hmodel, and \emph{Best Contrastive Plan Time} reports the time to find the best contrastive plan by iterative re-modeling with Bisection. NP denotes no plan produced within a given time-bound. NA denotes not applicable.}
    \label{ch3tab:table1}
\end{table}

\section{Related Works} \label{sec:ch3-related-work}
\noindent\textbf{\textit{Contrastive explanation:}}
(\cite{DBLP:journals/ker/Miller21}) extends the definition of explanation using \emph{structural causal models} to contrastive explanation. 
They distinguish contrastive explanations for two categories of questions: counterfactual questions like "why P rather than Q?" and bi-factual questions, which are of the form "why P but Q?" The work asserts that contrastive explanations are integral to understanding how people seek explanations, as evidenced by research in philosophy and social science. Specifically, the preference for contrastive questions like "Why P rather than Q?" over simple questions like "Why P?" highlights the importance of considering alternatives when seeking explanations. To produce contrastive explanations for differentiable models like deep neural networks, a method 
called the \emph{contrastive explanations method} (CEM) is suggested by (\cite{DBLP:conf/nips/DhurandharCLTTS18}).

\noindent\textbf{\textit{Iterative modeling:}}
In (\cite{Smith_2021}), the authors propose \emph{explainable planning as an iterative process} allowing a user to refine the hypothetical model and iterate the process by posing new questions. This permits the user to set extra constraints that can be incorporated into the hypothetical model if the explanation remains unsatisfactory.

\noindent\textbf{\textit{XAIP as a service:}} In (\cite{cashmore2019towards}), the authors propose a prototype framework for explainable planning as a service which can be used with any planner capable of reasoning with PDDL2.1. They present a console-based and a GUI-based interface, allowing users to view the plan and select a formal query from a list of questions. The system produces a visual representation of the original plan and a hypothetical Plan. This work extends such a tool interface to the hybrid planning domain. It implements the algorithms presented in (\cite{Sarwar2022Contrastive}).
\section{Conclusion} \label{sec:ch3-conclusion}


In this work, we present an interactive web interface for contrastive explanation of plans that implements the presented algorithms.
This tool provides a way to experiment with different planning domains in hybrid systems and to plug in different hybrid system planners as plan generation engines.
It explores a set of contrastive questions that a user of a planning tool may raise, and provides an interface for contrastive explanations of such questions.
It also provides provisions for finding a competitive contrastive plan based on the comparison metrics (e.g., plan makespan, plan length) that the underlying planner can provide.
We demonstrate experimental results on two planning domains using two state-of-the-art planners.

\chapter{Explaining Unsolvability of Planning Problems in Hybrid Systems with Model Reconciliation} \label{ch4}

\textbf{\textsc{Chapter Abstract:}} \textit{A recent problem of interest in Explainable AI Planning is that of explaining the unsolvability of planning problems. Though there has been a lot of research on generating explanations of solutions to planning problems, explaining the absence of solutions remains a largely open and understudied problem.
Model reconciliation has been a popular approach for generating explanations for such problems in recent literature, which involves an AI agent and a human planner, who have different models of the planning domain, and each explains to the other the differences they have in their domain representations and attempts to arrive at a consensus. More often than not, it is assumed that the AI agent has a correct and complete view of the domain, of which the human only has a partial view. Through reconciliation, the human domain is updated to be consistent with what the AI agent has. Most of the works in this direction are targeted toward classical planning problems on domains represented typically as discrete state transition systems or variants. In this chapter, we provide an approach towards model reconciliation for planning problems in hybrid systems represented as a mix of discrete and continuous domains.
We assume that the agent has a complete model of the environment, while the human has a partial or erroneous model and expects a plan for the planning problem when there is none. The explanation problem is presented as a process of continuous reconciliation between these two entities (agent and human) to make the human domain consistent with that of the agent. 
To this effect, we use a mix of graph traversal and path analysis, along with Linear programming to carry out the reconciliation process. In particular, we use the concept of Irreducible Infeasible Sets (IIS) to generate explanations. Experimental results on 2 representative hybrid domains show the efficacy of our approach.}

\section*{} \label{intro}
\lettrine[lraise=0.3, nindent=0em, slope=-.5em]Explainable planning is an active area of research in recent times, given the increasing number of application areas in which humans and autonomous agents collaborate. In such application areas, cooperative plans derived with mutual trust and understanding are important for achieving a desired objective. With automated planning being applied in safety-critical systems, the need for explanation and trust in the agent's behaviour has become ever more important to a human user. The ability to explain the rationale behind a decision of an autonomous agent is widely regarded as one of the precursors needed for humans to engage in trustworthy collaborations with autonomous agents. While there has been a lot of research on generating explanations to planning problems, most of the earlier works in explanation generation have focused on explaining why a given plan or action was chosen. However, explaining the unsolvability of a planning problem remains a largely open and understudied problem.

In the context of unsolvability explanation, recent works have looked at generating certificates or proofs of unsolvability~(\cite{DBLP:conf/aips/ErikssonRH17,DBLP:conf/aips/ErikssonRH18}). Finding counterfactual alterations to the original planning task such that the new planning task can be made solvable, i.e, an excuse for the unsolvability of the planning problem has been proposed in (\cite{DBLP:conf/aips/GobelbeckerKEBN10}).
Such certificates or proofs of unsolvability gear towards automatic verification rather than explaining unsolvability of planning problems. However, the excuses generated by altering the planning task may not always be adequate to explain unsolvability in complex planning domains. In most human-AI interaction scenarios for a given planning task on which human-agent collaboration is sought, the human may have a preconceived domain that may differ from the actual domain known to the agent. In such scenarios, the \emph{Model Reconciliation Problem} (MRP) (\cite{DBLP:conf/ijcai/ChakrabortiSZK17}) has been a popular approach to explain the agent's domain knowledge to the human where reconciliation between these two models is done by making the models equivalent such that a plan in one entails a plan in the other and vice-versa. To do so, MRP attempts to provide an explanation that can be used to update the human model such that the agent’s behaviour is also amenable to the human user. However, to the best of our knowledge, most of the works in MRP literature have been applied to planning problems in discrete domains.

\begin{figure*}[htbp]
    \centering
    \includegraphics[width=\textwidth]{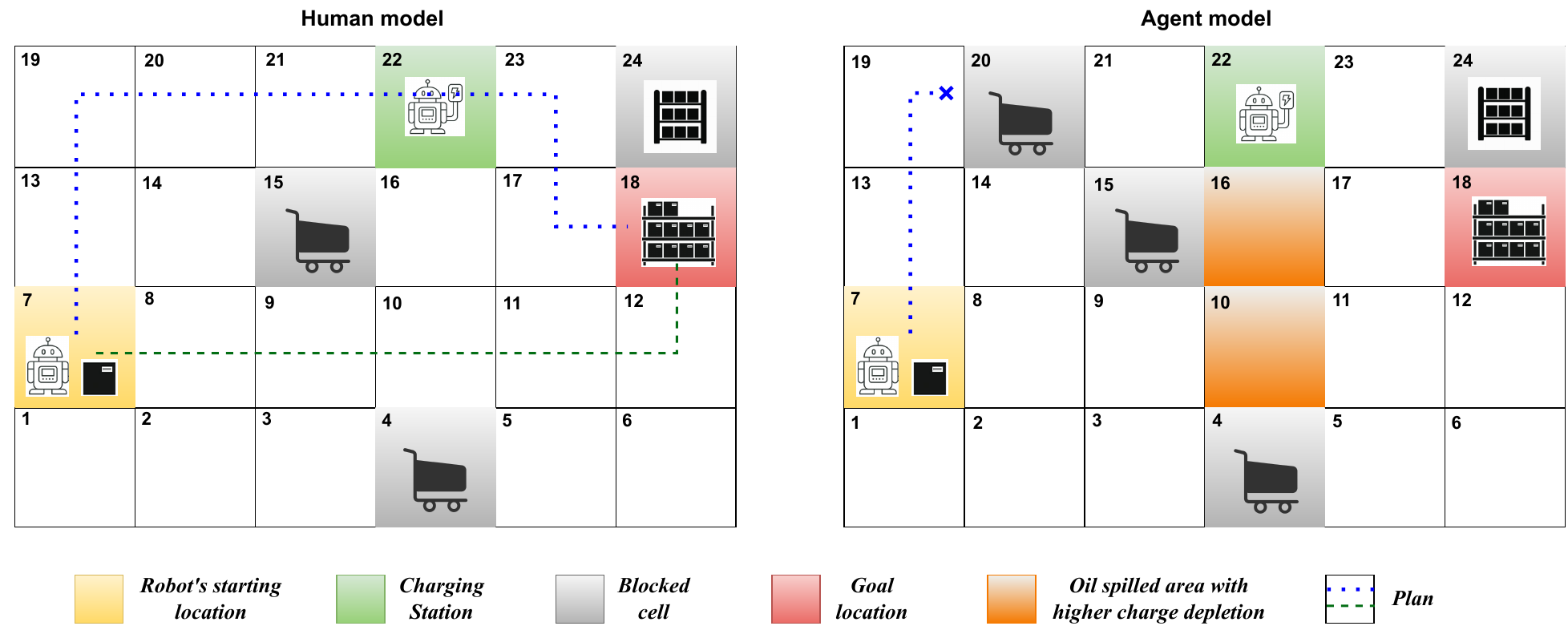}
    \caption{The human and the agent view of the warehouse environment is depicted. Each cell is associated with an identifier shown in the top-left corner of the cell. The Blue dotted line represents an expected plan by a human under its knowledge of the warehouse. Our explanation algorithm examines this plan in the agents' view of the world and detects the presence of an obstacle in cell number 20. This knowledge is reported to the user as an explanation of the infeasibility of the plan. The Green dashed line represents a candidate plan in the human model when abstracting its continuous dynamics. This plan is refuted by our algorithm when the feasibility of the path is checked against the continuous dynamics of the human model.}
    \label{fig:warehouse_automation}
\end{figure*}

In this chapter, we present a \emph{model reconciliation} framework for explaining unsolvability of planning problems in \emph{hybrid domains} that exhibit an interplay of discrete and continuous dynamics. In such a domain, a feasible plan has to satisfy not only the discrete dynamics but also the constraints imposed by the presence of the continuous dynamics of the domain. This makes the MRP problem more involved as well, since the reconciliation involves concurrence on both the discrete and continuous behaviors concerning the given planning task.
In our setting, we assume that the planning problem is solvable in the human model but unsolvable in the agent model. In other words, a feasible plan exists in the human model while no such plan is entailed by the agent's model of the domain.
We explain the unsolvability of the planning problem through a reconciliation process between the human and agent models. In other words, we examine the concurrence of each plan that the human model entails with its counterpart in the agent domain. To do so, we have a three-step procedure. In the first step, we examine the feasibility of a plan obtained by restricting only to the discrete dynamics of the human model while abstracting away the continuous one. If this fails, we provide an explanation of the plan's infeasibility by highlighting the difference in the discrete dynamics between the human and the agent model. We then continue examining the next plan for further reconciliations. On the other hand, where the plan is feasible in the agent model, we proceed to the next step, where we reintroduce the continuous dynamics together with the discrete one in the human model and re-examine the feasibility of the plan in the human model. When the plan is infeasible in the human model itself, it is discarded as far as reconciliation is concerned, being infeasible in both models and thus causing no conflict. On the other hand, if the plan is feasible, we proceed to the last step, where we check the feasibility of the same plan in the agent model. Note that the motivation behind the stratification of the human model based on its discrete and discrete-continuous dynamics is to efficiently reconcile the discrete dynamics with the agent model without needlessly spending planning effort in the entire complex model. As an outcome of the last step, the plan turns out to be infeasible, though feasible in the human model, since there are no feasible solutions to the planning problem at hand. In this case, the explanation is produced using the concept of irreducible infeasible sets (IIS) (\cite{DBLP:journals/informs/ChinneckD91}), and its application to bounded model checking of linear hybrid automata (\cite{DBLP:journals/entcs/FranzleH05, DBLP:conf/fmcad/BuLWL08}). For this, we construct a linear constraint set from the plan using linear programming (LP) such that the satisfiability of this constraint set implies the existence of a plan in the agent model. We then extract the IIS from the encoding using an underlying LP solver. This continues until we exhaustively enumerate all plans that the human can find in its model and the agent rules out each, either in the discrete setup or considering the continuous dynamics. The final explanation is a summary obtained in the three steps above. We demonstrate our approach on a warehouse automation and a water-level monitoring system model. 
The average time for model reconciliation is $1387.31$ seconds, where we consider all the possible plans for a planning problem.
The average time to generate explanations of unsolvability of a plan is $0.2$ seconds. This shows that our algorithm can quickly explain the causes of unsolvability. The stratification of the model dynamics and techniques to focus only on relevant dynamics of the model while searching for explanations expedite the explanation generation process. For instance, in the warehouse automation domain, $43.95\%$ of the plans are discarded for searching explanations. In the water-level monitoring domain, $72.22\%$ of plans are similarly pruned from generating any explanation. In summary, the contributions of this work are as follows.

\begin{itemize}
    \item We propose a path-based \emph{continuous model reconciliation framework} for explaining unsolvability of planning problems for hybrid systems. The explanation problem is formulated as a continuous reconciliation process between the agent and the human models to make the human knowledge base consistent with the agent.
    
    \item We show that a discrete and continuous path analysis approach can be leveraged to derive explanations of the unsolvability of a planning problem in hybrid domains. To this effect, we use a mix of \emph{graph traversal} and \emph{path analysis}, along with \emph{Linear programming} to carry out the reconciliation process. 
    
    \item The discrete path analysis falsifies a path by mapping each location and transition of the path in the human model to the agent model by a simple graph traversal. The continuous path analysis leverages reachability analysis, combining with minimal inconsistent constraint sets to find the infeasible path-segments for which the paths become unsatisfiable.

    \item To illustrate our explanation framework, we demonstrate a warehouse automation system as an example planning domain, along with a water-level monitoring system, and present results for problem instances on each domain with varying plan depth.

\end{itemize}

The rest of this chapter is organized as follows. In Section~\ref{motive}, we sketch a motivating example on which we demonstrate this work. Section~\ref{sec:problem-statement} provides an overview of the problem statement.  
Section~\ref{approach} illustrates our methodology and the framework of explanation. Section~\ref{evaluation} discusses implementation and results. Section~\ref{related_work} presents related literature. Finally, Section~\ref{conclusion} summarizes the contributions and findings of this work and discusses possible future directions.
\section{Motivating Example} \label{motive}

\noindent In this section, we present an MRP scenario in the context of warehouse automation where a robot operates to manage the inventories of the warehouse. The warehouse is divided into cells. The discrete dynamics here capture the connectivity of the cells along with the presence of objects in certain cells, which are interpreted as obstacles through which the robot cannot move. The movement of the robot is restricted to one of its adjacent cells, and movement to diagonal cells is prohibited. The continuous dynamics capture the battery charge depletion rate of the robot within a cell. Within each cell, the robot follows the dynamics particular to that cell. When the robot makes a transition from one cell to another, it starts to follow the dynamics of the new cell instantaneously. The robot is assigned the task of carrying a consignment to a designated cell while the number of cell visits is restricted to $\leq$ $D$ cells. The robot starts from the yellow-colored cell where the planning problem requires it to transport the black box to the goal cell (red-colored cell). The robot depletes its charge according to the cell dynamics while on the move. There is a charging station shown as a green-colored cell. The robot may visit this cell to recharge its battery.
The orange-colored cells represent the oil-spilled areas where the robot depletes battery charge at a higher rate due to the slippery floor condition. The grey-colored cells are blocked with obstacles. The robot is equipped with a rechargeable battery. The initial battery charge is 10 units. Each cell has a charge depletion rate of 2 units (modelling the continuous dynamics) except the orange cells, for which the corresponding value is 4. The robot has complete knowledge of the warehouse, whereas the human has only been 
exposed to a partial view. 

\begin{figure}[htbp]
    \centering
    \includegraphics[width= 0.85\textwidth]{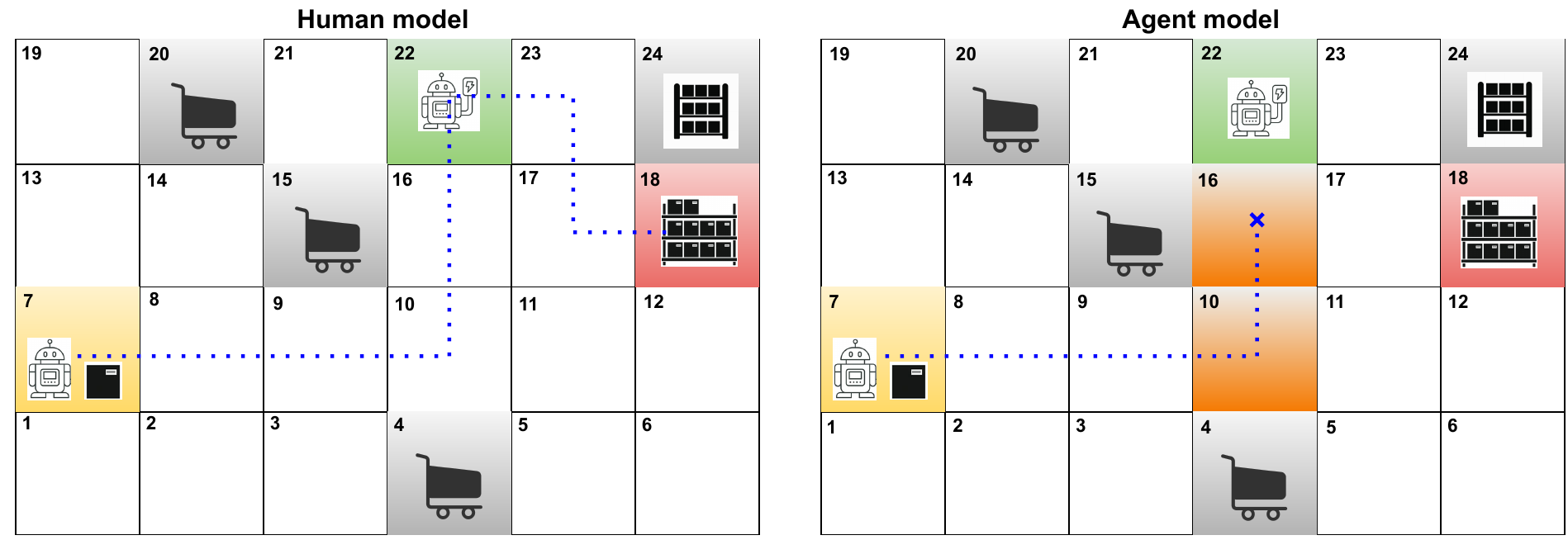}
    \caption{Infeasibility of the plan explained by continuous dynamics, particularly the higher charge depletion rate at orange cells}
    \label{fig:warehouse_automation2a}
\end{figure}

In Figure \ref{fig:warehouse_automation}, we have shown a representation of the human and the agent's view of the warehouse. Note that some of the obstacles in the warehouse are unknown to the human. Additionally, the human is unaware of the oil-spilled areas of the warehouse as well and is thereby not aware of the higher charge depletion rates in those cells.
Interestingly, there are a few candidate plans that are obtained when abstracting out the continuous dynamics of the human model and are detected as infeasible in the human model itself during the feasibility analysis with its continuous dynamics. In particular, the plans that avoid a visit to the recharge station are infeasible due to the battery drainage before reaching the goal, given that the charge depletion rate is at-least 2 units in each cell and the Manhattan distance (the distance between two points on a grid is the sum of the vertical and horizontal distances between them) to the destination for all such plans is $\geq$ 6.
One such infeasible plan is shown as a Green dashed line in the figure, whose Manhattan distance to the destination is 6.
Henceforth, the human would expect any plan via the recharge station as a solution to the planning task. The Blue dotted line in the figure represents such a plan. However, this plan goes through cell number 20, having an obstacle in the warehouse that the human is unaware of. When examining this plan in the agent's model of the warehouse, our explanation algorithm detects the presence of an obstacle in this cell and reports this to the human as an explanation of the plan's infeasibility. This knowledge transfer via an explanation results in a modified human view of the warehouse. As a result of this knowledge, the human would not enumerate any plan via this obstacle. Now, an alternate plan that the human may expect is shown as the Blue dotted line in Figure \ref{fig:warehouse_automation2a}.
When this plan is examined by our algorithm, it is found to be feasible in the agent's world when the continuous dynamics are abstracted in the first step. This is going to prompt our algorithm to now analyze the feasibility of this plan by considering the continuous dynamics as well. This is when our algorithm is going to detect that the plan goes via the cells with an oil-spill, and therefore, due to the higher battery depletion rate here (4 units instead of 2), the robot's battery drains completely before reaching the charging station. Consequently, our algorithm concludes the infeasibility of this plan.
Similarly, the remaining plans in the human's world would also be detected as infeasible due to the battery drainage before reaching the charging station.
The explanation will include the orange cell with oil-spill in the IIS to indicate it as a potential cause of the unsolvability in the agent's model.  



In the following section, we formally describe the MRP problem and the domain representation. We represent the agent model and the human model as hybrid automata  $\mathcal{HA}_A$ and $\mathcal{HA}_H$, respectively (\cite{ALUR19953}). Each cell of the domain is represented as a location in both the automata models $\mathcal{HA}_A$ and $\mathcal{HA}_H$, preserving the dynamics of that cell. 

\section{Problem Overview}\label{sec:problem-statement}

\noindent In the planning context, a transition $e \in \emph{Edge}$ in HA (see Defination~\ref{ch2def:HA}) depicts an action $act$ of the domain. The name of the action is the transition's label $e.a$, the pre-condition is the transition's guard $e.g$, and the post-condition is the transition's reset $e.r$. 
The \emph{Init} of the automaton defines the initial condition of the planning problem. 
The change in the discrete state of the domain due to the action is captured through the transition's source and target location $e.l$ and $e.l'$, respectively. A state of a \emph{HA} is a pair $(l,v)$ consisting of a location $l \in$ Loc and a valuation $v\in V$. $l$ denotes the discrete state, whereas $v$ denotes the continuous state of the system. A state can change either due to an application of an action or due to the passage of time following the continuous flow dynamics. The continuous state change is given by $flow_{l}(v, t)$, which is the solution of the differential equation Flow($l$). We refer to the former as \emph{action-transition} and the later as \emph{timed-transition}.

\noindent
A planning problem for a hybrid system consists of a hybrid automaton representation of the planning domain and a problem description, which is defined as:
\begin{definition} \label{ch4def:planning_problem}
A \emph{planning problem} $\Pi$ for a hybrid system HA is a tuple ($Dom$, $Prob$, $Depth$), where
\begin{itemize}
  \item $Dom$ is represented as a hybrid automation HA defining the planning domain (\cite{DBLP:journals/tecs/SarwarRB23}).
  \item $Prob$ is a tuple $\langle Init, Goal\rangle$ representing a problem description, where $Init$ and $Goal$ define the initial and the goal states. $Init$ is a tuple $\langle l_{init}, S_{init} \rangle$ such that $l_{init} \in Loc$ and $S_{init} \subseteq V$ and $Goal$ is a tuple $\langle l_{goal}, S_{goal} \rangle$ such that $l_{goal} \in Loc$ and $S_{goal} \subseteq V$. 
  \item \emph{Depth} $D$ defines the bound on the length of a plan.
  $\Box$
\end{itemize}
\end{definition}


\begin{definition}\label{ch4def:plan}
A plan $\phi$ is a tuple $\langle \lambda, makespan \rangle$. For a planning problem with a set of actions $A$, $\lambda$ is a finite set of pairs $\langle t, act \rangle$ {together with the plan \emph{makespan} $\in \mathbb{R}$}. In the pair, $t \in \mathbb{R}^+$ is the time instance of executing the action $act \in A$. The $makespan$ is the overall duration of the plan. $\Box$
\end{definition}

\noindent We refer to the length of the plan to be the number of actions appearing in $\lambda$.
The model reconciliation problem (MRP) is defined below.

\begin{definition}
    Given a human model $\mathcal{HA}_H$, an agent model $\mathcal{HA}_A$, a $Prob$ for which a plan $\phi$ exists for $(\mathcal{HA}_H, Prob)$ but no plan exists for $(\mathcal{HA}_A, Prob)$ within a plan of length $D$, 
    the model reconciliation problem is to provide an updated model $\mathcal{HA'}_H$ to the human  such that $\phi$ does not exist for $(\mathcal{HA'}_H, Prob)$. $\Box$
\end{definition}

\begin{figure*}[htp]
    \centering
    \includegraphics[width= 0.9\textwidth]{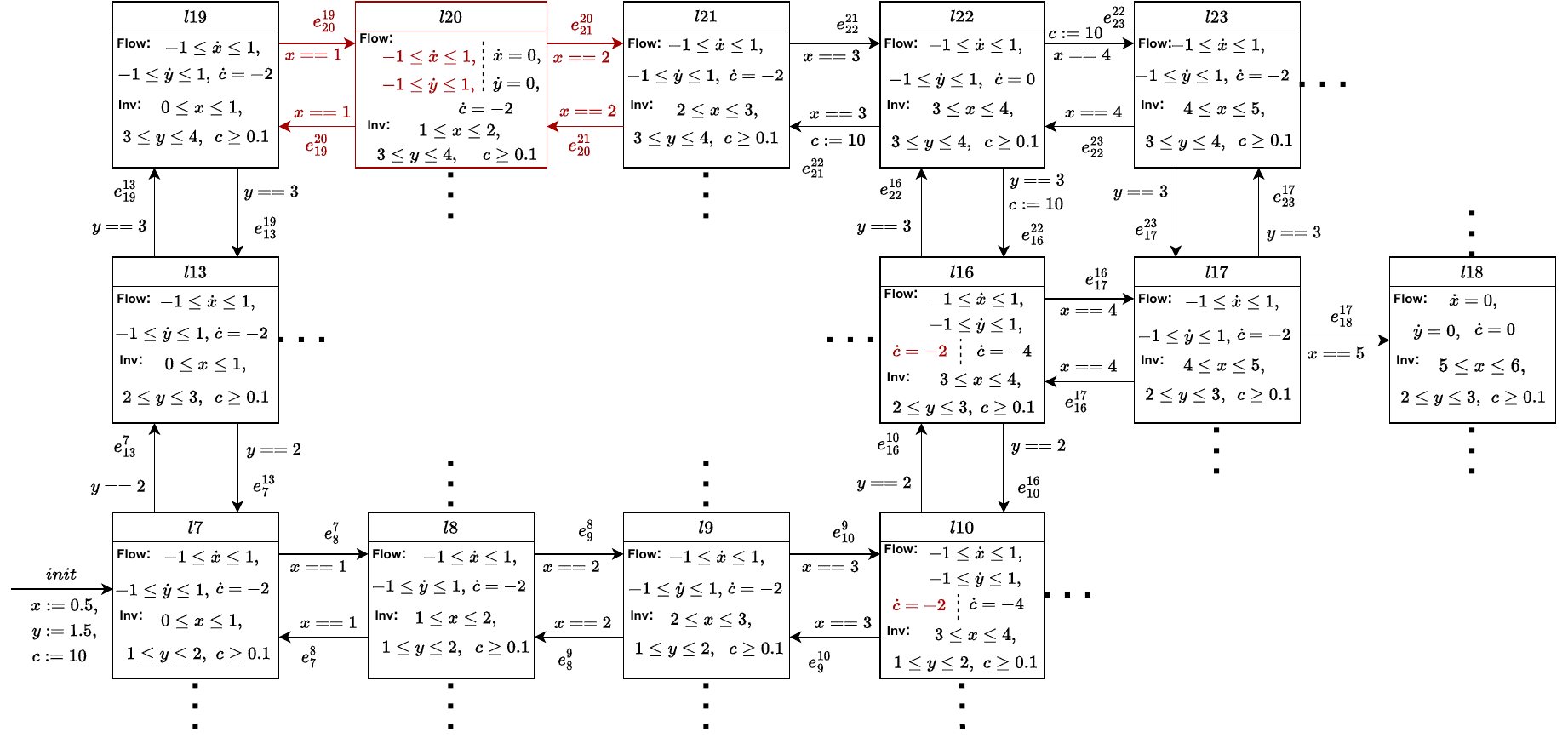}
    \caption{The HA model for the warehouse automation problem, where locations represent the corresponding cells shown in Figure \ref{fig:warehouse_automation}. The automaton provided here is a part of the warehouse automation domain.}
    \label{fig:warehouse_automata}
\end{figure*}

\noindent 
In our setup, as discussed earlier, we assume both $\mathcal{HA}_H$ and $\mathcal{HA}_A$ are represented as hybrid automata. Figure~\ref{fig:warehouse_automata} shows
the hybrid automata for the agent and human models for the problem in Figure~\ref{fig:warehouse_automation}. 
The differences between the human and the agent models are depicted in red. Location 20 ($l20$) represents the blocked cell 20 in the agent model, which the human assumes is free. The transitions, which are shown in red, are present in the human model but not in the agent model. The dynamics that are given in red represent the human's erroneous assumptions. 
The complete automaton could not be included as part of Figure~\ref{fig:warehouse_automata} due to space constraints.

\section{Our Reconciliation Methodology} \label{approach}
\noindent
In this section, we discuss the details of the reconciliation process. 
We assume that the human model and the agent model may differ on two points (a) the human may not have information about the 
inaccessibility of some location, and (b) the human may not have information 
about the correct continuous dynamics. 
In our setup, we consider an iterative MRP process. In each iteration, the human produces a plan in $\mathcal{HA}_H$, the agent provides an explanation to refute the existence of that plan in $\mathcal{HA}_A$, following which the human updates its $\mathcal{HA}_H$ and searches for another plan. This continues till no plan exists in the human model. To expedite this plan search-validate-reconcile process, the human first creates an abstraction of the hybrid domain 
$\mathcal{HA}_H$ by abstracting out the continuous dynamics and keeping only the abstract graph structure $\mathcal{G_H}$ 
consisting of the automaton locations and transitions underlying $\mathcal{HA}_H$. The initial location $l_{init}$ and goal location $l_{goal}$ of the planning problem $Prob$ are respectively marked as the initial and goal locations in $\mathcal{G_H}$,
and a path $p_H$ reaching the goal from the initial location within the specified length is extracted from $\mathcal{G_H}$. The path $p_H$ serves as a candidate plan when only the discrete dynamics are considered; however, it may or may not be feasible when the continuous dynamics are brought in. For the path $p_H$, the agent model is consulted to check if the same path can be reproduced from the corresponding initial to the goal location in the graph structure $\mathcal{G_A}$ underlying the 
agent domain automaton $\mathcal{HA}_A$. If not, the first edge that appears in the human plan but not in the agent is marked 
as an explanation, and also deleted from $\mathcal{G_H}$ such that no further plan involving this edge 
is generated from $\mathcal{G_H}$. Reconciliation then proceeds with the next initial-to-goal path in 
$\mathcal{G_H}$. However, if $p_H$ can be reproduced in the agent graph $\mathcal{G_A}$ (considering only the locations 
and transitions), the human checks to see if the continuous dynamics 
involved on $p_H$ (reconstructed by bringing in the location dynamics, invariants, flow, edge guards) entails a feasible solution 
in $\mathcal{HA}_H$. If not, reconciliation again proceeds with the next path in $\mathcal{G_H}$. However, if $p_H$ is 
feasible in $\mathcal{HA}_H$, it is passed on for validation in the agent model $\mathcal{HA}_A$, considering the full 
discrete + continuous setting. Evidently, since no plan for $Prob$ exists in $\mathcal{HA}_A$, the feasibility check for 
$p_H$ turns out to be negative, and the first location that is not able to meet the constraints 
is produced as an explanation. This continues till no further initial to goal paths can be generated from $\mathcal{G_H}$.


\noindent
Our method has 3 key steps as outlined below. 
\begin{itemize}
    \item Discrete path generation and reconciliation
    \item Path feasibility considering continuous dynamics
    \item Explanation and human model update
\end{itemize}
   
\begin{figure*}[htbp]
    \centering
    \includegraphics[width=0.85\textwidth]{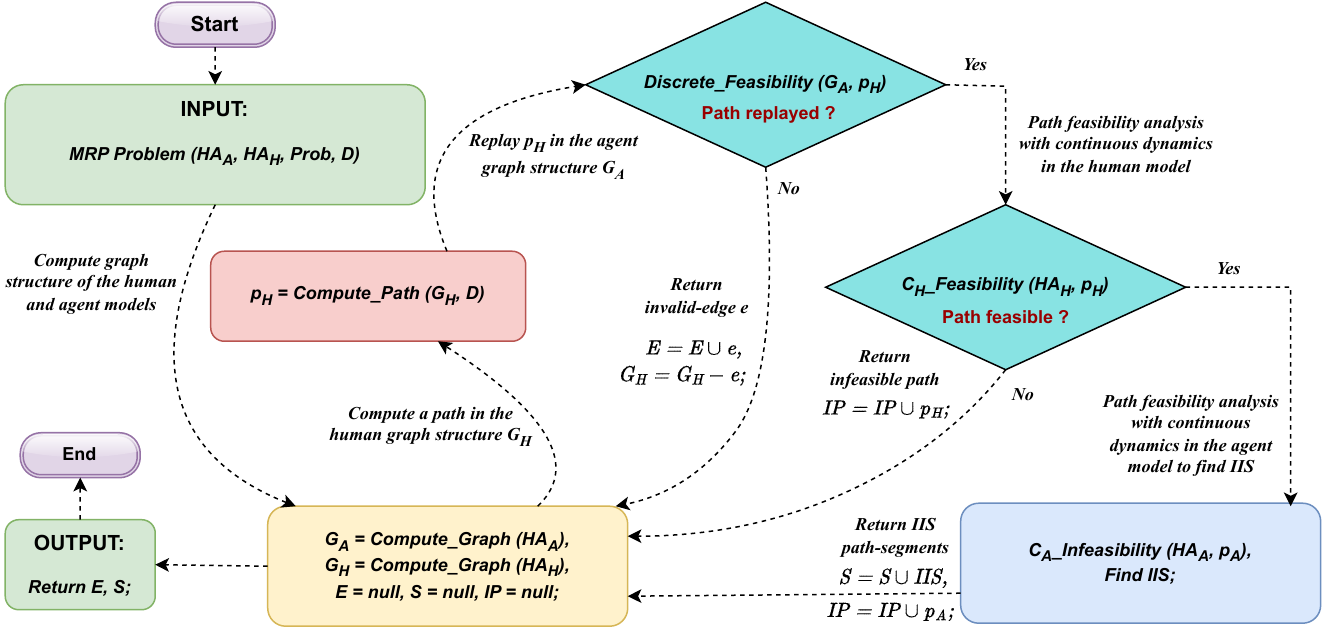}
    \caption{{The flow diagram of our model reconciliation framework}}
    \label{fig:explain_flow}
\end{figure*}

\noindent
We discuss each in detail below. The overall architecture of our explanation framework is shown in Figure~\ref{fig:explain_flow}.

\subsection{Discrete Path Generation and Reconciliation}
\noindent
The objective of this step is to examine whether a path in the human model is also valid in the agent model, considering only the discrete dynamics. The main motivation for this step is to quickly rule out invalid plans in the human model that are due to an incorrect understanding of the locations and transitions of the planning domain, without getting into the complex continuous dynamics. 
To this effect, we first define the concept of an abstract graph corresponding to a hybrid automaton obtained by abstracting out 
the continuous dynamics and considering only the location-location edge relationship. 

\begin{definition}
For a hybrid automaton HA = \big(\emph{Loc}, \emph{Var}, \emph{Flow}, \emph{Init}, \emph{Lab}, \emph{Edge}, \emph{Inv}\big), the graph of the automaton is defined as $G_{HA}$ = (Loc, $E$) where $E \subseteq \emph{Loc} \times \emph{Loc}$ such that for every ($l$, $a$, $g$, $r$, $l'$) $\in \emph{Edge}$, there is an edge ($l$, $l'$) $\in E$. $\Box$
\end{definition}

\noindent As the first step, the abstract graphs $\mathcal{G_A}$ and $\mathcal{G_H}$ corresponding to $\mathcal{HA}_A$ and $\mathcal{HA}_H$ respectively are obtained. Our approach then proceeds to extract a path $p_H$ from $\mathcal{G_H}$. In this graph, 
a path is a plan from the initial location to the goal location in the domain $Dom$ 
specified in the planning problem $\Pi$. We now define the structure of a path.

\begin{definition}
    A path $p$ in a graph structure $\mathcal{G}$ of a hybrid automaton HA is a sequence of $vertices$ and $edges$ 
    \begin{align*}
    l_0 \xrightarrow{e_0} l_1 \xrightarrow{e_1} l_2 \xrightarrow{e_2} \ldots \xrightarrow{e_{n-1}} l_N
    \end{align*}
    where $l_i$ $\in$ $Loc$, $e_i$ $\in$ $Edge$, $l_{i} = e_i.l$, $l_{i+1} = e_i.l'$. 
    $\Box$
\end{definition}

\noindent 
For reconciliation, we proceed to check if $p_H$ is a valid or spurious path in $\mathcal{G}_A$. 
{\em An important assumption we make here is that the location namespace of $\mathcal{G}_A$ and $\mathcal{G}_H$ match, in other words, 
for each location appearing on $p_H$, we can uniquely identify a corresponding location in $\mathcal{G}_A$.} 
Algorithm \ref{algo:d_replay} presents the pseudo-code for discrete feasibility analysis of a path, which takes $p_H$ and $\mathcal{G}_A$ as inputs. It checks for the existence of a path in $\mathcal{G}_A$, which is a counterpart of $p_H$. If such a path exists, we say $p_H$ is feasible in $\mathcal{G}_A$. If $p_H$ is not feasible, the algorithm returns the first \emph{invalid-edge} detected in $p_H$ not present in $\mathcal{G}_A$. We add the \emph{invalid-edge} to the 
\emph{explanation-edge} set \emph{E}, 
which we define as follows:

\begin{definition} \label{def:invalid_edge}
    An \emph{invalid-edge} $invalid_e$ is the first edge along a path $p_H$ which is present in $\mathcal{G}_H$ but not in $\mathcal{G}_A$. $invalid_e$ corresponds to a transition $e$ $\in$ $Edge$ in $\mathcal{HA}_H$ and not present in $\mathcal{HA}_A$.
\end{definition}

\begin{definition}
    The \emph{explanation-edge} set \emph{E} is the collection of \emph{invalid-edges} along different paths of the human graph structure $\mathcal{G_H}$ which are not feasible in the agent graph structure $\mathcal{G}_A$.
\end{definition}

\begin{algorithm}
  \caption{\textsf{discrete\_feasibility($p_H$, $\mathcal{G}_A$)}}
  \scriptsize
  \label{algo:d_replay}
  \LinesNumbered
  \SetAlgoLined
  \KwIn{A path $p_H$ from $\mathcal{G}_H$, and agent's graph $\mathcal{G}_A$.\\}
  \KwOut{The first \emph{invalid-edge} of $p_H$ if spurious, $True$ otherwise\\}
  \Begin{
    \For{$i$ = $0$ to $n-1$}{
        \eIf {($e_i.l == l_i$) \& ($e_i.l' == l_{i+1}$), where $e_i$ is the $i$-th edge on $p_H$, and $l_i$, $l_{i+1}$ $\in$ $Loc$ of $\mathcal{G}_A$.}
            {continue;}
            {Add $e_i$ to $E$; \\
            return $e_i$;}
    }
    return True
}
\end{algorithm}

\noindent 
Intuitively, we attempt to reconstruct the entire location-edge-location sequence in $p_H$ starting from the initial to the goal 
in the agent graph $G_A$. If this succeeds, we conclude $p_H$ is valid, spurious otherwise. In the latter case, we 
identify the first edge $e$ on $p_H$ that is invalid, i.e., does not have an existence in $G_A$. We mark $e$ as invalid 
and delete it from $G_H$ so that further paths generated from $G_H$ cannot include the spurious edge $e$. 

\begin{exmp}
Consider Figure \ref{fig:warehouse_automation} in which the blue dotted line represents a plan 
with the following sequence of locations and edges $l_7$-$e_{13}^7$-$l_{13}$-$e_{19}^{13}$-$l_{19}$-$e_{20}^{19}$-$l_{20}$-$e_{21}^{20}$-$l_{21}$-$e_{22}^{21}$-$l_{22}$-$e_{23}^{22}$-$l_{23}$-$e_{17}^{23}$-$l_{17}$-$e_{18}^{17}$-$l_{18}$. In the human model, the path goes through locations 19 and 20 as the human erroneously assumes location 20 is a free cell. Therefore, the transition from location 19 to location 20 is valid in the human model. However, the agent knows that location 20 is blocked. Thus, there is no such transition in the agent model, which is depicted by the cross-mark on the path there in the figure. Hence, when we consider discrete feasibility analysis of the path from the human abstract graph in the agent abstract graph, this transition $e_{20}^{19}$ from location 19 to location 20 is identified as an \emph{invalid-edge}. $\Box$ 
\end{exmp}

\noindent
The path generation and validation step is carried out first 
on the abstract graphs to expedite the reconciliation process. Paths that appear as candidate plans in the human model 
due to his lack of awareness of the complete location structure can be simply ruled out without 
considering the complex continuous dynamics, when they are attempted for 
reconstruction on $G_A$. Also, every path generated from $G_H$ does not correspond to a valid path in $\mathcal{HA}_H$. 
We now proceed with the next validation step with the continuous dynamics.

\subsection{Path Feasibility in the Continuous Dynamics}
\noindent
Once a path $p_H$ obtained from $G_H$ is identified as feasible in the abstract graph $G_A$, we proceed to check if 
the same is feasible in $\mathcal{HA}_H$ in the presence of the continuous dynamics. 
Algorithm \ref{algo:ch_replay} presents the pseudo-code for feasibility checking, which takes the path $p_H$ and the model $\mathcal{HA}_H$ as inputs. If $p_H$ is infeasible in $\mathcal{HA}_H$ itself, we mark it as \emph{infeasible} and add it to the \emph{infeasible-path} set \emph{IP}. Any \emph{infeasible path} and all its extensions are discarded 
to ensure that further paths are not generated involving such infeasible paths as sub-paths. This reduces the number 
of paths generated in our approach. However, if $p_H$ is feasible in $\mathcal{HA}_H$, it corresponds to a plan $\phi_H$ in the human model. The first step in our algorithm encodes the constraints associated with a path arising due to the continuous dynamics, while the second $Solve$ step invokes a solver to check if the constraints encoding the path entail a feasible solution. The encoding is such that a solution to the constraints implies the existence of a $run$ of the system that leads to the goal state from an initial state. The existence of a run in turn implies the existence of a plan to solve the planning problem. A path $p_H$ abstracts zero, one, or many \emph{runs} in $\mathcal{HA}_H$, mainly due to the non-determinism that it embodies with 
respect to the dwell time at a location. Corresponding to a path $p_H$ in $\mathcal{G_H}$, we first define the notion of a corresponding {\em run} in $\mathcal{HA}_H$.

\begin{algorithm}
  \caption{\textsf{c$_H$\_feasibility($p_H$, $\mathcal{HA}_H$).}}
  \scriptsize
  \label{algo:ch_replay}
  \LinesNumbered
  \SetAlgoLined
  \KwIn{A path $p_H$, human model $\mathcal{HA}_H$.\\}
  \KwOut{Return \textit{true} or \textit{false}.\\}
  \Begin{
    $c_H$ = Get constraints of $p_H$ in $\mathcal{HA}_H$;\\
    $result$ = Solve($c_H$) \tcc*{Use BACH}
    \eIf{$result$ == SAT}
            {return true \tcc*{When $p_H$ is feasible} }
            {add $p_H$ to $IP$;\\
            return false \tcc*{When $p_H$ is infeasible} }
  }
\end{algorithm}

\noindent

\begin{definition}\label{ch4def:run}
A run corresponding to a path $l_0 \xrightarrow{e_0} l_1 \xrightarrow{e_1} l_2 \xrightarrow{e_2} \ldots \xrightarrow{e_{n-1}} l_N$ 
is an alternating sequence of timed and action transitions of the hybrid automaton depicted as:
\begin{align*}
(l_{0},v_{0}) & \xrightarrow{\tau_0}(l_{0},v'_0) \xrightarrow{e_0} (l_1,v_{1}) \xrightarrow{\tau_1} (l_1,v'_{1})\xrightarrow{e_1} \ldots \xrightarrow{e_{n-1}}(l_{n},v_n)\xrightarrow{\tau_{n}} (l_{n},v'_{n})
\end{align*}
where (i) $v_0 \in S$, (ii) $e_i \in Edge$ such that $l_{i} = e_i.l$, $l_{i+1} = e_i.l'$, $v'_{i} \in e_i.g$, $v_{i+1} = e_i.r(v_{i})$, $\forall i \in \{0,\ldots,n-1\}$.
(ii) $\forall t\in [0, \tau_i]$, $flow_{l_i}(v_i, t) \in \emph{Inv}(l_i)$, $\forall i \in \{0,\ldots,n\}$.
(iii) $flow_{l_i}(v_i,\tau_i) = v'_i$;
(iv) $v_{i+1} = e_i.r(v_i)$.
(vi) $(l_n,v'_n)$ satisfies goal condition $G$.
$\Box$
\end{definition}
\noindent It can be seen that the existence of a run corresponding to a path also implies the existence of a feasible plan for the planning problem which can be obtained by extracting the actions associated with the transitions $e_i$ together with the time of taking the transitions $\tau_i$ as the action, time pairs of a valid plan. We now elaborate on the encoding of a path as constraints in detail below.
\subsection*{Constraint Encoding of a Path}
\noindent We encode a path as a set of linear constraints. Thus, the problem of checking the reachability of the location $l_n$ (representing the terminal location on a path) and thus validating the existence of a plan is reduced to a linear program encoded as a set of path constraints as discussed in (\cite{DBLP:conf/fmcad/BuLWL08}). This linear encoding is illustrated with a simple example.

\begin{exmp} \label{example:example}
  Consider Figure \ref{fig:warehouse_automata} which shows the hybrid automaton modeling of a part of the warehouse. Consider the path $p$ in the agent model $\langle \xrightarrow{init}  l_7\xrightarrow{e_{8}^{7}} l_{8}\xrightarrow{e_{9}^{8}} l_{9}\xrightarrow{e_{10}^{9}} l_{10}\xrightarrow{e_{16}^{10}} l_{16}\xrightarrow{e_{22}^{16}} l_{22}\xrightarrow{e_{23}^{22}} l_{23}\xrightarrow{e_{17}^{23}} l_{17}\xrightarrow{e_{18}^{17}} l_{18} \rangle$. 
\end{exmp}

\noindent The linear encoding of the path as constraints is defined below:

\textcolor{black}{
\begin{enumerate}
    \item A variable $t_i$ is associated with each location of the path, representing the timed transition in the respective location. For each $t_i$ $(0\leq i \leq 8)$, we add non-negativity constraints $t_i \geq 0$ in the linear program.
    \item For flow conditions in every location along the path, we generate the corresponding flow constraints. For instance, given the flow condition $-1\leq \dot{x} \leq 1$ in location $l_7$, we generate a constraint $x_{l_7}^{out} = x_{l_7}^{in} + t_0*\dot{x}$, where $x_{l_7}^{in}$ represents the value of $x$ when a run enters location $l_7$ and $x_{l_7}^{out}$ represents the value of $x$ when a run is to transition to $l_8$ after having stayed at location $l_7$ for $t_0$ units of time.
    \item For location invariants in every location along the path, we generate the invariant constraints. For instance, the invariant $0\leq x \leq 1$ in location $l_7$ is encoded as two constraints: $ 0 \leq x_{l_7}^{in} \leq 1$, and $0 \leq x_{l_7}^{out} \leq 1$.
    \item For reset on an edge along the path, we generate corresponding reset constraints. For instance, for $x=0.5$ on edge $init$, we get $x_{l_7}^{in} = 0.5$.
    \item For guard conditions along each edge in the path, we generate the corresponding constraints. For instance, the guard $x == 1$ in edge $e_{8}^{7}$ is represented as the constraint $x_{l_7}^{out} = 1$.  
\end{enumerate}
}

\noindent Conjunction of all the constraints constitutes the path formula, which we check for satisfiability. Below are the location-wise constraints for the path in the example. 

{\allowdisplaybreaks
\begin{align*}
    & \mathbf{Location\ l_7:}\ x_{l_7}^{in} = 0.5,\ y_{l_7}^{in} = 1.5,\ c_{l_7}^{in} = 10,\ t_0 \geq 0,\ x_{l_7}^{out} = x_{l_7}^{in} + t_0*\dot{x},\\
    & y_{l_7}^{out} = y_{l_7}^{in} + t_0*\dot{y},\ x_{l_7}^{out} = 1,\ c_{l_7}^{out} = c_{l_7}^{in} + (-2)*t_0,\ 0 \leq x_{l_7}^{in} \leq 1,\ 0 \leq x_{l_7}^{out} \leq 1,\\
    & 1 \leq y_{l_7}^{in} \leq 2,\ 1 \leq y_{l_7}^{out} \leq 2,\ c_{l_7}^{in} \geq 0.1,\ c_{l_7}^{out} \geq 0.1;\\
    & \mathbf{Location\ l_8:}\ x_{l_8}^{in} = x_{l_7}^{out},\ y_{l_8}^{in} =  y_{l_7}^{out},\ c_{l_8}^{in} = c_{l_7}^{out},\ t_1 \geq 0,\ x_{l_8}^{out} = x_{l_8}^{in} + t_1*\dot{x},\\
    & y_{l_8}^{out} = y_{l_8}^{in} + t_1*\dot{y},\ x_{l_8}^{out} = 2,\ c_{l_8}^{out} = c_{l_8}^{in} + (-2)*t_1,\ 1 \leq x_{l_8}^{in} \leq 2,\ 1 \leq x_{l_8}^{out} \leq 2,\\
    & 1 \leq y_{l_8}^{in} \leq 2,\ 1 \leq y_{l_8}^{out} \leq 2,\ c_{l_8}^{in} \geq 0.1,\ c_{l_8}^{out} \geq 0.1;\\
    & \mathbf{Location\ l_9:}\ x_{l_9}^{in} = x_{l_8}^{out},\ y_{l_9}^{in} =  y_{l_8}^{out},\ c_{l_9}^{in} = c_{l_8}^{out},\ t_2 \geq 0,\ x_{l_9}^{out} = x_{l_9}^{in} + t_2*\dot{x},\\
    & y_{l_9}^{out} = y_{l_9}^{in} + t_2*\dot{y},\ x_{l_9}^{out} = 3,\ c_{l_9}^{out} = c_{l_9}^{in} + (-2)*t_2,\ 2 \leq x_{l_9}^{in} \leq 3,\ 2 \leq x_{l_9}^{out} \leq 3,\\
    & 1 \leq y_{l_9}^{in} \leq 2,\ 1 \leq y_{l_9}^{out} \leq 2,\ c_{l_9}^{in} \geq 0.1,\ c_{l_9}^{out} \geq 0.1;\\
    & \mathbf{Location\ l_{10}:}\ x_{l_{10}}^{in} = x_{l_9}^{out},\ y_{l_{10}}^{in} =  y_{l_9}^{out},\ c_{l_{10}}^{in} = c_{l_9}^{out},\ t_3 \geq 0,\ x_{l_{10}}^{out} = x_{l_{10}}^{in} + t_3*\dot{x},\\
    & y_{l_{10}}^{out} = y_{l_{10}}^{in} + t_3*\dot{y},\ y_{l_{10}}^{out} = 2,\ c_{l_{10}}^{out} = c_{l_{10}}^{in} + (-4)*t_3,\ 3 \leq x_{l_{10}}^{in} \leq 4,\ 3 \leq x_{l_{10}}^{out} \leq 4,\\
    & 1 \leq y_{l_{10}}^{in} \leq 2,\ 1 \leq y_{l_{10}}^{out} \leq 2,\ c_{l_{10}}^{in} \geq 0.1,\ c_{l_{10}}^{out} \geq 0.1;\\
    & \mathbf{Location\ l_{16}:}\ x_{l_{16}}^{in} = x_{l_{10}}^{out},\ y_{l_{16}}^{in} =  y_{l_{10}}^{out},\ c_{l_{16}}^{in} = c_{l_{10}}^{out},\ t_4 \geq 0,\ x_{l_{16}}^{out} = x_{l_{16}}^{in} + t_4*\dot{x},\\ 
    & y_{l_{16}}^{out} = y_{l_{16}}^{in} + t_4*\dot{y},\ y_{l_{16}}^{out} = 3,\ c_{l_{16}}^{out} = c_{l_{16}}^{in} + (-4)*t_4,\ 3 \leq x_{l_{16}}^{in} \leq 4,\ 3 \leq x_{l_{16}}^{out} \leq 4,\\
    & 2 \leq y_{l_{16}}^{in} \leq 3,\ 2 \leq y_{l_{16}}^{out} \leq 3,\ c_{l_{16}}^{in} \geq 0.1,\ c_{l_{16}}^{out} \geq 0.1;\\
    & \mathbf{Location\ l_{22}:}\ x_{l_{22}}^{in} = x_{l_{16}}^{out},\ y_{l_{22}}^{in} =  y_{l_{16}}^{out},\ c_{l_{22}}^{in} = c_{l_{16}}^{out},\ t_5 \geq 0,\ x_{l_{22}}^{out} = x_{l_{22}}^{in} + t_5*\dot{x},\\
    & y_{l_{22}}^{out} = y_{l_{22}}^{in} + t_5*\dot{y},\ x_{l_{22}}^{out} = 4,\ c_{l_{22}}^{out} = c_{l_{22}}^{in} + t_5*0,\ 3 \leq x_{l_{22}}^{in} \leq 4,\ 3 \leq x_{l_{22}}^{out} \leq 4,\\
    & 3 \leq y_{l_{22}}^{in} \leq 4,\ 3 \leq y_{l_{22}}^{out} \leq 4,\ c_{l_{22}}^{in} \geq 0.1,\ c_{l_{22}}^{out} \geq 0.1;\\
    & \mathbf{Location\ l_{23}:}\ x_{l_{23}}^{in} = x_{l_{22}}^{out},\ y_{l_{23}}^{in} =  y_{l_{22}}^{out},\ c_{l_{23}}^{in} = 10,\ t_6 \geq 0,\ x_{l_{23}}^{out} = x_{l_{23}}^{in} + t_6*\dot{x},\\
    & y_{l_{23}}^{out} = y_{l_{23}}^{in} + t_6*\dot{y},\ c_{l_{23}}^{out} = c_{l_{23}}^{in} + (-2)*t_6,\ 4 \leq x_{l_{23}}^{in} \leq 5,\ 4 \leq x_{l_{23}}^{out} \leq 5,\\
    & 3 \leq y_{l_{23}}^{in} \leq 4,\ 3 \leq y_{l_{23}}^{out} \leq 4,\ c_{l_{23}}^{in} \geq 0.1,\ c_{l_{23}}^{out} \geq 0.1,\ y_{l_{23}}^{out} = 3;\\
    & \mathbf{Location\ l_{17}:}\ x_{l_{17}}^{in} = x_{l_{23}}^{out},\ y_{l_{17}}^{in} =  y_{l_{23}}^{out},\ c_{l_{17}}^{in} = c_{l_{23}}^{out},\ t_7 \geq 0,\ x_{l_{17}}^{out} = x_{l_{17}}^{in} + t_7*\dot{x},\\
    & y_{l_{17}}^{out} = y_{l_{17}}^{in} + t_7*\dot{y},\ c_{l_{17}}^{out} = c_{l_{17}}^{in} + (-2)*t_7,\ 4 \leq x_{l_{17}}^{in} \leq 5,\ 4 \leq x_{l_{17}}^{out} \leq 5,\\
    & 2 \leq y_{l_{17}}^{in} \leq 3,\ 2 \leq y_{l_{17}}^{out} \leq 3,\ c_{l_{17}}^{in} \geq 0.1,\ c_{l_{17}}^{out} \geq 0.1,\ x_{l_{17}}^{out} = 5;\\
    & \mathbf{Location\ l_{18}:}\ x_{l_{18}}^{in} = x_{l_{17}}^{out},\ y_{l_{18}}^{in} =  y_{l_{17}}^{out},\ c_{l_{18}}^{in} = c_{l_{17}}^{out},\ t_8 \geq 0,\ 5 \leq x_{l_{18}}^{in} \leq 6,\\
    & 2 \leq y_{l_{18}}^{in} \leq 3,\ c_{l_{18}}^{in} \geq 0.1,\\
\end{align*}
}

\subsection*{Path Feasibility Analysis in the Human model}
\noindent
For a feasible path $p_H$ obtained from $G_H$, we check if a feasible $run$ can be obtained by solving the path constraints for satisfiability. This is then solved by a \textit{Linear Programming} (LP) tool (\cite{DBLP:journals/entcs/LiAB07}). If the set of constraints is \textit{satisfiable} (SAT), a run exists which implies that the path $p_H$ has a plan.  
The result of executing this plan from the initial state $Init$ leads to the goal state $Goal$, and satisfies the goal condition, 
along with each location dynamics, beginning from the initial location, as specified in the problem $Prob$. 




\subsection*{Plan Infeasibility Analysis in the Agent model}
\noindent
Once a plan in the human model is obtained corresponding to $p_H$, we proceed to analyze it in the agent model $\mathcal{HA}_A$ using a similar constraint encoding strategy as shown above using the corresponding path $p_A$ obtained from $\mathcal{G}_A$. Evidently, since $Prob$ is unsolvable in the agent model, no valid $run$ corresponding to $p_A$ 
can be constructed in $\mathcal{HA}_A$. Thus, the $Solve$ step as done in Step 3 of Algorithm~\ref{algo:ca_replay} 
returns $unsatisfiable$  and we proceed to extract the IIS (\cite{DBLP:journals/informs/ChinneckD91}) to find the set of infeasible constraints that are collectively unsatisfiable for the given linear program encoding the run in the agent model.
IIS can be defined as below:

\begin{definition}
    IIS (\cite{DBLP:journals/informs/ChinneckD91}): An irreducible infeasible set (IIS) for a set of path constraints is a minimal set of inconsistent constraints. $\Box$
\end{definition}

\noindent Many software packages are available that support the efficient analysis of a linear constraint set and locating of the IIS, such as MINOS (\cite{CHINNECK19941}), IBM CPLEX (\cite{CPLEX}), and LINDO (\cite{LINDO}).
Now coming back to our discussion on the path constraints, the IIS of Example \ref{example:example} is of the path \{$c_{l_{10}}^{in} = c_{l_9}^{out}$, $c_{l_{10}}^{out} = c_{l_{10}}^{in} + (-4)*t_3$, $c_{l_{10}}^{in} \geq 0.1$, $c_{l_{10}}^{out} \geq 0.1$, $c_{l_{16}}^{in} = c_{l_{10}}^{out}$, $c_{l_{16}}^{out} = c_{l_{16}}^{in} + (-4)*t_4$, $c_{l_{16}}^{in} \geq 0.1$, $c_{l_{16}}^{out} \geq 0.1$, $c_{l_{22}}^{in} = c_{l_{16}}^{out}$, $c_{l_{22}}^{in} \geq 0.1$, $c_{l_{22}}^{out} \geq 0.1$\}
This IIS can be mapped back to the original elements in the path as illustrated in (\cite{DBLP:conf/hvc/BuYL11}) to find the path-segment for which the whole path becomes infeasible.
For example, the IIS constraints set of the above can be mapped to the path-segment $l_{10} \xrightarrow{e_{16}^{10}} l_{16} \xrightarrow{e_{22}^{16}} l_{22}$, which is actually infusing inconsistency for the whole path $p$ to become infeasible.
We denote these path-segments as \emph{IIS path-segments}.

\begin{definition} \label{def:IIS_path-segments}
    IIS path-segment: An {IIS path-segment} $ips$ is a segment of a path $p$ for which $p$ becomes infeasible as the underlying set of constraints of $ips$ is inconsistent. $\Box$
\end{definition}

\noindent The IIS path-segment $p''$ for our example path $p$ is $l_{10} \xrightarrow{e_{16}^{10}} l_{16} \xrightarrow{e_{22}^{16}} l_{22}$. Further, this path segment can be used for pruning paths in the human model in our algorithm. This is because any path $p'$ that contains the path segment $p''$ will certainly be infeasible~(\cite{DBLP:conf/hvc/BuYL11}) in the agent model, and the explanation of the infeasibility is derived in the IIS. We can therefore discard $p'$ for explanation generation in our algorithm.

Now coming back to our Algorithm \ref{algo:ca_replay}, since $p_A$ is infeasible in $\mathcal{HA}_A$, we add $p_A$ to the set \emph{IP}. All extensions of $p_A$ are also discarded to reduce the number of paths being checked.
We store the \emph{IIS path-segment} in the set \emph{S}. 

\begin{algorithm}
  \caption{\textsf{c$_A$\_infeasibility($p_A$, $\mathcal{HA}_A$)}}
  \scriptsize
  \label{algo:ca_replay}
  \LinesNumbered
  \SetAlgoLined
  \KwIn{A path $p_A$, and the agent's model $\mathcal{HA}_A$.\\}
  \KwOut{Return \emph{IIS path-segments}.\\}
  \Begin{
    $c_A$ = Get constraints of $p_A$ in $\mathcal{HA}_A$;\\
    Solve($c_H$); \\
    \emph{IIS-Path\_Segments} = Find\_IIS($c_A$) \tcc*{Use BACH}
    Add $p_A$ to $IP$;\\
    return \emph{IIS-Path\_Segments};
  }
\end{algorithm}

\noindent We give an explanatory example of continuous path analysis through which \emph{IIS path-segments} along a path can be identified. These \emph{IIS path-segments} are conveyed as explanations for the infeasibility of the path.

\begin{exmp}
    Consider the path in the human model of Example \ref{example:example} shown as a Blue dotted line in Figure \ref{fig:warehouse_automation2a}, which is feasible in the abstract graph structure $\mathcal{G}_A$ of the agent model. Continuous feasibility analysis in the human model reveals that the path is feasible, i.e., there is a corresponding plan in the human model. We take the corresponding path of the agent model and perform infeasibility analysis of the path with the continuous dynamics of the agent model which returns the \emph{IIS path-segment} $l_{10}$-$e_{16}^{10}$-$l_{16}$-$e_{22}^{16}$-$l_{22}$ for which the path becomes infeasible. We add the \emph{IIS path-segment} to the \emph{bad-segments} set \emph{S}. In following iterations, any path that consists of this \emph{IIS path-segment} is considered an infeasible path and not further assessed. For example, the path $l_7$-$e_{1}^{7}$-$l_1$-$e_{2}^{1}$-$l_2$-$e_{3}^{2}$-$l_{3}$-$e_{9}^{3}$-$l_9$-$e_{10}^{9}$-$l_{10}$-$e_{16}^{10}$-$l_{16}$-$e_{22}^{16}$-$l_{22}$-$e_{23}^{22}$-$l_{23}$-$e_{17}^{23}$-$l_{17}$-$e_{18}^{17}$-$l_{18}$ also consists of 
    the \emph{IIS path-segment} $l_{10}$-$e_{16}^{10}$-$l_{16}$-$e_{22}^{16}$-$l_{22}$ as a sub-path, and thus is known to be infeasible with IIS already extracted. Thus, this path will not be considered for explanation generation.
\end{exmp}

\subsection{Explanation and Human Model Update}

\noindent Our method provides explanations for the unsolvable planning problems in hybrid systems through a path-oriented continuous reconciliation process between the human model and the agent model.
The MRP involves the AI agent providing an explanation or model update to the human so that in the new updated human model, the unsolvability of the planning problem can be understood. As plans are projected to paths, the explanations provided by our method are of the following types:

\noindent
\paragraph{Path-wise explanation and model update:} A path-wise update of model differences, either in terms of transitions (through discrete feasibility analysis) or in terms of IIS path-segments (through continuous feasibility analysis) provides a short and precise explanation with respect to a single path.

\noindent
\paragraph{Unsolvability explanation by model reconciliation:} Providing a consolidated collection of model differences in terms of transitions and IIS path-segments along all paths.

\subsection*{Optimization}
\noindent
To optimize our technique, we prune paths from the human model while searching for a candidate path to reduce the number of path analysis steps, which is quite resource-consuming. We prune paths as follows:
\begin{itemize}
    \item We prune an \emph{invalid-edge} from the human graph structure to reduce the number of path computations.
    \item If a path is an extension of any \emph{infeasile path} from \emph{IP}, we discard the path.
    \item If a path consists of any of the \emph{IIS path-segments} as a sub-path from the set \emph{S}, we discard the path.
\end{itemize}

\section{Evaluation} \label{evaluation}
\noindent
In this section, we discuss the application of our framework on 2 case studies of planning problem domains of hybrid systems, (a) a warehouse automation system and (b) a water-level monitoring system. A brief description of the warehouse automation domain is presented in Section \ref{motive}, and a pictorial overview is shown in Figure \ref{fig:warehouse_automation}. It has 24 locations in both models (the human model and the agent model). We have used three planning problem instances on this domain, where they differ in the initial, goal, and blocked locations, as well as in the number of transitions in the domain. The water-level monitoring domain is presented with one planning problem instance.
This domain has 6 locations and 6 transitions in both models. 
All experiments are performed on a machine with 8 GB RAM, an Intel Core i5-8250U@1.60GHz, 8 8-core processors with Ubuntu 18.04 64-bit OS. The example warehouse automation domain, the problem files, and the code base can be found at: \url{https://gitlab.com/Sazwar/XSpeed-plan}.

Table \ref{ch4table: table1} shows results for three planning problem instances of the warehouse automation system and one planning problem instance of the water level monitoring system. Each planning problem instance is presented with varying plan depth to show the utility of this framework on larger plans, as well as on the plan space exploration. \emph{Benchmark} represents the planning domain, warehouse automation, and the water level monitoring system. \emph{Ins} represents the planning problem instances of the domains, while \emph{Dep} presents the bound on the plan length. \emph{Loc} shows the number of locations in the hybrid automata of the human and the agent models. \emph{Trans} denotes the number of edges in the hybrid automata of the human model (HM) and the agent model (AM), respectively. \emph{No. of paths} specifies the number of abstract paths obtained from the initial to the goal location for the planning problem instance in the human model obtained from the abstract graph structure. Now, recall that first, a path is to be checked in the agent’s discrete graph. The path may or may not be feasible with the continuous dynamics of the human model. Once the path is found in the agent’s discrete graph, we proceed to check for its feasibility in the continuous dynamics of the human model. \emph{No. of inf. paths} specifies the number of paths that are infeasible in the human model itself when the continuous dynamics are brought in. \emph{Paths replayed} presents the number of paths of the human model that are successfully replayed in the abstract graph structure of the agent model.
\emph{Invalid edges} shows the number of edges found that are present in the human model but not in the agent model. \emph{IIS segs} shows the number of IIS path-segments found. \emph{Time} and \emph{Mem} report the corresponding execution time and memory usage incurred by our framework for reconciliation.

\textcolor{black}{Table \ref{ch4table: table2} reports a path-wise average explanation generation time incurred by us. Since our reconciliation is path-based, our framework can also be used to analyze a single path to quickly pinpoint the infeasibility of a given plan. As earlier, \emph{Benchmark}, \emph{Ins}, and \emph{Depth} represent the domains, the planning problem instances, and the plan depth, respectively. The column \emph{Avg time} presents the average explanation generation time of a single path on those instances.}

\begin{table*}[]
  \centering\tiny
  \resizebox{\textwidth}{!}{
    \begin{tabular}{|c|c|c|c|c|c|c|c|c|c|c|c|c|}
    \hline
           & & & & \multicolumn{2}{|c|}{Trans} & \multicolumn{2}{|c|}{No. of (in HM)} & Paths & Invalid & IIS & Time & Mem \\\cline{5-8}
         Benchmark & Ins & Dep & {Loc} & HM & AM & paths & inf. paths & replayed & edges & segs & (in sec) & (in MB) \\
    \hline 
           & & 7 &  &  &  & 3 & 0 & 3 & 0 & 3 & 0.96 & 64.9 \\\cline{3-3} \cline{7-13}
           & Prob01 & 10 & 24 & 56 & 50 & 64 & 6 & 58 & 1 & 55 & 2.90 & 77.6 \\\cline{3-3} \cline{7-13}
           & & 15 &  &  &  & 98862 & 21352 & 77510 & 3 & 41513 & 13119.16 & 1993.7 \\\cline{2-13}
           
           \multirow{2}{*}{Warehouse} & & 7 &  &  &  & 2 & 2 & 0 & 1 & 0 & 1.20 & 71.1 \\\cline{3-3} \cline{7-13}
           \multirow{2}{*}{automation} & Prob02 & 10 & 24 & 56 & 50 & 38 & 32 & 6 & 3 & 3 & 1.48 & 70.3 \\\cline{3-3} \cline{7-13}
           & & 15 &  &  &  & 57694 & 41782 & 15912 & 3 & 3476 & 2837.01 & 1211.3 \\\cline{2-13}

           & & 7 &  &  &  & 2 & 2 & 0 & 1 & 0 & 0.72 & 75.7 \\\cline{3-3} \cline{7-13}
           & Prob03 & 10 & 24 & 54 & 48 & 262 & 198 & 64 & 3 & 56 & 3.36 & 75.5 \\\cline{3-3} \cline{7-13}
           & & 15 &  &  &  & 21836 & 15196 & 6640 & 3 & 3226 & 677.72 & 544.8\\
    \hline
           
           \multirow{2}{*}{Water-level} & & 6 &  &  &  & 1 & 0 & 1 & 0 & 1 & 0.53 & 61 \\\cline{3-3} \cline{7-13}
           \multirow{2}{*}{monitoring} & Prob01 & 20 & 6 & 6 & 6 & 5 & 0 & 2 & 0 & 2 & 0.96 & 60.2 \\\cline{3-3} \cline{7-13}
           & & 50 &  &  &  & 12 & 0 & 2 & 0 & 2 & 1.17 & 61.4 \\

    \hline
    \end{tabular}
  }
  \caption{Results on the warehouse automation and water-level monitoring systems. HM and AM represent Human Model and Agent Model, respectively.}
  \label{ch4table: table1}
\end{table*}

\begin{table}[]
  \centering\tiny
  \scalebox{1.5}{
    \begin{tabular}{|c|c|c|c|}
    \hline
        \multirow{2}{*}{Benchmark} & \multirow{2}{*}{Ins} & \multirow{2}{*}{Depth} & Avg time\\
         & & & (in sec)\\
    \hline 
        & \multirow{3}{*}{Prob01} & 7 & 0.32\\\cline{3-4}
        & & 10 & 0.045\\\cline{3-4}
        & & 15 & 0.132\\\cline{2-4}

        \multirow{2}{*}{Warehouse} & \multirow{3}{*}{Prob02} & 7 & 0.6\\\cline{3-4}
        \multirow{2}{*}{automation} & & 10 & 0.038\\\cline{3-4}
        & & 15 & 0.049\\\cline{2-4}

        & \multirow{3}{*}{Prob03} & 7 & 0.36\\\cline{3-4}
        & & 10 & 0.013\\\cline{3-4}
        & & 15 & 0.031\\\cline{2-4}        
    \hline
        \multirow{2}{*}{Water-level} & \multirow{3}{*}{Prob01} & 6 & 0.53\\\cline{3-4}
        \multirow{2}{*}{monitoring} & & 20 & 0.192\\\cline{3-4}
        & & 50 & 0.097\\
        
    \hline
    
    \end{tabular}
    }
  \caption{Average explanation generation time.}
  \label{ch4table: table2}
\end{table}

\emph{Analysis of results:} For all problem instances of the warehouse automation and the problem instance of the water-level monitoring system, the human model and the agent model have the same number of locations. However, the number of transitions in these two models varies in the warehouse automation domain, whereas they have the same number of transitions in the water-level monitoring domain. For each of the problem instances, we present results for varying plan depth. However, the number of paths explored increases exponentially with an increase in plan depth in the warehouse automation domain. 
The difference between the \emph{No. of paths in the human model (HM)} and the \emph{Paths replayed} in the agent model shows the number of paths pruned for optimization to speed up the process.
For the three planning problem instances in the warehouse automation domain, $21.58\%$, $72.42\%$, and $69.66\%$ of the paths are discarded, respectively, while searching for explanations. In the water-level monitoring domain, $72.22\%$ of paths are similarly pruned from generating any explanation.
The \emph{Invalid edges} presents only the explanation edge set that is not present in the agent model along the checked paths. Once an invalid-edge is found along a path, that path is not assessed further to speed up the execution. Hence, an invalid-edge that might appear later in the path is not checked.
The \emph{IIS segs} presents the number of distinct path-segments along which the planning problem is unsolvable.
The \emph{Time} and the \emph{Mem} usages indicate that our framework works reasonably well with time consumption and memory utilization. However, with the increase of plan depth, the time and memory usage increase as the number of paths 
enumerated increases exponentially.
The average time for model reconciliation to the MRP problem is $1387.31$ seconds, where we consider all the possible plans for a planning problem.
\textcolor{black}{The average explanation generation time of a path is $0.2$ second, which is quite small and signifies that we can quickly reconcile.}
\section{Related Works} \label{related_work}
\noindent
Our work leverages the intersection of reachability analysis and planning $-$ our techniques are inspired by path-oriented bounded model checking (BMC) (\cite{DBLP:journals/ac/BiereCCSZ03}) approach, and the concept of locating minimal infeasible constraint sets in linear programs (\cite{DBLP:journals/informs/ChinneckD91}) and its application to BMC of linear hybrid automata (\cite{DBLP:journals/entcs/FranzleH05, DBLP:conf/fmcad/BuLWL08}). We apply these techniques to explain unsolvability of the planning problem through model reconciliation (\cite{DBLP:conf/ijcai/ChakrabortiSZK17}), which was introduced by the planning community in the context of explainable AI planning (XAIP) (\cite{DBLP:conf/rweb/HoffmannM19}).

A significant amount of research in XAIP focuses on generating explanations of solutions to planning problems, i.e., the problem of explaining why a given plan or action was chosen. However, explaining unsolvability of a planning problem remains a largely open and understudied problem in this area of research.
Some notable works that have tried to address unsolvability of planning problems mostly looked at verifying the unsolvability by generating certificates~(\cite{DBLP:conf/aips/ErikssonH20}), (\cite{DBLP:conf/aips/ErikssonRH17}) or proofs (\cite{DBLP:conf/aips/ErikssonRH18}) rather than explaining the causalities of unsolvability of the planning problem. Such certificates or proofs of unsolvability are not enough to increase the human understandability of why the problem was unsolvable.
Additionally, most of these works are on classical planning problems on discrete domains. Verifying unsolvability of planning problems in hybrid systems comes with an additional challenge since these planning problems are undecidable in general (\cite{ALUR19953}). (\cite{DBLP:journals/tecs/SarwarRB23}) provides a way for addressing unsolvability of a planning problem in hybrid domains by $\delta$-approximate bounded reachability analysis (\cite{DBLP:journals/corr/GaoKCC14}).

Few notable works that are directed towards explaining the unsolvability of a planning problem are, similarly, limited to classical planning problems.
(\cite{DBLP:conf/aips/GobelbeckerKEBN10}) argues that excuses can be produced for why a plan cannot be found. This work proposes a formalization of counterfactual alterations to the original planning task, such that the new planning task 
turns out to be solvable, and provides an algorithm to find these excuses.
(\cite{DBLP:conf/ijcai/SreedharanSSK19}) uses hierarchical model abstractions to generate the reason for unsolvability of planning problems. These hierarchical model abstractions relax a planning problem until a solution can be found. Then, they look for landmarks of this relaxed problem that cannot be satisfied in less relaxed versions of the problem. The unsatisfiability of these landmarks provides a succinct description of critical propositions that cannot be satisfied.
(\cite{DBLP:conf/aaai/EiflerC0MS20}) has taken a somewhat different approach by deriving properties that must be exhibited by all possible plans that could serve as explanations in case of unsolvability.
However, generating excuses, hierarchically abstracting models, or deriving plan properties in terms of propositional formulas may not be enough to understand why a problem was unsolvable for complex domains like planning problems of hybrid systems, which encode mixed discrete and continuous dynamics. This motivates a reconciliation step.

MRP (\cite{DBLP:conf/ijcai/ChakrabortiSZK17}) has been a popular theme in this direction, where explanations are provided as the reconciliation between two models (the human model and the AI model) by bringing them closer. However, most of the works in the MRP literature have been applied to classical planning problem domains.
A logic-based extension to the MRP has been applied to mixed discrete-continuous domains~(\cite{DBLP:journals/jair/VasileiouYSKCM22}). In this work, the authors approach the MRP based on knowledge representation and reasoning. It provides a framework that finds a subset of the knowledge base of the agent with which to reconcile the human knowledge base for explanations. However, the explanation is provided for why a plan is feasible in a model rather than addressing the unsolvability of a planning problem head-on, as is done by us. 
In contrast, our approach to MRP is based on a path-based continuous model reconciliation process through communications between the agent and the human models that are presented as hybrid automata. Hybrid automata present a natural 
description formalism for hybrid domains, in addition to specialized language extensions (e.g., PDDL+). We believe 
that the hybrid automaton structure, being state-based, provides a useful design methodology for such otherwise 
complex domains. The adoption of hybrid automata and explanations produced thereof makes our work quite novel. 
To the best of our knowledge, explanations and reconciliations on hybrid automatons for hybrid domain models have not been addressed in the literature. Added to that is our approach 
of dealing with the reconciliation task in two different steps, first in the abstract graph, and then, in the 
continuous dynamics is new as well. This distinguishes our work from other approaches available in MRP literature. 




\section{Conclusion} \label{conclusion}
\noindent
In this work, we explore the area of explaining the unsolvability of planning problems for hybrid systems. We propose a path-based \emph{continuous model reconciliation} framework for explaining unsolvability. We show that a discrete and continuous path analysis approach can be leveraged to derive explanations of the unsolvability of a planning problem in hybrid domains.
While discrete path analysis falsifies a path by mapping each location and transition of the path in the human model to the agent model, continuous path analysis leverages reachability analysis, combining with minimal inconsistent constraint sets to find the infeasible path-segments for which the paths become unsatisfiable. We demonstrate a warehouse automation system as an example planning domain, along with a water-level monitoring system, and present results for problem instances on each domain with varying plan depth. We show that the average explanation generation time of this framework is significantly small, which can be utilized to quickly pinpoint any infeasibilities of a given plan. While the path-oriented analysis explores the plan space to find all the possible paths, it can track most of the causes that contribute to the unsolvability of the planning problem. As future work, we plan to include provisions to update and reconcile the continuous dynamics which is not considered in this version. Additionally, we intend to explore more domain models and instances going forward.

\chapter{Exploring Inevitable Waypoints for Unsolvability Explanation in Hybrid Planning Problems} \label{ch5}

\textbf{\textsc{Chapter Abstract:}} \textit{Explaining unsolvability of planning problems is of significant research interest in Explainable AI Planning. A number of research efforts on generating explanations of solutions to planning problems have been reported in AI planning literature. However, explaining the unsolvability of planning problems remains a largely open and understudied problem. A widely practiced approach to plan generation and automated problem solving, in general, is to decompose tasks into sub-problems that help progressively converge towards the goal. In this work, we propose to adopt the same philosophy of sub-problem identification as a mechanism for analyzing and explaining unsolvability of planning problems in hybrid systems. In particular, for a given unsolvable planning problem, we propose to identify common waypoints, which are universal obstacles to plan existence; in other words, they appear on every plan from the source to the planning goal. This work envisions such waypoints as sub-problems of the planning problem and the unreachability of any of these waypoints as an explanation for the unsolvability of the original planning problem. We propose a novel method of waypoint identification by casting the problem as an instance of the longest common subsequence problem, a widely popular problem in computer science, typically considered as an illustrative example for the dynamic programming paradigm. 
Once the waypoints are identified, we perform symbolic reachability analysis on them to identify the earliest unreachable waypoint and report it as the explanation of unsolvability. We present experimental results on unsolvable planning problems in hybrid domains. }

\section*{}
\lettrine[lraise=0.3, nindent=0em, slope=-.5em]In human-computer interaction (HCI), humans engage in trustworthy collaborations with autonomous agents, and one of the precursors of such collaborations is that an autonomous agent must explain the rationale behind its decision to the human. With the emergence of artificial intelligence (AI) and the multitude of application domains where AI planning is being envisioned to replace plans generated by humans, Explainable AI Planning (XAIP)~(\cite{DBLP:journals/corr/abs-1709-10256,DBLP:conf/rweb/HoffmannM19}) has emerged as an important connection between HCI and AI for designing explainable systems that bridges the gap between theoretical and algorithmic planning and real-world applications (\cite{DBLP:conf/ijcai/ChakrabortiSK20}).
While there has been a lot of research on generating explanations to planning problems, most of the earlier works in explanation generation have focused on explaining why a given plan or action was chosen (\cite{DBLP:conf/ijcai/ChakrabortiSZK17,DBLP:conf/aips/ChakrabortiKSSK19,DBLP:journals/jair/KrarupKMLC021,DBLP:journals/tecs/SarwarRB23}). However, explaining the unsolvability of a given planning problem remains a largely open and understudied problem.
The recent works that focus on explaining the unsolvability of planning problems have primarily concentrated on generating certificates or proofs of unsolvability (\cite{DBLP:conf/aips/ErikssonRH17,DBLP:conf/aips/ErikssonRH18}), or on identifying counterfactual alterations to the original planning task to make it solvable, often referred to as "excuses" (\cite{DBLP:conf/aips/GobelbeckerKEBN10}). These approaches, which are more oriented towards automatic verification, may fall short in adequately explaining unsolvability in complex planning domains.

A well-known insight into human thinking and problem-solving is that humans tend to decompose a problem into sub-problems that help in progressively converging towards the goal. Many AI systems mimic this notion in the way they solve problems. For instance, the main feature of the pioneering automated theorem prover, \emph{logic theorist}, is the use of problem-subproblem hierarchy (\cite{LT-1956}). An innovative technique for identification of sub-problems relevant for explaining unsolvability of a planning problem in domains with discrete dynamics has been proposed in (\cite{DBLP:conf/ijcai/SreedharanSSK19}). 
In this work, we propose adopting the same philosophy of sub-problem identification as an efficient mechanism for analyzing and explaining the unsolvability of planning problems in hybrid domains, which are domains with a combination of discrete and continuous dynamics. In particular, for a given unsolvable planning problem, we propose to identify a sequence of waypoints that are universal obstacles to plan existence; in other words, they appear on every path on every plan from the source to the planning goal. This work envisions such waypoints as sub-problems of the planning problem and the unreachability of any of these waypoints as an explanation for the unsolvability of the original problem at hand. We propose a novel method of waypoint identification by casting the problem as an instance of the longest common subsequence problem, a widely popular problem in computer science, typically considered as an illustrative example for the dynamic programming paradigm. Once the waypoints are identified, we perform symbolic reachability analysis on them to identify the earliest unreachable waypoint and report it as the explanation of unsolvability. We present experimental results on unsolvable planning problems in hybrid domains. In summary, the key contributions of this work are:
\begin{enumerate}
 \item[(a)] A proposal for an artifact for explaining unsolvability of hybrid planning problems based on the identification of inevitable waypoints.
 \item[(b)] A method to generate the explanation artifact by casting it as an instance of the longest common subsequence problem, and subsequently using symbolic reachability analysis on the hybrid automaton.
\end{enumerate}

The rest of this chapter is organized as follows. In Section~\ref{ch5motive}, we present a motivating example to demonstrate this work. Section~\ref{ch5problem-overview} provides a background and problem overview.  
Section~\ref{ch5methodology} illustrates our methodology and the framework of explanation. Section~\ref{ch5eval} discusses implementation and results. Section~\ref{ch5related_works} presents related literature. Finally, Section~\ref{ch5conclusion} summarizes the contributions and the findings of this work and discusses possible future directions.
\section{Motivating Example} \label{ch5motive}
\begin{figure}[htbp]
    \centering
    \includegraphics[width=0.66\textwidth]{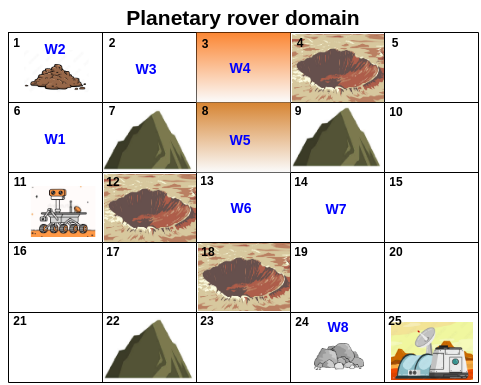}
    \caption{The rover domain is depicted. Initially, the rover is at cell 11. Mountains and craters are impassable regions of the terrain, such as cells 7, 12, etc. The rover needs to collect soil samples from cell 1 and rock samples from cell 24. The base station is at cell 25. The up-slope areas in the terrain are shown in orange. The inevitable waypoints in the domain are cells $w1$ to $w8$ marked in blue.}
    \label{fig:rover-domain}
\end{figure}

In this section, we present a motivating example in the context of a planning problem for a planetary rover that explores a planetary site and collects samples for experiments. The agent (an autonomous battery-powered rover) possesses knowledge about the topography of the exploration site as a planar grid.
Figure~\ref{fig:rover-domain} shows the topography as a 5$\times$5 grid of cells. The rover is initially positioned at cell 11, and there is a base-station at cell 25. The task of the rover is to collect soil and rock samples from designated sites in cells 1 and 24, respectively, and then reach the base-station. Mountainous regions and craters in the terrain are marked as impassable (cells 7, 12, etc). There are inclined areas in the terrain shown in orange cells. The motion dynamics of the rover is interpreted as a hybrid system. The rover's continuous motion and battery discharge dynamics can be different in each cell.
For instance, the motion and battery depletion dynamics in an inclined region is different from the dynamics in flat regions and regions where soil and rock samples are collected. When a rover makes a transition from one cell to another, it starts to follow the dynamics of the new cell instantaneously. The discrete dynamics here capture the connectivity of the cells in the presence of mountains and craters where the rover cannot move. The rover's movement is restricted to one of its adjacent cells, and movement to diagonal cells is prohibited.

When a planner reports the planning task as unsolvable, our algorithm identifies ordered waypoints that ought to be reached in any plan to achieve the task. For instance, in the discussed domain, our algorithm detects that the cells 6-1-2-3-8-13-14-24 must be visited by any valid plan to solve the planning task. These are marked as ordered waypoints $w1-w8$ in the figure. Our proposed algorithm envisions such waypoints as sub-goals of the planning problem. The unreachability of any of these waypoints under the domain dynamics is reported as an explanation for the unsolvability of the original planning problem. The computational challenge lies in finding the waypoints, finding an order between them, and lastly, finding the earliest waypoint that is unreachable under the dynamics.  We propose a method of waypoint identification by casting the problem as an instance of the longest common subsequence problem, a widely popular problem in computer science, typically considered as an illustrative example for the dynamic programming paradigm. Unreachability of a waypoint is determined using a bounded model checker. For instance, given the initial rover battery charge of 10 units and the battery depletion rates of the cells (depletion rate of 1 unit in cells except in cell 1 and cell 24 where soil and rock sampling depletes battery at a higher rate of 2 units, and in the cells with inclination having a depletion rate of 3 units), reachability analysis determines that cell 13 is the first unreachable waypoint and reports this as an explanation of unsolvability.

Some of the waypoints as sub-goals may be \textbf{explicitly} known from the planning problem description itself. For example, the planetary rover domain has two sub-goals explicitly mentioned, namely the collection of soil and rock samples from cell 1 and cell 24, respectively, marked as waypoints $w2$ and $w8$. There may be sub-goals that are not apparent from the problem description explicitly, but they are \textbf{implicitly} mandatory to complete the bigger planning task.
For example, the implicit waypoints in our planetary rover domain are marked $w1$ and $w3-w7$ in the figure. Our waypoint detection algorithm detects both the explicit as well as implicit waypoints.
We term these as inevitable waypoints.
In the following section, we formally describe the domain representation and the explanation problem we intend to solve.

\section{Problem Overview} \label{ch5problem-overview}

In this section, we provide a formal definition of a planning problem in a hybrid system. A hybrid system exhibits an interplay of discrete and continuous dynamics. Hybrid automata $\mathcal{H}$ (see Definition~\ref{ch2def:HA}) are a well-known mathematical model for such systems (\cite{ALUR19953,10.1007/3-540-57318-6_30}). 

\begin{figure*}[htbp]
    \centering
    \includegraphics[width = 0.95 \textwidth]{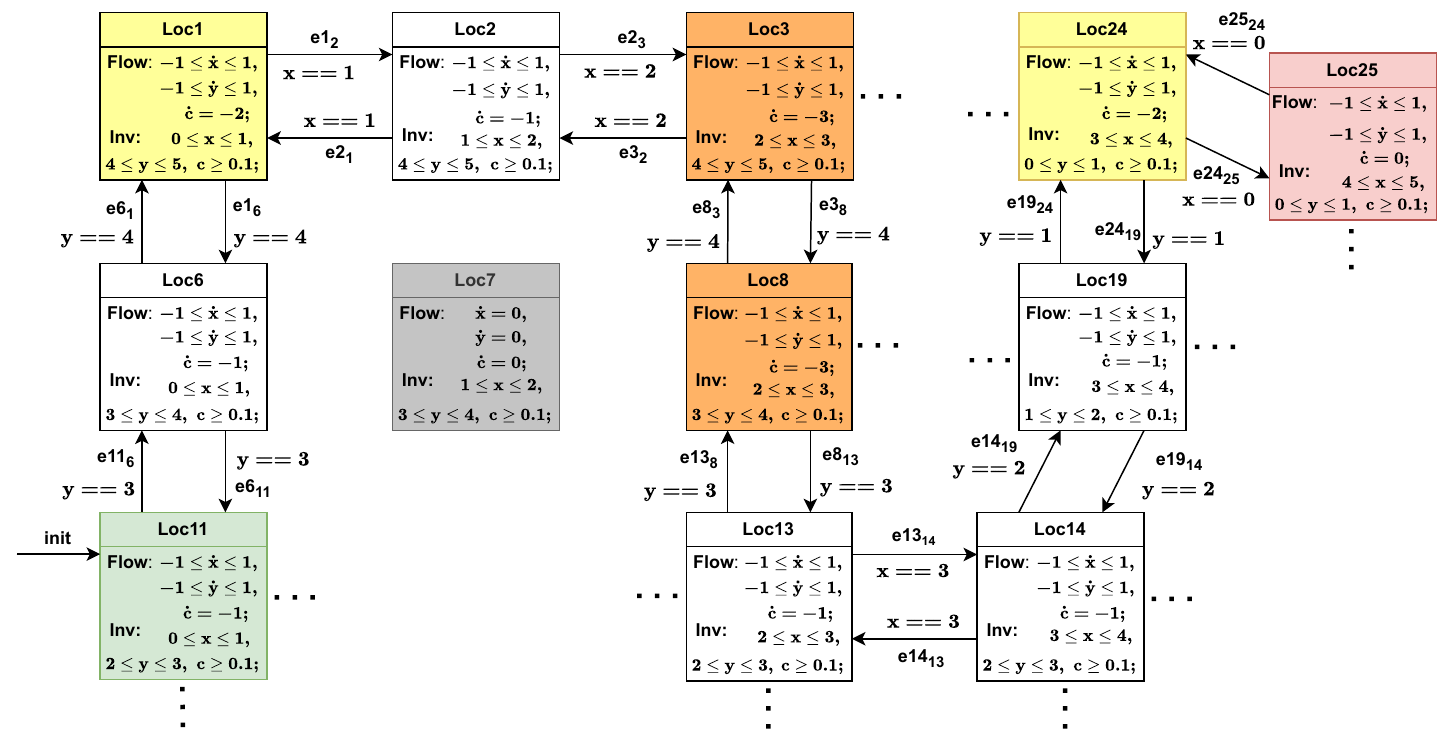}
    \caption{A hybrid automaton model of the planetary rover domain (partially shown).}
    \label{fig:rover-model}
\end{figure*}

A state of $\mathcal{H}$ is a pair $(l,v)$ consisting of a location $l \in Loc$ and a valuation $v\in $ \emph{Inv}($l$). The component $Init$ defines the set of initial states of the automaton. Figure~\ref{fig:rover-model} shows the hybrid automaton for the planetary rover domain, as an example. Each cell in Figure~\ref{fig:rover-domain} is represented as a location of the automaton with an invariant that is the region enclosed by the cell. The battery charge depletion rate $c$ and motion dynamics (described over the position variables $x$ and $y$) of the rover within the cell are modeled as flow equations of the location. The cell-to-cell movement of the rover is given as transitions between locations. The initial location is shown in green. The yellow locations represent the regions where the rover collects soil and rock samples. The orange locations represent the inclined regions. The base station, the rover's destination, is shown in red. An impassable location (loc7) is shown in grey with no incoming or outgoing edges. 

\begin{definition}
    \textcolor{black}{The graph of a hybrid automaton $\mathcal{H}$
    is defined as $\mathcal{G_{H}} = (V$, $E)$, where $V$ = \emph{Loc} and $E \subseteq \emph{Loc} \times \emph{Loc}$ such that for every $(l, a, g, r, l') \in \emph{Edge}$, there is an edge $(l, l') \in E$. $\Box$}
\end{definition}

\noindent A discrete evolution of the system is then represented as a path on the automaton graph. 

\begin{definition} \label{def:path}
    \textcolor{black}{A path $p$ between locations $l_0$ and $l_n$ in $\mathcal{G_{H}}$ is a sequence of $locations$ and $edges$ given as:
    \begin{align*}
    l_0 \xrightarrow{e_0} l_1 \xrightarrow{e_1} l_2 \xrightarrow{e_2} \ldots \xrightarrow{e_{n-1}} l_{n}
    \end{align*}
    where $l_i$ $\in$ $Loc$, $e_i$ $\in$ $Edge$, and $l_{i}$, $l_{i+1}$ are the source and destination of $e_i$ respectively. The length of a path is the number of edges it contains. 
    $\Box$}
\end{definition}

\noindent A planning problem typically consists of a domain description and the initial and goal states of the planning task. In this work, as part of the problem description, we also consider a bound on how many times the actions of the domain can be applied and a set of constraints that every valid plan must satisfy. As an underlying assumption, we consider the planning problem to have one initial and one goal location. This is not a significant limitation, since we can explain the unsolvability of a planning problem with multiple initial and goal locations by solving many explanation problems, each for a planning problem with an initial and goal location pair. We define a planning problem for a hybrid system as follows:

\begin{definition} \label{def:planning_problem}
A \emph{planning problem} $\Pi$ for a hybrid system is a three-tuple $(Dom$, $Prob$, $Depth)$, where
\begin{itemize}
  \item $Dom$ (\cite{DBLP:journals/tecs/SarwarRB23}) is represented as a hybrid automaton $\mathcal{H}$. 
  \item $Prob$ is a tuple $\langle Init, Goal \rangle$ representing a problem description, where $Init$ and $Goal$ define the initial and the goal states.
  \item \emph{Depth} defines the bound on how many actions can be applied in a plan. 
  $\Box$
\end{itemize}
\end{definition}

\noindent The set of initial states of the planning problem is given by $Init$, that is a tuple $\langle l_{0}, S_0 \rangle$ such that $l_{0} \in Loc$ and $S_{0} \subseteq$ \emph{Inv}($l_{0}$). The tuple represents initial states $\{(l_{0}, v) \mid v \in S_{0} \}$. Moreover, $Init$ must be a subset of the set of initial states of the automaton $\mathcal{H}$. The $Goal$ states of the planning problem are given as a tuple $\langle l_{goal}, S_{goal} \rangle$ such that $l_{goal} \in Loc$ and $S_{goal} \subseteq$ \emph{Inv}($l_{goal}$). The tuple represents goal states $\{(l_{goal}, v) \mid v \in S_{goal} \}$.
The set of labels of the hybrid automaton corresponds to the available actions for a plan. A state can change either due to a transition in the automaton or due to the passage of time, where the variables evolve according to the flow in a location. We refer to the former as \emph{discrete transition} and the latter as \emph{timed transition}. A discrete transition happens due to the application of an action given by the label of the transition, provided the valuation of the state satisfies the guard of the transition. On taking a discrete transition, the valuation of the new state must follow the reset map of the transition. We now define a plan for a planning problem $\Pi$:

\begin{definition}\label{def:plan}
A plan for a planning problem $\Pi$ is a tuple $\langle$ $\lambda_n$, $makespan$ $\rangle$ where $\lambda_n$ is a finite sequence of $n$ pairs $\langle t_i, a_i \rangle$. In the pair, $t_i \in \mathbb{R}^{\ge 0}$ is the time instance of executing the action $a_i \in$ \emph{Lab}. In the sequence, $t_i$ is non-decreasing. The $makespan$ is the duration of the plan. $\Box$
\end{definition}

\noindent We refer to the length of a plan to be the number of pairs $\langle t_i, a_i \rangle$ in $\lambda_n$. An \emph{executable} plan on a $\mathcal{H}$ is defined as follows:

\begin{definition}\label{def:run}
A plan $\langle \lambda_n, makespan \rangle$ for a planning problem $\Pi$ is called executable on the domain $\mathcal{H}$ of $\Pi$ if and only if the application of the plan on $\mathcal{H}$ results in an alternating sequence of timed and discrete transitions, called a run of the $\mathcal{H}$ depicted as:
\begin{align*}
(l_{0},v_{0}) &\xra{\tau_0}(l_{0},v'_0) \xrightarrow{a_1} (l_1,v_{1}) \xra{\tau_1} (l_1,v'_{1})\xrightarrow{a_2} \ldots \xrightarrow{a_{n}}(l_{n},v_n) \xra{\tau_{n}} (l_{n},v'_{n})
\end{align*}
where 
(i) $(l_{0},v_{0}) \in \emph{Init}$ 
(ii) $a_i$ is a label of some edge $e_i \in \emph{Edge}$ such that $l_{i-1}$ is the source and $l_{i}$ is the destination location of $e_i$, $v'_{i-1} \in g$ and $v_{i} \in r$ where $g$ is the guard and $r$ is the reset map of $e_i$, $\forall i \in \llbracket 1..n\rrbracket$ 
(iii) The transitions labeled $\tau_i \in \mathbb{R}^{\geq 0}$ represent timed transitions with $\tau_i$ being the time of dwelling in the location, with the constraint that $\forall t \in [0, \tau_i]$, the timed transition $(l_i,v_{i}) \xra{t} (l_i,v^+_{i})$ has  $v^+_{i} \in Inv(l_i)$, $\forall i \in \llbracket 0..n\rrbracket$ \}.
(iv) $\langle \sum_{j=0}^{i-1} {\tau_j}, a_i \rangle$ is a pair in $\lambda_n$, $\forall i \in \llbracket 1..n\rrbracket$ (v) $\sum_{i=0}^{n}{\tau_i}$ = \emph{makespan}.
$\Box$
\end{definition}

\noindent The length of a run is the number of discrete transitions it contains. In control-theoretic terms, a plan is a control strategy that acts on a plant, a hybrid automaton in our context. The application of a control strategy on a plant results in a controlled execution of the plant, which we call a \emph{run} in our context. Due to uncertainties modeled in a $\mathcal{H}$, an application of a plan may result in more than one run. A plan is called \emph{valid} if it is executable, i.e., its applications on $\mathcal{H}$ results in a run from a state in $Init$ to a state in $Goal$,
and the length of the plan is less than or equal to Depth. In the following text, we write "a run of a valid plan" as a short-form of saying "a resulting run of the domain $\mathcal{H}$ of $\Pi$ due to the application of a valid plan". A planning problem $\Pi$ is called solvable if it admits a valid plan. If no such plan exists, then the planning problem is said to be unsolvable.
\begin{definition}
    A planning problem $\Pi$ is unsolvable if it admits no valid plan. $\Box$
\end{definition}
\noindent The problem addressed in this work is as follows:

\begin{prob}
Given an unsolvable planning problem $\Pi$, generate an artifact automatically that explains why $\Pi$ is unsolvable.
\end{prob}

\noindent In the subsequent sections, we describe the details of the explanation artifact and the algorithm to generate the same.

\section{Methodology} \label{ch5methodology}
\begin{figure*}[htbp]
    \centering
    \includegraphics[width= \textwidth]{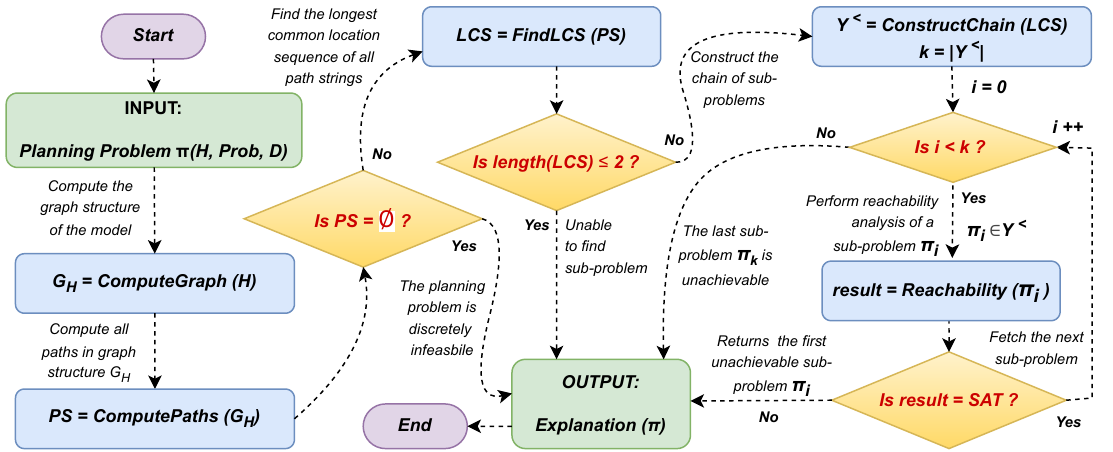}
    \caption{Proposed Explanation Framework.}
    \label{fig:framework}
\end{figure*}

\noindent In this section, we describe our explanation algorithm, which takes an unsolvable planning problem $\Pi$ as input and computes an artifact $Explanation (\Pi)$, which we define later in the text (Def.~\ref{def:explanation}). Our explanation algorithm attempts to divide an unsolvable planning problem $\Pi$ into several sub-problems, following the common divide-and-conquer paradigm of problem-solving. These sub-problems have the property that each must be solvable for $\Pi$ to be solvable. Identifying these sub-problems is computationally challenging and is the key to generating the explanation artifact. The proposed algorithm takes a layered approach. The sub-problems are determined by taking into consideration only the discrete dynamics of the hybrid automaton. Once the sub-problems are identified, the explanation artifact is generated by considering the hybrid dynamics in its entirety. The abstraction of the continuous dynamics in the first phase allows us to work in the domain of graphs, and consequently, we show a reduction from finding sub-problems to finding the longest common sub-sequence of a finite set of strings, a well-known problem in algorithms. Finally, the feasibility of the identified sub-problems is verified using symbolic reachability analysis, and the explanation is generated, which highlights which of the sub-problems is responsible for the unsolvability of $\Pi$. A schematic diagram of our explanation framework is shown in Figure~\ref{fig:framework}. We now present the details. 

\subsection{Decomposition into Sub-Problems}
We present here the central idea of this work, the binary relation among subproblems. Before presenting the idea of subproblems, we define below when a run of a valid plan of $\Pi$ is said to intersect with a set of states of the domain of the planning problem.   

\begin{definition} \label{intersection}
    \textcolor{black}{Given a run $r$ of a valid plan of $\Pi$ and a set of states $\langle l, S \rangle$ of the domain $\mathcal{H}$ of $\Pi$, the run $r$ is said to intersect with $\langle l, S \rangle$ when the following conditions hold: (i) there exists a timed transition $(l,v) \xra{\tau} (l,v')$ in $r$, (ii) there exists a dwelling time $t \in [0, \tau]$ such that the timed transition $(l,v) \xra{t} (l,x)$ has $x \in S$.}
\end{definition}

\noindent We now define the notion of a sub-problem as a relation between hybrid planning problems.
\begin{definition}\label{def:subproblem}
    Given planning problems $\Pi_i$ = $(Dom_i$, $Prob_i$, $Depth_i)$ and $\Pi_j$ = $(Dom_j$, $Prob_j$, $Depth_j)$, we say $\Pi_i$ is a sub-problem of $\Pi_j$ when the following conditions hold: (a) $Dom_i = Dom_j$, (b) $Depth_i = Depth_j$, and (c) $Prob_i$, $Prob_j$ are tuples $\langle$ $Init_i$, $Goal_i$ $\rangle$ and $\langle$ $Init_j$, $Goal_j$ $\rangle$ resp. where
    \begin{itemize}
        \item $Init_i = Init_j$ 
        
        \item For any run of a valid plan of $\Pi_j$ to exist, the run must intersect with $Goal_i$. 
    \end{itemize}
\end{definition}

\noindent Note that if $\Pi_i$ is a subproblem of $\Pi_j$, then $\Pi_i$ must be solvable for $\Pi_j$ to be solvable. In other words, it is impossible to have a run of a valid plan of $\Pi_j$ that does not intersect the goal states of $\Pi_i$. In the definition, there is no assumption of solvability or unsolvability of $\Pi_j$. For an unsolvable $\Pi_j$ where there exists no real run of a valid plan, the definition aims to convey that any "hypothetical" run of a valid plan of $\Pi_j$ must visit $Goal_i$.

\begin{figure*}[htbp]
  \centering
  \includegraphics[scale=0.50]{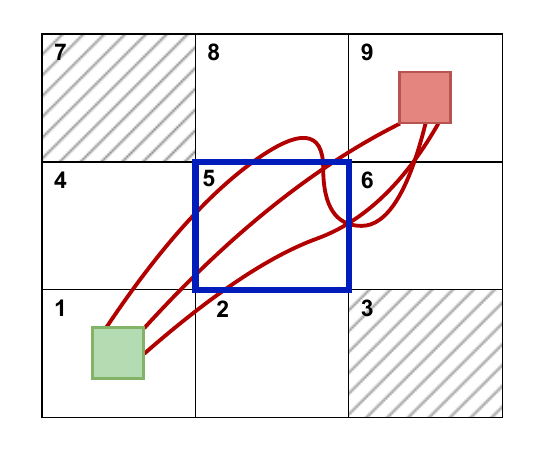}
  \caption{All runs of valid plans from the initial set to the goal set must pass through the blue doorway. We envision the blue region as an inevitable waypoint to the planning problem.}
  \label{fig:subgoal-example}
\end{figure*}

\begin{exmp} \label{exp:sub-problem-example}
    Consider a planning problem in a rover-like domain explained above, the state-space shown in Figure~\ref{fig:subgoal-example}. The states within each cell belong to the invariant of a distinct location of a $\mathcal{H}$ model with nine locations. The initial states are in the green region, and the goal states are in the red region. The shaded regions are impassable.
    Some of the runs of valid plans are shown as red trajectories from an initial state to a goal state.
    Observe that for any run of a valid plan to exist, it must pass through the region enclosed by the blue cell. This means that the region is an inevitable waypoint.
\end{exmp}

\noindent  In the following text, we interchangeably refer to a sub-problem as a \emph{waypoint}. We write $\Pi_i \le \Pi_j$ to say that $\Pi_i$ is a sub-problem of $\Pi_j$.  A planning problem $\Pi$ can have multiple sub-problems. We denote the set of all sub-problems of $\Pi$ by $\Pi^{\sqcup}$.
\begin{equation}
\Pi^{\sqcup} = \{ \Pi' \mid \Pi' \le \Pi \}
\end{equation}

\subsubsection{Abstraction over Planning Problems}

\noindent The cardinality of $\Pi^{\sqcup}$ can be potentially infinite. We therefore present the following construction of a finite set of planning problems $\Pi^*$ for a given $\Pi$ induced by its $\mathcal{H}$ domain. We also define an abstraction function $\alpha$ that maps each $\Pi' \in \Pi^{\sqcup}$ to an element of $\Pi^*$ such that if $\Pi' \le \Pi$ then $\alpha(\Pi') \le \Pi$. We then proceed with our analysis on this finite abstraction $\Pi^*$.

\begin{definition}\label{abstractPi}
Let $\Pi$ = ($\mathcal{H}$, $\langle Init$, $Goal \rangle$, $Depth$) be a planning problem. We define a finite set of planning problems $\Pi^*$ as follows:
\begin{equation}
    \Pi^* = \bigcup_{\ell \in Loc} \Pi_{\ell}
\end{equation}
where \emph{Loc} is the finite set of locations of $\mathcal{H}$, $\Pi_{\ell}$ = $\langle$ $\mathcal{H}$, $\langle Init$, $Goal_{\ell} \rangle$, $Depth$ $\rangle$ and $Goal_{\ell} = \langle \ell, Inv(\ell) \rangle$.
\end{definition}

\noindent The cardinality of $\Pi^*$ will be the cardinality of Loc of $\mathcal{H}$. For example, the cardinality of $\Pi^*$ for the problem $\Pi$ of Figure \ref{fig:subgoal-example} is nine, where each problem will have the tuple consisting of a location of the automaton and the corresponding invariant as its \emph{Goal}. The definition of $\alpha:\Pi^{\sqcup} \to \Pi^*$ for a given $\Pi'$ = ($\mathcal{H}$, $\langle Init$, $Goal\rangle$, $Depth$), where $Goal = \langle \ell_{goal}, S_{goal} \rangle$ is given as:   
\begin{equation}
\alpha(\Pi')=\Pi_{\ell_{goal}}
\end{equation}

\begin{proposition}
If $\Pi' \in \Pi^{\sqcup}$ then $\alpha(\Pi') \in \Pi^{\sqcup}$.
\end{proposition}
\begin{proof}
The goal states of $\Pi'$ constitute a subset of the goal states of $\alpha(\Pi')$ and therefore, if all valid runs of $\Pi$ intersect the goal states of $\Pi'$, then they also intersect the goal states of $\alpha(\Pi')$. Hence, $\alpha(\Pi') \le \Pi$ and thus $\alpha(\Pi') \in \Pi^{\sqcup}$. 
\end{proof}

\begin{proposition}
The ordered pair $\langle \Pi^*, \le \rangle$ is a partially ordered set (poset).
\end{proposition}

\begin{proof}
For $\langle \Pi^*, \le \rangle$ to be a poset, the binary relation $\le$ on $\Pi^*$ should be reflexive, anti-symmetric, and transitive, that is, $\le$ must be a partial order relation. From definition \ref{def:subproblem}, it is easy to see that every planning problem in $\Pi^*$ is a sub-problem to itself and hence reflexive. Transitivity and anti-symmetry also follow from the definition of sub-problem.
\end{proof}
\noindent \textbf{Chains} Recall that a subset $Y^< \subseteq \Pi^*$ of a partially ordered set $\langle \Pi^*, \le \rangle$ is a \emph{chain} if $\forall \Pi_i, \Pi_j \in Y^< : (\Pi_i \le \Pi_j) \lor (\Pi_j \le \Pi_i)$. The length of a chain is the number of elements in $Y^<$. A poset can have more than one chain of longest length.

\noindent We now describe the explanation artifact that we intend to generate with our explanation algorithm:

\begin{definition} \label{def:explanation}
    Given an unsolvable planning problem $\Pi$,
    an explanation artifact $Explanation(\Pi)$ is a planning problem $\Pi_i \in \Pi^*$ such that:
    \begin{enumerate}
        \item[(i)] $\Pi_i \in Y^{<}$ for some $Y^{<}$ such that $\Pi_i$ is unsolvable.

        \item[(ii)] $\forall$ $\Pi_j \in Y^{<}$ such that $\Pi_j \le \Pi_i$ and $\Pi_j \neq \Pi_i$, $\Pi_j$ is solvable.
    \end{enumerate} 
\end{definition}

\noindent The explanation is thus the first unsolvable sub-problem of $\Pi$ in a chain of sub-problems in the poset $\langle \Pi^*, \le \rangle$. As an illustration, assume that $Y^{<} = \{\Pi_1$, $\Pi_2$, $ \ldots$, $\Pi_n\}$ is a chain of sub-problems of $\Pi$ in $\langle \Pi^*,\le \rangle$ having a total order as $\Pi_1 \le \Pi_2 \le \ldots \le \Pi_n$, Explanation($\Pi$) is the $\Pi_i \in Y^<$ that is unsolvable where $\forall \Pi_j \in Y^<$ such that $\Pi_j \le \Pi_i$, $\Pi_j$ is solvable. For instance, consider the planning problem of the motivating example. Our abstraction will render 25 planning problems in $\Pi^*$, each having one of the cells as the goal. An example of a chain of sub-problems is the chain consisting of 8 sub-problems shown as waypoints $w1-w8$ with a total order $w1 \le w2 \le \ldots \le w8$ amongst them. Observe that this chain is also the longest possible chain of sub-problems in the poset. In this chain, the explanation generated will be the earliest $w_i$ that is unsolvable.

\noindent The goal of generating the proposed explanation artifact is to assist a human expert/control engineer in diagnosing the causes of unsolvability by \textbf{localizing the earliest cause of unsolvability}. The detection of the earliest waypoint, which is infeasible, localizes the primitive cause of unsolvability in that sense. The intuition behind finding a chain of sub-problems is to have a causal analysis of the unsolvability of the planning problem. 

We now present the algorithm to find Explanation($\Pi$) in the following section.
Our initial step for an unsolvable planning problem involves examining all graphically connected paths of a bounded depth, extending from the initial to the goal location within the domain's graph structure. Subsequently, we identify a common sequence of locations present on each of these paths. This location sequence is pivotal to constructing a chain of sub-problems for the unsolvable planning problem.
We first show a reduction from finding a chain of sub-problems to finding a longest-common-subsequence of finitely many strings.

\subsection{Reduction to Longest Common Subsequence (LCS) Problem}
The computation of Explanation($\Pi$) first requires finding a chain of sub-problems of $\Pi$ in $ \langle \Pi^*, \le \rangle$. We now show a reduction of this problem to the problem of finding an LCS of finitely many strings. Recall that \emph{reduction} is a way of converting one problem into another problem such that the solution of the second problem can be used to solve the first problem. 
To present the reduction to LCS, we use the graph of a hybrid automaton.

\noindent We represent a path in a graph as a string $ps$ of the location sequence while eliminating the edges. For example, a path $l_0 \xrightarrow{e_0} l_1 \xrightarrow{e_1} l_2 \xrightarrow{e_2} \ldots \xrightarrow{e_{n-1}} l_{n}$ is represented as a string "$l_0 l_1 l_2 \ldots l_n$". Now, for the given unsolvable problem $\Pi$ =  ($\mathcal{H}$, $\langle Init, Goal \rangle, Depth)$, we can compute all paths of length less than or equal to $Depth$ between $l_{0}$ and $l_{goal}$, the initial and the goal location in $Init$ and $Goal$ respectively. When $\Pi$ includes explicit sub-tasks of visiting a certain set of locations, we compute only those paths between $l_0$ to $l_{goal}$ that visit the given set of locations. Since we are interested in paths of bounded length, there will be finitely many such paths. The string representations of all such paths are denoted by the set $PS(\Pi)$.

The graph of the hybrid automaton provides a higher abstraction of the domain in the sense that if there is no path from $l_0$ to $l_{goal}$ in $\mathcal{G_{H}}$, then there cannot be any valid run of a plan from $Init$ to $Goal$ and hence the planning problem is unsolvable. We may then identify the cause of unsolvability to be in the discrete dynamics, oblivious to the continuous dynamics of the domain. More importantly, as we shall see now, a chain of sub-problems can be identified from the longest common subsequence of the strings in $PS(\Pi)$. Finding an LCS between strings is a classic computer science problem. An LCS measures the closeness of two or more strings by finding the maximum number of identical symbols in them in the same order (\cite{DBLP:conf/spire/BergrothHR00,DBLP:journals/jacm/Maier78,princeton1974finding}). Recall that a subsequence is different from a substring, which additionally requires that the common symbols present in the strings are without gaps.  We now present the main result of the work.

\begin{proposition}
Given a planning problem $\Pi$, computing a chain of sub-problems $Y^<$ in the poset $\langle \Pi^*, \le \rangle$ can be reduced to computing a longest common subsequence of $PS(\Pi)$.     
\end{proposition}

\begin{figure*}[htbp]
    \centering
    \includegraphics[width= 0.6\textwidth]{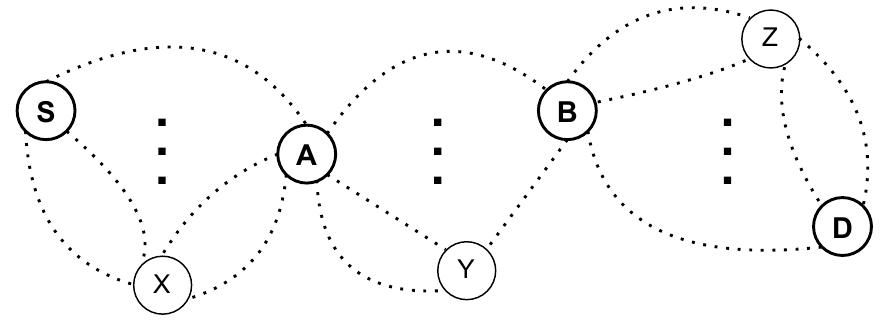}
    \caption{A depiction of a graph of a $\mathcal{H}$. Let S and D be the source and the goal locations in a planning problem $\Pi$. Every path from S to D visits locations A and B in sequence. Thus, S-A-B-D is the LCS of paths in PS($\Pi$). Clearly, if a valid run of a plan exists in the $\mathcal{H}$, the run must visit the invariants of S, A, B, and D sequentially. 
    }
    \label{fig:chain-example}
\end{figure*}

\begin{proof}
$\mathcal{G_{H}}$ = ($V$, $E$) is implicitly present in $\mathcal{H}$ of $\Pi$. Any graph search algorithm, such as breadth-first search, can compute paths in $PS(\Pi)$. Let "$l_0 l_i l_j \ldots l_n$" be an LCS of path strings in $PS(\Pi)$. Being a common subsequence, every path from $l_0$ to $l_n$ visits these nodes in sequence. This implies that every valid run of $\Pi$ must intersect the invariant of these locations in sequence, that is: $\Pi_{l_0} \le \Pi_{l_i} \le \Pi_{l_j} \le \ldots \le \Pi_{l_n}$. $\therefore Y^< = \{ \Pi_{l_0}, \Pi_{l_i}, \Pi_{l_j}, \ldots , \Pi_{n} \}$ is a chain in $\langle \Pi^*, \le \rangle$. Figure \ref{fig:chain-example} shows a sketch of the proof idea.       
\end{proof}

\begin{proposition}
 Given a planning problem $\Pi$, the length of LCS of $PS(\Pi)$ is bounded by the length of the shortest path in $PS(\Pi)$.
\label{prop:lcs-length}
\end{proposition}

\begin{proof}
 Let the length of the LCS be $l$, and the length of the shortest path $p$ in $PS(\Pi)$ be $|p|$, the number of locations in $p$. As the locations in LCS are common to all paths in $PS(\Pi)$. Therefore, the length of LCS cannot be greater than $p$, i.e., $l$ $\not>$ $|p|$.
\end{proof}

\noindent \textbf{Discussion:} If $PS(\Pi)$ is empty, our explanation algorithm terminates and reports that the planning problem is unsolvable due to the discrete dynamics, since there is no path between the initial and the goal locations. Observe that there is a connection between \emph{articulation points} or \emph{cut-vertices} of $\mathcal{G_{H}}$ and the locations in an LCS of $PS(\Pi)$. An articulation point is a vertex of a graph removal of which, along with its incident edges, results in an increase in the connected components in the graph. One can argue that every articulation point of $\mathcal{G_{H}}$ whose removal results in distinct components such that one contains $l_0$ and the other contains $l_{goal}$, will be a member of the LCS. This is because every path from $l_0$ to $l_{goal}$ must contain such articulation points and therefore will be captured in the LCS. Let us call such articulation points as \emph{disconnecting articulation points} in the sense that their removal disconnects $l_0$ and $l_{goal}$. Note that every vertex in an LCS need not be such an articulation point of $\mathcal{G_{H}}$. This is because an LCS is computed over paths in $PS(\Pi)$, which contains only paths of length bounded by a depth specified in the problem instance. There may be paths in $\mathcal{G_{H}}$ of longer length which does not pass through one or more vertices in the LCS. Such vertices in the LCS cannot be disconnecting articulation points whose removal disconnects $l_0$ and $l_{goal}$. Consequently, if our algorithm finds a trivial LCS string of length two, which is $l_0-l_{goal}$, then $\mathcal{G_{H}}$ has no disconnecting articulation point. Such an LCS is trivial because any valid plan of course must meet the invariant of $l_0$ followed by the invariant of $l_{goal}$. Therefore, Y$^<$ = $\{\Pi_{l_0}, \Pi_{l_{goal}} \}$ is always a chain of inevitable sub-problems for any planning problem $\Pi$. Although finding a trivial LCS does not mean there are no articulation points in $\mathcal{G_{H}}$. All articulation points in $\mathcal{G_{H}}$ can be computed in polynomial time, but computing Explanation($\Pi$) additionally requires finding \emph{disconnecting articulation points} and an ordering on them based on the sub-problem relation. Thus, the polynomial-time algorithm does not suffice. 

\noindent LCS of strings can be computed using the standard dynamic programming paradigm (\cite{DBLP:conf/spire/BergrothHR00}). For instance, in the motivating example of planetary rover domain, the longest common sequence for the strings of paths of length bounded by 15 from $l_{11}$ to $l_{25}$ is $l_{11}$-$l_6$-$l_1$-$l_2$-$l_3$-$l_8$-$l_{13}$-$l_{14}$-$l_{24}$-$l_{25}$. In the next section, we show the computation of Explanation($\Pi$).

\subsection{Explanation Generation by Reachability Analysis}
The computed LCS is converted to sub-problems from the locations in the LCS. For each location $l$ in the LCS, we construct a sub-problem $\Pi_{l}$ (recall definition \ref{abstractPi}). Thus, we have the chain $Y^< = \{ \Pi_{l_0}, \Pi_{l_i}, \Pi_{l_j}, \ldots , \Pi_{l_n} \}$ for an LCS say "$l_0 l_i l_j \ldots l_n$". We verify the solvability of the sub-problems in $Y^<$ (inevitable waypoints) using bounded reachability analysis on the hybrid automaton domain by reintroducing the continuous dynamics (by reintroducing the location invariants, flow, transition guards, and resets). Given a hybrid automaton, a set of initial and goal states, and a bound of analysis, say $d$, bounded reachability analysis is the method of computationally deciding whether any goal state is reachable from any initial state by a run of the automaton of length bounded by $d$. Therefore, bounded reachability of the goal states from initial states of a planning problem bounded by $Depth$ implies the existence of a run of a valid plan, which in turn implies the solvability of the given planning problem. In contrast, unreachability of the goal states implies the non-existence of any valid run and hence absence of a plan. As our planning problem under analysis is unsolvable, one or more of the waypoints in $Y^<$ must be unreachable. The reachability analysis of the waypoints is performed in the order in which they appear in the chain $Y^<$. If the sub-problem $\Pi_i$ is found to be reachable, we proceed to check the reachability of the next sub-problem $\Pi_{i+1}$ in the chain. The first $\Pi_i$ which is unreachable is returned as $Explanation(\Pi)$, implying that it is the first unreachable sub-problem/waypoint in a chain of sub-problems/waypoints.

We use a bounded reachability analysis tool \textsc{Bach} (\cite{DBLP:conf/fmcad/BuLWL08}) for reachability analysis of the waypoints. \textsc{Bach} can analyse linear hybrid automata (LHA) (\cite{DBLP:conf/lics/Henzinger96,DBLP:journals/entcs/LiAB07}) and reports a reachability problem instance as \emph{satisfiable} when a run exists from an initial state to a goal state in the corresponding hybrid automaton for the planning problem of length bounded by a given depth, deciding the solvability of the corresponding planning problem. Otherwise, \textsc{Bach} reports the instance as \emph{unsatisfiable}, when no such run exists. Algorithm~\ref{algo:reach} takes a chain of sub-problems $Y^<$ as input. It returns the first unreachable sub-problem in the chain as an explanation of unsolvability. 

\begin{algorithm}
\SetKwInOut{Input}{input}\SetKwInOut{Output}{output}
\Input{A chain of sub-problems $Y^<$ = $\langle \Pi_{l_0}, \Pi_{l_i}, \Pi_{l_j}, \ldots , \Pi_{l_n} \rangle$}
\Output{First unreachable sub-problem $\Pi_{l_k}$ in the chain.}
\DontPrintSemicolon
\caption{\textsf{Generating Explanation($\Pi$) using Reachability Analysis}}
\label{algo:reach}
\BlankLine
\LinesNumbered
  {$n \gets$ $length(Y^<)$ \small{\tcc*{No. of sub-problems in $Y^<$}} }
  \For{$k \gets 0$ to $n$} {
     {$\mathcal{R} \gets$ \textsc{reachabilityProblem}($\mathcal{H}$, $Init$, $Goal_{l_k}$, $Depth$), where $Goal_{l_k}$ is the goal of $\Pi_{l_k}$ \small{\tcc*{Reduced to reachability problem}} }
     {$result \gets$ \textsc{BACH} ($\mathcal{R}$) \small{\tcc*{Use BACH}} }
     \eIf{$result$ == $\textit{SAT}$}
        {\textbf{continue} \small{\tcc*{When sub-problem $\Pi_{l_k}$ is reachable.}}}
        {\KwRet{$\Pi_{l_k}$} \small{\tcc*{Returns first unreachable sub-problem as the explanation.}}}
  }
\end{algorithm}

\subsection{Complexity Analysis}

In a $\mathcal{H}$ with $n$ locations and given a source and destination location, the worst-case complexity of computing all paths in $\mathcal{G}_{H}$ of length at most $d$ from the given source to the destination is $O\big(n^d\big)$. 
The complexity of computing $LCS$ of strings corresponding to the paths (say $m$ many) of maximum length $d$ using standard dynamic programming is $O\big(d^m\big)$. It is known to be an NP-hard problem (\cite{DBLP:journals/asc/DjukanovicRB20,DBLP:journals/jacm/Maier78}). Although finding the inevitable waypoints with the proposed algorithm turns out to be inefficient asymptotically, for problem instances of small size ($\mathcal{H}$ with a few locations and for a small depth d), we show empirically that our algorithm can generate inevitable waypoints and the explanation artifact efficiently.

\paragraph{Complexity of Reachability Analysis :} The model checker \textsc{Bach} performs a path-oriented reachability analysis of a linear hybrid automaton. A path from the initial to the goal location is encoded into a set of linear constraints and consequently solved using a linear programming (LP) problem solver. Details of the path encoding can be found in (\cite{DBLP:conf/fmcad/BuLWL08,DBLP:conf/memocode/SarwarRB23}). Though practical LP solvers use variants of the Simplex algorithm, which runs efficiently on most practical problems, it is not a polynomial-time algorithm in general. In theory, linear programming has been shown to be in class P (\cite{khachiyan1979polynomial,KHACHIYAN198053}). Our algorithm calls \textsc{Bach} for each sub-problem in the computed chain. As the length of the computed chain, the LCS, is bounded by the length of the shortest path in $PS(\Pi)$ (proposition \ref{prop:lcs-length}), which is again bounded by $d$, the complexity is upper bounded by $d$ times the complexity of LP solving. 

\section{Results and Implementation} \label{ch5eval}
In this section, we present the performance of our framework on several unsolvable hybrid planning problem instances.

\subsection{Experimental Setup}

\paragraph{Benchmarks}: A brief description of the planning domains is given as follows: 
\textbf{Planetary rover domain} is presented in Section~\ref{ch5motive} and a pictorial overview is shown in Figure~\ref{fig:rover-domain}.
The planning task for the rover is to reach the base station from its initial location after collecting soil and rock samples from the designated sites. \textbf{City route-network domain} presents a route-network of a city where important places are given as junctions in the network. The planning problem for a battery-powered car is to navigate through the city's route network to reach its destination. \textbf{Warehouse automation domain}~(\cite{DBLP:conf/memocode/SarwarRB23}) represents a scenario where a robot operates to manage the inventories of a warehouse. The floor map of the warehouse is given as grid cells. Few cells in the warehouse are blocked, whereas on a few cells, the robot depletes more energy due to the condition of the surface, such as an oil spillage or being bumpy. The planning problem is for the robot to carry a consignment from its initial location to a goal location. We have crafted warehouse scenarios of varying grid dimensions for evaluating our algorithm. \textbf{Water-level monitor}~(\cite{DBLP:conf/fmcad/BuLWL08}) represents a system that controls the water level in a reservoir. The system goes into an unsafe state if the water level in the reservoir meets underflow or overflow conditions. The planning task is to drive the system to an unsafe state from a given initial state. \textbf{NAV}~(\cite{ARCH-COMP24:ARCH_COMP24_Category_Report}) models the motion of a point robot in a $2$-dimensional plane, partitioned into $3^2$ rectangular regions, and each such region is associated with a vector field described by the flow equations. The planning problem is to find a trajectory from an initial state to a goal state. \textbf{NRS}~(\cite{F-Wang-NRS, ARCH-COMP24:ARCH_COMP24_Category_Report}) represents a nuclear reactor system consisting of 2 rods that absorb neutrons from heavy water when inserted, and a controller that schedules the insertion of the rods into the heavy water.
The system is considered safe if there is exactly one rod absorbing neutrons in the heavy water at any instant of time. The planning problem is to find an unsafe execution of the system from a given initial state. A detailed description of the city route and warehouse automation domains is given in the Appendix. 

We constructed the \emph{Planetary rover} and the \emph{City route-network} domains for evaluating our algorithm. The \emph{Warehouse automation} domain is taken from (\cite{DBLP:conf/memocode/SarwarRB23}).
The rest of the domains are from verification problem instances in linear hybrid systems. For instance, the \emph{Water-level monitor} domain is a benchmark taken from (\cite{DBLP:conf/fmcad/BuLWL08}), whereas \emph{NAV} and \emph{NRS} are benchmarks taken from \textsc{Arch-comp 24 pcdb} category ~(\cite{ARCH-COMP24:ARCH_COMP24_Category_Report}).
In these domains, we pose the safety property verification problem as planning problem instances. The planning problems taken for evaluation are all known to be unsolvable.


\paragraph{Implementation}: All experiments are performed on a machine with 8 GB RAM, Intel Core i5-8250U@1.60GHz, and 8-core processor with Ubuntu 18.04 64-bit OS. All benchmark domains, the problem files, and the code base can be found at: \url{https://gitlab.com/Sazwar/Sub-goal-Construction}.

\subsection{Evaluation}

\begin{table*}
  \centering\tiny
  \resizebox{\textwidth}{!}{
    \begin{tabular}{|c|c|c|c|c|c|c|c|c|c|c|}
    \hline
         \multicolumn{2}{|c|}{\multirow{2}{*}{Benchmarks}} & \multirow{2}{*}{\#Locs} & \multirow{2}{*}{\#Trans} & \multirow{2}{*}{Depth} & \multirow{2}{*}{$\mid$PS($\Pi$)$\mid$}  & \multirow{2}{*}{$\lvert\text{Y}^<\rvert$}   & \#Feas. & \multirow{2}{*}{Exp($\Pi$)} & Time & Memory \\
         \multicolumn{2}{|c|}{} & &  & & & & wps & & (in sec) & (in MB) \\

    \hline
           \multicolumn{2}{|c|}{Planetary} & \multirow{2}{*}{25} & \multirow{2}{*}{40} & 15 & 244 & \multirow{2}{*}{10} & \multirow{2}{*}{6} & \multirow{2}{*}{Loc13} & 0.47 & 11.5 \\\cline{5-6}\cline{10-11}
           \multicolumn{2}{|c|}{rover (PR)} & & & 20 & 20477 & & & & 3.60 & 447.3 \\
    \hline 
           \multicolumn{2}{|c|}{City} & \multirow{2}{*}{10} & \multirow{2}{*}{25} & 10 & 468 & \multirow{2}{*}{4} & \multirow{2}{*}{2} & \multirow{2}{*}{Loc7} & 0.73 & 10.1 \\\cline{5-6}\cline{10-11}
           \multicolumn{2}{|c|}{route (CR)} & & & 15 & 92172 & & & & 5.93 & 1075.8 \\
         
    \hline 
           & \multirow{2}{*}{6x4} & \multirow{2}{*}{24} & \multirow{2}{*}{50} & 10 & 36 & \multirow{2}{*}{6} & \multirow{2}{*}{4} & \multirow{2}{*}{Loc17} & 1.57 & 8.8 \\\cline{5-6}\cline{10-11}
           & & & & 15 & 40998 & & & & 4.17 & 507.5 \\\cline{2-11}
           
           \multirow{4}{*}{Warehouse} & \multirow{2}{*}{6x6} & \multirow{2}{*}{36} & \multirow{2}{*}{78} & 12 & 12 & 8 & \multirow{2}{*}{4} & \multirow{2}{*}{Loc28} & 0.54 & 9.2 \\\cline{5-7}\cline{10-11}
           & & & & 17 & 5816 & 6 &  & & 4.47 & 884.9 \\\cline{2-11}
           
           \multirow{2}{*}{automation (WA)} & \multirow{2}{*}{8x8} & \multirow{2}{*}{64} & \multirow{2}{*}{100} & 12 & 16 & 9 & 7 & \multirow{2}{*}{Loc41} & 1.37 & 17.5 \\\cline{5-8}\cline{10-11}
           & & & & 17 & 10214 & 3 & 1 & & 8.62 & 1454.2 \\\cline{2-11}

           & \multirow{2}{*}{10x10} & \multirow{2}{*}{100} & \multirow{2}{*}{178} & 12 & 2 & 11 & 7 & \multirow{2}{*}{Loc57} & 4.13 & 133.9 \\\cline{5-8}\cline{10-11}
           & & & & 15 & 78 & 8 & 4 & & 25.14 & 1445.3 \\
    \hline
           \multicolumn{2}{|c|}{Water-level} & \multirow{2}{*}{6} & \multirow{2}{*}{6} & 20 & 5 & \multirow{2}{*}{3} & \multirow{2}{*}{2} & \multirow{2}{*}{Loc6} & 0.05 & 5.7 \\\cline{5-6}\cline{10-11}
           \multicolumn{2}{|c|}{monitor (WLM)} & & & 50 & 12 & & & & 0.05 & 5.7 \\
    \hline
          \multicolumn{2}{|c|}{\multirow{2}{*}{NAV}} & \multirow{2}{*}{9} & \multirow{2}{*}{24} & 10 & 2325 & \multirow{2}{*}{2} & \multirow{2}{*}{1} & \multirow{2}{*}{Loc6} & 0.39 & 9.5 \\\cline{5-6}\cline{10-11}
          \multicolumn{2}{|c|}{} & & & 15 & 149733 & & & & 3.41 & 773.2 \\
    \hline   
          \multicolumn{2}{|c|}{\multirow{2}{*}{NRS}} & \multirow{2}{*}{27} & \multirow{2}{*}{30} & 15 & 312 & \multirow{2}{*}{2} & \multirow{2}{*}{1} & \multirow{2}{*}{Loc25} & 0.03 & 7.6 \\\cline{5-6}\cline{10-11}
          \multicolumn{2}{|c|}{} & & & 20 & 7812 &  &  & & 0.14 & 22.1 \\
    \hline
    
    \end{tabular}
  }
  \caption{Explanation generation on unsolvable planning problem instances.}
  \label{table:table1}
\end{table*}

Table~\ref{table:table1} shows results of our framework on several hybrid systems benchmark domains. Each of the domains is presented with an unsolvable planning problem instance with varying bounds on the plan depth to test the scalability of the framework. \emph{Benchmark} represents the planning domains together with the planning problem, while \emph{\#Loc} and \emph{\#Trans} report the number of locations and edges in the hybrid automaton of the domain, respectively, showing the size of the domain.
\emph{Depth} presents the bound on the plan length. We have presented results for two plan depths on each domain.
\emph{$\mid$PS($\Pi$)$\mid$} specifies the number of path strings corresponding to the paths from the initial to the goal location of the planning problem instance in the graph structure of the $\mathcal{H}$ domain. Recall that we look at all these paths while computing the longest common location sub-sequence, which gives us the inevitable waypoints in the chain of sub-problems in Y$^<$.
\emph{$\mid$Y$^<$$\mid$} gives us the chain length, which emphasizes the number of inevitable sub-problems detected by our framework, and \emph{\#Feas. wps} denotes the number of solvable sub-problems/waypoints in the chain.
\emph{Exp($\Pi$)} presents the first unsolvable sub-problem in Y$^<$, and thereby, the first infeasible waypoint for the planning problem. A location $loc$ in the Exp($\Pi$) column represents the first sub-problem $\Pi_{loc}$ in the chain Y$^<$ that is unsolvable. For example, \emph{Loc13} corresponding to the entry of \emph{Planetary rover domain} reports the sub-problem $\Pi_{loc13}$ as the explanation of unsolvability of the problem instance. \emph{Time} and \emph{Memory} report the corresponding execution time and memory usage incurred by our framework.

\begin{table}
  \centering
    \begin{tabular}{|c|c|c|c|c|c|c|}
    \hline
        \multicolumn{2}{|c|}{\multirow{2}{*}{Benchmarks}} & \multirow{2}{*}{Depth} & \multicolumn{3}{c|}{Time (in secs)}  & \multirow{2}{*}{AT}\\\cline{4-6}
        \multicolumn{2}{|c|}{} & & (a) PE & (b) Finding Y$^<$ & (c) RA & \\

    \hline
           \multicolumn{2}{|c|}{\multirow{2}{*}{PR}} & 15 & 0.03 & 0.01 & 0.43 & 0.47 \\\cline{3-7}
           \multicolumn{2}{|c|}{} & 20 & 1.99 & 1.14 & 0.47 & 3.60\\
    \hline 
           \multicolumn{2}{|c|}{\multirow{2}{*}{CR}} & 10 & 0.03 & 0.01 & 0.69 & 0.73 \\\cline{3-7}
           \multicolumn{2}{|c|}{} & 15 & 4.72 & 0.23 & 0.98 & 5.93 \\
    \hline 
           \multirow{8}{*}{WA} & \multirow{2}{*}{6x4} & 10 & 0.01 & 0.01 & 1.54 & 1.57 \\\cline{3-7}
           & & 15 & 2.29 & 0.16 & 1.72 & 4.17 \\\cline{2-7}
           
           & \multirow{2}{*}{6x6} & 12 & 0.01 & 0.01 & 0.51 & 0.54 \\\cline{3-7}
           & & 17 & 3.67 & 0.04 & 0.76 & 4.47 \\\cline{2-7}
           
           & \multirow{2}{*}{8x8} & 12 & 0.05 & 0.01 & 1.31 & 1.37 \\\cline{3-7}
           & & 17 & 7.53 & 0.03 & 1.06 & 8.62 \\\cline{2-7}
           
           & \multirow{2}{*}{10x10} & 12 & 0.52 & 0.02 & 3.59 & 4.13 \\\cline{3-7}
           & & 15 & 21.68 & 0.02 & 3.44 & 25.14 \\
    \hline
           \multicolumn{2}{|c|}{\multirow{2}{*}{WLM}} & 20 & 0.01 & 0.01 & 0.03 & 0.05 \\\cline{3-7}
           \multicolumn{2}{|c|}{} & 50 & 0.01 & 0.01 & 0.03 & 0.05 \\
    \hline
           \multicolumn{2}{|c|}{\multirow{2}{*}{NAV}} & 10 & 0.01 & 0.01 & 0.37 & 0.39 \\\cline{3-7}
            \multicolumn{2}{|c|}{} & 15 & 2.78 & 0.21 & 0.42 & 3.41 \\
    \hline   
           \multicolumn{2}{|c|}{\multirow{2}{*}{NRS}} & 15 & 0.01 & 0.01 & 0.01 & 0.03 \\\cline{3-7}
            \multicolumn{2}{|c|}{} & 20 & 0.12 & 0.01 & 0.01 & 0.14 \\
    \hline
    
    \end{tabular}
  \caption{Performance analysis of our framework. \emph{PE} indicates Path-exploration time, \emph{RA} indicates Reachability analysis time, and \emph{AT} indicates Accumulative Time = (a) + (b) + (c).}
  \label{table:table2}
\end{table}
Table~\ref{table:table2} presents a detailed diagnosis of the execution time taken for explanation generation, showing the time taken for computing all initial to goal paths (PS), Computing a chain of inevitable sub-problems (Y$^<$), and reachability analysis to find the first unsolvable planning problem in the chain (Explanation($\Pi$)).

\subsection{Analysis of results}
Table~\ref{table:table1} shows that our framework identifies a chain of inevitable waypoints and an explanation of unsolvability efficiently. Performance degrades with an increase in the depth bound of the planning problem instance. This is clearly because increasing depth results in an exponential increase in the number of paths from the initial to the goal location, which also increases the time to compute LCS of path strings. In NAV and NRS, our algorithm reports the trivial chain of waypoints which is visiting the initial location followed by visiting the goal location as inevitable. Note that this is because the graph of these domains do not have any disconnecting articulation point (refer to the discussion section). 
Memory usage exceeds 500 MB in a few instances. 
This is because of a \emph{bfs} (breadth-first search) based path exploration where the size of the bfs queue increases exponentially at each level due to branching factor. Table~\ref{table:table2} shows the performance of the three major components of the algorithm. 
The path-exploration time and reachability analysis by the bounded model checker dominates the overall time taken by the algorithm. The results emphasize that it can quickly identify the sub-problems for a planning problem.
\begin{figure*}[htbp]
    \centering
    \begin{subfigure}[b]{0.48\textwidth}
        \centering
        \includegraphics[scale=0.4]{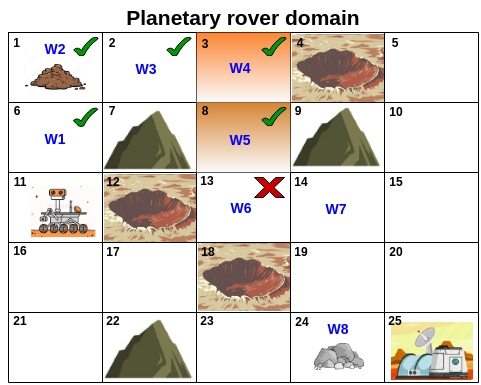}
        \caption{$W6$ is the first unreachable waypoint and serves as an explanation of the planning problem being unsolvable.}
        \label{fig:waypoints}
    \end{subfigure}
    \hfill
    \begin{subfigure}[b]{0.48\textwidth}
          \centering
         \includegraphics[height=165pt,width=0.97\textwidth]{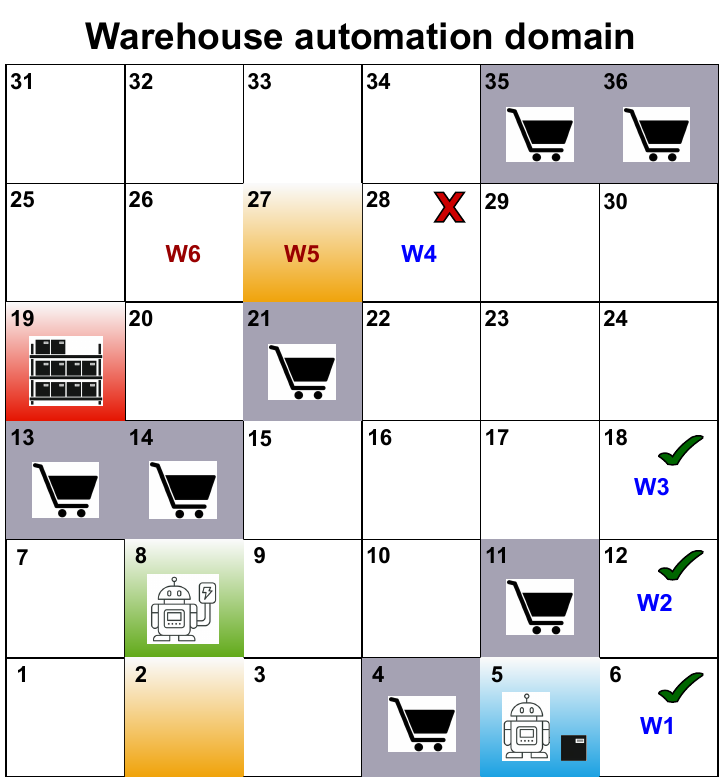}
         \caption{
         $W4$ is the first unreachable waypoint for the planning problem.
         }
         \label{fig:waypoints1}
    \end{subfigure}
    \caption{Illustration of results in the example scenarios.}
    \label{fig:res-way}
\end{figure*}


In Figure~\ref{fig:res-way}, we illustrate our method in the example scenarios of \emph{Planetary rover} and \emph{Warehouse automation} domains.
In the motivating example problem instance in \emph{Planetary rover} domain, the algorithm identified 8 sub-problems $\{W1, W2,\ldots,W8\}$ in Y$^<$ with the total order $W1 \le W2 \le \ldots  \le W8$, each representing an inevitable waypoint. Our explanation algorithm detects five waypoints ($W1$-$W5$) as reachable, depicted by green ticks, and reports $W6$ as the first unreachable waypoint shown by a red cross in Figure~\ref{fig:waypoints}. The unreachability of $W6$ can lead the human expert to deduce that the rover's initial battery charge is insufficient to drive it past the ascending regions in $W4$ and $W5$. Figure~\ref{fig:waypoints1} shows the identified waypoints and an explanation on a 6$\times$6 warehouse domain for a planning problem where a robot needs to carry a consignment from its initial location to the goal location.
In the domain, blue and red cells are the initial and goal locations, respectively. Yellow cells have surfaces with oil-spillage and therefore, the robot has a greater rate of battery depletion in these cells. Grey cells are blocked. The green cell is the only charging station.
The explanation algorithm identified 6 sub-problems $\{W1,W2,\ldots,W6\}$ for a planning problem depth bound of 12, and 4 sub-problems $W1$ to $W4$ for a depth bound of 17, respectively. The waypoints $W5$ and $W6$ are inevitable only when the depth bound is 12, since a path longer than 12 in length may not mandatorily visit these waypoints. Note that these vertices ($W5$ and $W6$) are not articulation points, whereas the other waypoints ($W1-W4$) are articulation points of the warehouse grid graph. In both problem instances, our algorithm reports that the robot cannot reach the waypoint $W4$ under the dynamics, which is an explanation of unsolvability. A control engineer can deduce that the initial battery charge and the charge capacity of the robot are not sufficient to reach the waypoint directly or via the recharging station. Therefore, a higher charge capacity or a better placement of the charging station close to the waypoint $W4$ may be a workaround to make the task solvable.

\noindent \textcolor{black}{\textbf{Discussion:} \emph{Depth} has a significant impact on the efficiency of our approach by restricting the number of paths explored in a planning domain. It also has an effect on identifying the waypoints for a problem. For example, the waypoints $w5$ and $w6$ are not the articulation points of the 6$\times$6 warehouse domain shown in Figure~\ref{fig:waypoints1}. They appear solely because of the given depth bound on the planning problem. 
The results presented in Table~\ref{table:table1} and Table~\ref{table:table2} highlight the effect of depth.}

\subsection{Domain Descriptions} \label{ch5dom-desc}
\begin{figure}[htbp]
    \centering
    \includegraphics[width =0.4\textwidth]{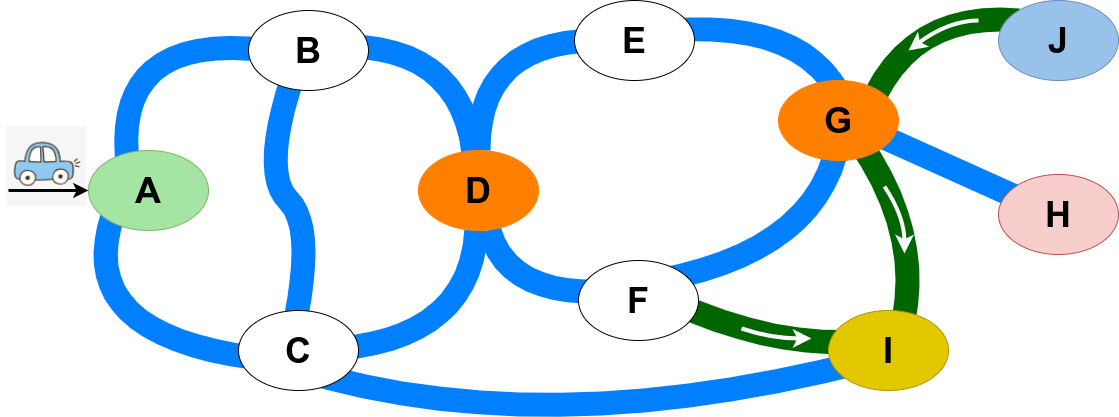}
    \caption{The city route network is depicted. Each node represents the important junctures of the city. The blue-colored routes represent both-way transportation between junctures. The green-colored routes represent one-way traffic. Initially, the car is at juncture A (shown as the green-colored node). The junctures I (yellow-colored), H (red-colored), and J (blue-colored) are the destinations of three different planning problems of the domain.
    The orange-colored nodes in the figure are the waypoints that appear in every source-to-destination path for a planning problem of A to H.}
    \label{fig:city_example}
\end{figure}

\textbf{City-network domain:}
The context for this domain is a car that wants to reach a destination through a route network of a city. Figure~\ref{fig:city_example} shows the route network of the city. It has 10 important junctures. Blue-colored and green-colored routes connect these junctures. They respectively represent both-way and one-way traffic in the city. The direction of the traffic in green routes is shown with a directed arrow. The car is initially at juncture A. The juncture A is shown as the green-colored node in the figure. The car has a battery that depletes energy at a constant rate represented by a variable $b$. Initially, it has 20 units of battery charge available. Similarly, the juncture-to-juncture movement delay for the car is represented by a variable $d$ where each route has a different delay. From a juncture, the car can only move to the adjacent junctures following the route between them. The junctures H (red-colored), I (yellow-colored), and J (blue-colored) are the destinations of three different planning problems of the domain. The orange-colored nodes in the figure are the waypoints that appear in every source-to-destination path for a planning problem of A to H. The routes represent the discrete dynamics of the domain that captures the connectivity of the junctures of the city. The continuous dynamics of the domain involve energy depletion and the juncture-to-juncture movement delay of the car due to different traffic patterns.

\begin{figure}[htbp]
    \centering
    \includegraphics[width =0.4\textwidth]{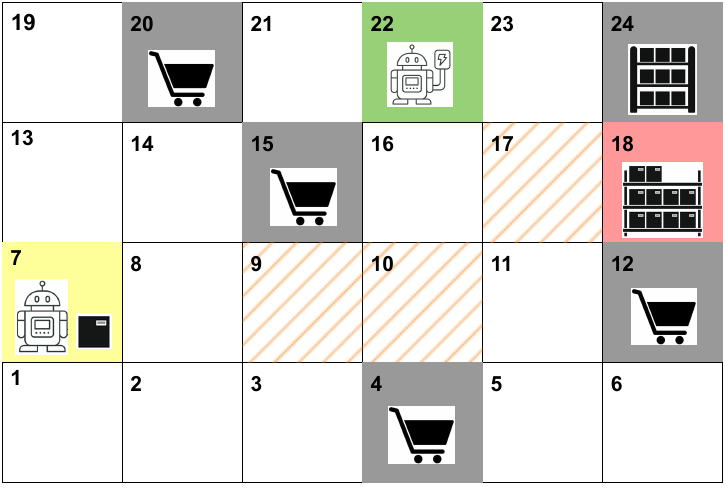}
    \caption{Warehouse automation domain.}
    \label{fig:warehouse-domain}
\end{figure}

\noindent \textbf{Warehouse automation domain:} We present the context of warehouse automation (\cite{DBLP:conf/memocode/SarwarRB23}) where a robot operates to manage the inventories of the warehouse. The warehouse is divided into cells. The discrete dynamics here capture the connectivity of the cells along with the presence of objects in certain cells, which are interpreted as obstacles through which the robot cannot move. The movement of the robot is restricted to one of its adjacent cells, and movement to diagonal cells is prohibited. The continuous dynamics capture the battery charge depletion rate of the robot within a cell. Within each cell, the robot follows the dynamics particular to that cell. When the robot makes a transition from one cell to another, it starts to follow the dynamics of the new cell instantaneously. The robot is assigned the task of carrying a consignment to a designated cell while the number of cell visits is restricted to $\leq$ $D$ cells. The robot starts from the yellow-colored cell where the planning problem requires it to transport the black box to the goal cell (red-colored cell). The robot depletes its charge according to the cell dynamics while on the move. There is a charging station shown as a green-colored cell. The robot may visit this cell to recharge its battery. The grey-colored cells are blocked with obstacles. The robot is equipped with a rechargeable battery. The initial battery charge is 10 units. Each cell has a charge depletion rate of 2 units (modeling the continuous dynamics). 
In Figure~\ref{fig:warehouse-domain}, we have shown a representation of the warehouse automation domain. Now, consider the planning problem where the robot needs to carry the consignment to the goal from its initial location. Every feasible path for the robot must go through the cells marked with hatched lines (orange-colored) as shown in the figure. We consider these cells as landmarks for the planning problem. A landmark here means a cell that a robot must visit on its way to the goal.
\section{Related Works} \label{ch5related_works}
Some notable works addressing the unsolvability of planning problems, mostly looked at verifying the unsolvability by generating certificates~(\cite{DBLP:conf/aips/ErikssonH20}), (\cite{DBLP:conf/aips/ErikssonRH17}) or proofs (\cite{DBLP:conf/aips/ErikssonRH18}) rather than explaining the causalities of unsolvability of the planning problem. Such certificates or proofs of unsolvability are not enough to increase the human understandability of why the problem was unsolvable.
Most of these works focus on planning problems in discrete domains. Verifying the unsolvability of planning problems in hybrid systems comes with an additional challenge since these planning problems are undecidable in general (\cite{ALUR19953}).
In~(\cite{DBLP:journals/tecs/SarwarRB23}), authors provide an approach to addressing the unsolvability of a planning problem in hybrid domains by $\delta$-approximate bounded reachability analysis (\cite{DBLP:journals/corr/GaoKCC14}). However, this work also verifies unsolvability rather than explaining it.
Few notable works that are directed towards explaining the unsolvability of a planning problem are, similarly, limited to classical planning problems.
Authors in~(\cite{DBLP:conf/aips/GobelbeckerKEBN10}) argue that excuses can be produced by counterfactual alterations to the original planning task such that the new planning task turns out to be solvable, and provides excuses for why a plan cannot be found.
In~(\cite{DBLP:conf/aaai/EiflerC0MS20}), authors derive properties of a plan which could serve as explanations in case of unsolvability.
However, generating excuses, or deriving plan properties in terms of propositional formulas may not be enough to understand why a problem was unsolvable for complex domains like planning problems of hybrid systems which encode mixed discrete and continuous dynamics.
In (\cite{DBLP:journals/jair/VasileiouYSKCM22}), an approach based on knowledge representation and reasoning has been applied to these domains. It provides explanations by finding a subset of the agent's knowledge base with which to reconcile the human knowledge base for explanations. However, it does not address unsolvability problems, rather, explains why a plan is feasible in a model.
A path-oriented reconciliation process between the agent and human models of hybrid systems is provided in~(\cite{DBLP:conf/memocode/SarwarRB23}). It performs the reachability analysis along a path and uses the concept of irreducible infeasible sets (IIS) to generate explanations for unsolvability.

In this work, we propose to decompose an unsolvable planning problem into sub-problems motivated by the well-known insight that humans tend to break down sequential planning problems in terms of the sub-problems they need to achieve (\cite{LT-1956,DBLP:journals/ai/VanLehn86}).
This has been a popular approach in many domains such as robotics (\cite{24200}) and AI (\cite{DBLP:journals/ai/SuttonPS99}) apart from planning (\cite{DBLP:journals/jair/HoffmannPS04,DBLP:conf/ecai/LipovetzkyG12,DBLP:conf/aaai/RichterHW08}). 
(\cite{DBLP:journals/jair/HoffmannPS04,DBLP:conf/ecai/LipovetzkyG12}) find sub-problems for a solvable planning problem of the discrete domains in terms of ordered landmarks. Landmarks are facts given as propositional formulas that must be true at some point in every valid solution plan.
In~(\cite{DBLP:conf/ijcai/SreedharanSSK19}), authors use hierarchical model abstractions to relax a planning problem until a solution can be found and looks for landmarks of this relaxed problem. They use these landmarks to identify the unachievable sub-problem for the planning problem. These works are in discrete domains. In contrast, our framework decomposes an unsolvable planning problem of hybrid domains into several smaller sub-problems by reducing it to an instance of longest common subsequence problem and consequently generating explanations using reachability analysis. 
\section{Conclusion} \label{ch5conclusion}
In this work, we explore the area of explaining the unsolvability of planning problems for hybrid systems by means of detecting the inevitable sub-problems that must be solvable in order for the bigger problem to be solvable.
We show a reduction from the problem of finding sub-problems and an ordering between them to finding the LCS of a finite set of path strings.  We present an explanation artifact through these sub-problems and by conducting reachability analysis. Results emphasize that our framework can efficiently identify inevitable sub-problems and the first infeasible one among them as an explanation for unsolvability of a planning problem. We believe that explanations reported by our algorithm can help a control engineer, an AI planner, or a human supervisor to comprehend the cause of unsolvability of the planning problem at hand.


\chapter{Conclusion and Future Work} \label{ch7}

\lettrine[lraise=0.3, nindent=0em, slope=-.5em]XAIP has garnered significant research interest due to its role in designing explainable systems. In this dissertation, we look into XAIP from a hybrid system perspective as these systems closely model real-world scenarios. We explain the behavior of such systems. Problems are discussed from two different directions: one that aims to explain when the planning problem is solvable, and there exists an automatically generated plan; the other case tries to explain when the planning problem is unsolvable, and there exists no such plan. We start our investigation with the motion planning problem for an autonomous robot.  


In chapter~\ref{ch6}, we demonstrate a case study on the motion planning problem. We present an integrated software framework for autonomous navigation of an \emph{lizard-inspired quadruped robot} in an unknown environment. While the framework has three components: \emph{SLAM}, \emph{motion-planning}, and \emph{control}, emphasis remains on motion planning and control.
The motion-planning problem is solved by reducing it to the constraint-satisfaction problem, which is solved with a state-of-the-art constraint solver, Z3. A hybrid controller then executes the solution to move the robot efficiently in the environment. We have experimented with several planning problem instances in various indoor simulation scenarios, and the results met our objectives. The proposed solution has many interesting applications, including surveillance in an unknown environment, information gathering about a hostage scenario, etc.


In chapter~\ref{ch2}, we provide a contrastive explanation framework that aims to generate explanations of a plan for a hybrid system planning problem.
Given a hybrid system model in PDDL+ and a plan describing the set of desirable actions on the same to achieve a desired goal:
\begin{enumerate}
    \item Our framework can integrate users' questions in PDDL+ and synthesize alternate plans using a hypothetical model (HModel) constructed by imposing constraints drawn from the questions.

    \item The framework incorporates a \emph{re-model and re-plan} approach to facilitate explanation to the iterative user's question.
\end{enumerate}
While the primary aim of these explanations is to build trust in AI-based systems, they also help to understand the inner dynamics of the planning domain and the planner.
Additionally, it can identify the modeling flaws to design a better planning model.
We present a detailed case study on our approach, with comparison metrics to compare the original plan with the alternate ones.
Furthermore, we provide a no-plan explanation algorithm for our unsolvable planning problem instances through bounded reachability analysis. We believe our framework can be of immense importance to the hybrid systems planning community for synthesizing better, explainable plans.
At the end of this chapter, we present a web-based contrastive explanation tool that implements the iterative re-modeling and re-planning algorithm, and provides a provision to experiment with different planning domains in hybrid systems and plug in different hybrid system planners as a plan generation engine.





Explaining unsolvability has been an interesting direction.
In chapter~\ref{ch4}, we explore this direction for hybrid system planning problems. We assume that the AI agent and the human have different knowledge bases.
While the agent has a complete model of the environment, the human has a partial or erroneous model and expects a plan for the planning problem when there is none.
We present a path-based \emph{continuous model reconciliation} framework that updates the human with the causes of unsolvability to make the human domain consistent with that of the agent.
\begin{enumerate}
    \item Our approach performs a discrete path analysis to quickly falsify a path by mapping each location and transition of the path in the human model to the agent model.
    \item In continuous path analysis, we leverage reachability analysis by combining it with minimal inconsistent constraint sets to find the infeasible path segments for which the paths become unsatisfiable.
\end{enumerate}
We demonstrate our work on two hybrid system planning domains (i.e., warehouse automation system and water-level monitoring system). Our framework generates explanations quickly, allowing for swift identification of the reasons behind a plan's failure. Furthermore, its path-oriented analysis thoroughly explores the planning space, enabling it to uncover the majority of factors contributing to the problem's unsolvability.


In chapter~\ref{ch5}, we aim to explain unsolvability by identifying the inevitable sub-problems for hybrid system planning problems. The proposed method performs the following:
\begin{enumerate}
    \item A graph traversal on the abstract graph of the domain to enumerate all feasible paths from source to goal.

    \item Identifies inevitable waypoints by casting it to a \emph{LCS} problem.

    \item Decomposes the planning problem into sub-problems based on waypoints.

    \item Finally, generate an explanation artifact through these sub-problems by conducting reachability analysis.
\end{enumerate}
We present our work on several hybrid system benchmarks, and results emphasize that it can efficiently identify inevitable sub-problems and generate explanation artifacts. We believe that explanations reported by our algorithm can help a control engineer, an AI planner, or a human supervisor to comprehend the cause of unsolvability of the planning problem at hand.

\paragraph{Limitation and Future Work.}
Planning for hybrid systems still holds many open challenges.
However, the key issues that are inherently present for such problems are \emph{complexity} and \emph{scalability}.
Our knowledge of the computational complexity of many practical planning fragments is still limited, even in the discrete-time setting.
A significant contributor to the complexity of planning for hybrid systems is the use of unbounded numeric and continuous variables, making it challenging to develop procedures with guaranteed termination over potentially infinite state spaces.
Despite the inherent complexity of unbounded numerics, many real-world scenarios can be effectively modeled using numeric variables with predefined bounds. Although initial studies exploring these bounded cases have begun to appear (\cite{DBLP:conf/aips/0001HJS23,DBLP:conf/ijcai/GiganteS23}), significant research is still required to understand how these theoretical results can be practically applied through new algorithms and heuristics that exploit these assumptions for more efficient reasoning, and expand to encompass temporal planning.

The works discussed so far primarily focus on generating explanations retrospectively, after a plan has been generated (or the search for one has failed). However, explanation can be integrated directly into an agent's decision-making process. Just as humans tend to make better choices when required to justify them, incorporating this philosophy into XAIP could lead to enhanced, more human-aware systems. This could be one interesting direction to look into in the future. For work presented in chapter \ref{ch4} and \ref{ch5}, we plan to include provisions for a more fine-grained analysis of the unsolvability and derive causes from the continuous dynamics that are not considered in the current versions. Furthermore, we aim to explore a more personalized way of generating explanations. In human interactions, explainers naturally adjust the level of detail and choose a conceptual model they believe will align with the listener's understanding. Consequently, explanations can be provided at varying levels of abstraction, relying on different conceptual models. For the work presented in chapter~\ref{ch6}, we plan to extend the software framework to address dynamic environments and motion planning of a swarm of robots.




\clearpage
\phantomsection
\addcontentsline{toc}{chapter}{Bibliography}
\bibliography{mybib}
\clearpage
\chapter{Appendix} \label{sec10}

\section{Car Domain} \label{appendix1}

\begin{lstlisting}[basicstyle=\footnotesize,caption={The planning problem instance 2 in the car domain.},captionpos=b,frame=single,label={Listing7},float=htbp]
(define (problem car_prob)
   (:domain car)
   (:init (running) (= (runningTime) 0) (= (upLimit) 2) (= (downLimit) -2)
         (= d 0) (= a 0) (= v 0))
   (:goal (and (goalReached) (not (engineBlown)) (<= (runningTime) 50)))
\end{lstlisting}

\section{Generator Events Domain} \label{appendix2}

\begin{lstlisting}[basicstyle=\footnotesize,caption={The generator-events domain in PDDL+.},captionpos=b,frame=single,label={lst:Listing8},float=htbp]
(define (domain generatorplus)
(:requirements :fluents :durative-actions :duration-inequalities 
               :adl :typing :time)
(:types generator tank)
(:predicates (generator-ran) (available ?t - tank)
             (using ?t - tank ?g - generator) (safe ?g - generator))
(:functions (fuelLevel ?g - generator) (capacity ?g - generator)
            (fuelInTank ?t - tank) (ptime ?t - tank))
(:durative-action generate
 :parameters (?g - generator)
 :duration (= ?duration 1000)
 :condition (and (over all (>= (fuelLevel ?g) 0)) (over all (safe ?g)))     
 :effect (and (decrease (fuelLevel ?g) (* #t 1)) (at end (generator-ran))))
(:action refuel
 :parameters (?g - generator ?t - tank)
 :precondition (and (not (using ?t ?g)) (available ?t))
 :effect (and (using ?t ?g) (not (available ?t))))
(:process refuelling
 :parameters (?g - generator ?t -tank)
 :precondition (and (using ?t ?g))
 :effect (and (increase (ptime ?t) (* #t 1))
    (decrease (fuelInTank ?t) (* #t (* 0.001 (* (ptime ?t) (ptime ?t)))))
    (increase (fuelLevel ?g) (* #t (* 0.001 (* (ptime ?t) (ptime ?t)))))))     
(:event tankEmpty
 :parameters (?g - generator ?t - tank)
 :precondition (and (using ?t ?g) (<= (fuelInTank ?t) 0))
 :effect (and (not (using ?t ?g))))
(:event generatorOverflow
 :parameters (?g - generator)
 :precondition (and (> (fuelLevel ?g) (capacity ?g)) (safe ?g))
 :effect (and (not (safe ?g)))))
\end{lstlisting}


\begin{lstlisting}[basicstyle=\footnotesize,caption={The planning problem instance 1 in the generator-events domain in PDDL+.},captionpos=b,frame=single,label={lst:Listing9},float=htbp]
(define (problem run-generatorplus)
(:domain generatorplus)
(:objects gen - generator t1 t2 - tank)
(:init (= (fuelLevel gen) 940) (= (capacity gen) 1600) (= (fuelInTank t1) 40)
           (= (fuelInTank t2) 40) (available t1) (available t2) (safe gen))  
(:goal (generator-ran)))
\end{lstlisting}

\begin{lstlisting}[basicstyle=\footnotesize,caption={The planning problem instance 2 in the generator-events domain in PDDL+.},captionpos=b,frame=single,label={lst:Listing10},float=htbp]
(define (problem run-generatorplus)
(:domain generatorplus)
(:objects gen - generator t1 t2 t3 - tank)
(:init (= (fuelLevel gen) 900) (= (capacity gen) 1600) (= (fuelInTank t1) 40)
           (= (fuelInTank t2) 40) (= (fuelInTank t1) 40) (available t1)
           (available t2) (available t3) (safe gen))  
(:goal (generator-ran)))
\end{lstlisting}

\noindent Hybrid Automatom Model of Generator-events Domain: To represent the generator-events domain (see \ref{sec:benchmarks}) for the planning problem instance 1 into a hybrid automaton model, the behaviour of the durative-action \emph{generate} is captured with the start-process-stop paradigm in PDDL+ \cite{PDDL+}. The durative-action \emph{generate} can be viewed as a sequence of four distinct but causally related parts: an instantaneous-action \emph{generateStart}, a process \emph{generateProcess}, another instantaneous-action \emph{generateEnd} and an event \emph{generateFail}. The \emph{generateStart} marks the start of the \emph{generate} action (shown by the name extension \emph{St} in Figure \ref{genModel}), it then activates the \emph{generateProcess} which in turn starts to decrease the \emph{fuelLevel} of the generator at a constant rate. The time elapsed is captured by the variable $d$. The \emph{generateEnd} updates \emph{(generatorRan)} to \emph{true} when $d$ becomes 1000. 
The event that is a part of the start-process-stop model monitors any violation of invariant of the durative-action. In our model, the event \emph{generateFail} is monitoring only the generator underflow condition (where \emph{fuelLevel} becomes zero), since the invariant \emph{(safe gen)} of the \emph{generate} action is already being monitored by the event \emph{generatorOverflow} that observes the generator overflow condition.
The corresponding hybrid automaton model for the planning problem instance is depicted in Figure \ref{genModel}. The locations in the automata are formed with subsets of predicate symbols that hold on that location along with the name extension \emph{St} (indicating the \emph{generate} action has started)
whenever applicable.

\begin{tikzpicture}
\node[rotate=90] at (0,0) {\begin{minipage}{0.8\textheight}
                   \centering
                   \includegraphics[width=1\textwidth]{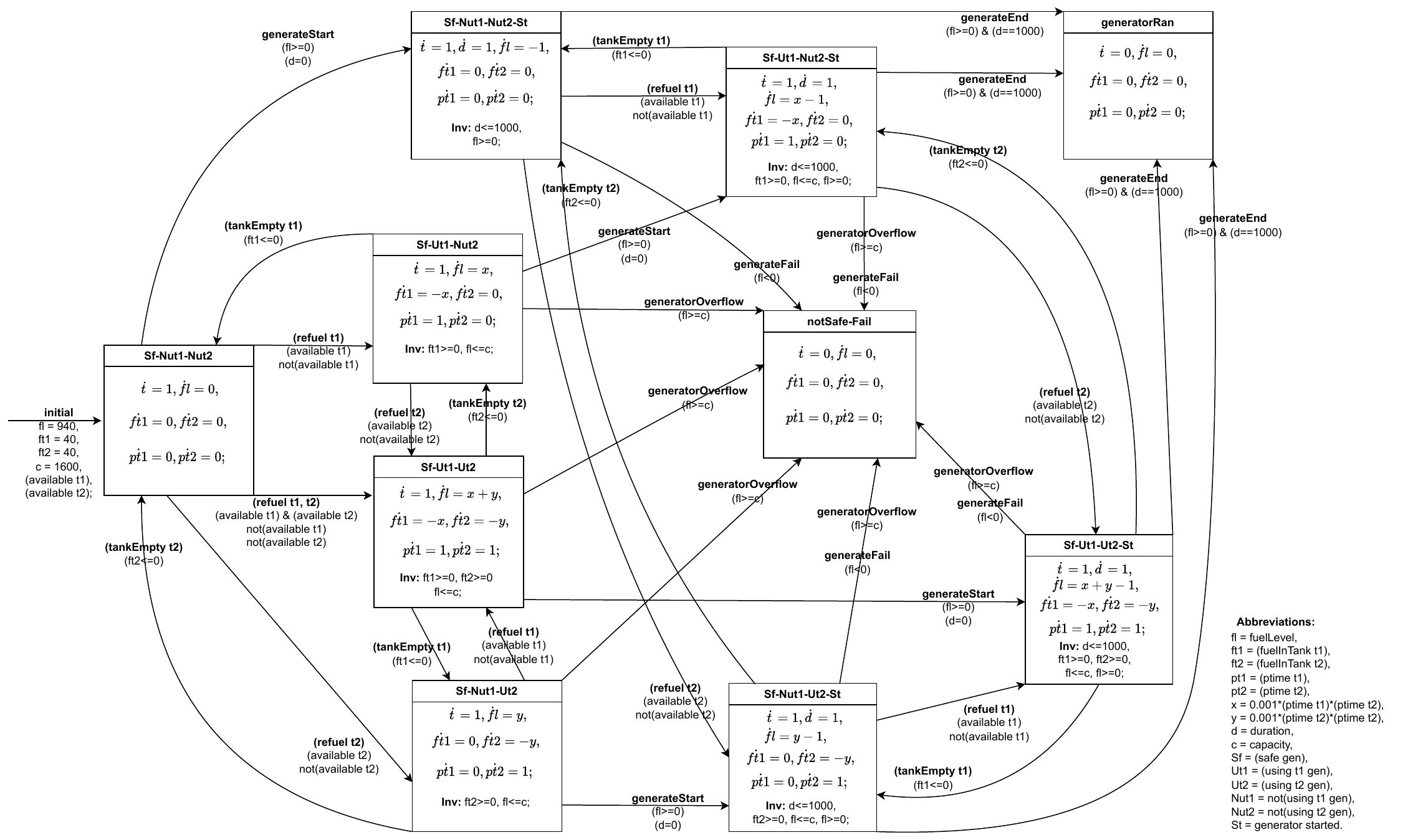}
                      \captionof{figure}{A hybrid automaton model of the generator-events domain for the planning problem instance 1.}
                      \label{genModel}
                \end{minipage}};
\end{tikzpicture}

\noindent Listing 11 and Listing 12 show the generated plan for the planning problem instances of Listing 9 and Listing 10 by \textsc{SMTPlan+}, respectively.

\begin{lstlisting}[basicstyle=\footnotesize,caption={A plan generated by \textsc{SMTPlan+} on the generator-events domain for the planning problem instance 1 in Listing 9. The plan duration is 1064.0 units.},captionpos=b,frame=single,label={Listing11},float=htbp]
    (@\textbf{\underline{Time}}@)         (@\textbf{\underline{Action}}@)              (@\textbf{\underline{Duration}}@)
    0.0:     (refuel gen t1)        [0.0]
    0.0:     (refuel gen t2)        [0.0]
    64.0:    (generate gen)        [1000.0]
\end{lstlisting}

\begin{lstlisting}[basicstyle=\footnotesize,caption={A plan generated by \textsc{SMTPlan+} on the generator-events domain for the planning problem instance 2 in Listing 10. The plan duration is 1064.0 units.},captionpos=b,frame=single,label={Listing12},float=htbp]
    (@\textbf{\underline{Time}}@)         (@\textbf{\underline{Action}}@)              (@\textbf{\underline{Duration}}@)
    0.0:     (refuel gen t1)        [0.0]
    0.0:     (refuel gen t2)        [0.0]
    0.0:     (refuel gen t3)        [0.0]    
    64.0:    (generate gen)        [1000.0]
\end{lstlisting}


\section{Planetary Lander Domain} \label{appendix3}

\begin{lstlisting}[basicstyle=\footnotesize,caption={The planetary-lander domain in PDDL+.},captionpos=b,frame=single,label={lst:Listing13}]
(define (domain power)
(:requirements :typing :durative-actions :fluents :time 
  :negative-preconditions :timed-initial-literals)
(:types equipment)
(:predicates (day) (commsOpen) (readyForObs1) (readyForObs2) (gotObs1)
  (gotObs2) (available ?e - equipment))
(:functions (demand) (supply) (soc) (charge_rate) (daytime) (heater_rate)
  (dusk) (dawn) (fullprepare_durtime) (prepareobs1_durtime)
  (prepareobs2_durtime) (observe1_durtime) (observe2_durtime) (obs1_rate)
  (obs2_rate) (A_rate) (B_rate) (C_rate) (D_rate) (safeLevel) (solar_const))
(:process charging
:parameters ()
:precondition (and (< (demand) (supply)) (day))
:effect (and (increase (soc) (* #t (* (* (- (supply) (demand))(charge_rate))
  (- 100 (soc)))))))
(:process discharging
:parameters ()
:precondition (> (demand) (supply))
:effect (decrease soc (* #t (- (demand) (supply)))))
(:process generating
:parameters ()
:precondition (day)
:effect (and (increase (supply) (* #t (* (* (solar_const) (daytime))
(+(*(daytime) (-(* 4 (daytime)) 90)) 450)))) (increase (daytime) (* #t 1))))
(:process night_operations
:parameters ()
:precondition (not (day))
:effect (and (increase (daytime) (* #t 1)) (decrease (soc)
  (* #t (heater_rate)))))
(:event nightfall
:parameters ()
:precondition (and (day) (>= (daytime) (dusk)))
:effect (and (assign (daytime) (- (dawn))) (not (day))))
(:event daybreak
:parameters ()
:precondition (and (not (day)) (>= (daytime) 0))
:effect (day))
(:durative-action fullPrepare
:parameters (?e - equipment)
:duration (= ?duration (fullprepare_durtime))
:condition (and (at start (available ?e)) (over all (> (soc) (safelevel))))
:effect (and (at start (not (available ?e))) (at start (increase (demand)
  (A_rate))) (at end (available ?e)) (at end (decrease (demand) (A_rate)))
  (at end (readyForObs1)) (at end (readyForObs2))))
(:durative-action prepareObs1
:parameters (?e - equipment)
:duration (= ?duration (prepareobs1_durtime))
:condition (and (at start (available ?e)) (over all (> (soc) (safelevel))))
:effect (and (at start (not (available ?e))) (at start (increase (demand)
  (B_rate))) (at end (available ?e)) (at end (decrease (demand) (B_rate)))
  (at end (readyForObs1))))
(:durative-action prepareObs2
:parameters (?e - equipment)
:duration (= ?duration (prepareobs2_durtime))
:condition (and (at start (available ?e)) (over all (> (soc) (safelevel))))
:effect (and (at start (not (available ?e))) (at start (increase (demand)
  (C_rate))) (at end (available ?e)) (at end (decrease (demand) (C_rate)))
  (at end (readyForObs2))))
(:durative-action observe1
:parameters (?e - equipment)
:duration (= ?duration (observe1_durtime))
:condition (and (at start (available ?e)) (at start (readyForObs1))
  (over all (> (soc) (safelevel))) (over all (not (commsOpen))))
:effect (and (at start (not (available ?e))) (at start (increase (demand)
  (obs1_rate))) (at end (available ?e)) (at end (decrease (demand)
  (obs1_rate))) (at end (not (readyForObs1))) (at end (gotObs1))))
(:durative-action observe2
:parameters (?e - equipment)
:duration (= ?duration (observe2_durtime))
:condition (and (at start (available ?e)) (at start (readyForObs2))
  (over all (> (soc) (safelevel))) (over all (not (commsOpen))))
:effect (and (at start (not (available ?e))) (at start (increase (demand)
  (obs2_rate))) (at end (available ?e)) (at end (decrease (demand)
  (obs2_rate))) (at end (not (readyForObs2))) (at end (gotObs2)))) )
\end{lstlisting}

\begin{lstlisting}[basicstyle=\footnotesize,caption={The planning problem instance 1 in the planetary-lander domain in PDDL+.},captionpos=b,frame=single,label={lst:Listing14},float=htbp]
(define (problem planety_prob)
(:domain planety) 
(:objects unit1 - equipment)
(:init  (at 0 (not (commsOpen))) (available unit1) (= A_rate 4.0)
(= B_rate 2.0) (= C_rate 2.5) (= D_rate 1.5) (= obs1_rate 4)
(= obs2_rate 5) (= fullprepare_durtime 3) (= observe1_durtime 7)
(= observe2_durtime 8) (= prepareobs1_durtime 1) (= prepareobs2_durtime 2)
(= demand 0) (= supply 0) (= soc 93) (= charge_rate 0.0095) (= daytime 4)
(= heater_rate 15.1) (= dusk 9.0) (= dawn 3.0) (= safeLevel 15.0)
(= solar_const 0.03) (not(gotObs1 )) (not(gotObs2 )) (not(readyforobs1))
(not(readyforobs2 )) (day))
(:goal (and (gotObs1) (gotObs2))) ) 
\end{lstlisting}

\begin{lstlisting}[basicstyle=\footnotesize,caption={The planning problem instance 2 in the planetary-lander domain in PDDL+.},captionpos=b,frame=single,label={lst:Listing15},float=htbp]
(define (problem planety_prob)
  (:domain planety) 
  (:objects unit1 - equipment)
  (:init  (at 0 (not (commsOpen))) (available unit1) (= A_rate 4.0)
    (= B_rate 2.0) (= C_rate 2.5) (= D_rate 1.5) (= obs1_rate 4)
    (= obs2_rate 5) (= fullprepare_durtime 3) (= observe1_durtime 7)
    (= observe2_durtime 8) (= prepareobs1_durtime 1) (= prepareobs2_durtime 2)
    (= demand 0) (= supply 0) (= soc 50) (= charge_rate 0.0095) (= daytime 4)
    (= heater_rate 15.1) (= dusk 9.0) (= dawn 3.0) (= safeLevel 25.0)
    (= solar_const 0.03) (not(gotObs1 )) (not(gotObs2 )) (not(readyforobs1))
    (not(readyforobs2 )) (day))
  (:goal (and (gotObs1) (gotObs2)))) 
\end{lstlisting}

\noindent{Listing 16 shows the generated plan on the planning problem instance 1 in Listing 14 by \textsc{UPMurphi}.}

\begin{lstlisting}[basicstyle=\footnotesize,caption={A plan generated by \textsc{UPMurphi} on the planetary-lander domain on planning problem in Listing 14. The plan duration is 18.002 units.},captionpos=b,frame=single,label={Listing16},float=htbp]
    (@\textbf{\underline{Time}}@)         (@\textbf{\underline{Action}}@)              (@\textbf{\underline{Duration}}@)
    0.000:   ( fullprepare1 unit1) [3.000]
    3.001:   ( observe1 unit1)     [7.000]
    10.002:  ( observe22 unit1)    [8.000]
\end{lstlisting}

\end{document}